%% file: main.tex
\documentclass[11pt]{article}

\usepackage[margin=1in]{geometry}
\usepackage{setspace}
\usepackage[T1]{fontenc}
\usepackage[utf8]{inputenc}
\usepackage{lmodern}

\usepackage{amsmath}
\usepackage{amssymb}
\usepackage{siunitx}
\usepackage{bm}  

\usepackage{graphicx}
\usepackage{wrapfig}
\usepackage{tabularx}
\usepackage{booktabs}
\usepackage{multirow}
\usepackage{longtable}
\usepackage{makecell}
\usepackage{tikz}
\usepackage{color, xcolor}
\usepackage{comment}

\usepackage{booktabs}
\usepackage{tabularx}
\usepackage{subcaption}

\usepackage{array}
\usepackage{rotating} 
\usepackage{pgfplots} 
\pgfplotsset{compat=1.18} 

\usepackage{algorithm}
\usepackage{algpseudocode}
\usepackage{hyperref}
\hypersetup{
    colorlinks=true,
    linkcolor=blue,
    citecolor=blue,
    urlcolor=blue
}
\usepackage[numbers,sort&compress]{natbib}
\usepackage{csquotes}
\usepackage{todonotes}

\newcommand{\mypar}[1]{\smallskip\noindent\textbf{#1.}}

\definecolor{ao(english)}{rgb}{0.0, 0.5, 0.0}
\definecolor{cadmiumgreen}{rgb}{0.0, 0.42, 0.24}
\definecolor{cinnamon}{rgb}{0.82, 0.41, 0.12}
\definecolor{burntumber}{rgb}{0.54, 0.2, 0.14}
\definecolor{circlegreen}{RGB}{78,145,6}
\definecolor{circlered}{RGB}{164,0,0}

\newcommand{\keywords}[1]{%
  \begin{center}
    \begin{minipage}{0.88\textwidth}
      \vspace{1em}
      \noindent\textbf{Keywords:} #1
    \end{minipage}
    \vspace{1.5em}
  \end{center}
}

\title{An Uncertainty-Aware ED-LSTM for Probabilistic Suffix Prediction}

\author{
  Henryk Mustroph \and
  Michel Kunkler \and
  Stefanie Rinderle-Ma
}

\date{} 

\begin{document}

\maketitle

\begin{center}
    \vspace{-1em}
    Technical University of Munich, TUM School of Computation, Information and Technology, Garching, Germany \\
    \texttt{\{henryk.mustroph, michel.kunkler, stefanie.rinderle-ma\}@tum.de}
    \vspace{1.5em}
\end{center}

\begin{abstract}
\noindent Suffix prediction of business processes forecasts the remaining sequence of events until process completion. Current approaches focus on predicting the most likely suffix, representing a single scenario. However, when the future course of a process is subject to uncertainty and high variability, the expressiveness of such a single scenario can be limited, since other possible scenarios, which together may have a higher overall probability, are overlooked. 
To address this limitation, we propose probabilistic suffix prediction, a novel approach that approximates a probability distribution of suffixes. The proposed approach is based on an Uncertainty-Aware Encoder-Decoder LSTM (U-ED-LSTM) and a Monte Carlo (MC) suffix sampling algorithm. We capture epistemic uncertainties via MC dropout and aleatoric uncertainties as learned loss attenuation. 
This technical report presents a comprehensive evaluation of the probabilistic suffix prediction approach's predictive performance and calibration under three different hyperparameter settings, using four real-life and one artificial event log.
The results show that: i) probabilistic suffix prediction can outperform most likely suffix prediction, the U-ED-LSTM has reasonable predictive performance, and ii) the model's predictions are well calibrated.

\end{abstract}

\keywords{Probabilistic Suffix Prediction \and Epistemic and Aleatoric Uncertainties \and Encoder-Decoder LSTM}

\section{Introduction} \label{sec:introduction}
In recent years, predicting the future course of a running business process (BP) using machine learning models has gained considerable attention in the field of Predictive Process Monitoring (PPM) \cite{maggi_2014_ppm}. 
Recent works have developed approaches that predict an entire sequence of remaining events, known as suffix prediction, using Neural Networks (NN) \cite{evermann_2017_lstm_suffix,tax_2017_ppm_lstm,camargo_2019_lstm_bpm,lin_2019_ppm,taymouri_2021_deep_adv_model,rama_maneiro_2024_exp_rnn,gunnarsson_2023_,wuyts_2024_sutran}.
Current suffix prediction approaches have focused on predicting a single most likely suffix. However, in certain domains, the future course of a business process is often subjected to uncertainties and high variability, making it unlikely that it will match exactly with the predicted most likely suffix.

Consider a drug development process in a pharmaceutical company. This process often involves uncertainty and variability, as unexpected events and human factors can influence how the process continues. The company aims to predict the future course of the process, for example, to plan resources, estimate how much time is remaining, and assess the chances of receiving market approval for the drug. Because the process can be highly variable, relying only on a single most likely prediction may not be sufficient. Instead, using confidence intervals can help the company better prepare for different possible outcomes by providing a range of likely scenarios.

Machine learning (ML) distinguishes epistemic and aleatoric uncertainties \cite{hullermeier_2021_alea_epi_unc}.
Epistemic uncertainties are reducible and stem from a lack of knowledge, e.g., training data.
Aleatoric uncertainties, conversely, are irreducible.
For instance, in the context of a business process, they can stem from external factors beyond the control of the organization running the process, such as delays in deliveries from external stakeholders or the involvement of humans in process execution.
%
In this work, instead of predicting a single most likely suffix, we consider epistemic and aleatoric uncertainties to predict a probability distribution of suffixes.
In line with the term probabilistic learning, used in ML to emphasize that not a single target is learned, but a target distribution \cite{klein_distributional}, we refer to our approach as \textit{Probabilistic Suffix Prediction}. 
We achieve probabilistic suffix prediction by first training an \textit{Uncertainty-Aware Encoder-Decoder Long Short-Term Memory} (U-ED-LSTM) NN on event log data and then by approximating a probability distribution of suffixes with our \textit{Monte-Carlo (MC) suffix sampling} algorithm.
The U-ED-LSTM captures epistemic uncertainties by applying MC dropout\cite{gal_2016_dropout,gal_2016_dropout_rnn,kendall_2017_caeu} and aleatoric uncertainties by learning probability distributions via learned loss attenuation \cite{kendall_2017_caeu}.
The MC suffix sampling algorithm generates suffix samples by performing multiple MC sampling iterations over the U-ED-LSTM. Each suffix sample is constructed by autoregressively drawing events from the probability distributions learned through loss attenuation.
By drawing sufficiently many samples, a posterior distribution of suffixes is approximated.
%

In this technical report, we evaluate the proposed probabilistic suffix prediction approach by analyzing the predictive performance and calibration of the U-ED-LSTM model across three different hyperparameter settings on four real-life and one artificial event log, preprocessed to datasets usable for PPM. 
To thoroughly assess the effectiveness of the probabilistic approach, we compare the probabilistic suffix predictions, by aggregating mean predictions out of all sampled suffixes, with most likely suffix predictions. 
The results demonstrate that i) the probabilistic suffix prediction approach can outperform most likely suffix prediction the U-ED-LSTM has reasonable predictive performance across various event logs, and ii) the model is capable of capturing temporal uncertainty in the event logs, shown by the calibration analysis of the remaining time predictions of the probabilistic suffix prediction approach.
This affirms the usage of probabilistic suffix prediction for more advanced tasks, such as confidence interval estimation based on sampled suffixes, an area not yet thoroughly explored in suffix prediction.
%

The technical report is outlined as follows: Section \ref{sec:prel} covers preliminaries, Section \ref{sec:meth} describes our probabilistic suffix prediction framework, Section \ref{sec:eval} presents the evaluation, Section \ref{sec:relwork} discusses related approaches, and Section \ref{sec:con} concludes the work.

\section{Preliminaries}\label{sec:prel}
This section introduces general uncertainty concepts, how to model uncertainty in NN, and a definition of suffix and remaining time prediction of BPs.

\subsection{Uncertainty in Machine Learning}\label{sec:prel:unc}
ML distinguishes epistemic and aleatoric uncertainties. \cite{hullermeier_2021_alea_epi_unc} defines and describes both types of uncertainties. 
Epistemic uncertainty is referred to as \enquote{uncertainty due to a lack of knowledge about the perfect predictor} \cite{hullermeier_2021_alea_epi_unc} and is reducible.
Epistemic uncertainty can be further divided into approximation uncertainty and model uncertainty.
Approximation uncertainty refers to a lack of data for selecting appropriate parameters for a predictor model and can generally be reduced by obtaining more training samples.
Model uncertainty refers to a model's insufficient approximation capabilities and can be reduced by training models with a higher capacity.
There is ongoing debate regarding how epistemic uncertainty should be captured, with one possibility being the use of probability distributions \cite{hullermeier_2021_alea_epi_unc}.
Aleatoric uncertainty is irreducible as it stems from inherently random effects in the underlying data. Aleatoric uncertainty is \enquote{appropriately modeled in terms of probability distributions} \cite{hullermeier_2021_alea_epi_unc} and can henceforth be learned in a probabilistic model.

\mypar{Uncertainty-Aware Neural Networks (NN)} For NNs, two common approaches for estimating a model's uncertainty in its prediction are Bayesian approximation and ensemble learning-based techniques \cite{abdar}.
Bayesian approximation can be conducted with Bayesian Neural Networks (BNNs).
BNNs assume their weights follow probability distributions, allowing a posterior distribution to be inferred, which can be used to quantify uncertainty. In most cases, obtaining an analytical solution for the posterior distribution is intractable due to neural networks' high non-linearity and dimensionality. Even techniques for approximating the posterior distribution, such as Markov Chain Monte Carlo or Variational Inference (VI) methods, can still be computationally expensive \cite{abdar,gal_2016_dropout}.
Ensemble techniques, on the other hand, achieve uncertainty quantification by aggregating the predictions of multiple models. This can also become computationally expensive, especially when numerous complex models are involved \cite{abdar}.

\mypar{Epistemic Uncertainty using Dropout as a Bayesian Approximation} 
Using Dropout during training and inference at every weight layer in an NN can be a simple and computationally efficient variational inference method for Bayesian approximation of a posterior distribution \cite{gal_2016_dropout}.
This approach is referred to as \textit{Monte Carlo (MC) dropout} because the posterior distribution $p(W \vert X, Y)$ is approximated with a variational distribution $q_{\theta}(W)$.
Masked weights are sampled from the variational distribution $\hat{W} \sim q_{\theta}(W)$, where $\theta$ denotes the set of the variational distribution’s parameters (weights and bias terms) to be optimized.
In practice, a dropout mask is often sampled from a Bernoulli distribution \(z \sim \operatorname{Bernoulli}(\text{p})\), where $\text{p}$ denotes the dropout probability. The dropout mask is then applied on the NN's weight matrices $W$ such that \(\hat{W} = W \operatorname{diag}(z)\).
During training, the $L_2$ regularization on the NN parameters $\theta$ ensures the method aligns with a probabilistic framework.

\mypar{Heteroscedastic Aleatoric Uncertainty as Learned Loss Attenuation}
Heteroscedastic models assume that observation noise, which follows a probability distribution, can vary with the input data $x$. This input-dependent observation noise, denoted as $\sigma(x)$, captures aleatoric uncertainty arising from inherent randomness in the data-generating process.
To explicitly model this irreducible uncertainty, NNs are extended to directly learn $\sigma(x)$. This is commonly done assuming that the observation noise follows a Normal distribution. 
To learn this observation noise using an NN $f^W(\cdot)$ parameterized by weights $W$, an additional output neuron $f^W_{\sigma}(x)$ is added to the (mean) output neuron $f^W_y(x)$.
However, in cases where both the inputs $x$ and the predicted outputs $f^W_y(x)$ are constrained to be strictly positive (e.g., time durations), assuming that the observation noise follows a Log-Normal distribution may be more appropriate. This is equivalent to assuming that the observation noise is Normal distributed over the input data in the log-transformed space, i.e., $\ln(x)$.
Training the standard deviation of a probability distribution by including it in the loss function is referred to as learned loss attenuation \cite{kendall_2017_caeu}.
In the regression case, the adapted loss function over $N$ training samples with target \(y\) can be written as the negative log-likelihood of the underlying probability density function\footnote{For a detailed derivation of the loss function, see \cite{bishop_1994_mixture_density}.}:
\begin{equation}\label{eq:loss_aleatoric_cont}
\begin{aligned}
   \mathcal{L}_{con} = 
   \frac{1}{N}\sum_{i=1}^{N} 
\frac{1}{2}\left( \frac{(y_i - f^W_y(x_i))^2}{f^W_{\sigma}(x_i)^2} + \log\bigl(f^W_{\sigma}(x_i)^2\bigr) \right)
\end{aligned}
\end{equation}

In the case of classification, NNs typically employ the Softmax function, which already outputs a categorical probability distribution.
However, this probability distribution might not capture model uncertainties \cite{kendall_2017_caeu}.
Therefore, means and variances can also be learned on the predicted logits.
Since the logits are passed in the Softmax function, MC integration has to be applied, i.e., averaging the cross-entropy loss of multiple draws from the logits distributions. We denote the number of MC trials with $T$, the categorical classes with $C$, and the ground truth class with $c$: \(\hat{z}_{i,t} = f^{W}_y(x_i) + f^{W}_\sigma(x_i) \epsilon_t,~~\epsilon_t \sim \mathcal{N}(0, I)\).
\begin{equation}\label{eq:loss_aleatoric_cat}
\begin{aligned}
   \mathcal{L}_{cat} &= 
   \frac{1}{N}\sum_{i=1}^{N}
   -\text{log} \left( \frac{1}{T} \sum_{t=1}^{T} \left( \text{exp}(\hat{z}_{i,t,c} - \text{log}(\sum_{c'}^{C}(\text{exp}(\hat{z}_{i,t,c'})))) \right) \right)
\end{aligned}
\end{equation}

The combination of epistemic uncertainty quantification using MC dropout and aleatoric uncertainty quantification via learned loss attenuation was first proposed by \cite{kendall_2017_caeu}, and this approach can be applied to any NN architecture.

\subsection{Suffix Prediction}\label{sec:prel:ppm}
We define an event log \(EL := \{t^{(1)}, t^{(2)}, ..., t^{(L)}\}\) as a set of cases, where \(L\) denotes the total number of cases.
A case is a sequence of events denoted by \(t^{(l)} := \langle e_1, e_2, ..., e_M\rangle\), where \(M\) is the number of events in case $l$.
An event is a tuple of event attributes, denoted \(e_m := (a_m, t_m, (d_{m_1}, ..., d_{m_k}))\).
In this work, we assume that an event has at least two attributes:
i) An event label $a_m$, which links the event to a class of event types, and ii) a timestamp attribute $t_m$, which expresses the time an event happened.
Additional event attributes are denoted as \((d_{m_1}, ..., d_{m_k})\).
We assume that event attributes are either categorical or continuous. 
A case can be split into several prefix and suffix pairs.
A prefix is defined as \(p_{\leq k} := \langle e_1, e_2, ..., e_k \rangle\), with \(1 \leq k < M\).
A suffix is defined as \(s_{> k} := \langle e_{k+1}, \dots, e_{M} \rangle\).
Suffix prediction involves predicting a suffix $\hat{s}$ based on an input prefix \(p_{\leq k}\).

\section{Probabilistic Suffix Prediction Framework} \label{sec:meth}
This section presents the probabilistic suffix prediction framework consisting of the \textit{U-ED-LSTM} model and the \textit{MC suffix sampling} algorithm.

\subsection{Uncertainty-Aware Encoder-Decoder LSTM}\label{sec:meth:model}
The U-ED-LSTM implementation comprises the data preparation, model architecture, and loss functions for training.

\mypar{Data Pre-processing and Embedding}
Given an event log, we first apply feature engineering techniques to the events’ timestamp attribute to derive additional features for the U-ED-LSTM. We introduce a \textit{case elapsed time} attribute, representing the time elapsed since the first event in the case, an \textit{event elapsed time} attribute, representing the time since the last event within the same case (with the value set to 0 for the first event), a \textit{day of the week} attribute, and a \textit{time of day} attribute. The latter two features are incorporated due to the potential influence of periodic trends on the future course of a process. For instance, in a company that operates only on weekdays, when an activity is completed on Friday evening, the next activity is unlikely to occur before Monday.
Missing values for continuous event attributes are encoded as 0.
For all continuous event attributes in the encoder input, decoder input, and decoder output, we apply standard scaling, excluding the raw timestamp, under the assumption that the observation noise follows a Normal distribution.
For all continuous event attributes in the decoder input and output, when assuming that the observation noise follows a Log-Normal distribution, we first transform the values into log-space using the natural logarithm function, $\ln(1 + x)$, to ensure only positive inputs. Then, again, Standard scaling is applied to the log-transformed values.
Following \cite{wuyts_2024_sutran}, we also apply input padding to facilitate batch training: Each case is padded with zeros at the beginning to a fixed length, determined by the maximum case length in the event log, excluding the top 1.5\% of the longest cases. This allows multiple prefixes, regardless of the actual prefix length, to be concatenated into a single batch tensor.
%
After the data pre-processing, all categorical event attributes are embedded using an embedding layer stack that maps each categorical event attribute into a vector of fixed dimensionality. For every event attribute with $K$ unique category classes, we add an additional NaN class and an unknown class (a category class not present in the training data). The embedding layer is defined as a learnable weight matrix of size \((K+2) \times D\), where \(D = \text{min}(600, \text{round}(1.6 (K+2)^{0.56}))\) is the chosen embedding dimension, following a common heuristic (see \cite{wuyts_2024_sutran}).

\begin{figure}[h]
\centering
    \includegraphics[width=0.8\textwidth]{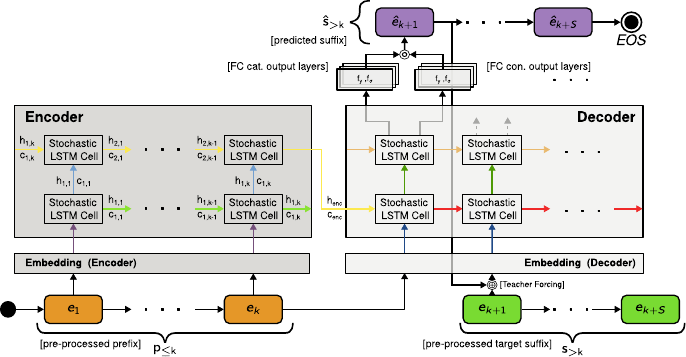}
    \caption{U-ED-LSTM Architecture and Training Pipeline}
\label{fig:enc_dec_architecture}
\end{figure}

\mypar{Model Architecture} 
The U-ED-LSTM employs an encoder-decoder (ED) architecture based on LSTMs \cite{hochreiter_1997_lstm}, similar as in \cite{taymouri_2021_deep_adv_model}.
LSTMs are well-suited for handling sequential data and have been proven effective for suffix prediction in business processes \cite{ceravolo_2024_ppm_trends}.
Additionally, ED architectures offer flexibility by decoupling tasks between the encoder and decoder and by handling different input and output event features: the encoder can focus on summarizing the prefix and can take all event attributes as input, while the decoder leverages these representations to predict target event attributes, e.g., only the event labels and time features (see \cite{lin_2019_ppm,taymouri_2021_deep_adv_model}). The U-ED-LSTM implementation allows users to flexibly select the encoder's input event attributes, as well as the decoder's input and output event attributes.
For both the encoder and decoder, we use LSTMs with stochastic LSTM cells, which apply dropout as a Bayesian approximation, as in \cite{weytjens_2022_unc_ppm}.
Rather than employing single output neurons to represent either a mean or logit value, each output is now represented by two neurons: one for the mean (or logit) and a second for the standard deviation.
Fig. \ref{fig:enc_dec_architecture} illustrates the U-ED-LSTM architecture during training.
%
Naive dropout has shown to be ineffective in recurrent NNs such as LSTMs \cite{gal_2016_dropout_rnn}.
Therefore, \cite{gal_2016_dropout_rnn} have proposed a different dropout variant in which the same dropout mask is applied across all time steps in an RNN, referred to variational dropout. We apply variational dropout in our U-ED-LSTM, indicated by the colored arrows in Figure \ref{fig:enc_dec_architecture}.

The \textbf{encoder} processes input prefixes to compress the event sequence information into a fixed-length representation, known as a latent vector. 
More formally, we define the encoder as a function \(f^{\hat{W}_{enc}}(\cdot)\), with masked weights sampled from the encoder's variational distribution \(\hat{W}_{enc} \sim q_{\theta_{enc}}(W_{enc})\).
For a given input prefix \(p_{\leq k}\), a latent vector tuple is predicted:
\(f^{\hat{W}_{enc}}(p_{\leq k}) := (h_{k}, c_{k}).\)
Thereby, \(h_{k}\) and \(c_{k}\) represent the last hidden and cell state in the encoder. 

The \textbf{decoder} receives at the first timestep the latent vector from the encoder $(h_{enc}, c_{enc})$ and the last event from the prefix $e_k$.
It then updates the latent vector and generates a new event.
At each subsequent timestep, the model uses the previously updated latent vector tuple and a previous event.
During training, teacher forcing is applied, selecting either the previous event from the target suffix or the last predicted event randomly.
Then the decoder autoregressively predicts $S$ events, where $S$ is a predefined fixed sequence length.
%
Similar to the encoder, we sample the decoders masked weights from its variational distribution $\hat{W}_{dec} \sim q_{\theta_{dec}}(W_{dec})$.
For a given time step \(s = 0, 1, \dots, S-1\), we define \(e_{k+s}\) as the current event, \((h_{k+s}, c_{k+s})\) as the current latent vector tuple, and the prediction of the next event and updated latent vector tuple as follows:
\(f^{\hat{W}_{dec}}\bigl(e_{k+s},(h_{k+s},c_{k+s})\bigr) 
:= \bigl(\hat{e}_{k+(s+1)}, (h_{k+(s+1)}, c_{k(s+1)})\bigr).
\)
A predicted event consists of the concatenation of all its event attributes.
These can be categorical or continuous, such that the predicted event is defined as:
 $\hat{e}_{k+(s+1)} := (\{f^{\hat{W}_{dec}}_{con_{1}}(\cdot), ..., f^{\hat{W}_{dec}}_{con_{D}}(\cdot)\}, \{f^{\hat{W}_{dec}}_{cat_{1}}(\cdot), ..., f^{\hat{W}_{dec}}_{cat_{K}}(\cdot)\})$, where $D$ the number of continuous and $K$ the number of categorical attributes.
 The continuous attribute predictions consist of a mean prediction $\hat{y}$ and a log-variance prediction $\hat{v}_{con}$, $f^{\hat{W}_{dec}}_{con}(\cdot) := (\hat{y}, \hat{v}_{con})$.
 The categorical attribute predictions consist of a mean $\hat{l}$ and a log-variance vector $\hat{v}_{cat}$, where each element in the respective vectors represents a mean and a log-variance value of a categorical class, $f^{\hat{W}_{dec}}_{cat}(\cdot) := (\hat{l}, \hat{v}_{cat})$.
 The log-variance is predicted, rather than the variance itself, to ensure numerical stability \cite{kendall_2017_caeu}.
 
\mypar{Loss Functions} To train the U-ED-LSTM, we use two distinct attenuated loss functions, one for continuous and another one for categorical event attributes. The loss is calculated for a batch of \(N\) prefix-suffix pair training samples, \(\{p^{(i)}_{\leq k}, s^{(i)}_{> k}\}_{i=1}^{N}\).

The continuous loss function based on Equation \ref{eq:loss_aleatoric_cont} is implemented as follows:
\begin{equation}\label{loss_num}
\begin{aligned}
    \mathcal{L}_{con} 
    &= \frac{1}{N \times S} \sum_{i=1}^{N} \sum_{s=0}^{S-1}
    \frac{1}{2}  \left( \frac{(y_{k+(s+1)}^{(i)} - \hat{y}_{k+(s+1)}^{(i)})^2}{\hat{\sigma}_{k+(s+1)}^{2^{(i)}}} + \text{log}({\hat{\sigma}_{k+(s+1)}^{2^{(i)}}}) \right),\\
    &= \frac{1}{N \times S} \sum_{i=1}^{N} \sum_{s=0}^{S-1}
    \frac{1}{2} \left( \text{exp}(-\hat{v}_{k+(s+1)}^{(i)})( y_{k+(s+1)}^{(i)} - \hat{y}_{k+(s+1)}^{(i)})^2 + \hat{v}_{k+(s+1)}^{(i)}) \right)
\end{aligned}
\end{equation}

For categorical event attributes, we calculate the loss using MC integration by averaging the cross-entropy loss (CEL) over multiple samples drawn from the logits distribution. \(\hat{z} = \hat{l} + \hat{\sigma} \, \epsilon_t\), where \(\epsilon_t \sim \mathcal{N}(0, I)\).  The loss function based on Equation \ref{eq:loss_aleatoric_cat} is implemented as follows:
\begin{equation}\label{loss_class}
\begin{aligned}
    \mathcal{L}_{cat} 
    &= \frac{1}{N \times S}\sum_{i=1}^{N} \sum_{s=0}^{S-1}
    -\text{log} 
    \left(\frac{1}{T} \sum_{t=1}^{T} 
    \left( \text{exp}(
    \hat{z}_{k+(s+1), t, c}^{(i)} - \text{log}\sum_{c'}^{C}\text{exp}(\hat{z}_{k+(s+1), t, c'}^{(i)})) 
    \right)\right),\\ 
    &= \frac{1}{N \times S}\sum_{i=1}^{N} \sum_{s=0}^{S-1} 
    \left(\frac{1}{T} \sum_{t=1}^{T} 
    \text{CEL}(y_{k+(s+1)}^{(i)}, \hat{z}_{k+(s+1),t}^{(i)})\right)
\end{aligned}
\end{equation}

The total loss consists of a weighted sum of continuous losses and categorical losses, weighted by weight coefficient vectors \(w_{con}\) and \(w_{cat}\) and the \(L_2\) regularization term of the encoder's and decoder's parameters. The total loss is implemented as follows:
\begin{equation}\label{loss_total}
\begin{aligned}
    \mathcal{L}
    = & \sum_{k=1}^{K} w^{k}_{con} \mathcal{L}^{k}_{con}
    + \sum_{d=1}^{D} w^{d}_{cat} \mathcal{L}^{d}_{cat} 
    + \lambda (\|\theta_{enc}\|^2_2 +\|\theta_{dec}\|^2_2 )
\end{aligned}
\end{equation}

\subsection{MC Suffix Sampling Algorithm}\label{sec:meth:prob_suff_pred}
The \textit{MC suffix sampling algorithm} approximates a posterior distribution of suffixes.
It is outlined in Algorithm \ref{alg:sampling} and combines elements from other MC sampling algorithms.
Algorithm \ref{alg:sampling} uses dropout as a Bayesian approximation to consider epistemic uncertainty, similar to \cite{zhu_2017_seq2seq_unc}, and considers aleatoric uncertainty by drawing samples from probability distributions learned via loss attenuation, following the approach of \cite{salinas}. 
The algorithm can be outlined as follows:
At each MC trial, a suffix is sampled.
For sampling a suffix, the prefix is first passed into the encoder to obtain a latent vector tuple $(h_{enc}, c_{enc})$.
Similar to \cite{zhu_2017_seq2seq_unc}, we employ variational dropout on the encoder.
The decoder is then employed for sampling a suffix auto-regressively.
At the first step, the latent vector tuple and the last event from the prefix are taken as input.
The event attributes of the first event of the suffix are then sampled, with categorical and continuous attributes handled differently:
For continuous event attributes, the event attribute values are drawn from Normal distributions with the predicted mean and variance pairs.
For categorical event attributes, the logit values are first drawn from Normal distributions and then passed through a Softmax function to obtain a categorical distribution.
In a subsequent step, the categorical class is drawn from this distribution.
The auto-regressive prediction of the next event continues until either \(EOS\) is predicted or the predefined maximum sequence length $M$ is reached.
The algorithm returns \(\tilde{S}\), a set of sampled suffixes with size \(T\).
\input{algorithm/03a_algorithm}
%

\section{Evaluation}\label{sec:eval}
The evaluation presents the predictive performance and calibration results of the probabilistic suffix prediction approach and the U-ED-LSTM under three different hyperparameter settings, using standard suffix prediction and calibration metrics on four real-life and one artificial dataset.
We compare probabilistic suffix prediction by comparing the aggregated mean of all sampled suffixes with the most likely suffix prediction.
Furthermore, we evaluated the calibration of the remaining time predictions to demonstrate the model’s ability to capture temporal uncertainties present in the event logs.
The goal is to: i) demonstrate that our probabilistic suffix prediction approach achieves reasonable performance and to identify the best-performing hyperparameter setting among the three evaluated, and ii) show that probabilistic suffix prediction outperforms most likely suffix prediction. This affirms the usage of probabilistic suffix prediction for more advanced tasks, such as confidence interval estimation based on sampled suffixes, an area not yet thoroughly explored in suffix prediction.
Our implementation and evaluation are publicly available.
\footnote{{\tiny Repository: \url{https://github.com/ProbabilisticSuffixPredictionLab/Probabilistic_Suffix_Prediction_U-ED-LSTM_pub}}}

\mypar{Datasets} We evaluated the U-ED-LSTM predictive performance on four real-life data sets. 
The Helpdesk\footnote{{\tiny Helpdesk:  \url{https://doi.org/10.4121/uuid:0c60edf1-6f83-4e75-9367-4c63b3e9d5bb}}} dataset is an event log from a ticket management system from an Italian software company.
The Sepsis\footnote{{\tiny Sepsis:  \url{https://doi.org/10.4121/uuid:915d2bfb-7e84-49ad-a286-dc35f063a460}}} dataset represents the pathway of patients diagnosed with Sepsis through a hospital.
The BPIC-2017\footnote{{\tiny BPIC-17: \url{https://doi.org/10.4121/uuid:5f3067df-f10b-45da-b98b-86ae4c7a310b}}} dataset is a loan application process from a Dutch bank and has been investigated in the Business Process Intelligence Competition (BPIC) 2017.
The PCR\footnote{{\tiny PCR: \url{https://doi.org/10.5281/zenodo.11617408}}} dataset contains process logs from laboratory SARS-CoV-2 RT-PCR tests over one year.
The artificial dataset represents a repair shop process and is referred to as Repair.
Properties for each dataset are presented in Table \ref{tab:datasets}.
The datasets were split at the case level into training and testing sets using an 80\%-20\% ratio, following the approach in \cite{ketyko_2022_lstm_suffix}.
The training data was further divided into a training and validation set, resulting in an overall 65\%-15\%-20\% training-validation-testing data split. We conducted our evaluation on the test set, starting from a prefix length of 1, i.e., $p_{\leq k}, \forall k \geq 1$.
\input{tables/data}

\mypar{Training and Sampling}
We trained our U-ED-LSTM model on an NVIDIA GTX 4090 GPU.
Several training optimization techniques were implemented:
Since \cite{gunnarsson_2023_} showed that sequence-to-sequence training is more effective than single-event training for suffix prediction, we calculated the loss and optimized the model based on sequences of $S = 5$ events.
Using a fixed sequence length $S$ performed better in sequence-to-sequence tasks and also reduced the impact of early prediction errors in the suffix.
Furthermore, we applied probabilistic teacher forcing, i.e., selecting either the last predicted or target event as input for the next event prediction \cite{taymouri_2021_deep_adv_model}.
We set the initial teacher forcing probability to $0.8$, meaning that $80\%$ of the input events came from the target suffix, and then decreased the ratio from $20\%$ of the training epochs onward.
Since suffix prediction is a multi-task learning problem, we implemented a task-balancing algorithm called GradNorm \cite{chen_2018_gradnorm}. GradNorm dynamically adjusts the gradient magnitudes and tunes the weight coefficient vectors \(w_{con}\) and \(w_{cat}\) after each optimization step based on the relative importance of each event attribute on the overall loss.
We set the MC dropout rate to a constant $\text{p} = 0.1$ during training and sampling, as recommended in \cite{weytjens_2022_unc_ppm}.
For each prefix in our testing datasets, we sampled \num{1000} suffixes.

\mypar{Hyperparameter Settings}
The U-ED-LSTM was trained using three distinct hyperparameter settings.
In the \textbf{first setting}, we trained the U-ED-LSTM with 2-layer encoder and decoder LSTMs, along with a fully-connected (FC) layer in the decoder containing separate mean and variance heads for each output event attribute. We assumed normal distributed noise and used all event attributes as input features for the encoder and input and output features for the decoder.
In the \textbf{second setting}, we trained the U-ED-LSTM with a 4-layer encoder and decoder LSTMs, along with an FC layer in the decoder containing separate mean and variance heads for each output event attribute. We assumed normal distributed noise. All event attributes were used as input features for the encoder, while only the event label (activity) and time attributes were used as input and output features for the decoder.
In the \textbf{third setting}, we trained the U-ED-LSTM with a 4-layer encoder and decoder LSTMs, along with an FC layer in the decoder containing separate mean and variance heads for each output event attribute. We assumed normal distributed noise and log-normal distributed noise on continuous time feature event attributes. All event attributes were used as input features for the encoder, while only the event label and time attributes were used as input and output features for the decoder.
%
The remaining hyperparameters were set as follows:
In each hyperparameter setting, the encoder and decoder LSTM have a hidden size of 128.
We used the standard Adam optimizer in the first setting, while the more advanced AdamW optimizer was applied in the second and third settings.
Learning rates were set, in each setting, the same, to $1 \times 10^{-4}$ for the smaller Helpdesk, PCR, and Repair datasets, $1 \times 10^{-5}$ for the Sepsis dataset, and $1 \times 10^{-6}$ for the larger BPIC-17 dataset. After extensive testing, we observed that larger datasets benefited from lower learning rates.
A batch size of 128 was used for all datasets except for the BPIC-17, which required a batch size 256 during training, to reduce training time.
All models were trained for 200 epochs in the first and 100 epochs in the second and third settings, without early stopping but with continuous monitoring of validation set performance. Across all settings and datasets, the learning curves decreased during the first 100 epochs and stagnated. Based on this observation, and to reduce training time, we consistently trained models for 100 epochs in the second and third settings.
Additionally, we applied a $L^2$ regularization parameter of $\lambda = 1 \times 10^{-4}$.
Table \ref{tab:hyperparameters} summarizes the hyperparameters in each setting.
\input{tables/hyperparameter}

\subsection{Predictive Performance}
We evaluated the predictive performance of the probabilistic suffix prediction approach by comparing the aggregated mean of all sampled suffixes with the most likely suffix prediction.
Additionally, we report predictive performance results from other state-of-the-art suffix prediction approaches using either ED-LSTMS or Transformers from existing literature, evaluated on the same datasets. Since the reported approaches and models were not re-implemented, the comparison is not meant to be direct but to provide an intuition of how other approaches performed, demonstrating that the U-ED-LSTM achieves competitive and reasonable predictive performance.

\mypar{Metrics and Other Approaches}
To evaluate the predictive performance, we adopted three commonly used evaluation metrics for suffix prediction:
The \textit{Damerau-Levenshtein Similarity} (DLS) metric to asses the predicted event labels, also known as activity sequence prediction\cite{rama_maneiro_2024_exp_rnn,wuyts_2024_sutran,taymouri_2021_deep_adv_model,Pasquadibisceglie_2024_lupin,camargo_2019_lstm_bpm}
and the \textit{Mean Average Error} (MAE) of the remaining time prediction 
\cite{wuyts_2024_sutran,taymouri_2021_deep_adv_model,tax_2017_ppm_lstm,camargo_2019_lstm_bpm}.
Previous studies \cite{tax_2017_ppm_lstm,camargo_2019_lstm_bpm} showed that some suffix prediction methods struggle to predict the correct suffix length.
To address this, we also evaluated the suffix length prediction by calculating the MAE.
The most likely suffix prediction was obtained by auto-regressively generating a suffix by sampling the most likely event label at each step until the $EOS$ token is reached, similar to other works \cite{wuyts_2024_sutran,camargo_2019_lstm_bpm,tax_2017_ppm_lstm,lin_2019_ppm,taymouri_2021_deep_adv_model,gunnarsson_2023_}. 

The DLS on the event labels is defined as a normalized DLS distance $\text{DLS}(\hat{s}, s):= 1 - \frac{\text{DL}(s, \hat{s})}{\text{max}({\vert s \vert, \vert \hat{s} \vert})}$ where $s$ and $\hat{s}$ denote the true and predicted sequence of event labels. Informally, $\text{DLS} = 1$ expresses that the two sequences are identical, while a $\text{DLS} = 0$ expresses that the two are entirely dissimilar. 
For the most likely suffixes, we computed the $\text{DLS}$ by comparing the prediction to the ground truth suffix.
For the probabilistic suffix, we calculated the $\text{DLS}$ for each sampled suffix and took the mean.
To obtain the most likely remaining time \textit{sum} MAE, we summed the event elapsed times of the most likely suffix and compared the result to the ground truth remaining time.
For the probabilistic remaining time \textit{sum} MAE, we calculated the mean of the sums of the event elapsed times across all sampled suffixes and compared it to the ground truth.
The remaining time \textit{last} MAE is computed by taking the last case elapsed time of the most likely predicted suffix and comparing it to the ground truth.
For the probabilistic remaining time \textit{last} MAE, the metric is calculated as the mean of the last case elapsed times across all sampled suffixes, which is then compared to the ground truth.
Most often in literature the event elapsed times are predicted to calculate the remaining time \textit{sum}.
%
For the most likely suffixes, we compared their lengths with the ground truth suffix lengths.
For the probabilistic, we computed the length of each sampled suffix, took the mean, and compared it to the ground truth.

As existing models from the literature, the following are selected:
\begin{itemize}
    \item \textit{MM-Pred}, form \cite{lin_2019_ppm}, is an ED-LSTM-based approach that incorporates a ``Modulator'' component to weight the importance of decoder input events. It was evaluated on the Helpdesk and BPIC-17 datasets. The authors used a 70\%-10\%-20\% train-validation-test split. The model consists of a 2-layer encoder and decoder LSTMs with a hidden size 32 and a dropout rate of 0.2 for regularization during training.
    \item \textit{ED-LSTM-GAN} from \cite{taymouri_2021_deep_adv_model} was evaluated on the Helpdesk and BPIC-17 datasets. The authors used a 70\%-10\%-20\% train-validation-test split. The model consists of a 5-layer encoder and decoder LSTMs with a hidden size 32 as generator and an FC layer as discriminator as part of the GAN architecture. Training is conducted for 500 epochs, with early stopping applied after 30 epochs. RMSprop is the optimizer, with a $5 \times 10^{-5}$ learning rate. Probabilistic teacher forcing is used with a ratio of 0.1. 
    \item \textit{AE} (inspired by \cite{lin_2019_ppm,taymouri_2021_deep_adv_model}) and \textit{AE-GAN} (inspired by \cite{taymouri_2021_deep_adv_model}) from \cite{ketyko_2022_lstm_suffix}, both ED-LSTMs, were evaluated on the Helpdesk, Sepsis, and BPIC-17 datasets. An 80\%-20\% train-test split was used. Both models consist of 4-layer encoder and decoder LSTMs with a hidden size of 128 and are trained for 400 epochs, with early stopping applied after 50 epochs. The Adam optimizer was used with a learning rate of $1 \times 10^{-4}$, and a dropout rate of 0.3 is applied for regularization.
    \item \textit{SuTraN}, along with an ED-LSTM re-implementation of \cite{taymouri_2021_deep_adv_model}, from \cite{wuyts_2024_sutran}, was evaluated on the BPIC-17 dataset. The authors used a 55\%-35\%-25\% train-validation-test split. The ED-LSTM architecture consists of a 4-layer encoder and decoder LSTMs with a hidden size 64. A batch size of 128 is used, and models are trained for 200 epochs with early stopping based on the validation set performance. The Adam optimizer is applied with an initial learning rate of $2 \times 10^{-4}$, adjusted dynamically via an exponential learning rate scheduler with a decay factor of 0.96. A $1 \times 10^{-4}$ weight decay is used, and teacher forcing is employed during training.
\end{itemize}

\mypar{Predictive Performance Results}
The results of the probabilistic and most likely on suffix length and suffix event labels prediction (categorical) are depicted in Table \ref{tab:performance_suffix_dls}. The results of the probabilistic and most likely on remaining time sum (sum of each event elapsed times) and last (last case elapsed time) (continuous) predictions are depicted in Table \ref{tab:performance_remaining_time}. Furthermore, Figure \ref{fig:performance_setting2-HSB} presents detailed predictive performance results for the Helpdesk and Sepsis datasets, visualized across different prefix and suffix lengths. Figure \ref{fig:performance_setting2-PR} shows the corresponding results for the PCR and artificial Repair datasets.

For the Helpdesk dataset, the best suffix event label DLS results were obtained in Setting 2, with a DLS of 0.82 for the most likely prediction and 0.65 for the probabilistic prediction. These results are comparable to those reported in the literature, where DLS values for most likely predictions range from 0.84 for ED-LSTM-GAN \cite{taymouri_2021_deep_adv_model}, 0.86 for AE and AE-GAN \cite{ketyko_2022_lstm_suffix} to 0.87 (MM-Pred \cite{lin_2019_ppm}). Improvements are also observed for the suffix length MAE when comparing Setting 1 to Setting 2.
One possible reason Setting 2 provided the best results on the Helpdesk dataset could be the deeper architecture with a 4-layer encoder and decoder LSTM. This enhanced the U-ED-LSTM's ability to abstract higher-level patterns, capture more complex dependencies, and generalize better to longer suffixes.

On the Sepsis dataset, the U-ED-LSTM achieved the best suffix length MAE and the highest suffix event label DLS in Setting 3, reaching a DLS of 0.18 for the probabilistic prediction, which outperformed the most likely prediction that achieved a DLS of only 0.09. However, while this seems to be still a relatively poor result, it is comparable to those reported in the literature, such as 0.14 for AE-GAN \cite{ketyko_2022_lstm_suffix} and 0.22 for AE \cite{ketyko_2022_lstm_suffix}. 
These results suggest that suffix prediction for the Sepsis dataset is challenging, likely due to high variability and many unique cases that do not follow a general pattern.

For the BPIC-17 dataset, the probabilistic prediction achieved the best results in terms of suffix length MAE and suffix event label DLS in Setting 2, reaching a DLS of 0.31 and outperforming the most likely prediction, which had a DLS of 0.21. However, in Setting 1, the most likely prediction yielded the highest DLS overall, with a value of 0.35. However, the probabilistic prediction results of Setting 2 are comparable to those reported in the literature, ranging from a DLS of 0.07 for AE and 0.14 for AE-GAN \cite{ketyko_2022_lstm_suffix}, to 0.30 for MM-Pred \cite{lin_2019_ppm}, 0.34 for ED-LSTM-GAN \cite{taymouri_2021_deep_adv_model}, 0.32 for the re-implemented ED-LSTM, and 0.38 for SuTraN \cite{wuyts_2024_sutran}.
This improvement can be attributed to the potentially long suffixes present in BPIC-17 cases. The probabilistic prediction performs better at estimating suffix lengths with greater length and variability, as illustrated in Figure \ref{fig:performance_setting2-HSP}. This result highlights the advantage of using a probabilistic approach for predicting long suffixes, where uncertainty tend to increase with each additional, possibly incorrect, predicted event.

For the PCR dataset, the Setting 1 yielded the best performance of the U-ED-LSTM in terms of both suffix length MAE and suffix event labels DLS. Moreover, across all hyperparameter settings, the most likely prediction consistently outperformed the probabilistic prediction in both suffix length and suffix event labels prediction for the PCR and the artificial Repair datasets.
The PCR dataset represents a highly automated process managed by a workflow engine, resulting in less variability than the other datasets. One possible reason why the 2-layer setting outperformed the 4-layer could be the relatively small size and low complexity of the PCR dataset.

On the Repair dataset, Setting 3 yielded the best results in terms of suffix length MAE and suffix event labels DLS.

Since the most likely suffix was obtained from auto-regressively sampling the mode event label with the highest softmax probability, it can be expected to have the best DLS. 
Interestingly, as shown in Figure \ref{fig:performance_setting2-HSB}, this observation does not hold for shorter prefix lengths in the Sepsis and BPIC-17 datasets under Setting 2. In particular, for small prefixes and longer suffixes, the suffix length MAE of the probabilistic prediction outperforms that of the most likely prediction in the Helpdesk, Sepsis, and Repair datasets, as shown in Figure \ref{fig:performance_setting2-HSB} and Figure \ref{fig:performance_setting2-PR}.

\input{tables/results_categorical}
\input{tables/results_continuous}
%

%
For evaluating the predictive performance on continuous event attributes, specifically remaining time \textit{sum} and remaining time \textit{last}, the results indicate that the remaining time sum provides a more accurate prediction. This is evident from the lower MAE values and its broader use in related literature. Therefore, the remaining time sum MAE results can be meaningfully compared to those reported by other approaches.
In most cases, the probabilistic prediction yields a lower remaining time sum MAE than the most likely prediction across all hyperparameter settings.

Focusing solely on the remaining time sum predictive performance, it can be observed that, for the Helpdesk dataset, Setting 1 yielded the best performance. The MAE was 11.21 for the most likely prediction and 8.74 for the probabilistic prediction, although the results are comparable to those of Setting 2. However, previously reported results in the literature are slightly better, with MAE of 6.21 for ED-LSTM-GAN \cite{taymouri_2021_deep_adv_model}, 3.88 for AE-GAN, and 3.83 for AE \cite{ketyko_2022_lstm_suffix}.

For the Sepsis dataset, the best result was achieved by the probabilistic prediction in Setting 2, with a remaining time \textit{sum} MAE of 31.18, outperforming the most likely prediction, which yielded an MAE of 34.50. This result is also significantly better than those reported in the literature, such as 735.04 for AE and 187.12 for AE-GAN \cite{ketyko_2022_lstm_suffix}. 
For the BPIC-17 dataset, Setting 2 also yielded the best remaining time \textit{sum} MAE, with 10.62 for the probabilistic prediction and 10.73 for the most likely prediction. These results are comparable to those reported in the literature, which range from 100.19 for AE-GAN \cite{ketyko_2022_lstm_suffix} to 5.5 for SuTraN \cite{wuyts_2024_sutran}. Notably, SuTraN directly predicts the remaining time using an additional fully connected output layer, rather than computing it by summing the predicted event elapsed times of the predicted events.

One possible reason why the probabilistic remaining time \textit{sum} prediction is more accurate than the most likely prediction is that the remaining time derived from the most likely suffix corresponds to the mode of the distribution of remaining times across all sampled suffixes. In contrast, the mean of these sampled remaining times is often closer to the true remaining time, especially in skewed distributions with high variance and therefore high uncertainty.
Interestingly, the assumed log-normal distribution returned significantly worse results than the normal distribution for all datasets except for the remaining time last MAE in the Sepsis dataset and the remaining time sum MAE in the Repair dateset. Learned log-normal distributed loss attenuation are sensitive to outlier predictions, which greatly influence the aggregated mean of all sampled suffixes as it can be seen in Figure \ref{fig:performance_setting2-HSB}, and Figure \ref{fig:performance_setting2-PR}. Consequently, after reversing the standardization and exponentiating, the remaining times of outliers can become extremely large when sampled in log space.
It is not surprising that the U-ED-LSTM in Setting 3 achieved the best remaining time sum MAE for the Repair dataset. Since the timestamps in this artificial dataset are sampled from a log-normal distribution, the model was able to learn accurately.
However, since the log-normal loss attenuation shows promising properties, such as ensuring non-negativity, the main challenge lies in extreme outlier sampled form the log-normal distribution, de-standardized and exponentiated. Therefore, in future work, we aim to clip the variance values during training and sampling once they exceed a certain threshold, ensuring that the exponentiated time values remain within a feasible range.

Overall, Setting 2 proved to be the most effective hyperparameter configuration. As previously mentioned, the detailed results for Setting 2 are visualized in Figure \ref{fig:performance_setting2-HSB} for the Helpdesk, Sepsis, and BPIC-17 datasets, and in Figure \ref{fig:performance_setting2-PR} for the PCR and Repair datasets.
The probabilistic prediction often outperforms the most likely prediction, particularly for short prefixes and long suffixes. Overall, the predictive performance results are reasonable but can be improved in future work. Nevertheless, we argue that probabilistic suffix prediction is effective and can therefore be used for more advanced applications such as confidence interval estimation.

\begin{figure}[H]
    \centering
    \begin{subfigure}{0.49\textwidth}
        \centering
        \resizebox{\linewidth}{!}{\input{figures/evaluation/Helpdesk/Helpdesk_norm_4layer.pgf}}
        \caption{Helpdesk}
    \end{subfigure}
    \begin{subfigure}{0.49\textwidth}
        \centering
        \resizebox{\linewidth}{!}{\input{figures/evaluation/Sepsis/Sepsis_norm_4layer.pgf}}
        \caption{Sepsis}
    \end{subfigure}
    \begin{subfigure}{0.49\textwidth}
        \centering
        \resizebox{\linewidth}{!}{\input{figures/evaluation/BPIC17/BPIC17_norm_4layer.pgf}}
        \caption{BPIC17}
    \end{subfigure}
    \caption{Predictive Performance - Setting 2 for the Datasets: Helpdesk, Sepsis, and BPIC-17}
    \label{fig:performance_setting2-HSB}
\end{figure}
\begin{figure}[H]
    \centering
    \begin{subfigure}{0.49\textwidth}
        \centering
        \resizebox{\linewidth}{!}{\input{figures/evaluation/PCR/PCR_norm_4layer.pgf}}
        \caption{PCR}
    \end{subfigure}
    \begin{subfigure}{0.49\textwidth}
        \centering
        \resizebox{\linewidth}{!}{\input{figures/evaluation/Repair/Repair_norm_4layer.pgf}}
        \caption{Repair}
    \end{subfigure}
    \caption{Predictive Performance - Setting 2 for the Datasets: PCR and Repair}
    \label{fig:performance_setting2-PR}
\end{figure}

\subsection{Calibration Results}
We evaluated the calibration of the U-ED-LSTM to demonstrate that the model can capture the variability in the respective event logs.

\mypar{Metrics} 
To evaluate the calibration of remaining time predictions (sum and last), we used the Probability Integral Transform (PIT). Given the predicted distribution of remaining times generated by the U-ED-LSTM for each test case, obtained by sampling multiple suffixes, we constructed an empirical cumulative distribution function (CDF) for each prediction.

For a given test case $i$, let $\hat{t}_{i,j}$ be the remaining time prediction of the $j$-th sampled suffix, and let $t_i$ be the ground-truth remaining time. The normalized PIT value per test case  $u_i$ is computed as: 
\(u_i= \frac{1}{T} \sum_{j=1}^{T} \textbf{1}\{\hat{t}_{i,j} \leq t_i\}\) , where $T=1000$ is the total number of MC samples and $\textbf{1}\{\cdot\}$ is the indicator function. The calculation is the same for the remaining time sum and last.
After computing the set of PIT values $u := \{u_1, \dots u_{D_{test}}\}$, PIT plots were constructed. The x-axis represents the PIT values between $[0,1]$, and the y-axis shows the probability density. The PIT plots can have different shapes, each indicating different model calibration:

A uniform distribution (i.e., a flat line at density 1) indicates perfect calibration. The predicted variance matches the true variability in the ground truth exactly.
A U-shape suggests that the model predicts too little variance and tends to underestimate outliers, a phenomenon referred to as underdispersion. For instance, if the true variability of remaining times spans ±5 days around the mean, but the model only predicts a variance of ±3 days, it will consistently fail to capture outliers.
Conversely, a bell shape indicates that the model predicts too much variance, a phenomenon referred to as overdispersion. For instance, if the true variability of remaining times spans ±5 days around the mean, and the model predicts a much higher variance of ±10 days, it will consistently predict, over all samples per case, remaining times below and above the ground truth.
A slope-shaped PIT plot indicates a systematic bias in the model’s predictions. In such cases, no reliable conclusions about the predicted variance can be drawn.

\mypar{Calibration Results}
Figure \ref{fig:calibration_s2-HSB} presents the PIT plots for the Helpdesk, Sepsis, and BPIC-17 datasets, illustrating the calibration of the remaining time predictions. Similarly, Figure \ref{fig:calibration_s2-PR} shows the PIT plots for the PCR and Repair datasets.
The PIT plots demonstrate that our approach shows no systematic bias across all data sets for the remaining time predictions: 

While the PIT plot from the Helpdesk test set shows that the predicted distributions are overdispered, the Sepsis and the BPIC-17 plots show peaks at 1, indicating that the actual remaining time values are often higher than the largest estimations. Conversely, the PCR plots show a peak at 0 and the Repair plots show a peak at around 0.2, indicating that the actual remaining time values often fall in the lower tail of a predicted distributions.

\begin{figure}[H]
    \centering
    \begin{minipage}{0.24\textwidth}
        \centering
        \resizebox{\linewidth}{!}{\input{figures/evaluation/Helpdesk/Helpdesk_PIT_event_elapsed_norm_4layer.pgf}}
    \end{minipage}
    \begin{minipage}{0.24\textwidth}
        \centering
        \resizebox{\linewidth}{!}{ \input{figures/evaluation/Helpdesk/Helpdesk_PIT_remaining_time_norm_4layer.pgf}}
    \end{minipage}
    \begin{minipage}{0.24\textwidth}
        \centering
        \resizebox{\linewidth}{!}{\input{figures/evaluation/Sepsis/Sepsis_PIT_event_elapsed_norm_4layer.pgf}}
    \end{minipage}
    \begin{minipage}{0.24\textwidth}
        \centering
        \resizebox{\linewidth}{!}{ \input{figures/evaluation/Sepsis/Sepsis_PIT_remaining_time_norm_4layer.pgf}}
    \end{minipage}
    \begin{minipage}{0.24\textwidth}
        \centering
        \resizebox{\linewidth}{!}{\input{figures/evaluation/BPIC17/BPIC17_PIT_event_elapsed_norm_4layer.pgf}}
    \end{minipage}
    \begin{minipage}{0.24\textwidth}
        \centering
        \resizebox{\linewidth}{!}{\input{figures/evaluation/BPIC17/BPIC17_PIT_remaining_time_norm_4layer.pgf}}
    \end{minipage}
    \caption{Model Calibration of Remaining Time Predictions - Setting 2 for the Datasets: Helpdesk, Sepsis, And BPIC-17}
    \label{fig:calibration_s2-HSB}
\end{figure}

\begin{figure}[H]
    \centering
    \begin{minipage}{0.24\textwidth}
        \centering
        \resizebox{\linewidth}{!}{\input{figures/evaluation/PCR/PCR_PIT_event_elapsed_norm_4layer.pgf}}
    \end{minipage}
    \begin{minipage}{0.24\textwidth}
        \centering
        \resizebox{\linewidth}{!}{ \input{figures/evaluation/PCR/PCR_PIT_remaining_time_norm_4layer.pgf}}
    \end{minipage}
    \begin{minipage}{0.24\textwidth}
        \centering
        \resizebox{\linewidth}{!}{\input{figures/evaluation/Repair/Repair_PIT_event_elapsed_norm_4layer.pgf}}
    \end{minipage}
    \begin{minipage}{0.24\textwidth}
        \centering
        \resizebox{\linewidth}{!}{\input{figures/evaluation/Repair/Repair_PIT_remaining_time_norm_4layer.pgf}}
    \end{minipage}
    \caption{Model Calibration of Remaining Time Predictions - Setting 2 for the Datasets: PCR and Repair}
    \label{fig:calibration_s2-PR}
\end{figure}

\section{Related Work}\label{sec:relwork}
Related works covering suffix prediction and uncertainty in PPM are presented.

\mypar{Suffix Prediction}
Current suffix prediction approaches focus on predicting the most likely suffix and improving predictive performance. The methods differ in the models used, predicted event attributes, and strategies to enhance training.
Early suffix prediction approaches use LSTMs \cite{evermann_2017_lstm_suffix,tax_2017_ppm_lstm,camargo_2019_lstm_bpm,gunnarsson_2023_}.
Predictive performance is improved by using encoder-decoder LSTMs \cite{lin_2019_ppm,taymouri_2021_deep_adv_model}. 
More recent encoder-decoders are enriched by more complex NN architectures such as combined General Recurrent Units, Graph NNs, and attention \cite{rama_maneiro_2024_exp_rnn} or transformers \cite{wuyts_2024_sutran}.
Recently, LLMs have been used for suffix prediction \cite{Pasquadibisceglie_2024_lupin}, facing challenges such as lack of interpretability or not all prefixes can simultaneously be passed into a prompt.
In addition, existing approaches can be categorized based on predicted event attributes. Some approaches predict only the sequence of activities \cite{evermann_2017_lstm_suffix,rama_maneiro_2024_exp_rnn,Pasquadibisceglie_2024_lupin} and lifecycle transitions \cite{lin_2019_ppm}. Other approaches predict the sequence of activities and time attributes \cite{tax_2017_ppm_lstm,taymouri_2021_deep_adv_model,wuyts_2024_sutran,gunnarsson_2023_}, and resource information \cite{camargo_2019_lstm_bpm}.
Special training considerations are applied to improve predictive performance.
\cite{taymouri_2021_deep_adv_model}, for example, introduce teacher forcing and enhance its encoder-decoder LSTM with adversarial training to improve performance and robustness.
\cite{gunnarsson_2023_,wuyts_2024_sutran} proposed CRTP, demonstrating that models trained this way outperform those optimized for single-event prediction.
For testing, \cite{camargo_2019_lstm_bpm} try random sampling from categorical distributions against an arg-max strategy to derive the best matching activities in a suffix. Similar to our approach, they observe better performance in suffix length prediction.

\mypar{Uncertainty in PPM}
For remaining time and next activity predictions, combined epistemic and aleatoric uncertainty for NNs is applied to PPM by \cite{weytjens_2022_unc_ppm}.
\cite{portolani_2022_unc_ppm} applies and compares deep ensemble and MC dropout in attention-based NNs for the next activity prediction.
Both approaches aim to improve single-event prediction performance and show how uncertainty and prediction accuracy correlate.
Most recently, \cite{mehdiyev_2025_post_hoc_unc_ppm} introduces Conformalized MC dropout, leveraging uncertainty and conformal predictions to construct prediction intervals for the next activity prediction to improve interpretability. However, they do not evaluate their approach on open-source, real-world datasets.
In \cite{pauwels_2020_bn_ppm,rauch_2024_pa_bayesian}, Bayesian Networks are used to predict the sequence of activities, but Bayesian networks cannot handle large and complex data. 

\section{Conclusion}\label{sec:con}
In this technical report, we presented an approach for \textit{probabilistic suffix prediction} that leverages our \textit{U-ED-LSTM} and \textit{MC suffix sampling} algorithm and an extensive performance evaluation of the proposed model.
Our approach captures epistemic uncertainty via MC dropout and aleatoric uncertainty as learned loss attenuation. No other work has yet addressed incorporating epistemic and aleatoric uncertainties for suffix predictions of business processes. Probabilistic suffix prediction can offer enhanced reliability and transparency by generating a distribution over possible future sequences rather than a single deterministic outcome. For instance, instead of predicting a single remaining time or a fixed number of activity loop executions, the model can provide a range of possible values and their associated probabilities.

\mypar{Future Work} We demonstrated the predictive performance and calibration of our U-ED-LSTM. However, to further improve the performance and calibration of our approach, we i) further experiment with different hyperparameters, ii) try to improve the log-normal distribution of assumed observation noise for loss attenuation to obtain results such as in \cite{salinas}, iii) try different methods to measure epistemic uncertainty such as deep ensembles instead of MC dropout, iv) choose different NN architectures for sequence predictions, especially for long-range sequences such as transformers. 
In the evaluation, we only assessed activity sequence and remaining time predictions. Since our approach can predict all event attributes, evaluating additional attributes could yield further insights into the model's predictive performance.


\end{document}

%% file: algorithm/03a_algorithm.tex
\begin{algorithm}
\scriptsize
\caption{MC Suffix Sampling}
\label{alg:sampling}
\begin{algorithmic}[1]
    
    \Require $T \in \mathbb{N}$: number of MC samples, $M \in \mathbb{N}$: max. suffix length to be sampled, p $\in [0,1]$: dropout probability, $p_{\leq k} = \langle e_1, e_2, ..., e_k \rangle$: prefix
    
    
    \Function{MCSuffixSampling}{T, M, p, $p_{\leq k}$}
        
        \State $\tilde{S} \leftarrow \emptyset$  \Comment{Set of Sampled Suffixes.}
    
        \For{$t = 1$ to $T$}
            \State $\tilde{s}_{>k} \leftarrow \langle \rangle$                            \Comment{Sampled Suffix}
            \State $\hat{W}_{enc} \leftarrow \text{VariationalDropout}(W_{enc}, \text{p})$
            \State $(h_{enc}, c_{enc}) \leftarrow f^{\hat{W}_{enc}}_{enc}(p_{\leq k})$  
            \State $\tilde{e} \leftarrow e_k$
            \State $i \leftarrow 0$
            
            \Repeat
                \State $\hat{W}_{dec} \leftarrow \text{NaiveDropout}(W_{dec}, \text{p})$
                \State $\hat{e}_{k+1}, (h,c) \leftarrow f^{\hat{W}_{dec}}_{dec}(\tilde{e}, (h_{enc}, c_{enc}))$
                \State $\hat{a}_{con}, \hat{a}_{cat} \leftarrow \hat{e}_{k+1}$

                \For{$j = 1$ to $ \vert D \vert $}
                    \State $\hat{y}, \hat{v}_{con} \leftarrow \hat{a}_{con}^{(j)}$
                    \State $\tilde{a}_{con}^{(j)} \sim \mathcal{N}(\hat{y}, exp(\hat{v}_{con}))$
                \EndFor
                \For{$j = 1$ to $ \vert K \vert $}
                    \State $\hat{l}, \hat{v}_{cat} \leftarrow \hat{a}_{cat}^{(j)}$
                    \State $\tilde{a}_{cat}^{(j)} \sim \text{Categorical}(\text{Softmax}(\mathcal{N}(\hat{l}, exp(\hat{v}_{cat}))))$
                \EndFor
                
                \State $\tilde{e} \leftarrow (\tilde{a}_{con}, \tilde{a}_{cat})$
    
            \State $\tilde{s}_{>k} \leftarrow \tilde{s}_{>k} \circ \tilde{e}$
            \State $i \leftarrow i + 1$
        \Until{$i = M$ or GetActivity($\tilde{a}_{cat}$) = $\text{'EOS'}$}
        
        \State $\tilde{S} \leftarrow \tilde{S} \cup \{ \tilde{s}_{>k} \}$
        \EndFor
        
        \State \Return $\tilde{S}$
    \EndFunction
\end{algorithmic}
\end{algorithm}

%% file: tables/data.tex
\begin{table}[ht]
\caption{Dataset Properties}
\label{tab:datasets}
\centering
\sisetup{group-separator={\,}, input-symbols = ()}
\scriptsize
\renewcommand{\arraystretch}{0.85}
\resizebox{\textwidth}{!}{%
\begin{tabular}{
    l|
    S[table-format=5.0]|
    S[table-format=7.0]|
    S[table-format=5.0]|
    S[table-format=2.0]|
    c|
    c|
    S[table-format=2.0]|
    S[table-format=2.0]
}
\toprule
\textbf{Dataset} & \textbf{Cases} & \textbf{Events} & \textbf{Variants} & \textbf{Activities} & \textbf{Mean–SD Case Length} & \textbf{Mean–SD Case Duration} & \textbf{Cat. Event Attr.} & \textbf{Con. Event Attr.} \\
\midrule
Helpdesk & \num{4580} & \num{21348} & \num{226} & \num{14} & \num{4.66} -- \num{1.18} & \num{40.86} -- \num{8.39} (days) & \num{12} & \num{4} \\
Sepsis & \num{1049} & \num{15214} & \num{845} & \num{16} & \num{14.48} -- \num{11.47} & \num{28.48} -- \num{60.54} (days) & \num{26} & \num{8} \\
BPIC17 & \num{31509} & \num{1202267} & \num{15930} & \num{26} & \num{38.16} -- \num{16.72} & \num{21.90} -- \num{13.17} (days) & \num{9} & \num{9} \\
PCR & \num{6166} & \num{117703} & \num{1213} & \num{8} & \num{19.09} -- \num{3.37} & \num{19872} -- \num{27864} (sec.) & \num{2} & \num{4} \\
Repair & \num{9896} & \num{79874} & \num{16} & \num{7} & \num{8.07} -- \num{2.14} & \num{2.27} -- \num{1.42} (days) & \num{1} & \num{4} \\

\bottomrule
\end{tabular}%
}
\end{table}

%% file: tables/hyperparameter.tex
\begin{table}[ht]
\caption{Hyperparameter Settings}
\label{tab:hyperparameters}
\centering
\sisetup{group-separator={\,}, input-symbols = ()}
\scriptsize
\renewcommand{\arraystretch}{0.75}
\resizebox{\textwidth}{!}{%
\begin{tabular}{
    l|
    c|
    c|
    c
}
\toprule
\textbf{Hyperparameter} & \textbf{Setting 1} & \textbf{Setting 2} & \textbf{Setting 3} \\
\midrule
U-ED-LSTM layers                                     & 2 & 4 & 4 \\
FC layers in decoder                                 & 1 & 1 & 1 \\
Assumed distribution for con. event attr.            & Normal & Normal & Log-Normal \\
Encoder features                                     & All & All & All \\
Decoder features                                     & All & Activity \& Time & Activity \& Time \\
Hidden size                                          & 128 & 128 & 128 \\
Optimizer                                            & Adam & AdamW & AdamW \\
Learning rate                                        & \num{1e-4} -- \num{1e-6} & \num{1e-4} -- \num{1e-6} & \num{1e-4} -- \num{1e-6} \\
Batch size                                           & \num{128} (BPIC17 256) & \num{128} (BPIC17 256) & \num{128} (BPIC17 256)\\
Epochs                                               & \num{200} & \num{100} & \num{100} \\
MC Dropout Probability (Train/ Test)                 & \num{0.1} & \num{0.1} & \num{0.1} \\
Weight-Decay/ Regularization                         & \num{1e-4} & \num{1e-4} & \num{1e-4} \\
\bottomrule
\end{tabular}%
}
\end{table}

%% file: tables/results_categorical.tex
\begin{table}[ht]
\centering
\caption{Predictive Performance (Categorical): Suffix Length MAE and Suffix Event Labels DLS}
\label{tab:performance_suffix_dls}
\scriptsize
\setlength{\tabcolsep}{3pt}
\renewcommand{\arraystretch}{0.9}
\begin{tabular}{l | c c c c c | c c c c c}
\toprule
\textbf{Method} &
\multicolumn{5}{c|}{\textbf{Suffix Length MAE $\downarrow$}} &
\multicolumn{5}{c}{\textbf{Suffix Event Labels DLS $\uparrow$}} \\
\cmidrule(lr){2-6} \cmidrule(lr){7-11}
& Helpdesk & Sepsis & BPIC17 & PCR & Repair
& Helpdesk & Sepsis & BPIC17 & PCR & Repair\\
\midrule

\midrule
\textbf{Own Results}
&  &  &  &  &
&  &  &  &  &\\
\midrule

Most likely - Setting 1
& 0.96 & 27.59 & 13.74 & \textbf{1.48} & 1.05
& 0.53 & 0.1 & \textbf{0.35} & \textbf{0.83} & 0.86\\

Probabilistic - Setting 1 
& 0.74 & 6.84 & 14.29 & 1.98 & 1.41
& 0.44 & 0.14 & 0.28 & 0.59 & 0.73\\

\midrule

Most likely - Setting 2
& \textbf{0.36} & 8.8 & 40.83  & 3.49 & 1.08
& \textbf{0.82} & 0.11 & 0.21 & 0.67 & 0.85\\

Probabilistic - Setting 2 
& 0.38 & 6.83 & \textbf{11.42} & 3.63 & 1.43
& 0.65 & 0.12 & 0.31 & 0.54 & 0.74\\

\midrule

Most likely - Setting 3
& 0.54 & 26.96 & 40.78 & 4.37 & \textbf{0.94}
& \textbf{0.82} & 0.09 & 0.16 & 0.62 & \textbf{0.87}\\

Probabilistic - Setting 3
& 0.53 & \textbf{6.16} & 33.71 & 4.43 & 1.24
& 0.52 & \textbf{0.18} & 0.2  & 0.54  & 0.75\\

\midrule
\textbf{ED-LSTM from Lit.}
&  &  &  & & 
&  &  &  & & \\
\midrule

MM-Pred  \cite{lin_2019_ppm}
& - & - & - & - & -
& 0.87 & - & 0.3 & - & -\\

\midrule

ED-LSTM-GAN \cite{taymouri_2021_deep_adv_model}
& - & - & - & - & -
& 0.84 & - & 0.34 & - & -\\

\midrule

AE \cite{ketyko_2022_lstm_suffix}
& - & - & - & - & -
& 0.86 & 0.22 & 0.14 & - & -\\

AE-GAN \cite{ketyko_2022_lstm_suffix}
& - & - & - & - & -
& 0.86 &  0.14 & 0.07 & - & -\\

\midrule

ED-LSTM \cite{wuyts_2024_sutran}
& - & - & - & - & -
& - & - & 0.32 & - & -\\

\midrule
\textbf{Transformer from Lit.}
&  &  &  &  &
&  &  &  &  &\\
\midrule

SuTraN \cite{wuyts_2024_sutran}
& - & - & - & - & -
& - & - & 0.38 & - & -\\

\bottomrule
\end{tabular}
\end{table}

%% file: tables/results_continuous.tex
\begin{table}[ht]
\centering
\caption{Predictive Performance (Continuous): Remaining Time MAE}
\label{tab:performance_remaining_time}
\scriptsize
\setlength{\tabcolsep}{3pt}
\renewcommand{\arraystretch}{0.9}
\begin{tabular}{l | c c c c c | c c c c c}
\toprule
\textbf{Method} &
\multicolumn{5}{c|}{\textbf{Remaining Time (sum) days MAE $\downarrow$}} &
\multicolumn{5}{c}{\textbf{Remaining Time (last) days MAE $\downarrow$}} \\
\cmidrule(lr){2-6} \cmidrule(lr){7-11}
& Helpdesk & Sepsis & BPIC17 & PCR (sec.) & Repair
& Helpdesk & Sepsis & BPIC17 & PCR (sec.) & Repair\\
\midrule

\midrule
\textbf{Own Results}
&  &  &  &  
&  & & &  \\
\midrule

Most likely Setting 1
& 11.21 & 38.09 & 11.76 & \textbf{159.19} & 0.78
& 18.35 & 29.21 & \textbf{9.95} & 9340.1 & \textbf{0.68}\\

Probabilistic Setting 1
& \textbf{8.74} & 31.21 & 12.45 & 165.91 & 0.9
& 14.02 & 34.3 & 10.83 & 19237.2 & 0.86 \\

\midrule

Most likely Setting 2
& 11.76 & 34.5 & 10.73 & 170.87 & 0.73
& \textbf{9.1} & 29.88 & 10.75 & \textbf{8871.74} & 0.7  \\

Probabilistic Setting 2
& 9.58 & \textbf{31.18} & \textbf{10.62}  & 170.44 & 0.78
& 10.99 & 31.41 & 14.4  & 12281.35 & 0.9 \\

\midrule

Most likely Setting 3
& 261.64 & 147.15 & 10.68 & 180.18 & 0.57  
& 553.82 & \textbf{24.54} & 2887.58 & 411704.04 & 315.58 \\

Probabilistic Setting 3
& 5789.51 & 135.05 & 1387.15 & 15262002.78 & \textbf{0.46}
& 557.32 & 24.87 & 4448.17 & 412893.69 & 315.71 \\

\midrule
\textbf{ED-LSTM from Lit.}
&  &  &  &  &
&  &  &  &  & \\
\midrule

MM-Pred  \cite{lin_2019_ppm}
& - & - & - & - & -
& - & - & - & - & - \\

\midrule

ED-LSTM-GAN \cite{taymouri_2021_deep_adv_model}
& 6.21 & - & 13.95 & - & -
& - & - & - & - & - \\

\midrule

AE \cite{ketyko_2022_lstm_suffix}
& 3.83 & 735.04 & 69.51 & - & - 
& - & - & - & - & - \\

AE-GAN \cite{ketyko_2022_lstm_suffix}
& 3.88 & 187.12 & 100.19 & - & -
& - &  - & - & - & - \\

\midrule

ED-LSTM \cite{wuyts_2024_sutran}
& - & - & 8.44 & - & - 
& - & - & - & - & - \\

\midrule
\textbf{Transformer from Lit.}
&  &  &  &  & 
&  &  &  &  & \\
\midrule

SuTraN \cite{wuyts_2024_sutran}
& - & - & 5.5 & - & -
& - & - & - & - & - \\

\bottomrule
\end{tabular}
\end{table}

%% file: figures/evaluation/Helpdesk/Helpdesk_PIT_event_elapsed_norm_4layer.pgf
\begingroup%
\makeatletter%
\begin{pgfpicture}%
\pgfpathrectangle{\pgfpointorigin}{\pgfqpoint{6.167556in}{4.557174in}}%
\pgfusepath{use as bounding box, clip}%
\begin{pgfscope}%
\pgfsetbuttcap%
\pgfsetmiterjoin%
\definecolor{currentfill}{rgb}{1.000000,1.000000,1.000000}%
\pgfsetfillcolor{currentfill}%
\pgfsetlinewidth{0.000000pt}%
\definecolor{currentstroke}{rgb}{1.000000,1.000000,1.000000}%
\pgfsetstrokecolor{currentstroke}%
\pgfsetdash{}{0pt}%
\pgfpathmoveto{\pgfqpoint{0.000000in}{0.000000in}}%
\pgfpathlineto{\pgfqpoint{6.167556in}{0.000000in}}%
\pgfpathlineto{\pgfqpoint{6.167556in}{4.557174in}}%
\pgfpathlineto{\pgfqpoint{0.000000in}{4.557174in}}%
\pgfpathlineto{\pgfqpoint{0.000000in}{0.000000in}}%
\pgfpathclose%
\pgfusepath{fill}%
\end{pgfscope}%
\begin{pgfscope}%
\pgfsetbuttcap%
\pgfsetmiterjoin%
\definecolor{currentfill}{rgb}{1.000000,1.000000,1.000000}%
\pgfsetfillcolor{currentfill}%
\pgfsetlinewidth{0.000000pt}%
\definecolor{currentstroke}{rgb}{0.000000,0.000000,0.000000}%
\pgfsetstrokecolor{currentstroke}%
\pgfsetstrokeopacity{0.000000}%
\pgfsetdash{}{0pt}%
\pgfpathmoveto{\pgfqpoint{0.802523in}{0.716028in}}%
\pgfpathlineto{\pgfqpoint{5.996275in}{0.716028in}}%
\pgfpathlineto{\pgfqpoint{5.996275in}{4.233871in}}%
\pgfpathlineto{\pgfqpoint{0.802523in}{4.233871in}}%
\pgfpathlineto{\pgfqpoint{0.802523in}{0.716028in}}%
\pgfpathclose%
\pgfusepath{fill}%
\end{pgfscope}%
\begin{pgfscope}%
\pgfpathrectangle{\pgfqpoint{0.802523in}{0.716028in}}{\pgfqpoint{5.193752in}{3.517843in}}%
\pgfusepath{clip}%
\pgfsetbuttcap%
\pgfsetmiterjoin%
\definecolor{currentfill}{rgb}{0.000000,0.000000,1.000000}%
\pgfsetfillcolor{currentfill}%
\pgfsetlinewidth{0.000000pt}%
\definecolor{currentstroke}{rgb}{0.000000,0.000000,0.000000}%
\pgfsetstrokecolor{currentstroke}%
\pgfsetdash{}{0pt}%
\pgfpathmoveto{\pgfqpoint{0.802523in}{0.716028in}}%
\pgfpathlineto{\pgfqpoint{0.981618in}{0.716028in}}%
\pgfpathlineto{\pgfqpoint{0.981618in}{0.716028in}}%
\pgfpathlineto{\pgfqpoint{0.802523in}{0.716028in}}%
\pgfpathlineto{\pgfqpoint{0.802523in}{0.716028in}}%
\pgfpathclose%
\pgfusepath{fill}%
\end{pgfscope}%
\begin{pgfscope}%
\pgfpathrectangle{\pgfqpoint{0.802523in}{0.716028in}}{\pgfqpoint{5.193752in}{3.517843in}}%
\pgfusepath{clip}%
\pgfsetbuttcap%
\pgfsetmiterjoin%
\definecolor{currentfill}{rgb}{0.000000,0.000000,1.000000}%
\pgfsetfillcolor{currentfill}%
\pgfsetlinewidth{0.000000pt}%
\definecolor{currentstroke}{rgb}{0.000000,0.000000,0.000000}%
\pgfsetstrokecolor{currentstroke}%
\pgfsetdash{}{0pt}%
\pgfpathmoveto{\pgfqpoint{0.981618in}{0.716028in}}%
\pgfpathlineto{\pgfqpoint{1.160713in}{0.716028in}}%
\pgfpathlineto{\pgfqpoint{1.160713in}{0.803050in}}%
\pgfpathlineto{\pgfqpoint{0.981618in}{0.803050in}}%
\pgfpathlineto{\pgfqpoint{0.981618in}{0.716028in}}%
\pgfpathclose%
\pgfusepath{fill}%
\end{pgfscope}%
\begin{pgfscope}%
\pgfpathrectangle{\pgfqpoint{0.802523in}{0.716028in}}{\pgfqpoint{5.193752in}{3.517843in}}%
\pgfusepath{clip}%
\pgfsetbuttcap%
\pgfsetmiterjoin%
\definecolor{currentfill}{rgb}{0.000000,0.000000,1.000000}%
\pgfsetfillcolor{currentfill}%
\pgfsetlinewidth{0.000000pt}%
\definecolor{currentstroke}{rgb}{0.000000,0.000000,0.000000}%
\pgfsetstrokecolor{currentstroke}%
\pgfsetdash{}{0pt}%
\pgfpathmoveto{\pgfqpoint{1.160713in}{0.716028in}}%
\pgfpathlineto{\pgfqpoint{1.339808in}{0.716028in}}%
\pgfpathlineto{\pgfqpoint{1.339808in}{0.716028in}}%
\pgfpathlineto{\pgfqpoint{1.160713in}{0.716028in}}%
\pgfpathlineto{\pgfqpoint{1.160713in}{0.716028in}}%
\pgfpathclose%
\pgfusepath{fill}%
\end{pgfscope}%
\begin{pgfscope}%
\pgfpathrectangle{\pgfqpoint{0.802523in}{0.716028in}}{\pgfqpoint{5.193752in}{3.517843in}}%
\pgfusepath{clip}%
\pgfsetbuttcap%
\pgfsetmiterjoin%
\definecolor{currentfill}{rgb}{0.000000,0.000000,1.000000}%
\pgfsetfillcolor{currentfill}%
\pgfsetlinewidth{0.000000pt}%
\definecolor{currentstroke}{rgb}{0.000000,0.000000,0.000000}%
\pgfsetstrokecolor{currentstroke}%
\pgfsetdash{}{0pt}%
\pgfpathmoveto{\pgfqpoint{1.339808in}{0.716028in}}%
\pgfpathlineto{\pgfqpoint{1.518902in}{0.716028in}}%
\pgfpathlineto{\pgfqpoint{1.518902in}{0.737784in}}%
\pgfpathlineto{\pgfqpoint{1.339808in}{0.737784in}}%
\pgfpathlineto{\pgfqpoint{1.339808in}{0.716028in}}%
\pgfpathclose%
\pgfusepath{fill}%
\end{pgfscope}%
\begin{pgfscope}%
\pgfpathrectangle{\pgfqpoint{0.802523in}{0.716028in}}{\pgfqpoint{5.193752in}{3.517843in}}%
\pgfusepath{clip}%
\pgfsetbuttcap%
\pgfsetmiterjoin%
\definecolor{currentfill}{rgb}{0.000000,0.000000,1.000000}%
\pgfsetfillcolor{currentfill}%
\pgfsetlinewidth{0.000000pt}%
\definecolor{currentstroke}{rgb}{0.000000,0.000000,0.000000}%
\pgfsetstrokecolor{currentstroke}%
\pgfsetdash{}{0pt}%
\pgfpathmoveto{\pgfqpoint{1.518902in}{0.716028in}}%
\pgfpathlineto{\pgfqpoint{1.697997in}{0.716028in}}%
\pgfpathlineto{\pgfqpoint{1.697997in}{0.759539in}}%
\pgfpathlineto{\pgfqpoint{1.518902in}{0.759539in}}%
\pgfpathlineto{\pgfqpoint{1.518902in}{0.716028in}}%
\pgfpathclose%
\pgfusepath{fill}%
\end{pgfscope}%
\begin{pgfscope}%
\pgfpathrectangle{\pgfqpoint{0.802523in}{0.716028in}}{\pgfqpoint{5.193752in}{3.517843in}}%
\pgfusepath{clip}%
\pgfsetbuttcap%
\pgfsetmiterjoin%
\definecolor{currentfill}{rgb}{0.000000,0.000000,1.000000}%
\pgfsetfillcolor{currentfill}%
\pgfsetlinewidth{0.000000pt}%
\definecolor{currentstroke}{rgb}{0.000000,0.000000,0.000000}%
\pgfsetstrokecolor{currentstroke}%
\pgfsetdash{}{0pt}%
\pgfpathmoveto{\pgfqpoint{1.697997in}{0.716028in}}%
\pgfpathlineto{\pgfqpoint{1.877092in}{0.716028in}}%
\pgfpathlineto{\pgfqpoint{1.877092in}{0.803050in}}%
\pgfpathlineto{\pgfqpoint{1.697997in}{0.803050in}}%
\pgfpathlineto{\pgfqpoint{1.697997in}{0.716028in}}%
\pgfpathclose%
\pgfusepath{fill}%
\end{pgfscope}%
\begin{pgfscope}%
\pgfpathrectangle{\pgfqpoint{0.802523in}{0.716028in}}{\pgfqpoint{5.193752in}{3.517843in}}%
\pgfusepath{clip}%
\pgfsetbuttcap%
\pgfsetmiterjoin%
\definecolor{currentfill}{rgb}{0.000000,0.000000,1.000000}%
\pgfsetfillcolor{currentfill}%
\pgfsetlinewidth{0.000000pt}%
\definecolor{currentstroke}{rgb}{0.000000,0.000000,0.000000}%
\pgfsetstrokecolor{currentstroke}%
\pgfsetdash{}{0pt}%
\pgfpathmoveto{\pgfqpoint{1.877092in}{0.716028in}}%
\pgfpathlineto{\pgfqpoint{2.056187in}{0.716028in}}%
\pgfpathlineto{\pgfqpoint{2.056187in}{1.412200in}}%
\pgfpathlineto{\pgfqpoint{1.877092in}{1.412200in}}%
\pgfpathlineto{\pgfqpoint{1.877092in}{0.716028in}}%
\pgfpathclose%
\pgfusepath{fill}%
\end{pgfscope}%
\begin{pgfscope}%
\pgfpathrectangle{\pgfqpoint{0.802523in}{0.716028in}}{\pgfqpoint{5.193752in}{3.517843in}}%
\pgfusepath{clip}%
\pgfsetbuttcap%
\pgfsetmiterjoin%
\definecolor{currentfill}{rgb}{0.000000,0.000000,1.000000}%
\pgfsetfillcolor{currentfill}%
\pgfsetlinewidth{0.000000pt}%
\definecolor{currentstroke}{rgb}{0.000000,0.000000,0.000000}%
\pgfsetstrokecolor{currentstroke}%
\pgfsetdash{}{0pt}%
\pgfpathmoveto{\pgfqpoint{2.056187in}{0.716028in}}%
\pgfpathlineto{\pgfqpoint{2.235282in}{0.716028in}}%
\pgfpathlineto{\pgfqpoint{2.235282in}{1.542732in}}%
\pgfpathlineto{\pgfqpoint{2.056187in}{1.542732in}}%
\pgfpathlineto{\pgfqpoint{2.056187in}{0.716028in}}%
\pgfpathclose%
\pgfusepath{fill}%
\end{pgfscope}%
\begin{pgfscope}%
\pgfpathrectangle{\pgfqpoint{0.802523in}{0.716028in}}{\pgfqpoint{5.193752in}{3.517843in}}%
\pgfusepath{clip}%
\pgfsetbuttcap%
\pgfsetmiterjoin%
\definecolor{currentfill}{rgb}{0.000000,0.000000,1.000000}%
\pgfsetfillcolor{currentfill}%
\pgfsetlinewidth{0.000000pt}%
\definecolor{currentstroke}{rgb}{0.000000,0.000000,0.000000}%
\pgfsetstrokecolor{currentstroke}%
\pgfsetdash{}{0pt}%
\pgfpathmoveto{\pgfqpoint{2.235282in}{0.716028in}}%
\pgfpathlineto{\pgfqpoint{2.414377in}{0.716028in}}%
\pgfpathlineto{\pgfqpoint{2.414377in}{1.716775in}}%
\pgfpathlineto{\pgfqpoint{2.235282in}{1.716775in}}%
\pgfpathlineto{\pgfqpoint{2.235282in}{0.716028in}}%
\pgfpathclose%
\pgfusepath{fill}%
\end{pgfscope}%
\begin{pgfscope}%
\pgfpathrectangle{\pgfqpoint{0.802523in}{0.716028in}}{\pgfqpoint{5.193752in}{3.517843in}}%
\pgfusepath{clip}%
\pgfsetbuttcap%
\pgfsetmiterjoin%
\definecolor{currentfill}{rgb}{0.000000,0.000000,1.000000}%
\pgfsetfillcolor{currentfill}%
\pgfsetlinewidth{0.000000pt}%
\definecolor{currentstroke}{rgb}{0.000000,0.000000,0.000000}%
\pgfsetstrokecolor{currentstroke}%
\pgfsetdash{}{0pt}%
\pgfpathmoveto{\pgfqpoint{2.414377in}{0.716028in}}%
\pgfpathlineto{\pgfqpoint{2.593472in}{0.716028in}}%
\pgfpathlineto{\pgfqpoint{2.593472in}{3.478960in}}%
\pgfpathlineto{\pgfqpoint{2.414377in}{3.478960in}}%
\pgfpathlineto{\pgfqpoint{2.414377in}{0.716028in}}%
\pgfpathclose%
\pgfusepath{fill}%
\end{pgfscope}%
\begin{pgfscope}%
\pgfpathrectangle{\pgfqpoint{0.802523in}{0.716028in}}{\pgfqpoint{5.193752in}{3.517843in}}%
\pgfusepath{clip}%
\pgfsetbuttcap%
\pgfsetmiterjoin%
\definecolor{currentfill}{rgb}{0.000000,0.000000,1.000000}%
\pgfsetfillcolor{currentfill}%
\pgfsetlinewidth{0.000000pt}%
\definecolor{currentstroke}{rgb}{0.000000,0.000000,0.000000}%
\pgfsetstrokecolor{currentstroke}%
\pgfsetdash{}{0pt}%
\pgfpathmoveto{\pgfqpoint{2.593472in}{0.716028in}}%
\pgfpathlineto{\pgfqpoint{2.772567in}{0.716028in}}%
\pgfpathlineto{\pgfqpoint{2.772567in}{2.260659in}}%
\pgfpathlineto{\pgfqpoint{2.593472in}{2.260659in}}%
\pgfpathlineto{\pgfqpoint{2.593472in}{0.716028in}}%
\pgfpathclose%
\pgfusepath{fill}%
\end{pgfscope}%
\begin{pgfscope}%
\pgfpathrectangle{\pgfqpoint{0.802523in}{0.716028in}}{\pgfqpoint{5.193752in}{3.517843in}}%
\pgfusepath{clip}%
\pgfsetbuttcap%
\pgfsetmiterjoin%
\definecolor{currentfill}{rgb}{0.000000,0.000000,1.000000}%
\pgfsetfillcolor{currentfill}%
\pgfsetlinewidth{0.000000pt}%
\definecolor{currentstroke}{rgb}{0.000000,0.000000,0.000000}%
\pgfsetstrokecolor{currentstroke}%
\pgfsetdash{}{0pt}%
\pgfpathmoveto{\pgfqpoint{2.772567in}{0.716028in}}%
\pgfpathlineto{\pgfqpoint{2.951662in}{0.716028in}}%
\pgfpathlineto{\pgfqpoint{2.951662in}{1.542732in}}%
\pgfpathlineto{\pgfqpoint{2.772567in}{1.542732in}}%
\pgfpathlineto{\pgfqpoint{2.772567in}{0.716028in}}%
\pgfpathclose%
\pgfusepath{fill}%
\end{pgfscope}%
\begin{pgfscope}%
\pgfpathrectangle{\pgfqpoint{0.802523in}{0.716028in}}{\pgfqpoint{5.193752in}{3.517843in}}%
\pgfusepath{clip}%
\pgfsetbuttcap%
\pgfsetmiterjoin%
\definecolor{currentfill}{rgb}{0.000000,0.000000,1.000000}%
\pgfsetfillcolor{currentfill}%
\pgfsetlinewidth{0.000000pt}%
\definecolor{currentstroke}{rgb}{0.000000,0.000000,0.000000}%
\pgfsetstrokecolor{currentstroke}%
\pgfsetdash{}{0pt}%
\pgfpathmoveto{\pgfqpoint{2.951662in}{0.716028in}}%
\pgfpathlineto{\pgfqpoint{3.130756in}{0.716028in}}%
\pgfpathlineto{\pgfqpoint{3.130756in}{2.804543in}}%
\pgfpathlineto{\pgfqpoint{2.951662in}{2.804543in}}%
\pgfpathlineto{\pgfqpoint{2.951662in}{0.716028in}}%
\pgfpathclose%
\pgfusepath{fill}%
\end{pgfscope}%
\begin{pgfscope}%
\pgfpathrectangle{\pgfqpoint{0.802523in}{0.716028in}}{\pgfqpoint{5.193752in}{3.517843in}}%
\pgfusepath{clip}%
\pgfsetbuttcap%
\pgfsetmiterjoin%
\definecolor{currentfill}{rgb}{0.000000,0.000000,1.000000}%
\pgfsetfillcolor{currentfill}%
\pgfsetlinewidth{0.000000pt}%
\definecolor{currentstroke}{rgb}{0.000000,0.000000,0.000000}%
\pgfsetstrokecolor{currentstroke}%
\pgfsetdash{}{0pt}%
\pgfpathmoveto{\pgfqpoint{3.130756in}{0.716028in}}%
\pgfpathlineto{\pgfqpoint{3.309851in}{0.716028in}}%
\pgfpathlineto{\pgfqpoint{3.309851in}{3.957578in}}%
\pgfpathlineto{\pgfqpoint{3.130756in}{3.957578in}}%
\pgfpathlineto{\pgfqpoint{3.130756in}{0.716028in}}%
\pgfpathclose%
\pgfusepath{fill}%
\end{pgfscope}%
\begin{pgfscope}%
\pgfpathrectangle{\pgfqpoint{0.802523in}{0.716028in}}{\pgfqpoint{5.193752in}{3.517843in}}%
\pgfusepath{clip}%
\pgfsetbuttcap%
\pgfsetmiterjoin%
\definecolor{currentfill}{rgb}{0.000000,0.000000,1.000000}%
\pgfsetfillcolor{currentfill}%
\pgfsetlinewidth{0.000000pt}%
\definecolor{currentstroke}{rgb}{0.000000,0.000000,0.000000}%
\pgfsetstrokecolor{currentstroke}%
\pgfsetdash{}{0pt}%
\pgfpathmoveto{\pgfqpoint{3.309851in}{0.716028in}}%
\pgfpathlineto{\pgfqpoint{3.488946in}{0.716028in}}%
\pgfpathlineto{\pgfqpoint{3.488946in}{4.066355in}}%
\pgfpathlineto{\pgfqpoint{3.309851in}{4.066355in}}%
\pgfpathlineto{\pgfqpoint{3.309851in}{0.716028in}}%
\pgfpathclose%
\pgfusepath{fill}%
\end{pgfscope}%
\begin{pgfscope}%
\pgfpathrectangle{\pgfqpoint{0.802523in}{0.716028in}}{\pgfqpoint{5.193752in}{3.517843in}}%
\pgfusepath{clip}%
\pgfsetbuttcap%
\pgfsetmiterjoin%
\definecolor{currentfill}{rgb}{0.000000,0.000000,1.000000}%
\pgfsetfillcolor{currentfill}%
\pgfsetlinewidth{0.000000pt}%
\definecolor{currentstroke}{rgb}{0.000000,0.000000,0.000000}%
\pgfsetstrokecolor{currentstroke}%
\pgfsetdash{}{0pt}%
\pgfpathmoveto{\pgfqpoint{3.488946in}{0.716028in}}%
\pgfpathlineto{\pgfqpoint{3.668041in}{0.716028in}}%
\pgfpathlineto{\pgfqpoint{3.668041in}{3.457204in}}%
\pgfpathlineto{\pgfqpoint{3.488946in}{3.457204in}}%
\pgfpathlineto{\pgfqpoint{3.488946in}{0.716028in}}%
\pgfpathclose%
\pgfusepath{fill}%
\end{pgfscope}%
\begin{pgfscope}%
\pgfpathrectangle{\pgfqpoint{0.802523in}{0.716028in}}{\pgfqpoint{5.193752in}{3.517843in}}%
\pgfusepath{clip}%
\pgfsetbuttcap%
\pgfsetmiterjoin%
\definecolor{currentfill}{rgb}{0.000000,0.000000,1.000000}%
\pgfsetfillcolor{currentfill}%
\pgfsetlinewidth{0.000000pt}%
\definecolor{currentstroke}{rgb}{0.000000,0.000000,0.000000}%
\pgfsetstrokecolor{currentstroke}%
\pgfsetdash{}{0pt}%
\pgfpathmoveto{\pgfqpoint{3.668041in}{0.716028in}}%
\pgfpathlineto{\pgfqpoint{3.847136in}{0.716028in}}%
\pgfpathlineto{\pgfqpoint{3.847136in}{2.891565in}}%
\pgfpathlineto{\pgfqpoint{3.668041in}{2.891565in}}%
\pgfpathlineto{\pgfqpoint{3.668041in}{0.716028in}}%
\pgfpathclose%
\pgfusepath{fill}%
\end{pgfscope}%
\begin{pgfscope}%
\pgfpathrectangle{\pgfqpoint{0.802523in}{0.716028in}}{\pgfqpoint{5.193752in}{3.517843in}}%
\pgfusepath{clip}%
\pgfsetbuttcap%
\pgfsetmiterjoin%
\definecolor{currentfill}{rgb}{0.000000,0.000000,1.000000}%
\pgfsetfillcolor{currentfill}%
\pgfsetlinewidth{0.000000pt}%
\definecolor{currentstroke}{rgb}{0.000000,0.000000,0.000000}%
\pgfsetstrokecolor{currentstroke}%
\pgfsetdash{}{0pt}%
\pgfpathmoveto{\pgfqpoint{3.847136in}{0.716028in}}%
\pgfpathlineto{\pgfqpoint{4.026231in}{0.716028in}}%
\pgfpathlineto{\pgfqpoint{4.026231in}{3.283162in}}%
\pgfpathlineto{\pgfqpoint{3.847136in}{3.283162in}}%
\pgfpathlineto{\pgfqpoint{3.847136in}{0.716028in}}%
\pgfpathclose%
\pgfusepath{fill}%
\end{pgfscope}%
\begin{pgfscope}%
\pgfpathrectangle{\pgfqpoint{0.802523in}{0.716028in}}{\pgfqpoint{5.193752in}{3.517843in}}%
\pgfusepath{clip}%
\pgfsetbuttcap%
\pgfsetmiterjoin%
\definecolor{currentfill}{rgb}{0.000000,0.000000,1.000000}%
\pgfsetfillcolor{currentfill}%
\pgfsetlinewidth{0.000000pt}%
\definecolor{currentstroke}{rgb}{0.000000,0.000000,0.000000}%
\pgfsetstrokecolor{currentstroke}%
\pgfsetdash{}{0pt}%
\pgfpathmoveto{\pgfqpoint{4.026231in}{0.716028in}}%
\pgfpathlineto{\pgfqpoint{4.205326in}{0.716028in}}%
\pgfpathlineto{\pgfqpoint{4.205326in}{3.631247in}}%
\pgfpathlineto{\pgfqpoint{4.026231in}{3.631247in}}%
\pgfpathlineto{\pgfqpoint{4.026231in}{0.716028in}}%
\pgfpathclose%
\pgfusepath{fill}%
\end{pgfscope}%
\begin{pgfscope}%
\pgfpathrectangle{\pgfqpoint{0.802523in}{0.716028in}}{\pgfqpoint{5.193752in}{3.517843in}}%
\pgfusepath{clip}%
\pgfsetbuttcap%
\pgfsetmiterjoin%
\definecolor{currentfill}{rgb}{0.000000,0.000000,1.000000}%
\pgfsetfillcolor{currentfill}%
\pgfsetlinewidth{0.000000pt}%
\definecolor{currentstroke}{rgb}{0.000000,0.000000,0.000000}%
\pgfsetstrokecolor{currentstroke}%
\pgfsetdash{}{0pt}%
\pgfpathmoveto{\pgfqpoint{4.205326in}{0.716028in}}%
\pgfpathlineto{\pgfqpoint{4.384421in}{0.716028in}}%
\pgfpathlineto{\pgfqpoint{4.384421in}{3.870556in}}%
\pgfpathlineto{\pgfqpoint{4.205326in}{3.870556in}}%
\pgfpathlineto{\pgfqpoint{4.205326in}{0.716028in}}%
\pgfpathclose%
\pgfusepath{fill}%
\end{pgfscope}%
\begin{pgfscope}%
\pgfpathrectangle{\pgfqpoint{0.802523in}{0.716028in}}{\pgfqpoint{5.193752in}{3.517843in}}%
\pgfusepath{clip}%
\pgfsetbuttcap%
\pgfsetmiterjoin%
\definecolor{currentfill}{rgb}{0.000000,0.000000,1.000000}%
\pgfsetfillcolor{currentfill}%
\pgfsetlinewidth{0.000000pt}%
\definecolor{currentstroke}{rgb}{0.000000,0.000000,0.000000}%
\pgfsetstrokecolor{currentstroke}%
\pgfsetdash{}{0pt}%
\pgfpathmoveto{\pgfqpoint{4.384421in}{0.716028in}}%
\pgfpathlineto{\pgfqpoint{4.563516in}{0.716028in}}%
\pgfpathlineto{\pgfqpoint{4.563516in}{3.000342in}}%
\pgfpathlineto{\pgfqpoint{4.384421in}{3.000342in}}%
\pgfpathlineto{\pgfqpoint{4.384421in}{0.716028in}}%
\pgfpathclose%
\pgfusepath{fill}%
\end{pgfscope}%
\begin{pgfscope}%
\pgfpathrectangle{\pgfqpoint{0.802523in}{0.716028in}}{\pgfqpoint{5.193752in}{3.517843in}}%
\pgfusepath{clip}%
\pgfsetbuttcap%
\pgfsetmiterjoin%
\definecolor{currentfill}{rgb}{0.000000,0.000000,1.000000}%
\pgfsetfillcolor{currentfill}%
\pgfsetlinewidth{0.000000pt}%
\definecolor{currentstroke}{rgb}{0.000000,0.000000,0.000000}%
\pgfsetstrokecolor{currentstroke}%
\pgfsetdash{}{0pt}%
\pgfpathmoveto{\pgfqpoint{4.563516in}{0.716028in}}%
\pgfpathlineto{\pgfqpoint{4.742611in}{0.716028in}}%
\pgfpathlineto{\pgfqpoint{4.742611in}{3.283162in}}%
\pgfpathlineto{\pgfqpoint{4.563516in}{3.283162in}}%
\pgfpathlineto{\pgfqpoint{4.563516in}{0.716028in}}%
\pgfpathclose%
\pgfusepath{fill}%
\end{pgfscope}%
\begin{pgfscope}%
\pgfpathrectangle{\pgfqpoint{0.802523in}{0.716028in}}{\pgfqpoint{5.193752in}{3.517843in}}%
\pgfusepath{clip}%
\pgfsetbuttcap%
\pgfsetmiterjoin%
\definecolor{currentfill}{rgb}{0.000000,0.000000,1.000000}%
\pgfsetfillcolor{currentfill}%
\pgfsetlinewidth{0.000000pt}%
\definecolor{currentstroke}{rgb}{0.000000,0.000000,0.000000}%
\pgfsetstrokecolor{currentstroke}%
\pgfsetdash{}{0pt}%
\pgfpathmoveto{\pgfqpoint{4.742611in}{0.716028in}}%
\pgfpathlineto{\pgfqpoint{4.921705in}{0.716028in}}%
\pgfpathlineto{\pgfqpoint{4.921705in}{2.869810in}}%
\pgfpathlineto{\pgfqpoint{4.742611in}{2.869810in}}%
\pgfpathlineto{\pgfqpoint{4.742611in}{0.716028in}}%
\pgfpathclose%
\pgfusepath{fill}%
\end{pgfscope}%
\begin{pgfscope}%
\pgfpathrectangle{\pgfqpoint{0.802523in}{0.716028in}}{\pgfqpoint{5.193752in}{3.517843in}}%
\pgfusepath{clip}%
\pgfsetbuttcap%
\pgfsetmiterjoin%
\definecolor{currentfill}{rgb}{0.000000,0.000000,1.000000}%
\pgfsetfillcolor{currentfill}%
\pgfsetlinewidth{0.000000pt}%
\definecolor{currentstroke}{rgb}{0.000000,0.000000,0.000000}%
\pgfsetstrokecolor{currentstroke}%
\pgfsetdash{}{0pt}%
\pgfpathmoveto{\pgfqpoint{4.921705in}{0.716028in}}%
\pgfpathlineto{\pgfqpoint{5.100800in}{0.716028in}}%
\pgfpathlineto{\pgfqpoint{5.100800in}{2.369436in}}%
\pgfpathlineto{\pgfqpoint{4.921705in}{2.369436in}}%
\pgfpathlineto{\pgfqpoint{4.921705in}{0.716028in}}%
\pgfpathclose%
\pgfusepath{fill}%
\end{pgfscope}%
\begin{pgfscope}%
\pgfpathrectangle{\pgfqpoint{0.802523in}{0.716028in}}{\pgfqpoint{5.193752in}{3.517843in}}%
\pgfusepath{clip}%
\pgfsetbuttcap%
\pgfsetmiterjoin%
\definecolor{currentfill}{rgb}{0.000000,0.000000,1.000000}%
\pgfsetfillcolor{currentfill}%
\pgfsetlinewidth{0.000000pt}%
\definecolor{currentstroke}{rgb}{0.000000,0.000000,0.000000}%
\pgfsetstrokecolor{currentstroke}%
\pgfsetdash{}{0pt}%
\pgfpathmoveto{\pgfqpoint{5.100800in}{0.716028in}}%
\pgfpathlineto{\pgfqpoint{5.279895in}{0.716028in}}%
\pgfpathlineto{\pgfqpoint{5.279895in}{2.021350in}}%
\pgfpathlineto{\pgfqpoint{5.100800in}{2.021350in}}%
\pgfpathlineto{\pgfqpoint{5.100800in}{0.716028in}}%
\pgfpathclose%
\pgfusepath{fill}%
\end{pgfscope}%
\begin{pgfscope}%
\pgfpathrectangle{\pgfqpoint{0.802523in}{0.716028in}}{\pgfqpoint{5.193752in}{3.517843in}}%
\pgfusepath{clip}%
\pgfsetbuttcap%
\pgfsetmiterjoin%
\definecolor{currentfill}{rgb}{0.000000,0.000000,1.000000}%
\pgfsetfillcolor{currentfill}%
\pgfsetlinewidth{0.000000pt}%
\definecolor{currentstroke}{rgb}{0.000000,0.000000,0.000000}%
\pgfsetstrokecolor{currentstroke}%
\pgfsetdash{}{0pt}%
\pgfpathmoveto{\pgfqpoint{5.279895in}{0.716028in}}%
\pgfpathlineto{\pgfqpoint{5.458990in}{0.716028in}}%
\pgfpathlineto{\pgfqpoint{5.458990in}{1.803797in}}%
\pgfpathlineto{\pgfqpoint{5.279895in}{1.803797in}}%
\pgfpathlineto{\pgfqpoint{5.279895in}{0.716028in}}%
\pgfpathclose%
\pgfusepath{fill}%
\end{pgfscope}%
\begin{pgfscope}%
\pgfpathrectangle{\pgfqpoint{0.802523in}{0.716028in}}{\pgfqpoint{5.193752in}{3.517843in}}%
\pgfusepath{clip}%
\pgfsetbuttcap%
\pgfsetmiterjoin%
\definecolor{currentfill}{rgb}{0.000000,0.000000,1.000000}%
\pgfsetfillcolor{currentfill}%
\pgfsetlinewidth{0.000000pt}%
\definecolor{currentstroke}{rgb}{0.000000,0.000000,0.000000}%
\pgfsetstrokecolor{currentstroke}%
\pgfsetdash{}{0pt}%
\pgfpathmoveto{\pgfqpoint{5.458990in}{0.716028in}}%
\pgfpathlineto{\pgfqpoint{5.638085in}{0.716028in}}%
\pgfpathlineto{\pgfqpoint{5.638085in}{1.368689in}}%
\pgfpathlineto{\pgfqpoint{5.458990in}{1.368689in}}%
\pgfpathlineto{\pgfqpoint{5.458990in}{0.716028in}}%
\pgfpathclose%
\pgfusepath{fill}%
\end{pgfscope}%
\begin{pgfscope}%
\pgfpathrectangle{\pgfqpoint{0.802523in}{0.716028in}}{\pgfqpoint{5.193752in}{3.517843in}}%
\pgfusepath{clip}%
\pgfsetbuttcap%
\pgfsetmiterjoin%
\definecolor{currentfill}{rgb}{0.000000,0.000000,1.000000}%
\pgfsetfillcolor{currentfill}%
\pgfsetlinewidth{0.000000pt}%
\definecolor{currentstroke}{rgb}{0.000000,0.000000,0.000000}%
\pgfsetstrokecolor{currentstroke}%
\pgfsetdash{}{0pt}%
\pgfpathmoveto{\pgfqpoint{5.638085in}{0.716028in}}%
\pgfpathlineto{\pgfqpoint{5.817180in}{0.716028in}}%
\pgfpathlineto{\pgfqpoint{5.817180in}{1.107625in}}%
\pgfpathlineto{\pgfqpoint{5.638085in}{1.107625in}}%
\pgfpathlineto{\pgfqpoint{5.638085in}{0.716028in}}%
\pgfpathclose%
\pgfusepath{fill}%
\end{pgfscope}%
\begin{pgfscope}%
\pgfpathrectangle{\pgfqpoint{0.802523in}{0.716028in}}{\pgfqpoint{5.193752in}{3.517843in}}%
\pgfusepath{clip}%
\pgfsetbuttcap%
\pgfsetmiterjoin%
\definecolor{currentfill}{rgb}{0.000000,0.000000,1.000000}%
\pgfsetfillcolor{currentfill}%
\pgfsetlinewidth{0.000000pt}%
\definecolor{currentstroke}{rgb}{0.000000,0.000000,0.000000}%
\pgfsetstrokecolor{currentstroke}%
\pgfsetdash{}{0pt}%
\pgfpathmoveto{\pgfqpoint{5.817180in}{0.716028in}}%
\pgfpathlineto{\pgfqpoint{5.996275in}{0.716028in}}%
\pgfpathlineto{\pgfqpoint{5.996275in}{0.824805in}}%
\pgfpathlineto{\pgfqpoint{5.817180in}{0.824805in}}%
\pgfpathlineto{\pgfqpoint{5.817180in}{0.716028in}}%
\pgfpathclose%
\pgfusepath{fill}%
\end{pgfscope}%
\begin{pgfscope}%
\pgfsetbuttcap%
\pgfsetroundjoin%
\definecolor{currentfill}{rgb}{0.000000,0.000000,0.000000}%
\pgfsetfillcolor{currentfill}%
\pgfsetlinewidth{0.200750pt}%
\definecolor{currentstroke}{rgb}{0.000000,0.000000,0.000000}%
\pgfsetstrokecolor{currentstroke}%
\pgfsetdash{}{0pt}%
\pgfsys@defobject{currentmarker}{\pgfqpoint{0.000000in}{-0.048611in}}{\pgfqpoint{0.000000in}{0.000000in}}{%
\pgfpathmoveto{\pgfqpoint{0.000000in}{0.000000in}}%
\pgfpathlineto{\pgfqpoint{0.000000in}{-0.048611in}}%
\pgfusepath{stroke,fill}%
}%
\begin{pgfscope}%
\pgfsys@transformshift{0.802523in}{0.716028in}%
\pgfsys@useobject{currentmarker}{}%
\end{pgfscope}%
\end{pgfscope}%
\begin{pgfscope}%
\definecolor{textcolor}{rgb}{0.000000,0.000000,0.000000}%
\pgfsetstrokecolor{textcolor}%
\pgfsetfillcolor{textcolor}%
\pgftext[x=0.802523in,y=0.618806in,,top]{\color{textcolor}{\rmfamily\fontsize{20.000000}{24.000000}\selectfont\catcode`\^=\active\def^{\ifmmode\sp\else\^{}\fi}\catcode`\%=\active\def
\end{pgfscope}%
\begin{pgfscope}%
\pgfsetbuttcap%
\pgfsetroundjoin%
\definecolor{currentfill}{rgb}{0.000000,0.000000,0.000000}%
\pgfsetfillcolor{currentfill}%
\pgfsetlinewidth{0.200750pt}%
\definecolor{currentstroke}{rgb}{0.000000,0.000000,0.000000}%
\pgfsetstrokecolor{currentstroke}%
\pgfsetdash{}{0pt}%
\pgfsys@defobject{currentmarker}{\pgfqpoint{0.000000in}{-0.048611in}}{\pgfqpoint{0.000000in}{0.000000in}}{%
\pgfpathmoveto{\pgfqpoint{0.000000in}{0.000000in}}%
\pgfpathlineto{\pgfqpoint{0.000000in}{-0.048611in}}%
\pgfusepath{stroke,fill}%
}%
\begin{pgfscope}%
\pgfsys@transformshift{1.841273in}{0.716028in}%
\pgfsys@useobject{currentmarker}{}%
\end{pgfscope}%
\end{pgfscope}%
\begin{pgfscope}%
\definecolor{textcolor}{rgb}{0.000000,0.000000,0.000000}%
\pgfsetstrokecolor{textcolor}%
\pgfsetfillcolor{textcolor}%
\pgftext[x=1.841273in,y=0.618806in,,top]{\color{textcolor}{\rmfamily\fontsize{20.000000}{24.000000}\selectfont\catcode`\^=\active\def^{\ifmmode\sp\else\^{}\fi}\catcode`\%=\active\def
\end{pgfscope}%
\begin{pgfscope}%
\pgfsetbuttcap%
\pgfsetroundjoin%
\definecolor{currentfill}{rgb}{0.000000,0.000000,0.000000}%
\pgfsetfillcolor{currentfill}%
\pgfsetlinewidth{0.200750pt}%
\definecolor{currentstroke}{rgb}{0.000000,0.000000,0.000000}%
\pgfsetstrokecolor{currentstroke}%
\pgfsetdash{}{0pt}%
\pgfsys@defobject{currentmarker}{\pgfqpoint{0.000000in}{-0.048611in}}{\pgfqpoint{0.000000in}{0.000000in}}{%
\pgfpathmoveto{\pgfqpoint{0.000000in}{0.000000in}}%
\pgfpathlineto{\pgfqpoint{0.000000in}{-0.048611in}}%
\pgfusepath{stroke,fill}%
}%
\begin{pgfscope}%
\pgfsys@transformshift{2.880024in}{0.716028in}%
\pgfsys@useobject{currentmarker}{}%
\end{pgfscope}%
\end{pgfscope}%
\begin{pgfscope}%
\definecolor{textcolor}{rgb}{0.000000,0.000000,0.000000}%
\pgfsetstrokecolor{textcolor}%
\pgfsetfillcolor{textcolor}%
\pgftext[x=2.880024in,y=0.618806in,,top]{\color{textcolor}{\rmfamily\fontsize{20.000000}{24.000000}\selectfont\catcode`\^=\active\def^{\ifmmode\sp\else\^{}\fi}\catcode`\%=\active\def
\end{pgfscope}%
\begin{pgfscope}%
\pgfsetbuttcap%
\pgfsetroundjoin%
\definecolor{currentfill}{rgb}{0.000000,0.000000,0.000000}%
\pgfsetfillcolor{currentfill}%
\pgfsetlinewidth{0.200750pt}%
\definecolor{currentstroke}{rgb}{0.000000,0.000000,0.000000}%
\pgfsetstrokecolor{currentstroke}%
\pgfsetdash{}{0pt}%
\pgfsys@defobject{currentmarker}{\pgfqpoint{0.000000in}{-0.048611in}}{\pgfqpoint{0.000000in}{0.000000in}}{%
\pgfpathmoveto{\pgfqpoint{0.000000in}{0.000000in}}%
\pgfpathlineto{\pgfqpoint{0.000000in}{-0.048611in}}%
\pgfusepath{stroke,fill}%
}%
\begin{pgfscope}%
\pgfsys@transformshift{3.918774in}{0.716028in}%
\pgfsys@useobject{currentmarker}{}%
\end{pgfscope}%
\end{pgfscope}%
\begin{pgfscope}%
\definecolor{textcolor}{rgb}{0.000000,0.000000,0.000000}%
\pgfsetstrokecolor{textcolor}%
\pgfsetfillcolor{textcolor}%
\pgftext[x=3.918774in,y=0.618806in,,top]{\color{textcolor}{\rmfamily\fontsize{20.000000}{24.000000}\selectfont\catcode`\^=\active\def^{\ifmmode\sp\else\^{}\fi}\catcode`\%=\active\def
\end{pgfscope}%
\begin{pgfscope}%
\pgfsetbuttcap%
\pgfsetroundjoin%
\definecolor{currentfill}{rgb}{0.000000,0.000000,0.000000}%
\pgfsetfillcolor{currentfill}%
\pgfsetlinewidth{0.200750pt}%
\definecolor{currentstroke}{rgb}{0.000000,0.000000,0.000000}%
\pgfsetstrokecolor{currentstroke}%
\pgfsetdash{}{0pt}%
\pgfsys@defobject{currentmarker}{\pgfqpoint{0.000000in}{-0.048611in}}{\pgfqpoint{0.000000in}{0.000000in}}{%
\pgfpathmoveto{\pgfqpoint{0.000000in}{0.000000in}}%
\pgfpathlineto{\pgfqpoint{0.000000in}{-0.048611in}}%
\pgfusepath{stroke,fill}%
}%
\begin{pgfscope}%
\pgfsys@transformshift{4.957524in}{0.716028in}%
\pgfsys@useobject{currentmarker}{}%
\end{pgfscope}%
\end{pgfscope}%
\begin{pgfscope}%
\definecolor{textcolor}{rgb}{0.000000,0.000000,0.000000}%
\pgfsetstrokecolor{textcolor}%
\pgfsetfillcolor{textcolor}%
\pgftext[x=4.957524in,y=0.618806in,,top]{\color{textcolor}{\rmfamily\fontsize{20.000000}{24.000000}\selectfont\catcode`\^=\active\def^{\ifmmode\sp\else\^{}\fi}\catcode`\%=\active\def
\end{pgfscope}%
\begin{pgfscope}%
\pgfsetbuttcap%
\pgfsetroundjoin%
\definecolor{currentfill}{rgb}{0.000000,0.000000,0.000000}%
\pgfsetfillcolor{currentfill}%
\pgfsetlinewidth{0.200750pt}%
\definecolor{currentstroke}{rgb}{0.000000,0.000000,0.000000}%
\pgfsetstrokecolor{currentstroke}%
\pgfsetdash{}{0pt}%
\pgfsys@defobject{currentmarker}{\pgfqpoint{0.000000in}{-0.048611in}}{\pgfqpoint{0.000000in}{0.000000in}}{%
\pgfpathmoveto{\pgfqpoint{0.000000in}{0.000000in}}%
\pgfpathlineto{\pgfqpoint{0.000000in}{-0.048611in}}%
\pgfusepath{stroke,fill}%
}%
\begin{pgfscope}%
\pgfsys@transformshift{5.996275in}{0.716028in}%
\pgfsys@useobject{currentmarker}{}%
\end{pgfscope}%
\end{pgfscope}%
\begin{pgfscope}%
\definecolor{textcolor}{rgb}{0.000000,0.000000,0.000000}%
\pgfsetstrokecolor{textcolor}%
\pgfsetfillcolor{textcolor}%
\pgftext[x=5.996275in,y=0.618806in,,top]{\color{textcolor}{\rmfamily\fontsize{20.000000}{24.000000}\selectfont\catcode`\^=\active\def^{\ifmmode\sp\else\^{}\fi}\catcode`\%=\active\def
\end{pgfscope}%
\begin{pgfscope}%
\definecolor{textcolor}{rgb}{0.000000,0.000000,0.000000}%
\pgfsetstrokecolor{textcolor}%
\pgfsetfillcolor{textcolor}%
\pgftext[x=3.399399in,y=0.307183in,,top]{\color{textcolor}{\rmfamily\fontsize{26.000000}{31.200000}\selectfont\catcode`\^=\active\def^{\ifmmode\sp\else\^{}\fi}\catcode`\%=\active\def
\end{pgfscope}%
\begin{pgfscope}%
\pgfsetbuttcap%
\pgfsetroundjoin%
\definecolor{currentfill}{rgb}{0.000000,0.000000,0.000000}%
\pgfsetfillcolor{currentfill}%
\pgfsetlinewidth{0.200750pt}%
\definecolor{currentstroke}{rgb}{0.000000,0.000000,0.000000}%
\pgfsetstrokecolor{currentstroke}%
\pgfsetdash{}{0pt}%
\pgfsys@defobject{currentmarker}{\pgfqpoint{-0.048611in}{0.000000in}}{\pgfqpoint{-0.000000in}{0.000000in}}{%
\pgfpathmoveto{\pgfqpoint{-0.000000in}{0.000000in}}%
\pgfpathlineto{\pgfqpoint{-0.048611in}{0.000000in}}%
\pgfusepath{stroke,fill}%
}%
\begin{pgfscope}%
\pgfsys@transformshift{0.802523in}{0.716028in}%
\pgfsys@useobject{currentmarker}{}%
\end{pgfscope}%
\end{pgfscope}%
\begin{pgfscope}%
\definecolor{textcolor}{rgb}{0.000000,0.000000,0.000000}%
\pgfsetstrokecolor{textcolor}%
\pgfsetfillcolor{textcolor}%
\pgftext[x=0.362738in, y=0.616009in, left, base]{\color{textcolor}{\rmfamily\fontsize{20.000000}{24.000000}\selectfont\catcode`\^=\active\def^{\ifmmode\sp\else\^{}\fi}\catcode`\%=\active\def
\end{pgfscope}%
\begin{pgfscope}%
\pgfsetbuttcap%
\pgfsetroundjoin%
\definecolor{currentfill}{rgb}{0.000000,0.000000,0.000000}%
\pgfsetfillcolor{currentfill}%
\pgfsetlinewidth{0.200750pt}%
\definecolor{currentstroke}{rgb}{0.000000,0.000000,0.000000}%
\pgfsetstrokecolor{currentstroke}%
\pgfsetdash{}{0pt}%
\pgfsys@defobject{currentmarker}{\pgfqpoint{-0.048611in}{0.000000in}}{\pgfqpoint{-0.000000in}{0.000000in}}{%
\pgfpathmoveto{\pgfqpoint{-0.000000in}{0.000000in}}%
\pgfpathlineto{\pgfqpoint{-0.048611in}{0.000000in}}%
\pgfusepath{stroke,fill}%
}%
\begin{pgfscope}%
\pgfsys@transformshift{0.802523in}{1.445958in}%
\pgfsys@useobject{currentmarker}{}%
\end{pgfscope}%
\end{pgfscope}%
\begin{pgfscope}%
\definecolor{textcolor}{rgb}{0.000000,0.000000,0.000000}%
\pgfsetstrokecolor{textcolor}%
\pgfsetfillcolor{textcolor}%
\pgftext[x=0.362738in, y=1.345939in, left, base]{\color{textcolor}{\rmfamily\fontsize{20.000000}{24.000000}\selectfont\catcode`\^=\active\def^{\ifmmode\sp\else\^{}\fi}\catcode`\%=\active\def
\end{pgfscope}%
\begin{pgfscope}%
\pgfsetbuttcap%
\pgfsetroundjoin%
\definecolor{currentfill}{rgb}{0.000000,0.000000,0.000000}%
\pgfsetfillcolor{currentfill}%
\pgfsetlinewidth{0.200750pt}%
\definecolor{currentstroke}{rgb}{0.000000,0.000000,0.000000}%
\pgfsetstrokecolor{currentstroke}%
\pgfsetdash{}{0pt}%
\pgfsys@defobject{currentmarker}{\pgfqpoint{-0.048611in}{0.000000in}}{\pgfqpoint{-0.000000in}{0.000000in}}{%
\pgfpathmoveto{\pgfqpoint{-0.000000in}{0.000000in}}%
\pgfpathlineto{\pgfqpoint{-0.048611in}{0.000000in}}%
\pgfusepath{stroke,fill}%
}%
\begin{pgfscope}%
\pgfsys@transformshift{0.802523in}{2.175888in}%
\pgfsys@useobject{currentmarker}{}%
\end{pgfscope}%
\end{pgfscope}%
\begin{pgfscope}%
\definecolor{textcolor}{rgb}{0.000000,0.000000,0.000000}%
\pgfsetstrokecolor{textcolor}%
\pgfsetfillcolor{textcolor}%
\pgftext[x=0.362738in, y=2.075869in, left, base]{\color{textcolor}{\rmfamily\fontsize{20.000000}{24.000000}\selectfont\catcode`\^=\active\def^{\ifmmode\sp\else\^{}\fi}\catcode`\%=\active\def
\end{pgfscope}%
\begin{pgfscope}%
\pgfsetbuttcap%
\pgfsetroundjoin%
\definecolor{currentfill}{rgb}{0.000000,0.000000,0.000000}%
\pgfsetfillcolor{currentfill}%
\pgfsetlinewidth{0.200750pt}%
\definecolor{currentstroke}{rgb}{0.000000,0.000000,0.000000}%
\pgfsetstrokecolor{currentstroke}%
\pgfsetdash{}{0pt}%
\pgfsys@defobject{currentmarker}{\pgfqpoint{-0.048611in}{0.000000in}}{\pgfqpoint{-0.000000in}{0.000000in}}{%
\pgfpathmoveto{\pgfqpoint{-0.000000in}{0.000000in}}%
\pgfpathlineto{\pgfqpoint{-0.048611in}{0.000000in}}%
\pgfusepath{stroke,fill}%
}%
\begin{pgfscope}%
\pgfsys@transformshift{0.802523in}{2.905818in}%
\pgfsys@useobject{currentmarker}{}%
\end{pgfscope}%
\end{pgfscope}%
\begin{pgfscope}%
\definecolor{textcolor}{rgb}{0.000000,0.000000,0.000000}%
\pgfsetstrokecolor{textcolor}%
\pgfsetfillcolor{textcolor}%
\pgftext[x=0.362738in, y=2.805799in, left, base]{\color{textcolor}{\rmfamily\fontsize{20.000000}{24.000000}\selectfont\catcode`\^=\active\def^{\ifmmode\sp\else\^{}\fi}\catcode`\%=\active\def
\end{pgfscope}%
\begin{pgfscope}%
\pgfsetbuttcap%
\pgfsetroundjoin%
\definecolor{currentfill}{rgb}{0.000000,0.000000,0.000000}%
\pgfsetfillcolor{currentfill}%
\pgfsetlinewidth{0.200750pt}%
\definecolor{currentstroke}{rgb}{0.000000,0.000000,0.000000}%
\pgfsetstrokecolor{currentstroke}%
\pgfsetdash{}{0pt}%
\pgfsys@defobject{currentmarker}{\pgfqpoint{-0.048611in}{0.000000in}}{\pgfqpoint{-0.000000in}{0.000000in}}{%
\pgfpathmoveto{\pgfqpoint{-0.000000in}{0.000000in}}%
\pgfpathlineto{\pgfqpoint{-0.048611in}{0.000000in}}%
\pgfusepath{stroke,fill}%
}%
\begin{pgfscope}%
\pgfsys@transformshift{0.802523in}{3.635749in}%
\pgfsys@useobject{currentmarker}{}%
\end{pgfscope}%
\end{pgfscope}%
\begin{pgfscope}%
\definecolor{textcolor}{rgb}{0.000000,0.000000,0.000000}%
\pgfsetstrokecolor{textcolor}%
\pgfsetfillcolor{textcolor}%
\pgftext[x=0.362738in, y=3.535729in, left, base]{\color{textcolor}{\rmfamily\fontsize{20.000000}{24.000000}\selectfont\catcode`\^=\active\def^{\ifmmode\sp\else\^{}\fi}\catcode`\%=\active\def
\end{pgfscope}%
\begin{pgfscope}%
\definecolor{textcolor}{rgb}{0.000000,0.000000,0.000000}%
\pgfsetstrokecolor{textcolor}%
\pgfsetfillcolor{textcolor}%
\pgftext[x=0.307183in,y=2.474950in,,bottom,rotate=90.000000]{\color{textcolor}{\rmfamily\fontsize{26.000000}{31.200000}\selectfont\catcode`\^=\active\def^{\ifmmode\sp\else\^{}\fi}\catcode`\%=\active\def
\end{pgfscope}%
\begin{pgfscope}%
\pgfpathrectangle{\pgfqpoint{0.802523in}{0.716028in}}{\pgfqpoint{5.193752in}{3.517843in}}%
\pgfusepath{clip}%
\pgfsetbuttcap%
\pgfsetroundjoin%
\pgfsetlinewidth{2.007500pt}%
\definecolor{currentstroke}{rgb}{0.501961,0.501961,0.501961}%
\pgfsetstrokecolor{currentstroke}%
\pgfsetdash{{7.400000pt}{3.200000pt}}{0.000000pt}%
\pgfpathmoveto{\pgfqpoint{0.802523in}{2.175888in}}%
\pgfpathlineto{\pgfqpoint{5.996275in}{2.175888in}}%
\pgfusepath{stroke}%
\end{pgfscope}%
\begin{pgfscope}%
\pgfsetrectcap%
\pgfsetmiterjoin%
\pgfsetlinewidth{0.803000pt}%
\definecolor{currentstroke}{rgb}{0.000000,0.000000,0.000000}%
\pgfsetstrokecolor{currentstroke}%
\pgfsetdash{}{0pt}%
\pgfpathmoveto{\pgfqpoint{0.802523in}{0.716028in}}%
\pgfpathlineto{\pgfqpoint{0.802523in}{4.233871in}}%
\pgfusepath{stroke}%
\end{pgfscope}%
\begin{pgfscope}%
\pgfsetrectcap%
\pgfsetmiterjoin%
\pgfsetlinewidth{0.803000pt}%
\definecolor{currentstroke}{rgb}{0.000000,0.000000,0.000000}%
\pgfsetstrokecolor{currentstroke}%
\pgfsetdash{}{0pt}%
\pgfpathmoveto{\pgfqpoint{5.996275in}{0.716028in}}%
\pgfpathlineto{\pgfqpoint{5.996275in}{4.233871in}}%
\pgfusepath{stroke}%
\end{pgfscope}%
\begin{pgfscope}%
\pgfsetrectcap%
\pgfsetmiterjoin%
\pgfsetlinewidth{0.803000pt}%
\definecolor{currentstroke}{rgb}{0.000000,0.000000,0.000000}%
\pgfsetstrokecolor{currentstroke}%
\pgfsetdash{}{0pt}%
\pgfpathmoveto{\pgfqpoint{0.802523in}{0.716028in}}%
\pgfpathlineto{\pgfqpoint{5.996275in}{0.716028in}}%
\pgfusepath{stroke}%
\end{pgfscope}%
\begin{pgfscope}%
\pgfsetrectcap%
\pgfsetmiterjoin%
\pgfsetlinewidth{0.803000pt}%
\definecolor{currentstroke}{rgb}{0.000000,0.000000,0.000000}%
\pgfsetstrokecolor{currentstroke}%
\pgfsetdash{}{0pt}%
\pgfpathmoveto{\pgfqpoint{0.802523in}{4.233871in}}%
\pgfpathlineto{\pgfqpoint{5.996275in}{4.233871in}}%
\pgfusepath{stroke}%
\end{pgfscope}%
\begin{pgfscope}%
\definecolor{textcolor}{rgb}{0.000000,0.000000,0.000000}%
\pgfsetstrokecolor{textcolor}%
\pgfsetfillcolor{textcolor}%
\pgftext[x=3.399399in,y=4.317204in,,base]{\color{textcolor}{\rmfamily\fontsize{25.000000}{30.000000}\selectfont\catcode`\^=\active\def^{\ifmmode\sp\else\^{}\fi}\catcode`\%=\active\def
\end{pgfscope}%
\begin{pgfscope}%
\pgfsetbuttcap%
\pgfsetmiterjoin%
\definecolor{currentfill}{rgb}{1.000000,1.000000,1.000000}%
\pgfsetfillcolor{currentfill}%
\pgfsetfillopacity{0.800000}%
\pgfsetlinewidth{1.003750pt}%
\definecolor{currentstroke}{rgb}{0.800000,0.800000,0.800000}%
\pgfsetstrokecolor{currentstroke}%
\pgfsetstrokeopacity{0.800000}%
\pgfsetdash{}{0pt}%
\pgfpathmoveto{\pgfqpoint{0.996967in}{3.616692in}}%
\pgfpathlineto{\pgfqpoint{2.538869in}{3.616692in}}%
\pgfpathquadraticcurveto{\pgfqpoint{2.594425in}{3.616692in}}{\pgfqpoint{2.594425in}{3.672248in}}%
\pgfpathlineto{\pgfqpoint{2.594425in}{4.039427in}}%
\pgfpathquadraticcurveto{\pgfqpoint{2.594425in}{4.094982in}}{\pgfqpoint{2.538869in}{4.094982in}}%
\pgfpathlineto{\pgfqpoint{0.996967in}{4.094982in}}%
\pgfpathquadraticcurveto{\pgfqpoint{0.941412in}{4.094982in}}{\pgfqpoint{0.941412in}{4.039427in}}%
\pgfpathlineto{\pgfqpoint{0.941412in}{3.672248in}}%
\pgfpathquadraticcurveto{\pgfqpoint{0.941412in}{3.616692in}}{\pgfqpoint{0.996967in}{3.616692in}}%
\pgfpathlineto{\pgfqpoint{0.996967in}{3.616692in}}%
\pgfpathclose%
\pgfusepath{stroke,fill}%
\end{pgfscope}%
\begin{pgfscope}%
\pgfsetbuttcap%
\pgfsetroundjoin%
\pgfsetlinewidth{2.007500pt}%
\definecolor{currentstroke}{rgb}{0.501961,0.501961,0.501961}%
\pgfsetstrokecolor{currentstroke}%
\pgfsetdash{{7.400000pt}{3.200000pt}}{0.000000pt}%
\pgfpathmoveto{\pgfqpoint{1.052523in}{3.881055in}}%
\pgfpathlineto{\pgfqpoint{1.330301in}{3.881055in}}%
\pgfpathlineto{\pgfqpoint{1.608078in}{3.881055in}}%
\pgfusepath{stroke}%
\end{pgfscope}%
\begin{pgfscope}%
\definecolor{textcolor}{rgb}{0.000000,0.000000,0.000000}%
\pgfsetstrokecolor{textcolor}%
\pgfsetfillcolor{textcolor}%
\pgftext[x=1.830301in,y=3.783833in,left,base]{\color{textcolor}{\rmfamily\fontsize{20.000000}{24.000000}\selectfont\catcode`\^=\active\def^{\ifmmode\sp\else\^{}\fi}\catcode`\%=\active\def
\end{pgfscope}%
\end{pgfpicture}%
\makeatother%
\endgroup%

%% file: figures/evaluation/Helpdesk/Helpdesk_PIT_remaining_time_norm_4layer.pgf
\begingroup%
\makeatletter%
\begin{pgfpicture}%
\pgfpathrectangle{\pgfpointorigin}{\pgfqpoint{6.167556in}{4.557174in}}%
\pgfusepath{use as bounding box, clip}%
\begin{pgfscope}%
\pgfsetbuttcap%
\pgfsetmiterjoin%
\definecolor{currentfill}{rgb}{1.000000,1.000000,1.000000}%
\pgfsetfillcolor{currentfill}%
\pgfsetlinewidth{0.000000pt}%
\definecolor{currentstroke}{rgb}{1.000000,1.000000,1.000000}%
\pgfsetstrokecolor{currentstroke}%
\pgfsetdash{}{0pt}%
\pgfpathmoveto{\pgfqpoint{0.000000in}{0.000000in}}%
\pgfpathlineto{\pgfqpoint{6.167556in}{0.000000in}}%
\pgfpathlineto{\pgfqpoint{6.167556in}{4.557174in}}%
\pgfpathlineto{\pgfqpoint{0.000000in}{4.557174in}}%
\pgfpathlineto{\pgfqpoint{0.000000in}{0.000000in}}%
\pgfpathclose%
\pgfusepath{fill}%
\end{pgfscope}%
\begin{pgfscope}%
\pgfsetbuttcap%
\pgfsetmiterjoin%
\definecolor{currentfill}{rgb}{1.000000,1.000000,1.000000}%
\pgfsetfillcolor{currentfill}%
\pgfsetlinewidth{0.000000pt}%
\definecolor{currentstroke}{rgb}{0.000000,0.000000,0.000000}%
\pgfsetstrokecolor{currentstroke}%
\pgfsetstrokeopacity{0.000000}%
\pgfsetdash{}{0pt}%
\pgfpathmoveto{\pgfqpoint{0.802523in}{0.716028in}}%
\pgfpathlineto{\pgfqpoint{5.996275in}{0.716028in}}%
\pgfpathlineto{\pgfqpoint{5.996275in}{4.233871in}}%
\pgfpathlineto{\pgfqpoint{0.802523in}{4.233871in}}%
\pgfpathlineto{\pgfqpoint{0.802523in}{0.716028in}}%
\pgfpathclose%
\pgfusepath{fill}%
\end{pgfscope}%
\begin{pgfscope}%
\pgfpathrectangle{\pgfqpoint{0.802523in}{0.716028in}}{\pgfqpoint{5.193752in}{3.517843in}}%
\pgfusepath{clip}%
\pgfsetbuttcap%
\pgfsetmiterjoin%
\definecolor{currentfill}{rgb}{0.000000,0.000000,1.000000}%
\pgfsetfillcolor{currentfill}%
\pgfsetlinewidth{0.000000pt}%
\definecolor{currentstroke}{rgb}{0.000000,0.000000,0.000000}%
\pgfsetstrokecolor{currentstroke}%
\pgfsetdash{}{0pt}%
\pgfpathmoveto{\pgfqpoint{0.802523in}{0.716028in}}%
\pgfpathlineto{\pgfqpoint{0.981618in}{0.716028in}}%
\pgfpathlineto{\pgfqpoint{0.981618in}{0.716028in}}%
\pgfpathlineto{\pgfqpoint{0.802523in}{0.716028in}}%
\pgfpathlineto{\pgfqpoint{0.802523in}{0.716028in}}%
\pgfpathclose%
\pgfusepath{fill}%
\end{pgfscope}%
\begin{pgfscope}%
\pgfpathrectangle{\pgfqpoint{0.802523in}{0.716028in}}{\pgfqpoint{5.193752in}{3.517843in}}%
\pgfusepath{clip}%
\pgfsetbuttcap%
\pgfsetmiterjoin%
\definecolor{currentfill}{rgb}{0.000000,0.000000,1.000000}%
\pgfsetfillcolor{currentfill}%
\pgfsetlinewidth{0.000000pt}%
\definecolor{currentstroke}{rgb}{0.000000,0.000000,0.000000}%
\pgfsetstrokecolor{currentstroke}%
\pgfsetdash{}{0pt}%
\pgfpathmoveto{\pgfqpoint{0.981618in}{0.716028in}}%
\pgfpathlineto{\pgfqpoint{1.160713in}{0.716028in}}%
\pgfpathlineto{\pgfqpoint{1.160713in}{0.716028in}}%
\pgfpathlineto{\pgfqpoint{0.981618in}{0.716028in}}%
\pgfpathlineto{\pgfqpoint{0.981618in}{0.716028in}}%
\pgfpathclose%
\pgfusepath{fill}%
\end{pgfscope}%
\begin{pgfscope}%
\pgfpathrectangle{\pgfqpoint{0.802523in}{0.716028in}}{\pgfqpoint{5.193752in}{3.517843in}}%
\pgfusepath{clip}%
\pgfsetbuttcap%
\pgfsetmiterjoin%
\definecolor{currentfill}{rgb}{0.000000,0.000000,1.000000}%
\pgfsetfillcolor{currentfill}%
\pgfsetlinewidth{0.000000pt}%
\definecolor{currentstroke}{rgb}{0.000000,0.000000,0.000000}%
\pgfsetstrokecolor{currentstroke}%
\pgfsetdash{}{0pt}%
\pgfpathmoveto{\pgfqpoint{1.160713in}{0.716028in}}%
\pgfpathlineto{\pgfqpoint{1.339808in}{0.716028in}}%
\pgfpathlineto{\pgfqpoint{1.339808in}{0.716028in}}%
\pgfpathlineto{\pgfqpoint{1.160713in}{0.716028in}}%
\pgfpathlineto{\pgfqpoint{1.160713in}{0.716028in}}%
\pgfpathclose%
\pgfusepath{fill}%
\end{pgfscope}%
\begin{pgfscope}%
\pgfpathrectangle{\pgfqpoint{0.802523in}{0.716028in}}{\pgfqpoint{5.193752in}{3.517843in}}%
\pgfusepath{clip}%
\pgfsetbuttcap%
\pgfsetmiterjoin%
\definecolor{currentfill}{rgb}{0.000000,0.000000,1.000000}%
\pgfsetfillcolor{currentfill}%
\pgfsetlinewidth{0.000000pt}%
\definecolor{currentstroke}{rgb}{0.000000,0.000000,0.000000}%
\pgfsetstrokecolor{currentstroke}%
\pgfsetdash{}{0pt}%
\pgfpathmoveto{\pgfqpoint{1.339808in}{0.716028in}}%
\pgfpathlineto{\pgfqpoint{1.518902in}{0.716028in}}%
\pgfpathlineto{\pgfqpoint{1.518902in}{0.716028in}}%
\pgfpathlineto{\pgfqpoint{1.339808in}{0.716028in}}%
\pgfpathlineto{\pgfqpoint{1.339808in}{0.716028in}}%
\pgfpathclose%
\pgfusepath{fill}%
\end{pgfscope}%
\begin{pgfscope}%
\pgfpathrectangle{\pgfqpoint{0.802523in}{0.716028in}}{\pgfqpoint{5.193752in}{3.517843in}}%
\pgfusepath{clip}%
\pgfsetbuttcap%
\pgfsetmiterjoin%
\definecolor{currentfill}{rgb}{0.000000,0.000000,1.000000}%
\pgfsetfillcolor{currentfill}%
\pgfsetlinewidth{0.000000pt}%
\definecolor{currentstroke}{rgb}{0.000000,0.000000,0.000000}%
\pgfsetstrokecolor{currentstroke}%
\pgfsetdash{}{0pt}%
\pgfpathmoveto{\pgfqpoint{1.518902in}{0.716028in}}%
\pgfpathlineto{\pgfqpoint{1.697997in}{0.716028in}}%
\pgfpathlineto{\pgfqpoint{1.697997in}{0.739294in}}%
\pgfpathlineto{\pgfqpoint{1.518902in}{0.739294in}}%
\pgfpathlineto{\pgfqpoint{1.518902in}{0.716028in}}%
\pgfpathclose%
\pgfusepath{fill}%
\end{pgfscope}%
\begin{pgfscope}%
\pgfpathrectangle{\pgfqpoint{0.802523in}{0.716028in}}{\pgfqpoint{5.193752in}{3.517843in}}%
\pgfusepath{clip}%
\pgfsetbuttcap%
\pgfsetmiterjoin%
\definecolor{currentfill}{rgb}{0.000000,0.000000,1.000000}%
\pgfsetfillcolor{currentfill}%
\pgfsetlinewidth{0.000000pt}%
\definecolor{currentstroke}{rgb}{0.000000,0.000000,0.000000}%
\pgfsetstrokecolor{currentstroke}%
\pgfsetdash{}{0pt}%
\pgfpathmoveto{\pgfqpoint{1.697997in}{0.716028in}}%
\pgfpathlineto{\pgfqpoint{1.877092in}{0.716028in}}%
\pgfpathlineto{\pgfqpoint{1.877092in}{1.320948in}}%
\pgfpathlineto{\pgfqpoint{1.697997in}{1.320948in}}%
\pgfpathlineto{\pgfqpoint{1.697997in}{0.716028in}}%
\pgfpathclose%
\pgfusepath{fill}%
\end{pgfscope}%
\begin{pgfscope}%
\pgfpathrectangle{\pgfqpoint{0.802523in}{0.716028in}}{\pgfqpoint{5.193752in}{3.517843in}}%
\pgfusepath{clip}%
\pgfsetbuttcap%
\pgfsetmiterjoin%
\definecolor{currentfill}{rgb}{0.000000,0.000000,1.000000}%
\pgfsetfillcolor{currentfill}%
\pgfsetlinewidth{0.000000pt}%
\definecolor{currentstroke}{rgb}{0.000000,0.000000,0.000000}%
\pgfsetstrokecolor{currentstroke}%
\pgfsetdash{}{0pt}%
\pgfpathmoveto{\pgfqpoint{1.877092in}{0.716028in}}%
\pgfpathlineto{\pgfqpoint{2.056187in}{0.716028in}}%
\pgfpathlineto{\pgfqpoint{2.056187in}{1.809538in}}%
\pgfpathlineto{\pgfqpoint{1.877092in}{1.809538in}}%
\pgfpathlineto{\pgfqpoint{1.877092in}{0.716028in}}%
\pgfpathclose%
\pgfusepath{fill}%
\end{pgfscope}%
\begin{pgfscope}%
\pgfpathrectangle{\pgfqpoint{0.802523in}{0.716028in}}{\pgfqpoint{5.193752in}{3.517843in}}%
\pgfusepath{clip}%
\pgfsetbuttcap%
\pgfsetmiterjoin%
\definecolor{currentfill}{rgb}{0.000000,0.000000,1.000000}%
\pgfsetfillcolor{currentfill}%
\pgfsetlinewidth{0.000000pt}%
\definecolor{currentstroke}{rgb}{0.000000,0.000000,0.000000}%
\pgfsetstrokecolor{currentstroke}%
\pgfsetdash{}{0pt}%
\pgfpathmoveto{\pgfqpoint{2.056187in}{0.716028in}}%
\pgfpathlineto{\pgfqpoint{2.235282in}{0.716028in}}%
\pgfpathlineto{\pgfqpoint{2.235282in}{1.507078in}}%
\pgfpathlineto{\pgfqpoint{2.056187in}{1.507078in}}%
\pgfpathlineto{\pgfqpoint{2.056187in}{0.716028in}}%
\pgfpathclose%
\pgfusepath{fill}%
\end{pgfscope}%
\begin{pgfscope}%
\pgfpathrectangle{\pgfqpoint{0.802523in}{0.716028in}}{\pgfqpoint{5.193752in}{3.517843in}}%
\pgfusepath{clip}%
\pgfsetbuttcap%
\pgfsetmiterjoin%
\definecolor{currentfill}{rgb}{0.000000,0.000000,1.000000}%
\pgfsetfillcolor{currentfill}%
\pgfsetlinewidth{0.000000pt}%
\definecolor{currentstroke}{rgb}{0.000000,0.000000,0.000000}%
\pgfsetstrokecolor{currentstroke}%
\pgfsetdash{}{0pt}%
\pgfpathmoveto{\pgfqpoint{2.235282in}{0.716028in}}%
\pgfpathlineto{\pgfqpoint{2.414377in}{0.716028in}}%
\pgfpathlineto{\pgfqpoint{2.414377in}{2.042199in}}%
\pgfpathlineto{\pgfqpoint{2.235282in}{2.042199in}}%
\pgfpathlineto{\pgfqpoint{2.235282in}{0.716028in}}%
\pgfpathclose%
\pgfusepath{fill}%
\end{pgfscope}%
\begin{pgfscope}%
\pgfpathrectangle{\pgfqpoint{0.802523in}{0.716028in}}{\pgfqpoint{5.193752in}{3.517843in}}%
\pgfusepath{clip}%
\pgfsetbuttcap%
\pgfsetmiterjoin%
\definecolor{currentfill}{rgb}{0.000000,0.000000,1.000000}%
\pgfsetfillcolor{currentfill}%
\pgfsetlinewidth{0.000000pt}%
\definecolor{currentstroke}{rgb}{0.000000,0.000000,0.000000}%
\pgfsetstrokecolor{currentstroke}%
\pgfsetdash{}{0pt}%
\pgfpathmoveto{\pgfqpoint{2.414377in}{0.716028in}}%
\pgfpathlineto{\pgfqpoint{2.593472in}{0.716028in}}%
\pgfpathlineto{\pgfqpoint{2.593472in}{3.345104in}}%
\pgfpathlineto{\pgfqpoint{2.414377in}{3.345104in}}%
\pgfpathlineto{\pgfqpoint{2.414377in}{0.716028in}}%
\pgfpathclose%
\pgfusepath{fill}%
\end{pgfscope}%
\begin{pgfscope}%
\pgfpathrectangle{\pgfqpoint{0.802523in}{0.716028in}}{\pgfqpoint{5.193752in}{3.517843in}}%
\pgfusepath{clip}%
\pgfsetbuttcap%
\pgfsetmiterjoin%
\definecolor{currentfill}{rgb}{0.000000,0.000000,1.000000}%
\pgfsetfillcolor{currentfill}%
\pgfsetlinewidth{0.000000pt}%
\definecolor{currentstroke}{rgb}{0.000000,0.000000,0.000000}%
\pgfsetstrokecolor{currentstroke}%
\pgfsetdash{}{0pt}%
\pgfpathmoveto{\pgfqpoint{2.593472in}{0.716028in}}%
\pgfpathlineto{\pgfqpoint{2.772567in}{0.716028in}}%
\pgfpathlineto{\pgfqpoint{2.772567in}{3.833693in}}%
\pgfpathlineto{\pgfqpoint{2.593472in}{3.833693in}}%
\pgfpathlineto{\pgfqpoint{2.593472in}{0.716028in}}%
\pgfpathclose%
\pgfusepath{fill}%
\end{pgfscope}%
\begin{pgfscope}%
\pgfpathrectangle{\pgfqpoint{0.802523in}{0.716028in}}{\pgfqpoint{5.193752in}{3.517843in}}%
\pgfusepath{clip}%
\pgfsetbuttcap%
\pgfsetmiterjoin%
\definecolor{currentfill}{rgb}{0.000000,0.000000,1.000000}%
\pgfsetfillcolor{currentfill}%
\pgfsetlinewidth{0.000000pt}%
\definecolor{currentstroke}{rgb}{0.000000,0.000000,0.000000}%
\pgfsetstrokecolor{currentstroke}%
\pgfsetdash{}{0pt}%
\pgfpathmoveto{\pgfqpoint{2.772567in}{0.716028in}}%
\pgfpathlineto{\pgfqpoint{2.951662in}{0.716028in}}%
\pgfpathlineto{\pgfqpoint{2.951662in}{3.577765in}}%
\pgfpathlineto{\pgfqpoint{2.772567in}{3.577765in}}%
\pgfpathlineto{\pgfqpoint{2.772567in}{0.716028in}}%
\pgfpathclose%
\pgfusepath{fill}%
\end{pgfscope}%
\begin{pgfscope}%
\pgfpathrectangle{\pgfqpoint{0.802523in}{0.716028in}}{\pgfqpoint{5.193752in}{3.517843in}}%
\pgfusepath{clip}%
\pgfsetbuttcap%
\pgfsetmiterjoin%
\definecolor{currentfill}{rgb}{0.000000,0.000000,1.000000}%
\pgfsetfillcolor{currentfill}%
\pgfsetlinewidth{0.000000pt}%
\definecolor{currentstroke}{rgb}{0.000000,0.000000,0.000000}%
\pgfsetstrokecolor{currentstroke}%
\pgfsetdash{}{0pt}%
\pgfpathmoveto{\pgfqpoint{2.951662in}{0.716028in}}%
\pgfpathlineto{\pgfqpoint{3.130756in}{0.716028in}}%
\pgfpathlineto{\pgfqpoint{3.130756in}{3.973290in}}%
\pgfpathlineto{\pgfqpoint{2.951662in}{3.973290in}}%
\pgfpathlineto{\pgfqpoint{2.951662in}{0.716028in}}%
\pgfpathclose%
\pgfusepath{fill}%
\end{pgfscope}%
\begin{pgfscope}%
\pgfpathrectangle{\pgfqpoint{0.802523in}{0.716028in}}{\pgfqpoint{5.193752in}{3.517843in}}%
\pgfusepath{clip}%
\pgfsetbuttcap%
\pgfsetmiterjoin%
\definecolor{currentfill}{rgb}{0.000000,0.000000,1.000000}%
\pgfsetfillcolor{currentfill}%
\pgfsetlinewidth{0.000000pt}%
\definecolor{currentstroke}{rgb}{0.000000,0.000000,0.000000}%
\pgfsetstrokecolor{currentstroke}%
\pgfsetdash{}{0pt}%
\pgfpathmoveto{\pgfqpoint{3.130756in}{0.716028in}}%
\pgfpathlineto{\pgfqpoint{3.309851in}{0.716028in}}%
\pgfpathlineto{\pgfqpoint{3.309851in}{2.298127in}}%
\pgfpathlineto{\pgfqpoint{3.130756in}{2.298127in}}%
\pgfpathlineto{\pgfqpoint{3.130756in}{0.716028in}}%
\pgfpathclose%
\pgfusepath{fill}%
\end{pgfscope}%
\begin{pgfscope}%
\pgfpathrectangle{\pgfqpoint{0.802523in}{0.716028in}}{\pgfqpoint{5.193752in}{3.517843in}}%
\pgfusepath{clip}%
\pgfsetbuttcap%
\pgfsetmiterjoin%
\definecolor{currentfill}{rgb}{0.000000,0.000000,1.000000}%
\pgfsetfillcolor{currentfill}%
\pgfsetlinewidth{0.000000pt}%
\definecolor{currentstroke}{rgb}{0.000000,0.000000,0.000000}%
\pgfsetstrokecolor{currentstroke}%
\pgfsetdash{}{0pt}%
\pgfpathmoveto{\pgfqpoint{3.309851in}{0.716028in}}%
\pgfpathlineto{\pgfqpoint{3.488946in}{0.716028in}}%
\pgfpathlineto{\pgfqpoint{3.488946in}{2.088731in}}%
\pgfpathlineto{\pgfqpoint{3.309851in}{2.088731in}}%
\pgfpathlineto{\pgfqpoint{3.309851in}{0.716028in}}%
\pgfpathclose%
\pgfusepath{fill}%
\end{pgfscope}%
\begin{pgfscope}%
\pgfpathrectangle{\pgfqpoint{0.802523in}{0.716028in}}{\pgfqpoint{5.193752in}{3.517843in}}%
\pgfusepath{clip}%
\pgfsetbuttcap%
\pgfsetmiterjoin%
\definecolor{currentfill}{rgb}{0.000000,0.000000,1.000000}%
\pgfsetfillcolor{currentfill}%
\pgfsetlinewidth{0.000000pt}%
\definecolor{currentstroke}{rgb}{0.000000,0.000000,0.000000}%
\pgfsetstrokecolor{currentstroke}%
\pgfsetdash{}{0pt}%
\pgfpathmoveto{\pgfqpoint{3.488946in}{0.716028in}}%
\pgfpathlineto{\pgfqpoint{3.668041in}{0.716028in}}%
\pgfpathlineto{\pgfqpoint{3.668041in}{2.321393in}}%
\pgfpathlineto{\pgfqpoint{3.488946in}{2.321393in}}%
\pgfpathlineto{\pgfqpoint{3.488946in}{0.716028in}}%
\pgfpathclose%
\pgfusepath{fill}%
\end{pgfscope}%
\begin{pgfscope}%
\pgfpathrectangle{\pgfqpoint{0.802523in}{0.716028in}}{\pgfqpoint{5.193752in}{3.517843in}}%
\pgfusepath{clip}%
\pgfsetbuttcap%
\pgfsetmiterjoin%
\definecolor{currentfill}{rgb}{0.000000,0.000000,1.000000}%
\pgfsetfillcolor{currentfill}%
\pgfsetlinewidth{0.000000pt}%
\definecolor{currentstroke}{rgb}{0.000000,0.000000,0.000000}%
\pgfsetstrokecolor{currentstroke}%
\pgfsetdash{}{0pt}%
\pgfpathmoveto{\pgfqpoint{3.668041in}{0.716028in}}%
\pgfpathlineto{\pgfqpoint{3.847136in}{0.716028in}}%
\pgfpathlineto{\pgfqpoint{3.847136in}{2.158530in}}%
\pgfpathlineto{\pgfqpoint{3.668041in}{2.158530in}}%
\pgfpathlineto{\pgfqpoint{3.668041in}{0.716028in}}%
\pgfpathclose%
\pgfusepath{fill}%
\end{pgfscope}%
\begin{pgfscope}%
\pgfpathrectangle{\pgfqpoint{0.802523in}{0.716028in}}{\pgfqpoint{5.193752in}{3.517843in}}%
\pgfusepath{clip}%
\pgfsetbuttcap%
\pgfsetmiterjoin%
\definecolor{currentfill}{rgb}{0.000000,0.000000,1.000000}%
\pgfsetfillcolor{currentfill}%
\pgfsetlinewidth{0.000000pt}%
\definecolor{currentstroke}{rgb}{0.000000,0.000000,0.000000}%
\pgfsetstrokecolor{currentstroke}%
\pgfsetdash{}{0pt}%
\pgfpathmoveto{\pgfqpoint{3.847136in}{0.716028in}}%
\pgfpathlineto{\pgfqpoint{4.026231in}{0.716028in}}%
\pgfpathlineto{\pgfqpoint{4.026231in}{1.856070in}}%
\pgfpathlineto{\pgfqpoint{3.847136in}{1.856070in}}%
\pgfpathlineto{\pgfqpoint{3.847136in}{0.716028in}}%
\pgfpathclose%
\pgfusepath{fill}%
\end{pgfscope}%
\begin{pgfscope}%
\pgfpathrectangle{\pgfqpoint{0.802523in}{0.716028in}}{\pgfqpoint{5.193752in}{3.517843in}}%
\pgfusepath{clip}%
\pgfsetbuttcap%
\pgfsetmiterjoin%
\definecolor{currentfill}{rgb}{0.000000,0.000000,1.000000}%
\pgfsetfillcolor{currentfill}%
\pgfsetlinewidth{0.000000pt}%
\definecolor{currentstroke}{rgb}{0.000000,0.000000,0.000000}%
\pgfsetstrokecolor{currentstroke}%
\pgfsetdash{}{0pt}%
\pgfpathmoveto{\pgfqpoint{4.026231in}{0.716028in}}%
\pgfpathlineto{\pgfqpoint{4.205326in}{0.716028in}}%
\pgfpathlineto{\pgfqpoint{4.205326in}{2.065465in}}%
\pgfpathlineto{\pgfqpoint{4.026231in}{2.065465in}}%
\pgfpathlineto{\pgfqpoint{4.026231in}{0.716028in}}%
\pgfpathclose%
\pgfusepath{fill}%
\end{pgfscope}%
\begin{pgfscope}%
\pgfpathrectangle{\pgfqpoint{0.802523in}{0.716028in}}{\pgfqpoint{5.193752in}{3.517843in}}%
\pgfusepath{clip}%
\pgfsetbuttcap%
\pgfsetmiterjoin%
\definecolor{currentfill}{rgb}{0.000000,0.000000,1.000000}%
\pgfsetfillcolor{currentfill}%
\pgfsetlinewidth{0.000000pt}%
\definecolor{currentstroke}{rgb}{0.000000,0.000000,0.000000}%
\pgfsetstrokecolor{currentstroke}%
\pgfsetdash{}{0pt}%
\pgfpathmoveto{\pgfqpoint{4.205326in}{0.716028in}}%
\pgfpathlineto{\pgfqpoint{4.384421in}{0.716028in}}%
\pgfpathlineto{\pgfqpoint{4.384421in}{2.856515in}}%
\pgfpathlineto{\pgfqpoint{4.205326in}{2.856515in}}%
\pgfpathlineto{\pgfqpoint{4.205326in}{0.716028in}}%
\pgfpathclose%
\pgfusepath{fill}%
\end{pgfscope}%
\begin{pgfscope}%
\pgfpathrectangle{\pgfqpoint{0.802523in}{0.716028in}}{\pgfqpoint{5.193752in}{3.517843in}}%
\pgfusepath{clip}%
\pgfsetbuttcap%
\pgfsetmiterjoin%
\definecolor{currentfill}{rgb}{0.000000,0.000000,1.000000}%
\pgfsetfillcolor{currentfill}%
\pgfsetlinewidth{0.000000pt}%
\definecolor{currentstroke}{rgb}{0.000000,0.000000,0.000000}%
\pgfsetstrokecolor{currentstroke}%
\pgfsetdash{}{0pt}%
\pgfpathmoveto{\pgfqpoint{4.384421in}{0.716028in}}%
\pgfpathlineto{\pgfqpoint{4.563516in}{0.716028in}}%
\pgfpathlineto{\pgfqpoint{4.563516in}{3.112442in}}%
\pgfpathlineto{\pgfqpoint{4.384421in}{3.112442in}}%
\pgfpathlineto{\pgfqpoint{4.384421in}{0.716028in}}%
\pgfpathclose%
\pgfusepath{fill}%
\end{pgfscope}%
\begin{pgfscope}%
\pgfpathrectangle{\pgfqpoint{0.802523in}{0.716028in}}{\pgfqpoint{5.193752in}{3.517843in}}%
\pgfusepath{clip}%
\pgfsetbuttcap%
\pgfsetmiterjoin%
\definecolor{currentfill}{rgb}{0.000000,0.000000,1.000000}%
\pgfsetfillcolor{currentfill}%
\pgfsetlinewidth{0.000000pt}%
\definecolor{currentstroke}{rgb}{0.000000,0.000000,0.000000}%
\pgfsetstrokecolor{currentstroke}%
\pgfsetdash{}{0pt}%
\pgfpathmoveto{\pgfqpoint{4.563516in}{0.716028in}}%
\pgfpathlineto{\pgfqpoint{4.742611in}{0.716028in}}%
\pgfpathlineto{\pgfqpoint{4.742611in}{4.066355in}}%
\pgfpathlineto{\pgfqpoint{4.563516in}{4.066355in}}%
\pgfpathlineto{\pgfqpoint{4.563516in}{0.716028in}}%
\pgfpathclose%
\pgfusepath{fill}%
\end{pgfscope}%
\begin{pgfscope}%
\pgfpathrectangle{\pgfqpoint{0.802523in}{0.716028in}}{\pgfqpoint{5.193752in}{3.517843in}}%
\pgfusepath{clip}%
\pgfsetbuttcap%
\pgfsetmiterjoin%
\definecolor{currentfill}{rgb}{0.000000,0.000000,1.000000}%
\pgfsetfillcolor{currentfill}%
\pgfsetlinewidth{0.000000pt}%
\definecolor{currentstroke}{rgb}{0.000000,0.000000,0.000000}%
\pgfsetstrokecolor{currentstroke}%
\pgfsetdash{}{0pt}%
\pgfpathmoveto{\pgfqpoint{4.742611in}{0.716028in}}%
\pgfpathlineto{\pgfqpoint{4.921705in}{0.716028in}}%
\pgfpathlineto{\pgfqpoint{4.921705in}{2.693652in}}%
\pgfpathlineto{\pgfqpoint{4.742611in}{2.693652in}}%
\pgfpathlineto{\pgfqpoint{4.742611in}{0.716028in}}%
\pgfpathclose%
\pgfusepath{fill}%
\end{pgfscope}%
\begin{pgfscope}%
\pgfpathrectangle{\pgfqpoint{0.802523in}{0.716028in}}{\pgfqpoint{5.193752in}{3.517843in}}%
\pgfusepath{clip}%
\pgfsetbuttcap%
\pgfsetmiterjoin%
\definecolor{currentfill}{rgb}{0.000000,0.000000,1.000000}%
\pgfsetfillcolor{currentfill}%
\pgfsetlinewidth{0.000000pt}%
\definecolor{currentstroke}{rgb}{0.000000,0.000000,0.000000}%
\pgfsetstrokecolor{currentstroke}%
\pgfsetdash{}{0pt}%
\pgfpathmoveto{\pgfqpoint{4.921705in}{0.716028in}}%
\pgfpathlineto{\pgfqpoint{5.100800in}{0.716028in}}%
\pgfpathlineto{\pgfqpoint{5.100800in}{2.391191in}}%
\pgfpathlineto{\pgfqpoint{4.921705in}{2.391191in}}%
\pgfpathlineto{\pgfqpoint{4.921705in}{0.716028in}}%
\pgfpathclose%
\pgfusepath{fill}%
\end{pgfscope}%
\begin{pgfscope}%
\pgfpathrectangle{\pgfqpoint{0.802523in}{0.716028in}}{\pgfqpoint{5.193752in}{3.517843in}}%
\pgfusepath{clip}%
\pgfsetbuttcap%
\pgfsetmiterjoin%
\definecolor{currentfill}{rgb}{0.000000,0.000000,1.000000}%
\pgfsetfillcolor{currentfill}%
\pgfsetlinewidth{0.000000pt}%
\definecolor{currentstroke}{rgb}{0.000000,0.000000,0.000000}%
\pgfsetstrokecolor{currentstroke}%
\pgfsetdash{}{0pt}%
\pgfpathmoveto{\pgfqpoint{5.100800in}{0.716028in}}%
\pgfpathlineto{\pgfqpoint{5.279895in}{0.716028in}}%
\pgfpathlineto{\pgfqpoint{5.279895in}{2.367925in}}%
\pgfpathlineto{\pgfqpoint{5.100800in}{2.367925in}}%
\pgfpathlineto{\pgfqpoint{5.100800in}{0.716028in}}%
\pgfpathclose%
\pgfusepath{fill}%
\end{pgfscope}%
\begin{pgfscope}%
\pgfpathrectangle{\pgfqpoint{0.802523in}{0.716028in}}{\pgfqpoint{5.193752in}{3.517843in}}%
\pgfusepath{clip}%
\pgfsetbuttcap%
\pgfsetmiterjoin%
\definecolor{currentfill}{rgb}{0.000000,0.000000,1.000000}%
\pgfsetfillcolor{currentfill}%
\pgfsetlinewidth{0.000000pt}%
\definecolor{currentstroke}{rgb}{0.000000,0.000000,0.000000}%
\pgfsetstrokecolor{currentstroke}%
\pgfsetdash{}{0pt}%
\pgfpathmoveto{\pgfqpoint{5.279895in}{0.716028in}}%
\pgfpathlineto{\pgfqpoint{5.458990in}{0.716028in}}%
\pgfpathlineto{\pgfqpoint{5.458990in}{2.577321in}}%
\pgfpathlineto{\pgfqpoint{5.279895in}{2.577321in}}%
\pgfpathlineto{\pgfqpoint{5.279895in}{0.716028in}}%
\pgfpathclose%
\pgfusepath{fill}%
\end{pgfscope}%
\begin{pgfscope}%
\pgfpathrectangle{\pgfqpoint{0.802523in}{0.716028in}}{\pgfqpoint{5.193752in}{3.517843in}}%
\pgfusepath{clip}%
\pgfsetbuttcap%
\pgfsetmiterjoin%
\definecolor{currentfill}{rgb}{0.000000,0.000000,1.000000}%
\pgfsetfillcolor{currentfill}%
\pgfsetlinewidth{0.000000pt}%
\definecolor{currentstroke}{rgb}{0.000000,0.000000,0.000000}%
\pgfsetstrokecolor{currentstroke}%
\pgfsetdash{}{0pt}%
\pgfpathmoveto{\pgfqpoint{5.458990in}{0.716028in}}%
\pgfpathlineto{\pgfqpoint{5.638085in}{0.716028in}}%
\pgfpathlineto{\pgfqpoint{5.638085in}{2.903047in}}%
\pgfpathlineto{\pgfqpoint{5.458990in}{2.903047in}}%
\pgfpathlineto{\pgfqpoint{5.458990in}{0.716028in}}%
\pgfpathclose%
\pgfusepath{fill}%
\end{pgfscope}%
\begin{pgfscope}%
\pgfpathrectangle{\pgfqpoint{0.802523in}{0.716028in}}{\pgfqpoint{5.193752in}{3.517843in}}%
\pgfusepath{clip}%
\pgfsetbuttcap%
\pgfsetmiterjoin%
\definecolor{currentfill}{rgb}{0.000000,0.000000,1.000000}%
\pgfsetfillcolor{currentfill}%
\pgfsetlinewidth{0.000000pt}%
\definecolor{currentstroke}{rgb}{0.000000,0.000000,0.000000}%
\pgfsetstrokecolor{currentstroke}%
\pgfsetdash{}{0pt}%
\pgfpathmoveto{\pgfqpoint{5.638085in}{0.716028in}}%
\pgfpathlineto{\pgfqpoint{5.817180in}{0.716028in}}%
\pgfpathlineto{\pgfqpoint{5.817180in}{3.135708in}}%
\pgfpathlineto{\pgfqpoint{5.638085in}{3.135708in}}%
\pgfpathlineto{\pgfqpoint{5.638085in}{0.716028in}}%
\pgfpathclose%
\pgfusepath{fill}%
\end{pgfscope}%
\begin{pgfscope}%
\pgfpathrectangle{\pgfqpoint{0.802523in}{0.716028in}}{\pgfqpoint{5.193752in}{3.517843in}}%
\pgfusepath{clip}%
\pgfsetbuttcap%
\pgfsetmiterjoin%
\definecolor{currentfill}{rgb}{0.000000,0.000000,1.000000}%
\pgfsetfillcolor{currentfill}%
\pgfsetlinewidth{0.000000pt}%
\definecolor{currentstroke}{rgb}{0.000000,0.000000,0.000000}%
\pgfsetstrokecolor{currentstroke}%
\pgfsetdash{}{0pt}%
\pgfpathmoveto{\pgfqpoint{5.817180in}{0.716028in}}%
\pgfpathlineto{\pgfqpoint{5.996275in}{0.716028in}}%
\pgfpathlineto{\pgfqpoint{5.996275in}{2.135264in}}%
\pgfpathlineto{\pgfqpoint{5.817180in}{2.135264in}}%
\pgfpathlineto{\pgfqpoint{5.817180in}{0.716028in}}%
\pgfpathclose%
\pgfusepath{fill}%
\end{pgfscope}%
\begin{pgfscope}%
\pgfsetbuttcap%
\pgfsetroundjoin%
\definecolor{currentfill}{rgb}{0.000000,0.000000,0.000000}%
\pgfsetfillcolor{currentfill}%
\pgfsetlinewidth{0.200750pt}%
\definecolor{currentstroke}{rgb}{0.000000,0.000000,0.000000}%
\pgfsetstrokecolor{currentstroke}%
\pgfsetdash{}{0pt}%
\pgfsys@defobject{currentmarker}{\pgfqpoint{0.000000in}{-0.048611in}}{\pgfqpoint{0.000000in}{0.000000in}}{%
\pgfpathmoveto{\pgfqpoint{0.000000in}{0.000000in}}%
\pgfpathlineto{\pgfqpoint{0.000000in}{-0.048611in}}%
\pgfusepath{stroke,fill}%
}%
\begin{pgfscope}%
\pgfsys@transformshift{0.802523in}{0.716028in}%
\pgfsys@useobject{currentmarker}{}%
\end{pgfscope}%
\end{pgfscope}%
\begin{pgfscope}%
\definecolor{textcolor}{rgb}{0.000000,0.000000,0.000000}%
\pgfsetstrokecolor{textcolor}%
\pgfsetfillcolor{textcolor}%
\pgftext[x=0.802523in,y=0.618806in,,top]{\color{textcolor}{\rmfamily\fontsize{20.000000}{24.000000}\selectfont\catcode`\^=\active\def^{\ifmmode\sp\else\^{}\fi}\catcode`\%=\active\def
\end{pgfscope}%
\begin{pgfscope}%
\pgfsetbuttcap%
\pgfsetroundjoin%
\definecolor{currentfill}{rgb}{0.000000,0.000000,0.000000}%
\pgfsetfillcolor{currentfill}%
\pgfsetlinewidth{0.200750pt}%
\definecolor{currentstroke}{rgb}{0.000000,0.000000,0.000000}%
\pgfsetstrokecolor{currentstroke}%
\pgfsetdash{}{0pt}%
\pgfsys@defobject{currentmarker}{\pgfqpoint{0.000000in}{-0.048611in}}{\pgfqpoint{0.000000in}{0.000000in}}{%
\pgfpathmoveto{\pgfqpoint{0.000000in}{0.000000in}}%
\pgfpathlineto{\pgfqpoint{0.000000in}{-0.048611in}}%
\pgfusepath{stroke,fill}%
}%
\begin{pgfscope}%
\pgfsys@transformshift{1.841273in}{0.716028in}%
\pgfsys@useobject{currentmarker}{}%
\end{pgfscope}%
\end{pgfscope}%
\begin{pgfscope}%
\definecolor{textcolor}{rgb}{0.000000,0.000000,0.000000}%
\pgfsetstrokecolor{textcolor}%
\pgfsetfillcolor{textcolor}%
\pgftext[x=1.841273in,y=0.618806in,,top]{\color{textcolor}{\rmfamily\fontsize{20.000000}{24.000000}\selectfont\catcode`\^=\active\def^{\ifmmode\sp\else\^{}\fi}\catcode`\%=\active\def
\end{pgfscope}%
\begin{pgfscope}%
\pgfsetbuttcap%
\pgfsetroundjoin%
\definecolor{currentfill}{rgb}{0.000000,0.000000,0.000000}%
\pgfsetfillcolor{currentfill}%
\pgfsetlinewidth{0.200750pt}%
\definecolor{currentstroke}{rgb}{0.000000,0.000000,0.000000}%
\pgfsetstrokecolor{currentstroke}%
\pgfsetdash{}{0pt}%
\pgfsys@defobject{currentmarker}{\pgfqpoint{0.000000in}{-0.048611in}}{\pgfqpoint{0.000000in}{0.000000in}}{%
\pgfpathmoveto{\pgfqpoint{0.000000in}{0.000000in}}%
\pgfpathlineto{\pgfqpoint{0.000000in}{-0.048611in}}%
\pgfusepath{stroke,fill}%
}%
\begin{pgfscope}%
\pgfsys@transformshift{2.880024in}{0.716028in}%
\pgfsys@useobject{currentmarker}{}%
\end{pgfscope}%
\end{pgfscope}%
\begin{pgfscope}%
\definecolor{textcolor}{rgb}{0.000000,0.000000,0.000000}%
\pgfsetstrokecolor{textcolor}%
\pgfsetfillcolor{textcolor}%
\pgftext[x=2.880024in,y=0.618806in,,top]{\color{textcolor}{\rmfamily\fontsize{20.000000}{24.000000}\selectfont\catcode`\^=\active\def^{\ifmmode\sp\else\^{}\fi}\catcode`\%=\active\def
\end{pgfscope}%
\begin{pgfscope}%
\pgfsetbuttcap%
\pgfsetroundjoin%
\definecolor{currentfill}{rgb}{0.000000,0.000000,0.000000}%
\pgfsetfillcolor{currentfill}%
\pgfsetlinewidth{0.200750pt}%
\definecolor{currentstroke}{rgb}{0.000000,0.000000,0.000000}%
\pgfsetstrokecolor{currentstroke}%
\pgfsetdash{}{0pt}%
\pgfsys@defobject{currentmarker}{\pgfqpoint{0.000000in}{-0.048611in}}{\pgfqpoint{0.000000in}{0.000000in}}{%
\pgfpathmoveto{\pgfqpoint{0.000000in}{0.000000in}}%
\pgfpathlineto{\pgfqpoint{0.000000in}{-0.048611in}}%
\pgfusepath{stroke,fill}%
}%
\begin{pgfscope}%
\pgfsys@transformshift{3.918774in}{0.716028in}%
\pgfsys@useobject{currentmarker}{}%
\end{pgfscope}%
\end{pgfscope}%
\begin{pgfscope}%
\definecolor{textcolor}{rgb}{0.000000,0.000000,0.000000}%
\pgfsetstrokecolor{textcolor}%
\pgfsetfillcolor{textcolor}%
\pgftext[x=3.918774in,y=0.618806in,,top]{\color{textcolor}{\rmfamily\fontsize{20.000000}{24.000000}\selectfont\catcode`\^=\active\def^{\ifmmode\sp\else\^{}\fi}\catcode`\%=\active\def
\end{pgfscope}%
\begin{pgfscope}%
\pgfsetbuttcap%
\pgfsetroundjoin%
\definecolor{currentfill}{rgb}{0.000000,0.000000,0.000000}%
\pgfsetfillcolor{currentfill}%
\pgfsetlinewidth{0.200750pt}%
\definecolor{currentstroke}{rgb}{0.000000,0.000000,0.000000}%
\pgfsetstrokecolor{currentstroke}%
\pgfsetdash{}{0pt}%
\pgfsys@defobject{currentmarker}{\pgfqpoint{0.000000in}{-0.048611in}}{\pgfqpoint{0.000000in}{0.000000in}}{%
\pgfpathmoveto{\pgfqpoint{0.000000in}{0.000000in}}%
\pgfpathlineto{\pgfqpoint{0.000000in}{-0.048611in}}%
\pgfusepath{stroke,fill}%
}%
\begin{pgfscope}%
\pgfsys@transformshift{4.957524in}{0.716028in}%
\pgfsys@useobject{currentmarker}{}%
\end{pgfscope}%
\end{pgfscope}%
\begin{pgfscope}%
\definecolor{textcolor}{rgb}{0.000000,0.000000,0.000000}%
\pgfsetstrokecolor{textcolor}%
\pgfsetfillcolor{textcolor}%
\pgftext[x=4.957524in,y=0.618806in,,top]{\color{textcolor}{\rmfamily\fontsize{20.000000}{24.000000}\selectfont\catcode`\^=\active\def^{\ifmmode\sp\else\^{}\fi}\catcode`\%=\active\def
\end{pgfscope}%
\begin{pgfscope}%
\pgfsetbuttcap%
\pgfsetroundjoin%
\definecolor{currentfill}{rgb}{0.000000,0.000000,0.000000}%
\pgfsetfillcolor{currentfill}%
\pgfsetlinewidth{0.200750pt}%
\definecolor{currentstroke}{rgb}{0.000000,0.000000,0.000000}%
\pgfsetstrokecolor{currentstroke}%
\pgfsetdash{}{0pt}%
\pgfsys@defobject{currentmarker}{\pgfqpoint{0.000000in}{-0.048611in}}{\pgfqpoint{0.000000in}{0.000000in}}{%
\pgfpathmoveto{\pgfqpoint{0.000000in}{0.000000in}}%
\pgfpathlineto{\pgfqpoint{0.000000in}{-0.048611in}}%
\pgfusepath{stroke,fill}%
}%
\begin{pgfscope}%
\pgfsys@transformshift{5.996275in}{0.716028in}%
\pgfsys@useobject{currentmarker}{}%
\end{pgfscope}%
\end{pgfscope}%
\begin{pgfscope}%
\definecolor{textcolor}{rgb}{0.000000,0.000000,0.000000}%
\pgfsetstrokecolor{textcolor}%
\pgfsetfillcolor{textcolor}%
\pgftext[x=5.996275in,y=0.618806in,,top]{\color{textcolor}{\rmfamily\fontsize{20.000000}{24.000000}\selectfont\catcode`\^=\active\def^{\ifmmode\sp\else\^{}\fi}\catcode`\%=\active\def
\end{pgfscope}%
\begin{pgfscope}%
\definecolor{textcolor}{rgb}{0.000000,0.000000,0.000000}%
\pgfsetstrokecolor{textcolor}%
\pgfsetfillcolor{textcolor}%
\pgftext[x=3.399399in,y=0.307183in,,top]{\color{textcolor}{\rmfamily\fontsize{26.000000}{31.200000}\selectfont\catcode`\^=\active\def^{\ifmmode\sp\else\^{}\fi}\catcode`\%=\active\def
\end{pgfscope}%
\begin{pgfscope}%
\pgfsetbuttcap%
\pgfsetroundjoin%
\definecolor{currentfill}{rgb}{0.000000,0.000000,0.000000}%
\pgfsetfillcolor{currentfill}%
\pgfsetlinewidth{0.200750pt}%
\definecolor{currentstroke}{rgb}{0.000000,0.000000,0.000000}%
\pgfsetstrokecolor{currentstroke}%
\pgfsetdash{}{0pt}%
\pgfsys@defobject{currentmarker}{\pgfqpoint{-0.048611in}{0.000000in}}{\pgfqpoint{-0.000000in}{0.000000in}}{%
\pgfpathmoveto{\pgfqpoint{-0.000000in}{0.000000in}}%
\pgfpathlineto{\pgfqpoint{-0.048611in}{0.000000in}}%
\pgfusepath{stroke,fill}%
}%
\begin{pgfscope}%
\pgfsys@transformshift{0.802523in}{0.716028in}%
\pgfsys@useobject{currentmarker}{}%
\end{pgfscope}%
\end{pgfscope}%
\begin{pgfscope}%
\definecolor{textcolor}{rgb}{0.000000,0.000000,0.000000}%
\pgfsetstrokecolor{textcolor}%
\pgfsetfillcolor{textcolor}%
\pgftext[x=0.362738in, y=0.616009in, left, base]{\color{textcolor}{\rmfamily\fontsize{20.000000}{24.000000}\selectfont\catcode`\^=\active\def^{\ifmmode\sp\else\^{}\fi}\catcode`\%=\active\def
\end{pgfscope}%
\begin{pgfscope}%
\pgfsetbuttcap%
\pgfsetroundjoin%
\definecolor{currentfill}{rgb}{0.000000,0.000000,0.000000}%
\pgfsetfillcolor{currentfill}%
\pgfsetlinewidth{0.200750pt}%
\definecolor{currentstroke}{rgb}{0.000000,0.000000,0.000000}%
\pgfsetstrokecolor{currentstroke}%
\pgfsetdash{}{0pt}%
\pgfsys@defobject{currentmarker}{\pgfqpoint{-0.048611in}{0.000000in}}{\pgfqpoint{-0.000000in}{0.000000in}}{%
\pgfpathmoveto{\pgfqpoint{-0.000000in}{0.000000in}}%
\pgfpathlineto{\pgfqpoint{-0.048611in}{0.000000in}}%
\pgfusepath{stroke,fill}%
}%
\begin{pgfscope}%
\pgfsys@transformshift{0.802523in}{1.496648in}%
\pgfsys@useobject{currentmarker}{}%
\end{pgfscope}%
\end{pgfscope}%
\begin{pgfscope}%
\definecolor{textcolor}{rgb}{0.000000,0.000000,0.000000}%
\pgfsetstrokecolor{textcolor}%
\pgfsetfillcolor{textcolor}%
\pgftext[x=0.362738in, y=1.396629in, left, base]{\color{textcolor}{\rmfamily\fontsize{20.000000}{24.000000}\selectfont\catcode`\^=\active\def^{\ifmmode\sp\else\^{}\fi}\catcode`\%=\active\def
\end{pgfscope}%
\begin{pgfscope}%
\pgfsetbuttcap%
\pgfsetroundjoin%
\definecolor{currentfill}{rgb}{0.000000,0.000000,0.000000}%
\pgfsetfillcolor{currentfill}%
\pgfsetlinewidth{0.200750pt}%
\definecolor{currentstroke}{rgb}{0.000000,0.000000,0.000000}%
\pgfsetstrokecolor{currentstroke}%
\pgfsetdash{}{0pt}%
\pgfsys@defobject{currentmarker}{\pgfqpoint{-0.048611in}{0.000000in}}{\pgfqpoint{-0.000000in}{0.000000in}}{%
\pgfpathmoveto{\pgfqpoint{-0.000000in}{0.000000in}}%
\pgfpathlineto{\pgfqpoint{-0.048611in}{0.000000in}}%
\pgfusepath{stroke,fill}%
}%
\begin{pgfscope}%
\pgfsys@transformshift{0.802523in}{2.277268in}%
\pgfsys@useobject{currentmarker}{}%
\end{pgfscope}%
\end{pgfscope}%
\begin{pgfscope}%
\definecolor{textcolor}{rgb}{0.000000,0.000000,0.000000}%
\pgfsetstrokecolor{textcolor}%
\pgfsetfillcolor{textcolor}%
\pgftext[x=0.362738in, y=2.177248in, left, base]{\color{textcolor}{\rmfamily\fontsize{20.000000}{24.000000}\selectfont\catcode`\^=\active\def^{\ifmmode\sp\else\^{}\fi}\catcode`\%=\active\def
\end{pgfscope}%
\begin{pgfscope}%
\pgfsetbuttcap%
\pgfsetroundjoin%
\definecolor{currentfill}{rgb}{0.000000,0.000000,0.000000}%
\pgfsetfillcolor{currentfill}%
\pgfsetlinewidth{0.200750pt}%
\definecolor{currentstroke}{rgb}{0.000000,0.000000,0.000000}%
\pgfsetstrokecolor{currentstroke}%
\pgfsetdash{}{0pt}%
\pgfsys@defobject{currentmarker}{\pgfqpoint{-0.048611in}{0.000000in}}{\pgfqpoint{-0.000000in}{0.000000in}}{%
\pgfpathmoveto{\pgfqpoint{-0.000000in}{0.000000in}}%
\pgfpathlineto{\pgfqpoint{-0.048611in}{0.000000in}}%
\pgfusepath{stroke,fill}%
}%
\begin{pgfscope}%
\pgfsys@transformshift{0.802523in}{3.057887in}%
\pgfsys@useobject{currentmarker}{}%
\end{pgfscope}%
\end{pgfscope}%
\begin{pgfscope}%
\definecolor{textcolor}{rgb}{0.000000,0.000000,0.000000}%
\pgfsetstrokecolor{textcolor}%
\pgfsetfillcolor{textcolor}%
\pgftext[x=0.362738in, y=2.957868in, left, base]{\color{textcolor}{\rmfamily\fontsize{20.000000}{24.000000}\selectfont\catcode`\^=\active\def^{\ifmmode\sp\else\^{}\fi}\catcode`\%=\active\def
\end{pgfscope}%
\begin{pgfscope}%
\pgfsetbuttcap%
\pgfsetroundjoin%
\definecolor{currentfill}{rgb}{0.000000,0.000000,0.000000}%
\pgfsetfillcolor{currentfill}%
\pgfsetlinewidth{0.200750pt}%
\definecolor{currentstroke}{rgb}{0.000000,0.000000,0.000000}%
\pgfsetstrokecolor{currentstroke}%
\pgfsetdash{}{0pt}%
\pgfsys@defobject{currentmarker}{\pgfqpoint{-0.048611in}{0.000000in}}{\pgfqpoint{-0.000000in}{0.000000in}}{%
\pgfpathmoveto{\pgfqpoint{-0.000000in}{0.000000in}}%
\pgfpathlineto{\pgfqpoint{-0.048611in}{0.000000in}}%
\pgfusepath{stroke,fill}%
}%
\begin{pgfscope}%
\pgfsys@transformshift{0.802523in}{3.838507in}%
\pgfsys@useobject{currentmarker}{}%
\end{pgfscope}%
\end{pgfscope}%
\begin{pgfscope}%
\definecolor{textcolor}{rgb}{0.000000,0.000000,0.000000}%
\pgfsetstrokecolor{textcolor}%
\pgfsetfillcolor{textcolor}%
\pgftext[x=0.362738in, y=3.738488in, left, base]{\color{textcolor}{\rmfamily\fontsize{20.000000}{24.000000}\selectfont\catcode`\^=\active\def^{\ifmmode\sp\else\^{}\fi}\catcode`\%=\active\def
\end{pgfscope}%
\begin{pgfscope}%
\definecolor{textcolor}{rgb}{0.000000,0.000000,0.000000}%
\pgfsetstrokecolor{textcolor}%
\pgfsetfillcolor{textcolor}%
\pgftext[x=0.307183in,y=2.474950in,,bottom,rotate=90.000000]{\color{textcolor}{\rmfamily\fontsize{26.000000}{31.200000}\selectfont\catcode`\^=\active\def^{\ifmmode\sp\else\^{}\fi}\catcode`\%=\active\def
\end{pgfscope}%
\begin{pgfscope}%
\pgfpathrectangle{\pgfqpoint{0.802523in}{0.716028in}}{\pgfqpoint{5.193752in}{3.517843in}}%
\pgfusepath{clip}%
\pgfsetbuttcap%
\pgfsetroundjoin%
\pgfsetlinewidth{2.007500pt}%
\definecolor{currentstroke}{rgb}{0.501961,0.501961,0.501961}%
\pgfsetstrokecolor{currentstroke}%
\pgfsetdash{{7.400000pt}{3.200000pt}}{0.000000pt}%
\pgfpathmoveto{\pgfqpoint{0.802523in}{2.277268in}}%
\pgfpathlineto{\pgfqpoint{5.996275in}{2.277268in}}%
\pgfusepath{stroke}%
\end{pgfscope}%
\begin{pgfscope}%
\pgfsetrectcap%
\pgfsetmiterjoin%
\pgfsetlinewidth{0.803000pt}%
\definecolor{currentstroke}{rgb}{0.000000,0.000000,0.000000}%
\pgfsetstrokecolor{currentstroke}%
\pgfsetdash{}{0pt}%
\pgfpathmoveto{\pgfqpoint{0.802523in}{0.716028in}}%
\pgfpathlineto{\pgfqpoint{0.802523in}{4.233871in}}%
\pgfusepath{stroke}%
\end{pgfscope}%
\begin{pgfscope}%
\pgfsetrectcap%
\pgfsetmiterjoin%
\pgfsetlinewidth{0.803000pt}%
\definecolor{currentstroke}{rgb}{0.000000,0.000000,0.000000}%
\pgfsetstrokecolor{currentstroke}%
\pgfsetdash{}{0pt}%
\pgfpathmoveto{\pgfqpoint{5.996275in}{0.716028in}}%
\pgfpathlineto{\pgfqpoint{5.996275in}{4.233871in}}%
\pgfusepath{stroke}%
\end{pgfscope}%
\begin{pgfscope}%
\pgfsetrectcap%
\pgfsetmiterjoin%
\pgfsetlinewidth{0.803000pt}%
\definecolor{currentstroke}{rgb}{0.000000,0.000000,0.000000}%
\pgfsetstrokecolor{currentstroke}%
\pgfsetdash{}{0pt}%
\pgfpathmoveto{\pgfqpoint{0.802523in}{0.716028in}}%
\pgfpathlineto{\pgfqpoint{5.996275in}{0.716028in}}%
\pgfusepath{stroke}%
\end{pgfscope}%
\begin{pgfscope}%
\pgfsetrectcap%
\pgfsetmiterjoin%
\pgfsetlinewidth{0.803000pt}%
\definecolor{currentstroke}{rgb}{0.000000,0.000000,0.000000}%
\pgfsetstrokecolor{currentstroke}%
\pgfsetdash{}{0pt}%
\pgfpathmoveto{\pgfqpoint{0.802523in}{4.233871in}}%
\pgfpathlineto{\pgfqpoint{5.996275in}{4.233871in}}%
\pgfusepath{stroke}%
\end{pgfscope}%
\begin{pgfscope}%
\definecolor{textcolor}{rgb}{0.000000,0.000000,0.000000}%
\pgfsetstrokecolor{textcolor}%
\pgfsetfillcolor{textcolor}%
\pgftext[x=3.399399in,y=4.317204in,,base]{\color{textcolor}{\rmfamily\fontsize{25.000000}{30.000000}\selectfont\catcode`\^=\active\def^{\ifmmode\sp\else\^{}\fi}\catcode`\%=\active\def
\end{pgfscope}%
\begin{pgfscope}%
\pgfsetbuttcap%
\pgfsetmiterjoin%
\definecolor{currentfill}{rgb}{1.000000,1.000000,1.000000}%
\pgfsetfillcolor{currentfill}%
\pgfsetfillopacity{0.800000}%
\pgfsetlinewidth{1.003750pt}%
\definecolor{currentstroke}{rgb}{0.800000,0.800000,0.800000}%
\pgfsetstrokecolor{currentstroke}%
\pgfsetstrokeopacity{0.800000}%
\pgfsetdash{}{0pt}%
\pgfpathmoveto{\pgfqpoint{4.259928in}{3.616692in}}%
\pgfpathlineto{\pgfqpoint{5.801830in}{3.616692in}}%
\pgfpathquadraticcurveto{\pgfqpoint{5.857386in}{3.616692in}}{\pgfqpoint{5.857386in}{3.672248in}}%
\pgfpathlineto{\pgfqpoint{5.857386in}{4.039427in}}%
\pgfpathquadraticcurveto{\pgfqpoint{5.857386in}{4.094982in}}{\pgfqpoint{5.801830in}{4.094982in}}%
\pgfpathlineto{\pgfqpoint{4.259928in}{4.094982in}}%
\pgfpathquadraticcurveto{\pgfqpoint{4.204373in}{4.094982in}}{\pgfqpoint{4.204373in}{4.039427in}}%
\pgfpathlineto{\pgfqpoint{4.204373in}{3.672248in}}%
\pgfpathquadraticcurveto{\pgfqpoint{4.204373in}{3.616692in}}{\pgfqpoint{4.259928in}{3.616692in}}%
\pgfpathlineto{\pgfqpoint{4.259928in}{3.616692in}}%
\pgfpathclose%
\pgfusepath{stroke,fill}%
\end{pgfscope}%
\begin{pgfscope}%
\pgfsetbuttcap%
\pgfsetroundjoin%
\pgfsetlinewidth{2.007500pt}%
\definecolor{currentstroke}{rgb}{0.501961,0.501961,0.501961}%
\pgfsetstrokecolor{currentstroke}%
\pgfsetdash{{7.400000pt}{3.200000pt}}{0.000000pt}%
\pgfpathmoveto{\pgfqpoint{4.315484in}{3.881055in}}%
\pgfpathlineto{\pgfqpoint{4.593262in}{3.881055in}}%
\pgfpathlineto{\pgfqpoint{4.871039in}{3.881055in}}%
\pgfusepath{stroke}%
\end{pgfscope}%
\begin{pgfscope}%
\definecolor{textcolor}{rgb}{0.000000,0.000000,0.000000}%
\pgfsetstrokecolor{textcolor}%
\pgfsetfillcolor{textcolor}%
\pgftext[x=5.093262in,y=3.783833in,left,base]{\color{textcolor}{\rmfamily\fontsize{20.000000}{24.000000}\selectfont\catcode`\^=\active\def^{\ifmmode\sp\else\^{}\fi}\catcode`\%=\active\def
\end{pgfscope}%
\end{pgfpicture}%
\makeatother%
\endgroup%

%% file: figures/evaluation/Sepsis/Sepsis_PIT_event_elapsed_norm_4layer.pgf
\begingroup%
\makeatletter%
\begin{pgfpicture}%
\pgfpathrectangle{\pgfpointorigin}{\pgfqpoint{6.159289in}{4.557174in}}%
\pgfusepath{use as bounding box, clip}%
\begin{pgfscope}%
\pgfsetbuttcap%
\pgfsetmiterjoin%
\definecolor{currentfill}{rgb}{1.000000,1.000000,1.000000}%
\pgfsetfillcolor{currentfill}%
\pgfsetlinewidth{0.000000pt}%
\definecolor{currentstroke}{rgb}{1.000000,1.000000,1.000000}%
\pgfsetstrokecolor{currentstroke}%
\pgfsetdash{}{0pt}%
\pgfpathmoveto{\pgfqpoint{0.000000in}{0.000000in}}%
\pgfpathlineto{\pgfqpoint{6.159289in}{0.000000in}}%
\pgfpathlineto{\pgfqpoint{6.159289in}{4.557174in}}%
\pgfpathlineto{\pgfqpoint{0.000000in}{4.557174in}}%
\pgfpathlineto{\pgfqpoint{0.000000in}{0.000000in}}%
\pgfpathclose%
\pgfusepath{fill}%
\end{pgfscope}%
\begin{pgfscope}%
\pgfsetbuttcap%
\pgfsetmiterjoin%
\definecolor{currentfill}{rgb}{1.000000,1.000000,1.000000}%
\pgfsetfillcolor{currentfill}%
\pgfsetlinewidth{0.000000pt}%
\definecolor{currentstroke}{rgb}{0.000000,0.000000,0.000000}%
\pgfsetstrokecolor{currentstroke}%
\pgfsetstrokeopacity{0.000000}%
\pgfsetdash{}{0pt}%
\pgfpathmoveto{\pgfqpoint{0.592068in}{0.716028in}}%
\pgfpathlineto{\pgfqpoint{5.988008in}{0.716028in}}%
\pgfpathlineto{\pgfqpoint{5.988008in}{4.233871in}}%
\pgfpathlineto{\pgfqpoint{0.592068in}{4.233871in}}%
\pgfpathlineto{\pgfqpoint{0.592068in}{0.716028in}}%
\pgfpathclose%
\pgfusepath{fill}%
\end{pgfscope}%
\begin{pgfscope}%
\pgfpathrectangle{\pgfqpoint{0.592068in}{0.716028in}}{\pgfqpoint{5.395940in}{3.517843in}}%
\pgfusepath{clip}%
\pgfsetbuttcap%
\pgfsetmiterjoin%
\definecolor{currentfill}{rgb}{0.000000,0.000000,1.000000}%
\pgfsetfillcolor{currentfill}%
\pgfsetlinewidth{0.000000pt}%
\definecolor{currentstroke}{rgb}{0.000000,0.000000,0.000000}%
\pgfsetstrokecolor{currentstroke}%
\pgfsetdash{}{0pt}%
\pgfpathmoveto{\pgfqpoint{0.592068in}{0.716028in}}%
\pgfpathlineto{\pgfqpoint{0.778135in}{0.716028in}}%
\pgfpathlineto{\pgfqpoint{0.778135in}{0.716028in}}%
\pgfpathlineto{\pgfqpoint{0.592068in}{0.716028in}}%
\pgfpathlineto{\pgfqpoint{0.592068in}{0.716028in}}%
\pgfpathclose%
\pgfusepath{fill}%
\end{pgfscope}%
\begin{pgfscope}%
\pgfpathrectangle{\pgfqpoint{0.592068in}{0.716028in}}{\pgfqpoint{5.395940in}{3.517843in}}%
\pgfusepath{clip}%
\pgfsetbuttcap%
\pgfsetmiterjoin%
\definecolor{currentfill}{rgb}{0.000000,0.000000,1.000000}%
\pgfsetfillcolor{currentfill}%
\pgfsetlinewidth{0.000000pt}%
\definecolor{currentstroke}{rgb}{0.000000,0.000000,0.000000}%
\pgfsetstrokecolor{currentstroke}%
\pgfsetdash{}{0pt}%
\pgfpathmoveto{\pgfqpoint{0.778135in}{0.716028in}}%
\pgfpathlineto{\pgfqpoint{0.964202in}{0.716028in}}%
\pgfpathlineto{\pgfqpoint{0.964202in}{0.716028in}}%
\pgfpathlineto{\pgfqpoint{0.778135in}{0.716028in}}%
\pgfpathlineto{\pgfqpoint{0.778135in}{0.716028in}}%
\pgfpathclose%
\pgfusepath{fill}%
\end{pgfscope}%
\begin{pgfscope}%
\pgfpathrectangle{\pgfqpoint{0.592068in}{0.716028in}}{\pgfqpoint{5.395940in}{3.517843in}}%
\pgfusepath{clip}%
\pgfsetbuttcap%
\pgfsetmiterjoin%
\definecolor{currentfill}{rgb}{0.000000,0.000000,1.000000}%
\pgfsetfillcolor{currentfill}%
\pgfsetlinewidth{0.000000pt}%
\definecolor{currentstroke}{rgb}{0.000000,0.000000,0.000000}%
\pgfsetstrokecolor{currentstroke}%
\pgfsetdash{}{0pt}%
\pgfpathmoveto{\pgfqpoint{0.964202in}{0.716028in}}%
\pgfpathlineto{\pgfqpoint{1.150269in}{0.716028in}}%
\pgfpathlineto{\pgfqpoint{1.150269in}{0.716028in}}%
\pgfpathlineto{\pgfqpoint{0.964202in}{0.716028in}}%
\pgfpathlineto{\pgfqpoint{0.964202in}{0.716028in}}%
\pgfpathclose%
\pgfusepath{fill}%
\end{pgfscope}%
\begin{pgfscope}%
\pgfpathrectangle{\pgfqpoint{0.592068in}{0.716028in}}{\pgfqpoint{5.395940in}{3.517843in}}%
\pgfusepath{clip}%
\pgfsetbuttcap%
\pgfsetmiterjoin%
\definecolor{currentfill}{rgb}{0.000000,0.000000,1.000000}%
\pgfsetfillcolor{currentfill}%
\pgfsetlinewidth{0.000000pt}%
\definecolor{currentstroke}{rgb}{0.000000,0.000000,0.000000}%
\pgfsetstrokecolor{currentstroke}%
\pgfsetdash{}{0pt}%
\pgfpathmoveto{\pgfqpoint{1.150269in}{0.716028in}}%
\pgfpathlineto{\pgfqpoint{1.336335in}{0.716028in}}%
\pgfpathlineto{\pgfqpoint{1.336335in}{0.716028in}}%
\pgfpathlineto{\pgfqpoint{1.150269in}{0.716028in}}%
\pgfpathlineto{\pgfqpoint{1.150269in}{0.716028in}}%
\pgfpathclose%
\pgfusepath{fill}%
\end{pgfscope}%
\begin{pgfscope}%
\pgfpathrectangle{\pgfqpoint{0.592068in}{0.716028in}}{\pgfqpoint{5.395940in}{3.517843in}}%
\pgfusepath{clip}%
\pgfsetbuttcap%
\pgfsetmiterjoin%
\definecolor{currentfill}{rgb}{0.000000,0.000000,1.000000}%
\pgfsetfillcolor{currentfill}%
\pgfsetlinewidth{0.000000pt}%
\definecolor{currentstroke}{rgb}{0.000000,0.000000,0.000000}%
\pgfsetstrokecolor{currentstroke}%
\pgfsetdash{}{0pt}%
\pgfpathmoveto{\pgfqpoint{1.336335in}{0.716028in}}%
\pgfpathlineto{\pgfqpoint{1.522402in}{0.716028in}}%
\pgfpathlineto{\pgfqpoint{1.522402in}{0.716028in}}%
\pgfpathlineto{\pgfqpoint{1.336335in}{0.716028in}}%
\pgfpathlineto{\pgfqpoint{1.336335in}{0.716028in}}%
\pgfpathclose%
\pgfusepath{fill}%
\end{pgfscope}%
\begin{pgfscope}%
\pgfpathrectangle{\pgfqpoint{0.592068in}{0.716028in}}{\pgfqpoint{5.395940in}{3.517843in}}%
\pgfusepath{clip}%
\pgfsetbuttcap%
\pgfsetmiterjoin%
\definecolor{currentfill}{rgb}{0.000000,0.000000,1.000000}%
\pgfsetfillcolor{currentfill}%
\pgfsetlinewidth{0.000000pt}%
\definecolor{currentstroke}{rgb}{0.000000,0.000000,0.000000}%
\pgfsetstrokecolor{currentstroke}%
\pgfsetdash{}{0pt}%
\pgfpathmoveto{\pgfqpoint{1.522402in}{0.716028in}}%
\pgfpathlineto{\pgfqpoint{1.708469in}{0.716028in}}%
\pgfpathlineto{\pgfqpoint{1.708469in}{0.716028in}}%
\pgfpathlineto{\pgfqpoint{1.522402in}{0.716028in}}%
\pgfpathlineto{\pgfqpoint{1.522402in}{0.716028in}}%
\pgfpathclose%
\pgfusepath{fill}%
\end{pgfscope}%
\begin{pgfscope}%
\pgfpathrectangle{\pgfqpoint{0.592068in}{0.716028in}}{\pgfqpoint{5.395940in}{3.517843in}}%
\pgfusepath{clip}%
\pgfsetbuttcap%
\pgfsetmiterjoin%
\definecolor{currentfill}{rgb}{0.000000,0.000000,1.000000}%
\pgfsetfillcolor{currentfill}%
\pgfsetlinewidth{0.000000pt}%
\definecolor{currentstroke}{rgb}{0.000000,0.000000,0.000000}%
\pgfsetstrokecolor{currentstroke}%
\pgfsetdash{}{0pt}%
\pgfpathmoveto{\pgfqpoint{1.708469in}{0.716028in}}%
\pgfpathlineto{\pgfqpoint{1.894536in}{0.716028in}}%
\pgfpathlineto{\pgfqpoint{1.894536in}{0.716028in}}%
\pgfpathlineto{\pgfqpoint{1.708469in}{0.716028in}}%
\pgfpathlineto{\pgfqpoint{1.708469in}{0.716028in}}%
\pgfpathclose%
\pgfusepath{fill}%
\end{pgfscope}%
\begin{pgfscope}%
\pgfpathrectangle{\pgfqpoint{0.592068in}{0.716028in}}{\pgfqpoint{5.395940in}{3.517843in}}%
\pgfusepath{clip}%
\pgfsetbuttcap%
\pgfsetmiterjoin%
\definecolor{currentfill}{rgb}{0.000000,0.000000,1.000000}%
\pgfsetfillcolor{currentfill}%
\pgfsetlinewidth{0.000000pt}%
\definecolor{currentstroke}{rgb}{0.000000,0.000000,0.000000}%
\pgfsetstrokecolor{currentstroke}%
\pgfsetdash{}{0pt}%
\pgfpathmoveto{\pgfqpoint{1.894536in}{0.716028in}}%
\pgfpathlineto{\pgfqpoint{2.080603in}{0.716028in}}%
\pgfpathlineto{\pgfqpoint{2.080603in}{0.716028in}}%
\pgfpathlineto{\pgfqpoint{1.894536in}{0.716028in}}%
\pgfpathlineto{\pgfqpoint{1.894536in}{0.716028in}}%
\pgfpathclose%
\pgfusepath{fill}%
\end{pgfscope}%
\begin{pgfscope}%
\pgfpathrectangle{\pgfqpoint{0.592068in}{0.716028in}}{\pgfqpoint{5.395940in}{3.517843in}}%
\pgfusepath{clip}%
\pgfsetbuttcap%
\pgfsetmiterjoin%
\definecolor{currentfill}{rgb}{0.000000,0.000000,1.000000}%
\pgfsetfillcolor{currentfill}%
\pgfsetlinewidth{0.000000pt}%
\definecolor{currentstroke}{rgb}{0.000000,0.000000,0.000000}%
\pgfsetstrokecolor{currentstroke}%
\pgfsetdash{}{0pt}%
\pgfpathmoveto{\pgfqpoint{2.080603in}{0.716028in}}%
\pgfpathlineto{\pgfqpoint{2.266670in}{0.716028in}}%
\pgfpathlineto{\pgfqpoint{2.266670in}{0.716028in}}%
\pgfpathlineto{\pgfqpoint{2.080603in}{0.716028in}}%
\pgfpathlineto{\pgfqpoint{2.080603in}{0.716028in}}%
\pgfpathclose%
\pgfusepath{fill}%
\end{pgfscope}%
\begin{pgfscope}%
\pgfpathrectangle{\pgfqpoint{0.592068in}{0.716028in}}{\pgfqpoint{5.395940in}{3.517843in}}%
\pgfusepath{clip}%
\pgfsetbuttcap%
\pgfsetmiterjoin%
\definecolor{currentfill}{rgb}{0.000000,0.000000,1.000000}%
\pgfsetfillcolor{currentfill}%
\pgfsetlinewidth{0.000000pt}%
\definecolor{currentstroke}{rgb}{0.000000,0.000000,0.000000}%
\pgfsetstrokecolor{currentstroke}%
\pgfsetdash{}{0pt}%
\pgfpathmoveto{\pgfqpoint{2.266670in}{0.716028in}}%
\pgfpathlineto{\pgfqpoint{2.452737in}{0.716028in}}%
\pgfpathlineto{\pgfqpoint{2.452737in}{0.716028in}}%
\pgfpathlineto{\pgfqpoint{2.266670in}{0.716028in}}%
\pgfpathlineto{\pgfqpoint{2.266670in}{0.716028in}}%
\pgfpathclose%
\pgfusepath{fill}%
\end{pgfscope}%
\begin{pgfscope}%
\pgfpathrectangle{\pgfqpoint{0.592068in}{0.716028in}}{\pgfqpoint{5.395940in}{3.517843in}}%
\pgfusepath{clip}%
\pgfsetbuttcap%
\pgfsetmiterjoin%
\definecolor{currentfill}{rgb}{0.000000,0.000000,1.000000}%
\pgfsetfillcolor{currentfill}%
\pgfsetlinewidth{0.000000pt}%
\definecolor{currentstroke}{rgb}{0.000000,0.000000,0.000000}%
\pgfsetstrokecolor{currentstroke}%
\pgfsetdash{}{0pt}%
\pgfpathmoveto{\pgfqpoint{2.452737in}{0.716028in}}%
\pgfpathlineto{\pgfqpoint{2.638804in}{0.716028in}}%
\pgfpathlineto{\pgfqpoint{2.638804in}{0.716028in}}%
\pgfpathlineto{\pgfqpoint{2.452737in}{0.716028in}}%
\pgfpathlineto{\pgfqpoint{2.452737in}{0.716028in}}%
\pgfpathclose%
\pgfusepath{fill}%
\end{pgfscope}%
\begin{pgfscope}%
\pgfpathrectangle{\pgfqpoint{0.592068in}{0.716028in}}{\pgfqpoint{5.395940in}{3.517843in}}%
\pgfusepath{clip}%
\pgfsetbuttcap%
\pgfsetmiterjoin%
\definecolor{currentfill}{rgb}{0.000000,0.000000,1.000000}%
\pgfsetfillcolor{currentfill}%
\pgfsetlinewidth{0.000000pt}%
\definecolor{currentstroke}{rgb}{0.000000,0.000000,0.000000}%
\pgfsetstrokecolor{currentstroke}%
\pgfsetdash{}{0pt}%
\pgfpathmoveto{\pgfqpoint{2.638804in}{0.716028in}}%
\pgfpathlineto{\pgfqpoint{2.824871in}{0.716028in}}%
\pgfpathlineto{\pgfqpoint{2.824871in}{0.716028in}}%
\pgfpathlineto{\pgfqpoint{2.638804in}{0.716028in}}%
\pgfpathlineto{\pgfqpoint{2.638804in}{0.716028in}}%
\pgfpathclose%
\pgfusepath{fill}%
\end{pgfscope}%
\begin{pgfscope}%
\pgfpathrectangle{\pgfqpoint{0.592068in}{0.716028in}}{\pgfqpoint{5.395940in}{3.517843in}}%
\pgfusepath{clip}%
\pgfsetbuttcap%
\pgfsetmiterjoin%
\definecolor{currentfill}{rgb}{0.000000,0.000000,1.000000}%
\pgfsetfillcolor{currentfill}%
\pgfsetlinewidth{0.000000pt}%
\definecolor{currentstroke}{rgb}{0.000000,0.000000,0.000000}%
\pgfsetstrokecolor{currentstroke}%
\pgfsetdash{}{0pt}%
\pgfpathmoveto{\pgfqpoint{2.824871in}{0.716028in}}%
\pgfpathlineto{\pgfqpoint{3.010937in}{0.716028in}}%
\pgfpathlineto{\pgfqpoint{3.010937in}{0.727621in}}%
\pgfpathlineto{\pgfqpoint{2.824871in}{0.727621in}}%
\pgfpathlineto{\pgfqpoint{2.824871in}{0.716028in}}%
\pgfpathclose%
\pgfusepath{fill}%
\end{pgfscope}%
\begin{pgfscope}%
\pgfpathrectangle{\pgfqpoint{0.592068in}{0.716028in}}{\pgfqpoint{5.395940in}{3.517843in}}%
\pgfusepath{clip}%
\pgfsetbuttcap%
\pgfsetmiterjoin%
\definecolor{currentfill}{rgb}{0.000000,0.000000,1.000000}%
\pgfsetfillcolor{currentfill}%
\pgfsetlinewidth{0.000000pt}%
\definecolor{currentstroke}{rgb}{0.000000,0.000000,0.000000}%
\pgfsetstrokecolor{currentstroke}%
\pgfsetdash{}{0pt}%
\pgfpathmoveto{\pgfqpoint{3.010937in}{0.716028in}}%
\pgfpathlineto{\pgfqpoint{3.197004in}{0.716028in}}%
\pgfpathlineto{\pgfqpoint{3.197004in}{1.110184in}}%
\pgfpathlineto{\pgfqpoint{3.010937in}{1.110184in}}%
\pgfpathlineto{\pgfqpoint{3.010937in}{0.716028in}}%
\pgfpathclose%
\pgfusepath{fill}%
\end{pgfscope}%
\begin{pgfscope}%
\pgfpathrectangle{\pgfqpoint{0.592068in}{0.716028in}}{\pgfqpoint{5.395940in}{3.517843in}}%
\pgfusepath{clip}%
\pgfsetbuttcap%
\pgfsetmiterjoin%
\definecolor{currentfill}{rgb}{0.000000,0.000000,1.000000}%
\pgfsetfillcolor{currentfill}%
\pgfsetlinewidth{0.000000pt}%
\definecolor{currentstroke}{rgb}{0.000000,0.000000,0.000000}%
\pgfsetstrokecolor{currentstroke}%
\pgfsetdash{}{0pt}%
\pgfpathmoveto{\pgfqpoint{3.197004in}{0.716028in}}%
\pgfpathlineto{\pgfqpoint{3.383071in}{0.716028in}}%
\pgfpathlineto{\pgfqpoint{3.383071in}{2.779551in}}%
\pgfpathlineto{\pgfqpoint{3.197004in}{2.779551in}}%
\pgfpathlineto{\pgfqpoint{3.197004in}{0.716028in}}%
\pgfpathclose%
\pgfusepath{fill}%
\end{pgfscope}%
\begin{pgfscope}%
\pgfpathrectangle{\pgfqpoint{0.592068in}{0.716028in}}{\pgfqpoint{5.395940in}{3.517843in}}%
\pgfusepath{clip}%
\pgfsetbuttcap%
\pgfsetmiterjoin%
\definecolor{currentfill}{rgb}{0.000000,0.000000,1.000000}%
\pgfsetfillcolor{currentfill}%
\pgfsetlinewidth{0.000000pt}%
\definecolor{currentstroke}{rgb}{0.000000,0.000000,0.000000}%
\pgfsetstrokecolor{currentstroke}%
\pgfsetdash{}{0pt}%
\pgfpathmoveto{\pgfqpoint{3.383071in}{0.716028in}}%
\pgfpathlineto{\pgfqpoint{3.569138in}{0.716028in}}%
\pgfpathlineto{\pgfqpoint{3.569138in}{2.744773in}}%
\pgfpathlineto{\pgfqpoint{3.383071in}{2.744773in}}%
\pgfpathlineto{\pgfqpoint{3.383071in}{0.716028in}}%
\pgfpathclose%
\pgfusepath{fill}%
\end{pgfscope}%
\begin{pgfscope}%
\pgfpathrectangle{\pgfqpoint{0.592068in}{0.716028in}}{\pgfqpoint{5.395940in}{3.517843in}}%
\pgfusepath{clip}%
\pgfsetbuttcap%
\pgfsetmiterjoin%
\definecolor{currentfill}{rgb}{0.000000,0.000000,1.000000}%
\pgfsetfillcolor{currentfill}%
\pgfsetlinewidth{0.000000pt}%
\definecolor{currentstroke}{rgb}{0.000000,0.000000,0.000000}%
\pgfsetstrokecolor{currentstroke}%
\pgfsetdash{}{0pt}%
\pgfpathmoveto{\pgfqpoint{3.569138in}{0.716028in}}%
\pgfpathlineto{\pgfqpoint{3.755205in}{0.716028in}}%
\pgfpathlineto{\pgfqpoint{3.755205in}{2.965036in}}%
\pgfpathlineto{\pgfqpoint{3.569138in}{2.965036in}}%
\pgfpathlineto{\pgfqpoint{3.569138in}{0.716028in}}%
\pgfpathclose%
\pgfusepath{fill}%
\end{pgfscope}%
\begin{pgfscope}%
\pgfpathrectangle{\pgfqpoint{0.592068in}{0.716028in}}{\pgfqpoint{5.395940in}{3.517843in}}%
\pgfusepath{clip}%
\pgfsetbuttcap%
\pgfsetmiterjoin%
\definecolor{currentfill}{rgb}{0.000000,0.000000,1.000000}%
\pgfsetfillcolor{currentfill}%
\pgfsetlinewidth{0.000000pt}%
\definecolor{currentstroke}{rgb}{0.000000,0.000000,0.000000}%
\pgfsetstrokecolor{currentstroke}%
\pgfsetdash{}{0pt}%
\pgfpathmoveto{\pgfqpoint{3.755205in}{0.716028in}}%
\pgfpathlineto{\pgfqpoint{3.941272in}{0.716028in}}%
\pgfpathlineto{\pgfqpoint{3.941272in}{2.663623in}}%
\pgfpathlineto{\pgfqpoint{3.755205in}{2.663623in}}%
\pgfpathlineto{\pgfqpoint{3.755205in}{0.716028in}}%
\pgfpathclose%
\pgfusepath{fill}%
\end{pgfscope}%
\begin{pgfscope}%
\pgfpathrectangle{\pgfqpoint{0.592068in}{0.716028in}}{\pgfqpoint{5.395940in}{3.517843in}}%
\pgfusepath{clip}%
\pgfsetbuttcap%
\pgfsetmiterjoin%
\definecolor{currentfill}{rgb}{0.000000,0.000000,1.000000}%
\pgfsetfillcolor{currentfill}%
\pgfsetlinewidth{0.000000pt}%
\definecolor{currentstroke}{rgb}{0.000000,0.000000,0.000000}%
\pgfsetstrokecolor{currentstroke}%
\pgfsetdash{}{0pt}%
\pgfpathmoveto{\pgfqpoint{3.941272in}{0.716028in}}%
\pgfpathlineto{\pgfqpoint{4.127339in}{0.716028in}}%
\pgfpathlineto{\pgfqpoint{4.127339in}{1.991239in}}%
\pgfpathlineto{\pgfqpoint{3.941272in}{1.991239in}}%
\pgfpathlineto{\pgfqpoint{3.941272in}{0.716028in}}%
\pgfpathclose%
\pgfusepath{fill}%
\end{pgfscope}%
\begin{pgfscope}%
\pgfpathrectangle{\pgfqpoint{0.592068in}{0.716028in}}{\pgfqpoint{5.395940in}{3.517843in}}%
\pgfusepath{clip}%
\pgfsetbuttcap%
\pgfsetmiterjoin%
\definecolor{currentfill}{rgb}{0.000000,0.000000,1.000000}%
\pgfsetfillcolor{currentfill}%
\pgfsetlinewidth{0.000000pt}%
\definecolor{currentstroke}{rgb}{0.000000,0.000000,0.000000}%
\pgfsetstrokecolor{currentstroke}%
\pgfsetdash{}{0pt}%
\pgfpathmoveto{\pgfqpoint{4.127339in}{0.716028in}}%
\pgfpathlineto{\pgfqpoint{4.313406in}{0.716028in}}%
\pgfpathlineto{\pgfqpoint{4.313406in}{1.968053in}}%
\pgfpathlineto{\pgfqpoint{4.127339in}{1.968053in}}%
\pgfpathlineto{\pgfqpoint{4.127339in}{0.716028in}}%
\pgfpathclose%
\pgfusepath{fill}%
\end{pgfscope}%
\begin{pgfscope}%
\pgfpathrectangle{\pgfqpoint{0.592068in}{0.716028in}}{\pgfqpoint{5.395940in}{3.517843in}}%
\pgfusepath{clip}%
\pgfsetbuttcap%
\pgfsetmiterjoin%
\definecolor{currentfill}{rgb}{0.000000,0.000000,1.000000}%
\pgfsetfillcolor{currentfill}%
\pgfsetlinewidth{0.000000pt}%
\definecolor{currentstroke}{rgb}{0.000000,0.000000,0.000000}%
\pgfsetstrokecolor{currentstroke}%
\pgfsetdash{}{0pt}%
\pgfpathmoveto{\pgfqpoint{4.313406in}{0.716028in}}%
\pgfpathlineto{\pgfqpoint{4.499473in}{0.716028in}}%
\pgfpathlineto{\pgfqpoint{4.499473in}{1.585490in}}%
\pgfpathlineto{\pgfqpoint{4.313406in}{1.585490in}}%
\pgfpathlineto{\pgfqpoint{4.313406in}{0.716028in}}%
\pgfpathclose%
\pgfusepath{fill}%
\end{pgfscope}%
\begin{pgfscope}%
\pgfpathrectangle{\pgfqpoint{0.592068in}{0.716028in}}{\pgfqpoint{5.395940in}{3.517843in}}%
\pgfusepath{clip}%
\pgfsetbuttcap%
\pgfsetmiterjoin%
\definecolor{currentfill}{rgb}{0.000000,0.000000,1.000000}%
\pgfsetfillcolor{currentfill}%
\pgfsetlinewidth{0.000000pt}%
\definecolor{currentstroke}{rgb}{0.000000,0.000000,0.000000}%
\pgfsetstrokecolor{currentstroke}%
\pgfsetdash{}{0pt}%
\pgfpathmoveto{\pgfqpoint{4.499473in}{0.716028in}}%
\pgfpathlineto{\pgfqpoint{4.685539in}{0.716028in}}%
\pgfpathlineto{\pgfqpoint{4.685539in}{1.481155in}}%
\pgfpathlineto{\pgfqpoint{4.499473in}{1.481155in}}%
\pgfpathlineto{\pgfqpoint{4.499473in}{0.716028in}}%
\pgfpathclose%
\pgfusepath{fill}%
\end{pgfscope}%
\begin{pgfscope}%
\pgfpathrectangle{\pgfqpoint{0.592068in}{0.716028in}}{\pgfqpoint{5.395940in}{3.517843in}}%
\pgfusepath{clip}%
\pgfsetbuttcap%
\pgfsetmiterjoin%
\definecolor{currentfill}{rgb}{0.000000,0.000000,1.000000}%
\pgfsetfillcolor{currentfill}%
\pgfsetlinewidth{0.000000pt}%
\definecolor{currentstroke}{rgb}{0.000000,0.000000,0.000000}%
\pgfsetstrokecolor{currentstroke}%
\pgfsetdash{}{0pt}%
\pgfpathmoveto{\pgfqpoint{4.685539in}{0.716028in}}%
\pgfpathlineto{\pgfqpoint{4.871606in}{0.716028in}}%
\pgfpathlineto{\pgfqpoint{4.871606in}{1.400005in}}%
\pgfpathlineto{\pgfqpoint{4.685539in}{1.400005in}}%
\pgfpathlineto{\pgfqpoint{4.685539in}{0.716028in}}%
\pgfpathclose%
\pgfusepath{fill}%
\end{pgfscope}%
\begin{pgfscope}%
\pgfpathrectangle{\pgfqpoint{0.592068in}{0.716028in}}{\pgfqpoint{5.395940in}{3.517843in}}%
\pgfusepath{clip}%
\pgfsetbuttcap%
\pgfsetmiterjoin%
\definecolor{currentfill}{rgb}{0.000000,0.000000,1.000000}%
\pgfsetfillcolor{currentfill}%
\pgfsetlinewidth{0.000000pt}%
\definecolor{currentstroke}{rgb}{0.000000,0.000000,0.000000}%
\pgfsetstrokecolor{currentstroke}%
\pgfsetdash{}{0pt}%
\pgfpathmoveto{\pgfqpoint{4.871606in}{0.716028in}}%
\pgfpathlineto{\pgfqpoint{5.057673in}{0.716028in}}%
\pgfpathlineto{\pgfqpoint{5.057673in}{1.666640in}}%
\pgfpathlineto{\pgfqpoint{4.871606in}{1.666640in}}%
\pgfpathlineto{\pgfqpoint{4.871606in}{0.716028in}}%
\pgfpathclose%
\pgfusepath{fill}%
\end{pgfscope}%
\begin{pgfscope}%
\pgfpathrectangle{\pgfqpoint{0.592068in}{0.716028in}}{\pgfqpoint{5.395940in}{3.517843in}}%
\pgfusepath{clip}%
\pgfsetbuttcap%
\pgfsetmiterjoin%
\definecolor{currentfill}{rgb}{0.000000,0.000000,1.000000}%
\pgfsetfillcolor{currentfill}%
\pgfsetlinewidth{0.000000pt}%
\definecolor{currentstroke}{rgb}{0.000000,0.000000,0.000000}%
\pgfsetstrokecolor{currentstroke}%
\pgfsetdash{}{0pt}%
\pgfpathmoveto{\pgfqpoint{5.057673in}{0.716028in}}%
\pgfpathlineto{\pgfqpoint{5.243740in}{0.716028in}}%
\pgfpathlineto{\pgfqpoint{5.243740in}{1.249298in}}%
\pgfpathlineto{\pgfqpoint{5.057673in}{1.249298in}}%
\pgfpathlineto{\pgfqpoint{5.057673in}{0.716028in}}%
\pgfpathclose%
\pgfusepath{fill}%
\end{pgfscope}%
\begin{pgfscope}%
\pgfpathrectangle{\pgfqpoint{0.592068in}{0.716028in}}{\pgfqpoint{5.395940in}{3.517843in}}%
\pgfusepath{clip}%
\pgfsetbuttcap%
\pgfsetmiterjoin%
\definecolor{currentfill}{rgb}{0.000000,0.000000,1.000000}%
\pgfsetfillcolor{currentfill}%
\pgfsetlinewidth{0.000000pt}%
\definecolor{currentstroke}{rgb}{0.000000,0.000000,0.000000}%
\pgfsetstrokecolor{currentstroke}%
\pgfsetdash{}{0pt}%
\pgfpathmoveto{\pgfqpoint{5.243740in}{0.716028in}}%
\pgfpathlineto{\pgfqpoint{5.429807in}{0.716028in}}%
\pgfpathlineto{\pgfqpoint{5.429807in}{1.110184in}}%
\pgfpathlineto{\pgfqpoint{5.243740in}{1.110184in}}%
\pgfpathlineto{\pgfqpoint{5.243740in}{0.716028in}}%
\pgfpathclose%
\pgfusepath{fill}%
\end{pgfscope}%
\begin{pgfscope}%
\pgfpathrectangle{\pgfqpoint{0.592068in}{0.716028in}}{\pgfqpoint{5.395940in}{3.517843in}}%
\pgfusepath{clip}%
\pgfsetbuttcap%
\pgfsetmiterjoin%
\definecolor{currentfill}{rgb}{0.000000,0.000000,1.000000}%
\pgfsetfillcolor{currentfill}%
\pgfsetlinewidth{0.000000pt}%
\definecolor{currentstroke}{rgb}{0.000000,0.000000,0.000000}%
\pgfsetstrokecolor{currentstroke}%
\pgfsetdash{}{0pt}%
\pgfpathmoveto{\pgfqpoint{5.429807in}{0.716028in}}%
\pgfpathlineto{\pgfqpoint{5.615874in}{0.716028in}}%
\pgfpathlineto{\pgfqpoint{5.615874in}{1.156556in}}%
\pgfpathlineto{\pgfqpoint{5.429807in}{1.156556in}}%
\pgfpathlineto{\pgfqpoint{5.429807in}{0.716028in}}%
\pgfpathclose%
\pgfusepath{fill}%
\end{pgfscope}%
\begin{pgfscope}%
\pgfpathrectangle{\pgfqpoint{0.592068in}{0.716028in}}{\pgfqpoint{5.395940in}{3.517843in}}%
\pgfusepath{clip}%
\pgfsetbuttcap%
\pgfsetmiterjoin%
\definecolor{currentfill}{rgb}{0.000000,0.000000,1.000000}%
\pgfsetfillcolor{currentfill}%
\pgfsetlinewidth{0.000000pt}%
\definecolor{currentstroke}{rgb}{0.000000,0.000000,0.000000}%
\pgfsetstrokecolor{currentstroke}%
\pgfsetdash{}{0pt}%
\pgfpathmoveto{\pgfqpoint{5.615874in}{0.716028in}}%
\pgfpathlineto{\pgfqpoint{5.801941in}{0.716028in}}%
\pgfpathlineto{\pgfqpoint{5.801941in}{1.214520in}}%
\pgfpathlineto{\pgfqpoint{5.615874in}{1.214520in}}%
\pgfpathlineto{\pgfqpoint{5.615874in}{0.716028in}}%
\pgfpathclose%
\pgfusepath{fill}%
\end{pgfscope}%
\begin{pgfscope}%
\pgfpathrectangle{\pgfqpoint{0.592068in}{0.716028in}}{\pgfqpoint{5.395940in}{3.517843in}}%
\pgfusepath{clip}%
\pgfsetbuttcap%
\pgfsetmiterjoin%
\definecolor{currentfill}{rgb}{0.000000,0.000000,1.000000}%
\pgfsetfillcolor{currentfill}%
\pgfsetlinewidth{0.000000pt}%
\definecolor{currentstroke}{rgb}{0.000000,0.000000,0.000000}%
\pgfsetstrokecolor{currentstroke}%
\pgfsetdash{}{0pt}%
\pgfpathmoveto{\pgfqpoint{5.801941in}{0.716028in}}%
\pgfpathlineto{\pgfqpoint{5.988008in}{0.716028in}}%
\pgfpathlineto{\pgfqpoint{5.988008in}{4.066355in}}%
\pgfpathlineto{\pgfqpoint{5.801941in}{4.066355in}}%
\pgfpathlineto{\pgfqpoint{5.801941in}{0.716028in}}%
\pgfpathclose%
\pgfusepath{fill}%
\end{pgfscope}%
\begin{pgfscope}%
\pgfsetbuttcap%
\pgfsetroundjoin%
\definecolor{currentfill}{rgb}{0.000000,0.000000,0.000000}%
\pgfsetfillcolor{currentfill}%
\pgfsetlinewidth{0.200750pt}%
\definecolor{currentstroke}{rgb}{0.000000,0.000000,0.000000}%
\pgfsetstrokecolor{currentstroke}%
\pgfsetdash{}{0pt}%
\pgfsys@defobject{currentmarker}{\pgfqpoint{0.000000in}{-0.048611in}}{\pgfqpoint{0.000000in}{0.000000in}}{%
\pgfpathmoveto{\pgfqpoint{0.000000in}{0.000000in}}%
\pgfpathlineto{\pgfqpoint{0.000000in}{-0.048611in}}%
\pgfusepath{stroke,fill}%
}%
\begin{pgfscope}%
\pgfsys@transformshift{0.592068in}{0.716028in}%
\pgfsys@useobject{currentmarker}{}%
\end{pgfscope}%
\end{pgfscope}%
\begin{pgfscope}%
\definecolor{textcolor}{rgb}{0.000000,0.000000,0.000000}%
\pgfsetstrokecolor{textcolor}%
\pgfsetfillcolor{textcolor}%
\pgftext[x=0.592068in,y=0.618806in,,top]{\color{textcolor}{\rmfamily\fontsize{20.000000}{24.000000}\selectfont\catcode`\^=\active\def^{\ifmmode\sp\else\^{}\fi}\catcode`\%=\active\def
\end{pgfscope}%
\begin{pgfscope}%
\pgfsetbuttcap%
\pgfsetroundjoin%
\definecolor{currentfill}{rgb}{0.000000,0.000000,0.000000}%
\pgfsetfillcolor{currentfill}%
\pgfsetlinewidth{0.200750pt}%
\definecolor{currentstroke}{rgb}{0.000000,0.000000,0.000000}%
\pgfsetstrokecolor{currentstroke}%
\pgfsetdash{}{0pt}%
\pgfsys@defobject{currentmarker}{\pgfqpoint{0.000000in}{-0.048611in}}{\pgfqpoint{0.000000in}{0.000000in}}{%
\pgfpathmoveto{\pgfqpoint{0.000000in}{0.000000in}}%
\pgfpathlineto{\pgfqpoint{0.000000in}{-0.048611in}}%
\pgfusepath{stroke,fill}%
}%
\begin{pgfscope}%
\pgfsys@transformshift{1.671256in}{0.716028in}%
\pgfsys@useobject{currentmarker}{}%
\end{pgfscope}%
\end{pgfscope}%
\begin{pgfscope}%
\definecolor{textcolor}{rgb}{0.000000,0.000000,0.000000}%
\pgfsetstrokecolor{textcolor}%
\pgfsetfillcolor{textcolor}%
\pgftext[x=1.671256in,y=0.618806in,,top]{\color{textcolor}{\rmfamily\fontsize{20.000000}{24.000000}\selectfont\catcode`\^=\active\def^{\ifmmode\sp\else\^{}\fi}\catcode`\%=\active\def
\end{pgfscope}%
\begin{pgfscope}%
\pgfsetbuttcap%
\pgfsetroundjoin%
\definecolor{currentfill}{rgb}{0.000000,0.000000,0.000000}%
\pgfsetfillcolor{currentfill}%
\pgfsetlinewidth{0.200750pt}%
\definecolor{currentstroke}{rgb}{0.000000,0.000000,0.000000}%
\pgfsetstrokecolor{currentstroke}%
\pgfsetdash{}{0pt}%
\pgfsys@defobject{currentmarker}{\pgfqpoint{0.000000in}{-0.048611in}}{\pgfqpoint{0.000000in}{0.000000in}}{%
\pgfpathmoveto{\pgfqpoint{0.000000in}{0.000000in}}%
\pgfpathlineto{\pgfqpoint{0.000000in}{-0.048611in}}%
\pgfusepath{stroke,fill}%
}%
\begin{pgfscope}%
\pgfsys@transformshift{2.750444in}{0.716028in}%
\pgfsys@useobject{currentmarker}{}%
\end{pgfscope}%
\end{pgfscope}%
\begin{pgfscope}%
\definecolor{textcolor}{rgb}{0.000000,0.000000,0.000000}%
\pgfsetstrokecolor{textcolor}%
\pgfsetfillcolor{textcolor}%
\pgftext[x=2.750444in,y=0.618806in,,top]{\color{textcolor}{\rmfamily\fontsize{20.000000}{24.000000}\selectfont\catcode`\^=\active\def^{\ifmmode\sp\else\^{}\fi}\catcode`\%=\active\def
\end{pgfscope}%
\begin{pgfscope}%
\pgfsetbuttcap%
\pgfsetroundjoin%
\definecolor{currentfill}{rgb}{0.000000,0.000000,0.000000}%
\pgfsetfillcolor{currentfill}%
\pgfsetlinewidth{0.200750pt}%
\definecolor{currentstroke}{rgb}{0.000000,0.000000,0.000000}%
\pgfsetstrokecolor{currentstroke}%
\pgfsetdash{}{0pt}%
\pgfsys@defobject{currentmarker}{\pgfqpoint{0.000000in}{-0.048611in}}{\pgfqpoint{0.000000in}{0.000000in}}{%
\pgfpathmoveto{\pgfqpoint{0.000000in}{0.000000in}}%
\pgfpathlineto{\pgfqpoint{0.000000in}{-0.048611in}}%
\pgfusepath{stroke,fill}%
}%
\begin{pgfscope}%
\pgfsys@transformshift{3.829632in}{0.716028in}%
\pgfsys@useobject{currentmarker}{}%
\end{pgfscope}%
\end{pgfscope}%
\begin{pgfscope}%
\definecolor{textcolor}{rgb}{0.000000,0.000000,0.000000}%
\pgfsetstrokecolor{textcolor}%
\pgfsetfillcolor{textcolor}%
\pgftext[x=3.829632in,y=0.618806in,,top]{\color{textcolor}{\rmfamily\fontsize{20.000000}{24.000000}\selectfont\catcode`\^=\active\def^{\ifmmode\sp\else\^{}\fi}\catcode`\%=\active\def
\end{pgfscope}%
\begin{pgfscope}%
\pgfsetbuttcap%
\pgfsetroundjoin%
\definecolor{currentfill}{rgb}{0.000000,0.000000,0.000000}%
\pgfsetfillcolor{currentfill}%
\pgfsetlinewidth{0.200750pt}%
\definecolor{currentstroke}{rgb}{0.000000,0.000000,0.000000}%
\pgfsetstrokecolor{currentstroke}%
\pgfsetdash{}{0pt}%
\pgfsys@defobject{currentmarker}{\pgfqpoint{0.000000in}{-0.048611in}}{\pgfqpoint{0.000000in}{0.000000in}}{%
\pgfpathmoveto{\pgfqpoint{0.000000in}{0.000000in}}%
\pgfpathlineto{\pgfqpoint{0.000000in}{-0.048611in}}%
\pgfusepath{stroke,fill}%
}%
\begin{pgfscope}%
\pgfsys@transformshift{4.908820in}{0.716028in}%
\pgfsys@useobject{currentmarker}{}%
\end{pgfscope}%
\end{pgfscope}%
\begin{pgfscope}%
\definecolor{textcolor}{rgb}{0.000000,0.000000,0.000000}%
\pgfsetstrokecolor{textcolor}%
\pgfsetfillcolor{textcolor}%
\pgftext[x=4.908820in,y=0.618806in,,top]{\color{textcolor}{\rmfamily\fontsize{20.000000}{24.000000}\selectfont\catcode`\^=\active\def^{\ifmmode\sp\else\^{}\fi}\catcode`\%=\active\def
\end{pgfscope}%
\begin{pgfscope}%
\pgfsetbuttcap%
\pgfsetroundjoin%
\definecolor{currentfill}{rgb}{0.000000,0.000000,0.000000}%
\pgfsetfillcolor{currentfill}%
\pgfsetlinewidth{0.200750pt}%
\definecolor{currentstroke}{rgb}{0.000000,0.000000,0.000000}%
\pgfsetstrokecolor{currentstroke}%
\pgfsetdash{}{0pt}%
\pgfsys@defobject{currentmarker}{\pgfqpoint{0.000000in}{-0.048611in}}{\pgfqpoint{0.000000in}{0.000000in}}{%
\pgfpathmoveto{\pgfqpoint{0.000000in}{0.000000in}}%
\pgfpathlineto{\pgfqpoint{0.000000in}{-0.048611in}}%
\pgfusepath{stroke,fill}%
}%
\begin{pgfscope}%
\pgfsys@transformshift{5.988008in}{0.716028in}%
\pgfsys@useobject{currentmarker}{}%
\end{pgfscope}%
\end{pgfscope}%
\begin{pgfscope}%
\definecolor{textcolor}{rgb}{0.000000,0.000000,0.000000}%
\pgfsetstrokecolor{textcolor}%
\pgfsetfillcolor{textcolor}%
\pgftext[x=5.988008in,y=0.618806in,,top]{\color{textcolor}{\rmfamily\fontsize{20.000000}{24.000000}\selectfont\catcode`\^=\active\def^{\ifmmode\sp\else\^{}\fi}\catcode`\%=\active\def
\end{pgfscope}%
\begin{pgfscope}%
\definecolor{textcolor}{rgb}{0.000000,0.000000,0.000000}%
\pgfsetstrokecolor{textcolor}%
\pgfsetfillcolor{textcolor}%
\pgftext[x=3.290038in,y=0.307183in,,top]{\color{textcolor}{\rmfamily\fontsize{26.000000}{31.200000}\selectfont\catcode`\^=\active\def^{\ifmmode\sp\else\^{}\fi}\catcode`\%=\active\def
\end{pgfscope}%
\begin{pgfscope}%
\pgfsetbuttcap%
\pgfsetroundjoin%
\definecolor{currentfill}{rgb}{0.000000,0.000000,0.000000}%
\pgfsetfillcolor{currentfill}%
\pgfsetlinewidth{0.200750pt}%
\definecolor{currentstroke}{rgb}{0.000000,0.000000,0.000000}%
\pgfsetstrokecolor{currentstroke}%
\pgfsetdash{}{0pt}%
\pgfsys@defobject{currentmarker}{\pgfqpoint{-0.048611in}{0.000000in}}{\pgfqpoint{-0.000000in}{0.000000in}}{%
\pgfpathmoveto{\pgfqpoint{-0.000000in}{0.000000in}}%
\pgfpathlineto{\pgfqpoint{-0.048611in}{0.000000in}}%
\pgfusepath{stroke,fill}%
}%
\begin{pgfscope}%
\pgfsys@transformshift{0.592068in}{0.716028in}%
\pgfsys@useobject{currentmarker}{}%
\end{pgfscope}%
\end{pgfscope}%
\begin{pgfscope}%
\definecolor{textcolor}{rgb}{0.000000,0.000000,0.000000}%
\pgfsetstrokecolor{textcolor}%
\pgfsetfillcolor{textcolor}%
\pgftext[x=0.362738in, y=0.616009in, left, base]{\color{textcolor}{\rmfamily\fontsize{20.000000}{24.000000}\selectfont\catcode`\^=\active\def^{\ifmmode\sp\else\^{}\fi}\catcode`\%=\active\def
\end{pgfscope}%
\begin{pgfscope}%
\pgfsetbuttcap%
\pgfsetroundjoin%
\definecolor{currentfill}{rgb}{0.000000,0.000000,0.000000}%
\pgfsetfillcolor{currentfill}%
\pgfsetlinewidth{0.200750pt}%
\definecolor{currentstroke}{rgb}{0.000000,0.000000,0.000000}%
\pgfsetstrokecolor{currentstroke}%
\pgfsetdash{}{0pt}%
\pgfsys@defobject{currentmarker}{\pgfqpoint{-0.048611in}{0.000000in}}{\pgfqpoint{-0.000000in}{0.000000in}}{%
\pgfpathmoveto{\pgfqpoint{-0.000000in}{0.000000in}}%
\pgfpathlineto{\pgfqpoint{-0.048611in}{0.000000in}}%
\pgfusepath{stroke,fill}%
}%
\begin{pgfscope}%
\pgfsys@transformshift{0.592068in}{1.395608in}%
\pgfsys@useobject{currentmarker}{}%
\end{pgfscope}%
\end{pgfscope}%
\begin{pgfscope}%
\definecolor{textcolor}{rgb}{0.000000,0.000000,0.000000}%
\pgfsetstrokecolor{textcolor}%
\pgfsetfillcolor{textcolor}%
\pgftext[x=0.362738in, y=1.295588in, left, base]{\color{textcolor}{\rmfamily\fontsize{20.000000}{24.000000}\selectfont\catcode`\^=\active\def^{\ifmmode\sp\else\^{}\fi}\catcode`\%=\active\def
\end{pgfscope}%
\begin{pgfscope}%
\pgfsetbuttcap%
\pgfsetroundjoin%
\definecolor{currentfill}{rgb}{0.000000,0.000000,0.000000}%
\pgfsetfillcolor{currentfill}%
\pgfsetlinewidth{0.200750pt}%
\definecolor{currentstroke}{rgb}{0.000000,0.000000,0.000000}%
\pgfsetstrokecolor{currentstroke}%
\pgfsetdash{}{0pt}%
\pgfsys@defobject{currentmarker}{\pgfqpoint{-0.048611in}{0.000000in}}{\pgfqpoint{-0.000000in}{0.000000in}}{%
\pgfpathmoveto{\pgfqpoint{-0.000000in}{0.000000in}}%
\pgfpathlineto{\pgfqpoint{-0.048611in}{0.000000in}}%
\pgfusepath{stroke,fill}%
}%
\begin{pgfscope}%
\pgfsys@transformshift{0.592068in}{2.075187in}%
\pgfsys@useobject{currentmarker}{}%
\end{pgfscope}%
\end{pgfscope}%
\begin{pgfscope}%
\definecolor{textcolor}{rgb}{0.000000,0.000000,0.000000}%
\pgfsetstrokecolor{textcolor}%
\pgfsetfillcolor{textcolor}%
\pgftext[x=0.362738in, y=1.975168in, left, base]{\color{textcolor}{\rmfamily\fontsize{20.000000}{24.000000}\selectfont\catcode`\^=\active\def^{\ifmmode\sp\else\^{}\fi}\catcode`\%=\active\def
\end{pgfscope}%
\begin{pgfscope}%
\pgfsetbuttcap%
\pgfsetroundjoin%
\definecolor{currentfill}{rgb}{0.000000,0.000000,0.000000}%
\pgfsetfillcolor{currentfill}%
\pgfsetlinewidth{0.200750pt}%
\definecolor{currentstroke}{rgb}{0.000000,0.000000,0.000000}%
\pgfsetstrokecolor{currentstroke}%
\pgfsetdash{}{0pt}%
\pgfsys@defobject{currentmarker}{\pgfqpoint{-0.048611in}{0.000000in}}{\pgfqpoint{-0.000000in}{0.000000in}}{%
\pgfpathmoveto{\pgfqpoint{-0.000000in}{0.000000in}}%
\pgfpathlineto{\pgfqpoint{-0.048611in}{0.000000in}}%
\pgfusepath{stroke,fill}%
}%
\begin{pgfscope}%
\pgfsys@transformshift{0.592068in}{2.754766in}%
\pgfsys@useobject{currentmarker}{}%
\end{pgfscope}%
\end{pgfscope}%
\begin{pgfscope}%
\definecolor{textcolor}{rgb}{0.000000,0.000000,0.000000}%
\pgfsetstrokecolor{textcolor}%
\pgfsetfillcolor{textcolor}%
\pgftext[x=0.362738in, y=2.654747in, left, base]{\color{textcolor}{\rmfamily\fontsize{20.000000}{24.000000}\selectfont\catcode`\^=\active\def^{\ifmmode\sp\else\^{}\fi}\catcode`\%=\active\def
\end{pgfscope}%
\begin{pgfscope}%
\pgfsetbuttcap%
\pgfsetroundjoin%
\definecolor{currentfill}{rgb}{0.000000,0.000000,0.000000}%
\pgfsetfillcolor{currentfill}%
\pgfsetlinewidth{0.200750pt}%
\definecolor{currentstroke}{rgb}{0.000000,0.000000,0.000000}%
\pgfsetstrokecolor{currentstroke}%
\pgfsetdash{}{0pt}%
\pgfsys@defobject{currentmarker}{\pgfqpoint{-0.048611in}{0.000000in}}{\pgfqpoint{-0.000000in}{0.000000in}}{%
\pgfpathmoveto{\pgfqpoint{-0.000000in}{0.000000in}}%
\pgfpathlineto{\pgfqpoint{-0.048611in}{0.000000in}}%
\pgfusepath{stroke,fill}%
}%
\begin{pgfscope}%
\pgfsys@transformshift{0.592068in}{3.434346in}%
\pgfsys@useobject{currentmarker}{}%
\end{pgfscope}%
\end{pgfscope}%
\begin{pgfscope}%
\definecolor{textcolor}{rgb}{0.000000,0.000000,0.000000}%
\pgfsetstrokecolor{textcolor}%
\pgfsetfillcolor{textcolor}%
\pgftext[x=0.362738in, y=3.334327in, left, base]{\color{textcolor}{\rmfamily\fontsize{20.000000}{24.000000}\selectfont\catcode`\^=\active\def^{\ifmmode\sp\else\^{}\fi}\catcode`\%=\active\def
\end{pgfscope}%
\begin{pgfscope}%
\pgfsetbuttcap%
\pgfsetroundjoin%
\definecolor{currentfill}{rgb}{0.000000,0.000000,0.000000}%
\pgfsetfillcolor{currentfill}%
\pgfsetlinewidth{0.200750pt}%
\definecolor{currentstroke}{rgb}{0.000000,0.000000,0.000000}%
\pgfsetstrokecolor{currentstroke}%
\pgfsetdash{}{0pt}%
\pgfsys@defobject{currentmarker}{\pgfqpoint{-0.048611in}{0.000000in}}{\pgfqpoint{-0.000000in}{0.000000in}}{%
\pgfpathmoveto{\pgfqpoint{-0.000000in}{0.000000in}}%
\pgfpathlineto{\pgfqpoint{-0.048611in}{0.000000in}}%
\pgfusepath{stroke,fill}%
}%
\begin{pgfscope}%
\pgfsys@transformshift{0.592068in}{4.113925in}%
\pgfsys@useobject{currentmarker}{}%
\end{pgfscope}%
\end{pgfscope}%
\begin{pgfscope}%
\definecolor{textcolor}{rgb}{0.000000,0.000000,0.000000}%
\pgfsetstrokecolor{textcolor}%
\pgfsetfillcolor{textcolor}%
\pgftext[x=0.362738in, y=4.013906in, left, base]{\color{textcolor}{\rmfamily\fontsize{20.000000}{24.000000}\selectfont\catcode`\^=\active\def^{\ifmmode\sp\else\^{}\fi}\catcode`\%=\active\def
\end{pgfscope}%
\begin{pgfscope}%
\definecolor{textcolor}{rgb}{0.000000,0.000000,0.000000}%
\pgfsetstrokecolor{textcolor}%
\pgfsetfillcolor{textcolor}%
\pgftext[x=0.307183in,y=2.474950in,,bottom,rotate=90.000000]{\color{textcolor}{\rmfamily\fontsize{26.000000}{31.200000}\selectfont\catcode`\^=\active\def^{\ifmmode\sp\else\^{}\fi}\catcode`\%=\active\def
\end{pgfscope}%
\begin{pgfscope}%
\pgfpathrectangle{\pgfqpoint{0.592068in}{0.716028in}}{\pgfqpoint{5.395940in}{3.517843in}}%
\pgfusepath{clip}%
\pgfsetbuttcap%
\pgfsetroundjoin%
\pgfsetlinewidth{2.007500pt}%
\definecolor{currentstroke}{rgb}{0.501961,0.501961,0.501961}%
\pgfsetstrokecolor{currentstroke}%
\pgfsetdash{{7.400000pt}{3.200000pt}}{0.000000pt}%
\pgfpathmoveto{\pgfqpoint{0.592068in}{1.395608in}}%
\pgfpathlineto{\pgfqpoint{5.988008in}{1.395608in}}%
\pgfusepath{stroke}%
\end{pgfscope}%
\begin{pgfscope}%
\pgfsetrectcap%
\pgfsetmiterjoin%
\pgfsetlinewidth{0.803000pt}%
\definecolor{currentstroke}{rgb}{0.000000,0.000000,0.000000}%
\pgfsetstrokecolor{currentstroke}%
\pgfsetdash{}{0pt}%
\pgfpathmoveto{\pgfqpoint{0.592068in}{0.716028in}}%
\pgfpathlineto{\pgfqpoint{0.592068in}{4.233871in}}%
\pgfusepath{stroke}%
\end{pgfscope}%
\begin{pgfscope}%
\pgfsetrectcap%
\pgfsetmiterjoin%
\pgfsetlinewidth{0.803000pt}%
\definecolor{currentstroke}{rgb}{0.000000,0.000000,0.000000}%
\pgfsetstrokecolor{currentstroke}%
\pgfsetdash{}{0pt}%
\pgfpathmoveto{\pgfqpoint{5.988008in}{0.716028in}}%
\pgfpathlineto{\pgfqpoint{5.988008in}{4.233871in}}%
\pgfusepath{stroke}%
\end{pgfscope}%
\begin{pgfscope}%
\pgfsetrectcap%
\pgfsetmiterjoin%
\pgfsetlinewidth{0.803000pt}%
\definecolor{currentstroke}{rgb}{0.000000,0.000000,0.000000}%
\pgfsetstrokecolor{currentstroke}%
\pgfsetdash{}{0pt}%
\pgfpathmoveto{\pgfqpoint{0.592068in}{0.716028in}}%
\pgfpathlineto{\pgfqpoint{5.988008in}{0.716028in}}%
\pgfusepath{stroke}%
\end{pgfscope}%
\begin{pgfscope}%
\pgfsetrectcap%
\pgfsetmiterjoin%
\pgfsetlinewidth{0.803000pt}%
\definecolor{currentstroke}{rgb}{0.000000,0.000000,0.000000}%
\pgfsetstrokecolor{currentstroke}%
\pgfsetdash{}{0pt}%
\pgfpathmoveto{\pgfqpoint{0.592068in}{4.233871in}}%
\pgfpathlineto{\pgfqpoint{5.988008in}{4.233871in}}%
\pgfusepath{stroke}%
\end{pgfscope}%
\begin{pgfscope}%
\definecolor{textcolor}{rgb}{0.000000,0.000000,0.000000}%
\pgfsetstrokecolor{textcolor}%
\pgfsetfillcolor{textcolor}%
\pgftext[x=3.290038in,y=4.317204in,,base]{\color{textcolor}{\rmfamily\fontsize{25.000000}{30.000000}\selectfont\catcode`\^=\active\def^{\ifmmode\sp\else\^{}\fi}\catcode`\%=\active\def
\end{pgfscope}%
\begin{pgfscope}%
\pgfsetbuttcap%
\pgfsetmiterjoin%
\definecolor{currentfill}{rgb}{1.000000,1.000000,1.000000}%
\pgfsetfillcolor{currentfill}%
\pgfsetfillopacity{0.800000}%
\pgfsetlinewidth{1.003750pt}%
\definecolor{currentstroke}{rgb}{0.800000,0.800000,0.800000}%
\pgfsetstrokecolor{currentstroke}%
\pgfsetstrokeopacity{0.800000}%
\pgfsetdash{}{0pt}%
\pgfpathmoveto{\pgfqpoint{0.786512in}{3.616692in}}%
\pgfpathlineto{\pgfqpoint{2.328414in}{3.616692in}}%
\pgfpathquadraticcurveto{\pgfqpoint{2.383970in}{3.616692in}}{\pgfqpoint{2.383970in}{3.672248in}}%
\pgfpathlineto{\pgfqpoint{2.383970in}{4.039427in}}%
\pgfpathquadraticcurveto{\pgfqpoint{2.383970in}{4.094982in}}{\pgfqpoint{2.328414in}{4.094982in}}%
\pgfpathlineto{\pgfqpoint{0.786512in}{4.094982in}}%
\pgfpathquadraticcurveto{\pgfqpoint{0.730957in}{4.094982in}}{\pgfqpoint{0.730957in}{4.039427in}}%
\pgfpathlineto{\pgfqpoint{0.730957in}{3.672248in}}%
\pgfpathquadraticcurveto{\pgfqpoint{0.730957in}{3.616692in}}{\pgfqpoint{0.786512in}{3.616692in}}%
\pgfpathlineto{\pgfqpoint{0.786512in}{3.616692in}}%
\pgfpathclose%
\pgfusepath{stroke,fill}%
\end{pgfscope}%
\begin{pgfscope}%
\pgfsetbuttcap%
\pgfsetroundjoin%
\pgfsetlinewidth{2.007500pt}%
\definecolor{currentstroke}{rgb}{0.501961,0.501961,0.501961}%
\pgfsetstrokecolor{currentstroke}%
\pgfsetdash{{7.400000pt}{3.200000pt}}{0.000000pt}%
\pgfpathmoveto{\pgfqpoint{0.842068in}{3.881055in}}%
\pgfpathlineto{\pgfqpoint{1.119846in}{3.881055in}}%
\pgfpathlineto{\pgfqpoint{1.397623in}{3.881055in}}%
\pgfusepath{stroke}%
\end{pgfscope}%
\begin{pgfscope}%
\definecolor{textcolor}{rgb}{0.000000,0.000000,0.000000}%
\pgfsetstrokecolor{textcolor}%
\pgfsetfillcolor{textcolor}%
\pgftext[x=1.619846in,y=3.783833in,left,base]{\color{textcolor}{\rmfamily\fontsize{20.000000}{24.000000}\selectfont\catcode`\^=\active\def^{\ifmmode\sp\else\^{}\fi}\catcode`\%=\active\def
\end{pgfscope}%
\end{pgfpicture}%
\makeatother%
\endgroup%

%% file: figures/evaluation/Sepsis/Sepsis_PIT_remaining_time_norm_4layer.pgf
\begingroup%
\makeatletter%
\begin{pgfpicture}%
\pgfpathrectangle{\pgfpointorigin}{\pgfqpoint{6.159289in}{4.557174in}}%
\pgfusepath{use as bounding box, clip}%
\begin{pgfscope}%
\pgfsetbuttcap%
\pgfsetmiterjoin%
\definecolor{currentfill}{rgb}{1.000000,1.000000,1.000000}%
\pgfsetfillcolor{currentfill}%
\pgfsetlinewidth{0.000000pt}%
\definecolor{currentstroke}{rgb}{1.000000,1.000000,1.000000}%
\pgfsetstrokecolor{currentstroke}%
\pgfsetdash{}{0pt}%
\pgfpathmoveto{\pgfqpoint{0.000000in}{0.000000in}}%
\pgfpathlineto{\pgfqpoint{6.159289in}{0.000000in}}%
\pgfpathlineto{\pgfqpoint{6.159289in}{4.557174in}}%
\pgfpathlineto{\pgfqpoint{0.000000in}{4.557174in}}%
\pgfpathlineto{\pgfqpoint{0.000000in}{0.000000in}}%
\pgfpathclose%
\pgfusepath{fill}%
\end{pgfscope}%
\begin{pgfscope}%
\pgfsetbuttcap%
\pgfsetmiterjoin%
\definecolor{currentfill}{rgb}{1.000000,1.000000,1.000000}%
\pgfsetfillcolor{currentfill}%
\pgfsetlinewidth{0.000000pt}%
\definecolor{currentstroke}{rgb}{0.000000,0.000000,0.000000}%
\pgfsetstrokecolor{currentstroke}%
\pgfsetstrokeopacity{0.000000}%
\pgfsetdash{}{0pt}%
\pgfpathmoveto{\pgfqpoint{0.592068in}{0.716028in}}%
\pgfpathlineto{\pgfqpoint{5.988008in}{0.716028in}}%
\pgfpathlineto{\pgfqpoint{5.988008in}{4.233871in}}%
\pgfpathlineto{\pgfqpoint{0.592068in}{4.233871in}}%
\pgfpathlineto{\pgfqpoint{0.592068in}{0.716028in}}%
\pgfpathclose%
\pgfusepath{fill}%
\end{pgfscope}%
\begin{pgfscope}%
\pgfpathrectangle{\pgfqpoint{0.592068in}{0.716028in}}{\pgfqpoint{5.395940in}{3.517843in}}%
\pgfusepath{clip}%
\pgfsetbuttcap%
\pgfsetmiterjoin%
\definecolor{currentfill}{rgb}{0.000000,0.000000,1.000000}%
\pgfsetfillcolor{currentfill}%
\pgfsetlinewidth{0.000000pt}%
\definecolor{currentstroke}{rgb}{0.000000,0.000000,0.000000}%
\pgfsetstrokecolor{currentstroke}%
\pgfsetdash{}{0pt}%
\pgfpathmoveto{\pgfqpoint{0.592068in}{0.716028in}}%
\pgfpathlineto{\pgfqpoint{0.778135in}{0.716028in}}%
\pgfpathlineto{\pgfqpoint{0.778135in}{0.716028in}}%
\pgfpathlineto{\pgfqpoint{0.592068in}{0.716028in}}%
\pgfpathlineto{\pgfqpoint{0.592068in}{0.716028in}}%
\pgfpathclose%
\pgfusepath{fill}%
\end{pgfscope}%
\begin{pgfscope}%
\pgfpathrectangle{\pgfqpoint{0.592068in}{0.716028in}}{\pgfqpoint{5.395940in}{3.517843in}}%
\pgfusepath{clip}%
\pgfsetbuttcap%
\pgfsetmiterjoin%
\definecolor{currentfill}{rgb}{0.000000,0.000000,1.000000}%
\pgfsetfillcolor{currentfill}%
\pgfsetlinewidth{0.000000pt}%
\definecolor{currentstroke}{rgb}{0.000000,0.000000,0.000000}%
\pgfsetstrokecolor{currentstroke}%
\pgfsetdash{}{0pt}%
\pgfpathmoveto{\pgfqpoint{0.778135in}{0.716028in}}%
\pgfpathlineto{\pgfqpoint{0.964202in}{0.716028in}}%
\pgfpathlineto{\pgfqpoint{0.964202in}{0.716028in}}%
\pgfpathlineto{\pgfqpoint{0.778135in}{0.716028in}}%
\pgfpathlineto{\pgfqpoint{0.778135in}{0.716028in}}%
\pgfpathclose%
\pgfusepath{fill}%
\end{pgfscope}%
\begin{pgfscope}%
\pgfpathrectangle{\pgfqpoint{0.592068in}{0.716028in}}{\pgfqpoint{5.395940in}{3.517843in}}%
\pgfusepath{clip}%
\pgfsetbuttcap%
\pgfsetmiterjoin%
\definecolor{currentfill}{rgb}{0.000000,0.000000,1.000000}%
\pgfsetfillcolor{currentfill}%
\pgfsetlinewidth{0.000000pt}%
\definecolor{currentstroke}{rgb}{0.000000,0.000000,0.000000}%
\pgfsetstrokecolor{currentstroke}%
\pgfsetdash{}{0pt}%
\pgfpathmoveto{\pgfqpoint{0.964202in}{0.716028in}}%
\pgfpathlineto{\pgfqpoint{1.150269in}{0.716028in}}%
\pgfpathlineto{\pgfqpoint{1.150269in}{0.716028in}}%
\pgfpathlineto{\pgfqpoint{0.964202in}{0.716028in}}%
\pgfpathlineto{\pgfqpoint{0.964202in}{0.716028in}}%
\pgfpathclose%
\pgfusepath{fill}%
\end{pgfscope}%
\begin{pgfscope}%
\pgfpathrectangle{\pgfqpoint{0.592068in}{0.716028in}}{\pgfqpoint{5.395940in}{3.517843in}}%
\pgfusepath{clip}%
\pgfsetbuttcap%
\pgfsetmiterjoin%
\definecolor{currentfill}{rgb}{0.000000,0.000000,1.000000}%
\pgfsetfillcolor{currentfill}%
\pgfsetlinewidth{0.000000pt}%
\definecolor{currentstroke}{rgb}{0.000000,0.000000,0.000000}%
\pgfsetstrokecolor{currentstroke}%
\pgfsetdash{}{0pt}%
\pgfpathmoveto{\pgfqpoint{1.150269in}{0.716028in}}%
\pgfpathlineto{\pgfqpoint{1.336335in}{0.716028in}}%
\pgfpathlineto{\pgfqpoint{1.336335in}{0.716028in}}%
\pgfpathlineto{\pgfqpoint{1.150269in}{0.716028in}}%
\pgfpathlineto{\pgfqpoint{1.150269in}{0.716028in}}%
\pgfpathclose%
\pgfusepath{fill}%
\end{pgfscope}%
\begin{pgfscope}%
\pgfpathrectangle{\pgfqpoint{0.592068in}{0.716028in}}{\pgfqpoint{5.395940in}{3.517843in}}%
\pgfusepath{clip}%
\pgfsetbuttcap%
\pgfsetmiterjoin%
\definecolor{currentfill}{rgb}{0.000000,0.000000,1.000000}%
\pgfsetfillcolor{currentfill}%
\pgfsetlinewidth{0.000000pt}%
\definecolor{currentstroke}{rgb}{0.000000,0.000000,0.000000}%
\pgfsetstrokecolor{currentstroke}%
\pgfsetdash{}{0pt}%
\pgfpathmoveto{\pgfqpoint{1.336335in}{0.716028in}}%
\pgfpathlineto{\pgfqpoint{1.522402in}{0.716028in}}%
\pgfpathlineto{\pgfqpoint{1.522402in}{0.716028in}}%
\pgfpathlineto{\pgfqpoint{1.336335in}{0.716028in}}%
\pgfpathlineto{\pgfqpoint{1.336335in}{0.716028in}}%
\pgfpathclose%
\pgfusepath{fill}%
\end{pgfscope}%
\begin{pgfscope}%
\pgfpathrectangle{\pgfqpoint{0.592068in}{0.716028in}}{\pgfqpoint{5.395940in}{3.517843in}}%
\pgfusepath{clip}%
\pgfsetbuttcap%
\pgfsetmiterjoin%
\definecolor{currentfill}{rgb}{0.000000,0.000000,1.000000}%
\pgfsetfillcolor{currentfill}%
\pgfsetlinewidth{0.000000pt}%
\definecolor{currentstroke}{rgb}{0.000000,0.000000,0.000000}%
\pgfsetstrokecolor{currentstroke}%
\pgfsetdash{}{0pt}%
\pgfpathmoveto{\pgfqpoint{1.522402in}{0.716028in}}%
\pgfpathlineto{\pgfqpoint{1.708469in}{0.716028in}}%
\pgfpathlineto{\pgfqpoint{1.708469in}{0.716028in}}%
\pgfpathlineto{\pgfqpoint{1.522402in}{0.716028in}}%
\pgfpathlineto{\pgfqpoint{1.522402in}{0.716028in}}%
\pgfpathclose%
\pgfusepath{fill}%
\end{pgfscope}%
\begin{pgfscope}%
\pgfpathrectangle{\pgfqpoint{0.592068in}{0.716028in}}{\pgfqpoint{5.395940in}{3.517843in}}%
\pgfusepath{clip}%
\pgfsetbuttcap%
\pgfsetmiterjoin%
\definecolor{currentfill}{rgb}{0.000000,0.000000,1.000000}%
\pgfsetfillcolor{currentfill}%
\pgfsetlinewidth{0.000000pt}%
\definecolor{currentstroke}{rgb}{0.000000,0.000000,0.000000}%
\pgfsetstrokecolor{currentstroke}%
\pgfsetdash{}{0pt}%
\pgfpathmoveto{\pgfqpoint{1.708469in}{0.716028in}}%
\pgfpathlineto{\pgfqpoint{1.894536in}{0.716028in}}%
\pgfpathlineto{\pgfqpoint{1.894536in}{0.716028in}}%
\pgfpathlineto{\pgfqpoint{1.708469in}{0.716028in}}%
\pgfpathlineto{\pgfqpoint{1.708469in}{0.716028in}}%
\pgfpathclose%
\pgfusepath{fill}%
\end{pgfscope}%
\begin{pgfscope}%
\pgfpathrectangle{\pgfqpoint{0.592068in}{0.716028in}}{\pgfqpoint{5.395940in}{3.517843in}}%
\pgfusepath{clip}%
\pgfsetbuttcap%
\pgfsetmiterjoin%
\definecolor{currentfill}{rgb}{0.000000,0.000000,1.000000}%
\pgfsetfillcolor{currentfill}%
\pgfsetlinewidth{0.000000pt}%
\definecolor{currentstroke}{rgb}{0.000000,0.000000,0.000000}%
\pgfsetstrokecolor{currentstroke}%
\pgfsetdash{}{0pt}%
\pgfpathmoveto{\pgfqpoint{1.894536in}{0.716028in}}%
\pgfpathlineto{\pgfqpoint{2.080603in}{0.716028in}}%
\pgfpathlineto{\pgfqpoint{2.080603in}{0.870510in}}%
\pgfpathlineto{\pgfqpoint{1.894536in}{0.870510in}}%
\pgfpathlineto{\pgfqpoint{1.894536in}{0.716028in}}%
\pgfpathclose%
\pgfusepath{fill}%
\end{pgfscope}%
\begin{pgfscope}%
\pgfpathrectangle{\pgfqpoint{0.592068in}{0.716028in}}{\pgfqpoint{5.395940in}{3.517843in}}%
\pgfusepath{clip}%
\pgfsetbuttcap%
\pgfsetmiterjoin%
\definecolor{currentfill}{rgb}{0.000000,0.000000,1.000000}%
\pgfsetfillcolor{currentfill}%
\pgfsetlinewidth{0.000000pt}%
\definecolor{currentstroke}{rgb}{0.000000,0.000000,0.000000}%
\pgfsetstrokecolor{currentstroke}%
\pgfsetdash{}{0pt}%
\pgfpathmoveto{\pgfqpoint{2.080603in}{0.716028in}}%
\pgfpathlineto{\pgfqpoint{2.266670in}{0.716028in}}%
\pgfpathlineto{\pgfqpoint{2.266670in}{2.328433in}}%
\pgfpathlineto{\pgfqpoint{2.080603in}{2.328433in}}%
\pgfpathlineto{\pgfqpoint{2.080603in}{0.716028in}}%
\pgfpathclose%
\pgfusepath{fill}%
\end{pgfscope}%
\begin{pgfscope}%
\pgfpathrectangle{\pgfqpoint{0.592068in}{0.716028in}}{\pgfqpoint{5.395940in}{3.517843in}}%
\pgfusepath{clip}%
\pgfsetbuttcap%
\pgfsetmiterjoin%
\definecolor{currentfill}{rgb}{0.000000,0.000000,1.000000}%
\pgfsetfillcolor{currentfill}%
\pgfsetlinewidth{0.000000pt}%
\definecolor{currentstroke}{rgb}{0.000000,0.000000,0.000000}%
\pgfsetstrokecolor{currentstroke}%
\pgfsetdash{}{0pt}%
\pgfpathmoveto{\pgfqpoint{2.266670in}{0.716028in}}%
\pgfpathlineto{\pgfqpoint{2.452737in}{0.716028in}}%
\pgfpathlineto{\pgfqpoint{2.452737in}{2.975326in}}%
\pgfpathlineto{\pgfqpoint{2.266670in}{2.975326in}}%
\pgfpathlineto{\pgfqpoint{2.266670in}{0.716028in}}%
\pgfpathclose%
\pgfusepath{fill}%
\end{pgfscope}%
\begin{pgfscope}%
\pgfpathrectangle{\pgfqpoint{0.592068in}{0.716028in}}{\pgfqpoint{5.395940in}{3.517843in}}%
\pgfusepath{clip}%
\pgfsetbuttcap%
\pgfsetmiterjoin%
\definecolor{currentfill}{rgb}{0.000000,0.000000,1.000000}%
\pgfsetfillcolor{currentfill}%
\pgfsetlinewidth{0.000000pt}%
\definecolor{currentstroke}{rgb}{0.000000,0.000000,0.000000}%
\pgfsetstrokecolor{currentstroke}%
\pgfsetdash{}{0pt}%
\pgfpathmoveto{\pgfqpoint{2.452737in}{0.716028in}}%
\pgfpathlineto{\pgfqpoint{2.638804in}{0.716028in}}%
\pgfpathlineto{\pgfqpoint{2.638804in}{3.670495in}}%
\pgfpathlineto{\pgfqpoint{2.452737in}{3.670495in}}%
\pgfpathlineto{\pgfqpoint{2.452737in}{0.716028in}}%
\pgfpathclose%
\pgfusepath{fill}%
\end{pgfscope}%
\begin{pgfscope}%
\pgfpathrectangle{\pgfqpoint{0.592068in}{0.716028in}}{\pgfqpoint{5.395940in}{3.517843in}}%
\pgfusepath{clip}%
\pgfsetbuttcap%
\pgfsetmiterjoin%
\definecolor{currentfill}{rgb}{0.000000,0.000000,1.000000}%
\pgfsetfillcolor{currentfill}%
\pgfsetlinewidth{0.000000pt}%
\definecolor{currentstroke}{rgb}{0.000000,0.000000,0.000000}%
\pgfsetstrokecolor{currentstroke}%
\pgfsetdash{}{0pt}%
\pgfpathmoveto{\pgfqpoint{2.638804in}{0.716028in}}%
\pgfpathlineto{\pgfqpoint{2.824871in}{0.716028in}}%
\pgfpathlineto{\pgfqpoint{2.824871in}{1.498093in}}%
\pgfpathlineto{\pgfqpoint{2.638804in}{1.498093in}}%
\pgfpathlineto{\pgfqpoint{2.638804in}{0.716028in}}%
\pgfpathclose%
\pgfusepath{fill}%
\end{pgfscope}%
\begin{pgfscope}%
\pgfpathrectangle{\pgfqpoint{0.592068in}{0.716028in}}{\pgfqpoint{5.395940in}{3.517843in}}%
\pgfusepath{clip}%
\pgfsetbuttcap%
\pgfsetmiterjoin%
\definecolor{currentfill}{rgb}{0.000000,0.000000,1.000000}%
\pgfsetfillcolor{currentfill}%
\pgfsetlinewidth{0.000000pt}%
\definecolor{currentstroke}{rgb}{0.000000,0.000000,0.000000}%
\pgfsetstrokecolor{currentstroke}%
\pgfsetdash{}{0pt}%
\pgfpathmoveto{\pgfqpoint{2.824871in}{0.716028in}}%
\pgfpathlineto{\pgfqpoint{3.010937in}{0.716028in}}%
\pgfpathlineto{\pgfqpoint{3.010937in}{1.594644in}}%
\pgfpathlineto{\pgfqpoint{2.824871in}{1.594644in}}%
\pgfpathlineto{\pgfqpoint{2.824871in}{0.716028in}}%
\pgfpathclose%
\pgfusepath{fill}%
\end{pgfscope}%
\begin{pgfscope}%
\pgfpathrectangle{\pgfqpoint{0.592068in}{0.716028in}}{\pgfqpoint{5.395940in}{3.517843in}}%
\pgfusepath{clip}%
\pgfsetbuttcap%
\pgfsetmiterjoin%
\definecolor{currentfill}{rgb}{0.000000,0.000000,1.000000}%
\pgfsetfillcolor{currentfill}%
\pgfsetlinewidth{0.000000pt}%
\definecolor{currentstroke}{rgb}{0.000000,0.000000,0.000000}%
\pgfsetstrokecolor{currentstroke}%
\pgfsetdash{}{0pt}%
\pgfpathmoveto{\pgfqpoint{3.010937in}{0.716028in}}%
\pgfpathlineto{\pgfqpoint{3.197004in}{0.716028in}}%
\pgfpathlineto{\pgfqpoint{3.197004in}{1.884298in}}%
\pgfpathlineto{\pgfqpoint{3.010937in}{1.884298in}}%
\pgfpathlineto{\pgfqpoint{3.010937in}{0.716028in}}%
\pgfpathclose%
\pgfusepath{fill}%
\end{pgfscope}%
\begin{pgfscope}%
\pgfpathrectangle{\pgfqpoint{0.592068in}{0.716028in}}{\pgfqpoint{5.395940in}{3.517843in}}%
\pgfusepath{clip}%
\pgfsetbuttcap%
\pgfsetmiterjoin%
\definecolor{currentfill}{rgb}{0.000000,0.000000,1.000000}%
\pgfsetfillcolor{currentfill}%
\pgfsetlinewidth{0.000000pt}%
\definecolor{currentstroke}{rgb}{0.000000,0.000000,0.000000}%
\pgfsetstrokecolor{currentstroke}%
\pgfsetdash{}{0pt}%
\pgfpathmoveto{\pgfqpoint{3.197004in}{0.716028in}}%
\pgfpathlineto{\pgfqpoint{3.383071in}{0.716028in}}%
\pgfpathlineto{\pgfqpoint{3.383071in}{0.986372in}}%
\pgfpathlineto{\pgfqpoint{3.197004in}{0.986372in}}%
\pgfpathlineto{\pgfqpoint{3.197004in}{0.716028in}}%
\pgfpathclose%
\pgfusepath{fill}%
\end{pgfscope}%
\begin{pgfscope}%
\pgfpathrectangle{\pgfqpoint{0.592068in}{0.716028in}}{\pgfqpoint{5.395940in}{3.517843in}}%
\pgfusepath{clip}%
\pgfsetbuttcap%
\pgfsetmiterjoin%
\definecolor{currentfill}{rgb}{0.000000,0.000000,1.000000}%
\pgfsetfillcolor{currentfill}%
\pgfsetlinewidth{0.000000pt}%
\definecolor{currentstroke}{rgb}{0.000000,0.000000,0.000000}%
\pgfsetstrokecolor{currentstroke}%
\pgfsetdash{}{0pt}%
\pgfpathmoveto{\pgfqpoint{3.383071in}{0.716028in}}%
\pgfpathlineto{\pgfqpoint{3.569138in}{0.716028in}}%
\pgfpathlineto{\pgfqpoint{3.569138in}{0.793269in}}%
\pgfpathlineto{\pgfqpoint{3.383071in}{0.793269in}}%
\pgfpathlineto{\pgfqpoint{3.383071in}{0.716028in}}%
\pgfpathclose%
\pgfusepath{fill}%
\end{pgfscope}%
\begin{pgfscope}%
\pgfpathrectangle{\pgfqpoint{0.592068in}{0.716028in}}{\pgfqpoint{5.395940in}{3.517843in}}%
\pgfusepath{clip}%
\pgfsetbuttcap%
\pgfsetmiterjoin%
\definecolor{currentfill}{rgb}{0.000000,0.000000,1.000000}%
\pgfsetfillcolor{currentfill}%
\pgfsetlinewidth{0.000000pt}%
\definecolor{currentstroke}{rgb}{0.000000,0.000000,0.000000}%
\pgfsetstrokecolor{currentstroke}%
\pgfsetdash{}{0pt}%
\pgfpathmoveto{\pgfqpoint{3.569138in}{0.716028in}}%
\pgfpathlineto{\pgfqpoint{3.755205in}{0.716028in}}%
\pgfpathlineto{\pgfqpoint{3.755205in}{0.947751in}}%
\pgfpathlineto{\pgfqpoint{3.569138in}{0.947751in}}%
\pgfpathlineto{\pgfqpoint{3.569138in}{0.716028in}}%
\pgfpathclose%
\pgfusepath{fill}%
\end{pgfscope}%
\begin{pgfscope}%
\pgfpathrectangle{\pgfqpoint{0.592068in}{0.716028in}}{\pgfqpoint{5.395940in}{3.517843in}}%
\pgfusepath{clip}%
\pgfsetbuttcap%
\pgfsetmiterjoin%
\definecolor{currentfill}{rgb}{0.000000,0.000000,1.000000}%
\pgfsetfillcolor{currentfill}%
\pgfsetlinewidth{0.000000pt}%
\definecolor{currentstroke}{rgb}{0.000000,0.000000,0.000000}%
\pgfsetstrokecolor{currentstroke}%
\pgfsetdash{}{0pt}%
\pgfpathmoveto{\pgfqpoint{3.755205in}{0.716028in}}%
\pgfpathlineto{\pgfqpoint{3.941272in}{0.716028in}}%
\pgfpathlineto{\pgfqpoint{3.941272in}{1.198784in}}%
\pgfpathlineto{\pgfqpoint{3.755205in}{1.198784in}}%
\pgfpathlineto{\pgfqpoint{3.755205in}{0.716028in}}%
\pgfpathclose%
\pgfusepath{fill}%
\end{pgfscope}%
\begin{pgfscope}%
\pgfpathrectangle{\pgfqpoint{0.592068in}{0.716028in}}{\pgfqpoint{5.395940in}{3.517843in}}%
\pgfusepath{clip}%
\pgfsetbuttcap%
\pgfsetmiterjoin%
\definecolor{currentfill}{rgb}{0.000000,0.000000,1.000000}%
\pgfsetfillcolor{currentfill}%
\pgfsetlinewidth{0.000000pt}%
\definecolor{currentstroke}{rgb}{0.000000,0.000000,0.000000}%
\pgfsetstrokecolor{currentstroke}%
\pgfsetdash{}{0pt}%
\pgfpathmoveto{\pgfqpoint{3.941272in}{0.716028in}}%
\pgfpathlineto{\pgfqpoint{4.127339in}{0.716028in}}%
\pgfpathlineto{\pgfqpoint{4.127339in}{0.967061in}}%
\pgfpathlineto{\pgfqpoint{3.941272in}{0.967061in}}%
\pgfpathlineto{\pgfqpoint{3.941272in}{0.716028in}}%
\pgfpathclose%
\pgfusepath{fill}%
\end{pgfscope}%
\begin{pgfscope}%
\pgfpathrectangle{\pgfqpoint{0.592068in}{0.716028in}}{\pgfqpoint{5.395940in}{3.517843in}}%
\pgfusepath{clip}%
\pgfsetbuttcap%
\pgfsetmiterjoin%
\definecolor{currentfill}{rgb}{0.000000,0.000000,1.000000}%
\pgfsetfillcolor{currentfill}%
\pgfsetlinewidth{0.000000pt}%
\definecolor{currentstroke}{rgb}{0.000000,0.000000,0.000000}%
\pgfsetstrokecolor{currentstroke}%
\pgfsetdash{}{0pt}%
\pgfpathmoveto{\pgfqpoint{4.127339in}{0.716028in}}%
\pgfpathlineto{\pgfqpoint{4.313406in}{0.716028in}}%
\pgfpathlineto{\pgfqpoint{4.313406in}{0.938096in}}%
\pgfpathlineto{\pgfqpoint{4.127339in}{0.938096in}}%
\pgfpathlineto{\pgfqpoint{4.127339in}{0.716028in}}%
\pgfpathclose%
\pgfusepath{fill}%
\end{pgfscope}%
\begin{pgfscope}%
\pgfpathrectangle{\pgfqpoint{0.592068in}{0.716028in}}{\pgfqpoint{5.395940in}{3.517843in}}%
\pgfusepath{clip}%
\pgfsetbuttcap%
\pgfsetmiterjoin%
\definecolor{currentfill}{rgb}{0.000000,0.000000,1.000000}%
\pgfsetfillcolor{currentfill}%
\pgfsetlinewidth{0.000000pt}%
\definecolor{currentstroke}{rgb}{0.000000,0.000000,0.000000}%
\pgfsetstrokecolor{currentstroke}%
\pgfsetdash{}{0pt}%
\pgfpathmoveto{\pgfqpoint{4.313406in}{0.716028in}}%
\pgfpathlineto{\pgfqpoint{4.499473in}{0.716028in}}%
\pgfpathlineto{\pgfqpoint{4.499473in}{1.053957in}}%
\pgfpathlineto{\pgfqpoint{4.313406in}{1.053957in}}%
\pgfpathlineto{\pgfqpoint{4.313406in}{0.716028in}}%
\pgfpathclose%
\pgfusepath{fill}%
\end{pgfscope}%
\begin{pgfscope}%
\pgfpathrectangle{\pgfqpoint{0.592068in}{0.716028in}}{\pgfqpoint{5.395940in}{3.517843in}}%
\pgfusepath{clip}%
\pgfsetbuttcap%
\pgfsetmiterjoin%
\definecolor{currentfill}{rgb}{0.000000,0.000000,1.000000}%
\pgfsetfillcolor{currentfill}%
\pgfsetlinewidth{0.000000pt}%
\definecolor{currentstroke}{rgb}{0.000000,0.000000,0.000000}%
\pgfsetstrokecolor{currentstroke}%
\pgfsetdash{}{0pt}%
\pgfpathmoveto{\pgfqpoint{4.499473in}{0.716028in}}%
\pgfpathlineto{\pgfqpoint{4.685539in}{0.716028in}}%
\pgfpathlineto{\pgfqpoint{4.685539in}{0.773959in}}%
\pgfpathlineto{\pgfqpoint{4.499473in}{0.773959in}}%
\pgfpathlineto{\pgfqpoint{4.499473in}{0.716028in}}%
\pgfpathclose%
\pgfusepath{fill}%
\end{pgfscope}%
\begin{pgfscope}%
\pgfpathrectangle{\pgfqpoint{0.592068in}{0.716028in}}{\pgfqpoint{5.395940in}{3.517843in}}%
\pgfusepath{clip}%
\pgfsetbuttcap%
\pgfsetmiterjoin%
\definecolor{currentfill}{rgb}{0.000000,0.000000,1.000000}%
\pgfsetfillcolor{currentfill}%
\pgfsetlinewidth{0.000000pt}%
\definecolor{currentstroke}{rgb}{0.000000,0.000000,0.000000}%
\pgfsetstrokecolor{currentstroke}%
\pgfsetdash{}{0pt}%
\pgfpathmoveto{\pgfqpoint{4.685539in}{0.716028in}}%
\pgfpathlineto{\pgfqpoint{4.871606in}{0.716028in}}%
\pgfpathlineto{\pgfqpoint{4.871606in}{1.092578in}}%
\pgfpathlineto{\pgfqpoint{4.685539in}{1.092578in}}%
\pgfpathlineto{\pgfqpoint{4.685539in}{0.716028in}}%
\pgfpathclose%
\pgfusepath{fill}%
\end{pgfscope}%
\begin{pgfscope}%
\pgfpathrectangle{\pgfqpoint{0.592068in}{0.716028in}}{\pgfqpoint{5.395940in}{3.517843in}}%
\pgfusepath{clip}%
\pgfsetbuttcap%
\pgfsetmiterjoin%
\definecolor{currentfill}{rgb}{0.000000,0.000000,1.000000}%
\pgfsetfillcolor{currentfill}%
\pgfsetlinewidth{0.000000pt}%
\definecolor{currentstroke}{rgb}{0.000000,0.000000,0.000000}%
\pgfsetstrokecolor{currentstroke}%
\pgfsetdash{}{0pt}%
\pgfpathmoveto{\pgfqpoint{4.871606in}{0.716028in}}%
\pgfpathlineto{\pgfqpoint{5.057673in}{0.716028in}}%
\pgfpathlineto{\pgfqpoint{5.057673in}{1.005682in}}%
\pgfpathlineto{\pgfqpoint{4.871606in}{1.005682in}}%
\pgfpathlineto{\pgfqpoint{4.871606in}{0.716028in}}%
\pgfpathclose%
\pgfusepath{fill}%
\end{pgfscope}%
\begin{pgfscope}%
\pgfpathrectangle{\pgfqpoint{0.592068in}{0.716028in}}{\pgfqpoint{5.395940in}{3.517843in}}%
\pgfusepath{clip}%
\pgfsetbuttcap%
\pgfsetmiterjoin%
\definecolor{currentfill}{rgb}{0.000000,0.000000,1.000000}%
\pgfsetfillcolor{currentfill}%
\pgfsetlinewidth{0.000000pt}%
\definecolor{currentstroke}{rgb}{0.000000,0.000000,0.000000}%
\pgfsetstrokecolor{currentstroke}%
\pgfsetdash{}{0pt}%
\pgfpathmoveto{\pgfqpoint{5.057673in}{0.716028in}}%
\pgfpathlineto{\pgfqpoint{5.243740in}{0.716028in}}%
\pgfpathlineto{\pgfqpoint{5.243740in}{0.880165in}}%
\pgfpathlineto{\pgfqpoint{5.057673in}{0.880165in}}%
\pgfpathlineto{\pgfqpoint{5.057673in}{0.716028in}}%
\pgfpathclose%
\pgfusepath{fill}%
\end{pgfscope}%
\begin{pgfscope}%
\pgfpathrectangle{\pgfqpoint{0.592068in}{0.716028in}}{\pgfqpoint{5.395940in}{3.517843in}}%
\pgfusepath{clip}%
\pgfsetbuttcap%
\pgfsetmiterjoin%
\definecolor{currentfill}{rgb}{0.000000,0.000000,1.000000}%
\pgfsetfillcolor{currentfill}%
\pgfsetlinewidth{0.000000pt}%
\definecolor{currentstroke}{rgb}{0.000000,0.000000,0.000000}%
\pgfsetstrokecolor{currentstroke}%
\pgfsetdash{}{0pt}%
\pgfpathmoveto{\pgfqpoint{5.243740in}{0.716028in}}%
\pgfpathlineto{\pgfqpoint{5.429807in}{0.716028in}}%
\pgfpathlineto{\pgfqpoint{5.429807in}{0.735338in}}%
\pgfpathlineto{\pgfqpoint{5.243740in}{0.735338in}}%
\pgfpathlineto{\pgfqpoint{5.243740in}{0.716028in}}%
\pgfpathclose%
\pgfusepath{fill}%
\end{pgfscope}%
\begin{pgfscope}%
\pgfpathrectangle{\pgfqpoint{0.592068in}{0.716028in}}{\pgfqpoint{5.395940in}{3.517843in}}%
\pgfusepath{clip}%
\pgfsetbuttcap%
\pgfsetmiterjoin%
\definecolor{currentfill}{rgb}{0.000000,0.000000,1.000000}%
\pgfsetfillcolor{currentfill}%
\pgfsetlinewidth{0.000000pt}%
\definecolor{currentstroke}{rgb}{0.000000,0.000000,0.000000}%
\pgfsetstrokecolor{currentstroke}%
\pgfsetdash{}{0pt}%
\pgfpathmoveto{\pgfqpoint{5.429807in}{0.716028in}}%
\pgfpathlineto{\pgfqpoint{5.615874in}{0.716028in}}%
\pgfpathlineto{\pgfqpoint{5.615874in}{0.822235in}}%
\pgfpathlineto{\pgfqpoint{5.429807in}{0.822235in}}%
\pgfpathlineto{\pgfqpoint{5.429807in}{0.716028in}}%
\pgfpathclose%
\pgfusepath{fill}%
\end{pgfscope}%
\begin{pgfscope}%
\pgfpathrectangle{\pgfqpoint{0.592068in}{0.716028in}}{\pgfqpoint{5.395940in}{3.517843in}}%
\pgfusepath{clip}%
\pgfsetbuttcap%
\pgfsetmiterjoin%
\definecolor{currentfill}{rgb}{0.000000,0.000000,1.000000}%
\pgfsetfillcolor{currentfill}%
\pgfsetlinewidth{0.000000pt}%
\definecolor{currentstroke}{rgb}{0.000000,0.000000,0.000000}%
\pgfsetstrokecolor{currentstroke}%
\pgfsetdash{}{0pt}%
\pgfpathmoveto{\pgfqpoint{5.615874in}{0.716028in}}%
\pgfpathlineto{\pgfqpoint{5.801941in}{0.716028in}}%
\pgfpathlineto{\pgfqpoint{5.801941in}{1.082923in}}%
\pgfpathlineto{\pgfqpoint{5.615874in}{1.082923in}}%
\pgfpathlineto{\pgfqpoint{5.615874in}{0.716028in}}%
\pgfpathclose%
\pgfusepath{fill}%
\end{pgfscope}%
\begin{pgfscope}%
\pgfpathrectangle{\pgfqpoint{0.592068in}{0.716028in}}{\pgfqpoint{5.395940in}{3.517843in}}%
\pgfusepath{clip}%
\pgfsetbuttcap%
\pgfsetmiterjoin%
\definecolor{currentfill}{rgb}{0.000000,0.000000,1.000000}%
\pgfsetfillcolor{currentfill}%
\pgfsetlinewidth{0.000000pt}%
\definecolor{currentstroke}{rgb}{0.000000,0.000000,0.000000}%
\pgfsetstrokecolor{currentstroke}%
\pgfsetdash{}{0pt}%
\pgfpathmoveto{\pgfqpoint{5.801941in}{0.716028in}}%
\pgfpathlineto{\pgfqpoint{5.988008in}{0.716028in}}%
\pgfpathlineto{\pgfqpoint{5.988008in}{4.066355in}}%
\pgfpathlineto{\pgfqpoint{5.801941in}{4.066355in}}%
\pgfpathlineto{\pgfqpoint{5.801941in}{0.716028in}}%
\pgfpathclose%
\pgfusepath{fill}%
\end{pgfscope}%
\begin{pgfscope}%
\pgfsetbuttcap%
\pgfsetroundjoin%
\definecolor{currentfill}{rgb}{0.000000,0.000000,0.000000}%
\pgfsetfillcolor{currentfill}%
\pgfsetlinewidth{0.200750pt}%
\definecolor{currentstroke}{rgb}{0.000000,0.000000,0.000000}%
\pgfsetstrokecolor{currentstroke}%
\pgfsetdash{}{0pt}%
\pgfsys@defobject{currentmarker}{\pgfqpoint{0.000000in}{-0.048611in}}{\pgfqpoint{0.000000in}{0.000000in}}{%
\pgfpathmoveto{\pgfqpoint{0.000000in}{0.000000in}}%
\pgfpathlineto{\pgfqpoint{0.000000in}{-0.048611in}}%
\pgfusepath{stroke,fill}%
}%
\begin{pgfscope}%
\pgfsys@transformshift{0.592068in}{0.716028in}%
\pgfsys@useobject{currentmarker}{}%
\end{pgfscope}%
\end{pgfscope}%
\begin{pgfscope}%
\definecolor{textcolor}{rgb}{0.000000,0.000000,0.000000}%
\pgfsetstrokecolor{textcolor}%
\pgfsetfillcolor{textcolor}%
\pgftext[x=0.592068in,y=0.618806in,,top]{\color{textcolor}{\rmfamily\fontsize{20.000000}{24.000000}\selectfont\catcode`\^=\active\def^{\ifmmode\sp\else\^{}\fi}\catcode`\%=\active\def
\end{pgfscope}%
\begin{pgfscope}%
\pgfsetbuttcap%
\pgfsetroundjoin%
\definecolor{currentfill}{rgb}{0.000000,0.000000,0.000000}%
\pgfsetfillcolor{currentfill}%
\pgfsetlinewidth{0.200750pt}%
\definecolor{currentstroke}{rgb}{0.000000,0.000000,0.000000}%
\pgfsetstrokecolor{currentstroke}%
\pgfsetdash{}{0pt}%
\pgfsys@defobject{currentmarker}{\pgfqpoint{0.000000in}{-0.048611in}}{\pgfqpoint{0.000000in}{0.000000in}}{%
\pgfpathmoveto{\pgfqpoint{0.000000in}{0.000000in}}%
\pgfpathlineto{\pgfqpoint{0.000000in}{-0.048611in}}%
\pgfusepath{stroke,fill}%
}%
\begin{pgfscope}%
\pgfsys@transformshift{1.671256in}{0.716028in}%
\pgfsys@useobject{currentmarker}{}%
\end{pgfscope}%
\end{pgfscope}%
\begin{pgfscope}%
\definecolor{textcolor}{rgb}{0.000000,0.000000,0.000000}%
\pgfsetstrokecolor{textcolor}%
\pgfsetfillcolor{textcolor}%
\pgftext[x=1.671256in,y=0.618806in,,top]{\color{textcolor}{\rmfamily\fontsize{20.000000}{24.000000}\selectfont\catcode`\^=\active\def^{\ifmmode\sp\else\^{}\fi}\catcode`\%=\active\def
\end{pgfscope}%
\begin{pgfscope}%
\pgfsetbuttcap%
\pgfsetroundjoin%
\definecolor{currentfill}{rgb}{0.000000,0.000000,0.000000}%
\pgfsetfillcolor{currentfill}%
\pgfsetlinewidth{0.200750pt}%
\definecolor{currentstroke}{rgb}{0.000000,0.000000,0.000000}%
\pgfsetstrokecolor{currentstroke}%
\pgfsetdash{}{0pt}%
\pgfsys@defobject{currentmarker}{\pgfqpoint{0.000000in}{-0.048611in}}{\pgfqpoint{0.000000in}{0.000000in}}{%
\pgfpathmoveto{\pgfqpoint{0.000000in}{0.000000in}}%
\pgfpathlineto{\pgfqpoint{0.000000in}{-0.048611in}}%
\pgfusepath{stroke,fill}%
}%
\begin{pgfscope}%
\pgfsys@transformshift{2.750444in}{0.716028in}%
\pgfsys@useobject{currentmarker}{}%
\end{pgfscope}%
\end{pgfscope}%
\begin{pgfscope}%
\definecolor{textcolor}{rgb}{0.000000,0.000000,0.000000}%
\pgfsetstrokecolor{textcolor}%
\pgfsetfillcolor{textcolor}%
\pgftext[x=2.750444in,y=0.618806in,,top]{\color{textcolor}{\rmfamily\fontsize{20.000000}{24.000000}\selectfont\catcode`\^=\active\def^{\ifmmode\sp\else\^{}\fi}\catcode`\%=\active\def
\end{pgfscope}%
\begin{pgfscope}%
\pgfsetbuttcap%
\pgfsetroundjoin%
\definecolor{currentfill}{rgb}{0.000000,0.000000,0.000000}%
\pgfsetfillcolor{currentfill}%
\pgfsetlinewidth{0.200750pt}%
\definecolor{currentstroke}{rgb}{0.000000,0.000000,0.000000}%
\pgfsetstrokecolor{currentstroke}%
\pgfsetdash{}{0pt}%
\pgfsys@defobject{currentmarker}{\pgfqpoint{0.000000in}{-0.048611in}}{\pgfqpoint{0.000000in}{0.000000in}}{%
\pgfpathmoveto{\pgfqpoint{0.000000in}{0.000000in}}%
\pgfpathlineto{\pgfqpoint{0.000000in}{-0.048611in}}%
\pgfusepath{stroke,fill}%
}%
\begin{pgfscope}%
\pgfsys@transformshift{3.829632in}{0.716028in}%
\pgfsys@useobject{currentmarker}{}%
\end{pgfscope}%
\end{pgfscope}%
\begin{pgfscope}%
\definecolor{textcolor}{rgb}{0.000000,0.000000,0.000000}%
\pgfsetstrokecolor{textcolor}%
\pgfsetfillcolor{textcolor}%
\pgftext[x=3.829632in,y=0.618806in,,top]{\color{textcolor}{\rmfamily\fontsize{20.000000}{24.000000}\selectfont\catcode`\^=\active\def^{\ifmmode\sp\else\^{}\fi}\catcode`\%=\active\def
\end{pgfscope}%
\begin{pgfscope}%
\pgfsetbuttcap%
\pgfsetroundjoin%
\definecolor{currentfill}{rgb}{0.000000,0.000000,0.000000}%
\pgfsetfillcolor{currentfill}%
\pgfsetlinewidth{0.200750pt}%
\definecolor{currentstroke}{rgb}{0.000000,0.000000,0.000000}%
\pgfsetstrokecolor{currentstroke}%
\pgfsetdash{}{0pt}%
\pgfsys@defobject{currentmarker}{\pgfqpoint{0.000000in}{-0.048611in}}{\pgfqpoint{0.000000in}{0.000000in}}{%
\pgfpathmoveto{\pgfqpoint{0.000000in}{0.000000in}}%
\pgfpathlineto{\pgfqpoint{0.000000in}{-0.048611in}}%
\pgfusepath{stroke,fill}%
}%
\begin{pgfscope}%
\pgfsys@transformshift{4.908820in}{0.716028in}%
\pgfsys@useobject{currentmarker}{}%
\end{pgfscope}%
\end{pgfscope}%
\begin{pgfscope}%
\definecolor{textcolor}{rgb}{0.000000,0.000000,0.000000}%
\pgfsetstrokecolor{textcolor}%
\pgfsetfillcolor{textcolor}%
\pgftext[x=4.908820in,y=0.618806in,,top]{\color{textcolor}{\rmfamily\fontsize{20.000000}{24.000000}\selectfont\catcode`\^=\active\def^{\ifmmode\sp\else\^{}\fi}\catcode`\%=\active\def
\end{pgfscope}%
\begin{pgfscope}%
\pgfsetbuttcap%
\pgfsetroundjoin%
\definecolor{currentfill}{rgb}{0.000000,0.000000,0.000000}%
\pgfsetfillcolor{currentfill}%
\pgfsetlinewidth{0.200750pt}%
\definecolor{currentstroke}{rgb}{0.000000,0.000000,0.000000}%
\pgfsetstrokecolor{currentstroke}%
\pgfsetdash{}{0pt}%
\pgfsys@defobject{currentmarker}{\pgfqpoint{0.000000in}{-0.048611in}}{\pgfqpoint{0.000000in}{0.000000in}}{%
\pgfpathmoveto{\pgfqpoint{0.000000in}{0.000000in}}%
\pgfpathlineto{\pgfqpoint{0.000000in}{-0.048611in}}%
\pgfusepath{stroke,fill}%
}%
\begin{pgfscope}%
\pgfsys@transformshift{5.988008in}{0.716028in}%
\pgfsys@useobject{currentmarker}{}%
\end{pgfscope}%
\end{pgfscope}%
\begin{pgfscope}%
\definecolor{textcolor}{rgb}{0.000000,0.000000,0.000000}%
\pgfsetstrokecolor{textcolor}%
\pgfsetfillcolor{textcolor}%
\pgftext[x=5.988008in,y=0.618806in,,top]{\color{textcolor}{\rmfamily\fontsize{20.000000}{24.000000}\selectfont\catcode`\^=\active\def^{\ifmmode\sp\else\^{}\fi}\catcode`\%=\active\def
\end{pgfscope}%
\begin{pgfscope}%
\definecolor{textcolor}{rgb}{0.000000,0.000000,0.000000}%
\pgfsetstrokecolor{textcolor}%
\pgfsetfillcolor{textcolor}%
\pgftext[x=3.290038in,y=0.307183in,,top]{\color{textcolor}{\rmfamily\fontsize{26.000000}{31.200000}\selectfont\catcode`\^=\active\def^{\ifmmode\sp\else\^{}\fi}\catcode`\%=\active\def
\end{pgfscope}%
\begin{pgfscope}%
\pgfsetbuttcap%
\pgfsetroundjoin%
\definecolor{currentfill}{rgb}{0.000000,0.000000,0.000000}%
\pgfsetfillcolor{currentfill}%
\pgfsetlinewidth{0.200750pt}%
\definecolor{currentstroke}{rgb}{0.000000,0.000000,0.000000}%
\pgfsetstrokecolor{currentstroke}%
\pgfsetdash{}{0pt}%
\pgfsys@defobject{currentmarker}{\pgfqpoint{-0.048611in}{0.000000in}}{\pgfqpoint{-0.000000in}{0.000000in}}{%
\pgfpathmoveto{\pgfqpoint{-0.000000in}{0.000000in}}%
\pgfpathlineto{\pgfqpoint{-0.048611in}{0.000000in}}%
\pgfusepath{stroke,fill}%
}%
\begin{pgfscope}%
\pgfsys@transformshift{0.592068in}{0.716028in}%
\pgfsys@useobject{currentmarker}{}%
\end{pgfscope}%
\end{pgfscope}%
\begin{pgfscope}%
\definecolor{textcolor}{rgb}{0.000000,0.000000,0.000000}%
\pgfsetstrokecolor{textcolor}%
\pgfsetfillcolor{textcolor}%
\pgftext[x=0.362738in, y=0.616009in, left, base]{\color{textcolor}{\rmfamily\fontsize{20.000000}{24.000000}\selectfont\catcode`\^=\active\def^{\ifmmode\sp\else\^{}\fi}\catcode`\%=\active\def
\end{pgfscope}%
\begin{pgfscope}%
\pgfsetbuttcap%
\pgfsetroundjoin%
\definecolor{currentfill}{rgb}{0.000000,0.000000,0.000000}%
\pgfsetfillcolor{currentfill}%
\pgfsetlinewidth{0.200750pt}%
\definecolor{currentstroke}{rgb}{0.000000,0.000000,0.000000}%
\pgfsetstrokecolor{currentstroke}%
\pgfsetdash{}{0pt}%
\pgfsys@defobject{currentmarker}{\pgfqpoint{-0.048611in}{0.000000in}}{\pgfqpoint{-0.000000in}{0.000000in}}{%
\pgfpathmoveto{\pgfqpoint{-0.000000in}{0.000000in}}%
\pgfpathlineto{\pgfqpoint{-0.048611in}{0.000000in}}%
\pgfusepath{stroke,fill}%
}%
\begin{pgfscope}%
\pgfsys@transformshift{0.592068in}{1.848008in}%
\pgfsys@useobject{currentmarker}{}%
\end{pgfscope}%
\end{pgfscope}%
\begin{pgfscope}%
\definecolor{textcolor}{rgb}{0.000000,0.000000,0.000000}%
\pgfsetstrokecolor{textcolor}%
\pgfsetfillcolor{textcolor}%
\pgftext[x=0.362738in, y=1.747989in, left, base]{\color{textcolor}{\rmfamily\fontsize{20.000000}{24.000000}\selectfont\catcode`\^=\active\def^{\ifmmode\sp\else\^{}\fi}\catcode`\%=\active\def
\end{pgfscope}%
\begin{pgfscope}%
\pgfsetbuttcap%
\pgfsetroundjoin%
\definecolor{currentfill}{rgb}{0.000000,0.000000,0.000000}%
\pgfsetfillcolor{currentfill}%
\pgfsetlinewidth{0.200750pt}%
\definecolor{currentstroke}{rgb}{0.000000,0.000000,0.000000}%
\pgfsetstrokecolor{currentstroke}%
\pgfsetdash{}{0pt}%
\pgfsys@defobject{currentmarker}{\pgfqpoint{-0.048611in}{0.000000in}}{\pgfqpoint{-0.000000in}{0.000000in}}{%
\pgfpathmoveto{\pgfqpoint{-0.000000in}{0.000000in}}%
\pgfpathlineto{\pgfqpoint{-0.048611in}{0.000000in}}%
\pgfusepath{stroke,fill}%
}%
\begin{pgfscope}%
\pgfsys@transformshift{0.592068in}{2.979987in}%
\pgfsys@useobject{currentmarker}{}%
\end{pgfscope}%
\end{pgfscope}%
\begin{pgfscope}%
\definecolor{textcolor}{rgb}{0.000000,0.000000,0.000000}%
\pgfsetstrokecolor{textcolor}%
\pgfsetfillcolor{textcolor}%
\pgftext[x=0.362738in, y=2.879968in, left, base]{\color{textcolor}{\rmfamily\fontsize{20.000000}{24.000000}\selectfont\catcode`\^=\active\def^{\ifmmode\sp\else\^{}\fi}\catcode`\%=\active\def
\end{pgfscope}%
\begin{pgfscope}%
\pgfsetbuttcap%
\pgfsetroundjoin%
\definecolor{currentfill}{rgb}{0.000000,0.000000,0.000000}%
\pgfsetfillcolor{currentfill}%
\pgfsetlinewidth{0.200750pt}%
\definecolor{currentstroke}{rgb}{0.000000,0.000000,0.000000}%
\pgfsetstrokecolor{currentstroke}%
\pgfsetdash{}{0pt}%
\pgfsys@defobject{currentmarker}{\pgfqpoint{-0.048611in}{0.000000in}}{\pgfqpoint{-0.000000in}{0.000000in}}{%
\pgfpathmoveto{\pgfqpoint{-0.000000in}{0.000000in}}%
\pgfpathlineto{\pgfqpoint{-0.048611in}{0.000000in}}%
\pgfusepath{stroke,fill}%
}%
\begin{pgfscope}%
\pgfsys@transformshift{0.592068in}{4.111967in}%
\pgfsys@useobject{currentmarker}{}%
\end{pgfscope}%
\end{pgfscope}%
\begin{pgfscope}%
\definecolor{textcolor}{rgb}{0.000000,0.000000,0.000000}%
\pgfsetstrokecolor{textcolor}%
\pgfsetfillcolor{textcolor}%
\pgftext[x=0.362738in, y=4.011948in, left, base]{\color{textcolor}{\rmfamily\fontsize{20.000000}{24.000000}\selectfont\catcode`\^=\active\def^{\ifmmode\sp\else\^{}\fi}\catcode`\%=\active\def
\end{pgfscope}%
\begin{pgfscope}%
\definecolor{textcolor}{rgb}{0.000000,0.000000,0.000000}%
\pgfsetstrokecolor{textcolor}%
\pgfsetfillcolor{textcolor}%
\pgftext[x=0.307183in,y=2.474950in,,bottom,rotate=90.000000]{\color{textcolor}{\rmfamily\fontsize{26.000000}{31.200000}\selectfont\catcode`\^=\active\def^{\ifmmode\sp\else\^{}\fi}\catcode`\%=\active\def
\end{pgfscope}%
\begin{pgfscope}%
\pgfpathrectangle{\pgfqpoint{0.592068in}{0.716028in}}{\pgfqpoint{5.395940in}{3.517843in}}%
\pgfusepath{clip}%
\pgfsetbuttcap%
\pgfsetroundjoin%
\pgfsetlinewidth{2.007500pt}%
\definecolor{currentstroke}{rgb}{0.501961,0.501961,0.501961}%
\pgfsetstrokecolor{currentstroke}%
\pgfsetdash{{7.400000pt}{3.200000pt}}{0.000000pt}%
\pgfpathmoveto{\pgfqpoint{0.592068in}{1.282018in}}%
\pgfpathlineto{\pgfqpoint{5.988008in}{1.282018in}}%
\pgfusepath{stroke}%
\end{pgfscope}%
\begin{pgfscope}%
\pgfsetrectcap%
\pgfsetmiterjoin%
\pgfsetlinewidth{0.803000pt}%
\definecolor{currentstroke}{rgb}{0.000000,0.000000,0.000000}%
\pgfsetstrokecolor{currentstroke}%
\pgfsetdash{}{0pt}%
\pgfpathmoveto{\pgfqpoint{0.592068in}{0.716028in}}%
\pgfpathlineto{\pgfqpoint{0.592068in}{4.233871in}}%
\pgfusepath{stroke}%
\end{pgfscope}%
\begin{pgfscope}%
\pgfsetrectcap%
\pgfsetmiterjoin%
\pgfsetlinewidth{0.803000pt}%
\definecolor{currentstroke}{rgb}{0.000000,0.000000,0.000000}%
\pgfsetstrokecolor{currentstroke}%
\pgfsetdash{}{0pt}%
\pgfpathmoveto{\pgfqpoint{5.988008in}{0.716028in}}%
\pgfpathlineto{\pgfqpoint{5.988008in}{4.233871in}}%
\pgfusepath{stroke}%
\end{pgfscope}%
\begin{pgfscope}%
\pgfsetrectcap%
\pgfsetmiterjoin%
\pgfsetlinewidth{0.803000pt}%
\definecolor{currentstroke}{rgb}{0.000000,0.000000,0.000000}%
\pgfsetstrokecolor{currentstroke}%
\pgfsetdash{}{0pt}%
\pgfpathmoveto{\pgfqpoint{0.592068in}{0.716028in}}%
\pgfpathlineto{\pgfqpoint{5.988008in}{0.716028in}}%
\pgfusepath{stroke}%
\end{pgfscope}%
\begin{pgfscope}%
\pgfsetrectcap%
\pgfsetmiterjoin%
\pgfsetlinewidth{0.803000pt}%
\definecolor{currentstroke}{rgb}{0.000000,0.000000,0.000000}%
\pgfsetstrokecolor{currentstroke}%
\pgfsetdash{}{0pt}%
\pgfpathmoveto{\pgfqpoint{0.592068in}{4.233871in}}%
\pgfpathlineto{\pgfqpoint{5.988008in}{4.233871in}}%
\pgfusepath{stroke}%
\end{pgfscope}%
\begin{pgfscope}%
\definecolor{textcolor}{rgb}{0.000000,0.000000,0.000000}%
\pgfsetstrokecolor{textcolor}%
\pgfsetfillcolor{textcolor}%
\pgftext[x=3.290038in,y=4.317204in,,base]{\color{textcolor}{\rmfamily\fontsize{25.000000}{30.000000}\selectfont\catcode`\^=\active\def^{\ifmmode\sp\else\^{}\fi}\catcode`\%=\active\def
\end{pgfscope}%
\begin{pgfscope}%
\pgfsetbuttcap%
\pgfsetmiterjoin%
\definecolor{currentfill}{rgb}{1.000000,1.000000,1.000000}%
\pgfsetfillcolor{currentfill}%
\pgfsetfillopacity{0.800000}%
\pgfsetlinewidth{1.003750pt}%
\definecolor{currentstroke}{rgb}{0.800000,0.800000,0.800000}%
\pgfsetstrokecolor{currentstroke}%
\pgfsetstrokeopacity{0.800000}%
\pgfsetdash{}{0pt}%
\pgfpathmoveto{\pgfqpoint{0.786512in}{3.616692in}}%
\pgfpathlineto{\pgfqpoint{2.328414in}{3.616692in}}%
\pgfpathquadraticcurveto{\pgfqpoint{2.383970in}{3.616692in}}{\pgfqpoint{2.383970in}{3.672248in}}%
\pgfpathlineto{\pgfqpoint{2.383970in}{4.039427in}}%
\pgfpathquadraticcurveto{\pgfqpoint{2.383970in}{4.094982in}}{\pgfqpoint{2.328414in}{4.094982in}}%
\pgfpathlineto{\pgfqpoint{0.786512in}{4.094982in}}%
\pgfpathquadraticcurveto{\pgfqpoint{0.730957in}{4.094982in}}{\pgfqpoint{0.730957in}{4.039427in}}%
\pgfpathlineto{\pgfqpoint{0.730957in}{3.672248in}}%
\pgfpathquadraticcurveto{\pgfqpoint{0.730957in}{3.616692in}}{\pgfqpoint{0.786512in}{3.616692in}}%
\pgfpathlineto{\pgfqpoint{0.786512in}{3.616692in}}%
\pgfpathclose%
\pgfusepath{stroke,fill}%
\end{pgfscope}%
\begin{pgfscope}%
\pgfsetbuttcap%
\pgfsetroundjoin%
\pgfsetlinewidth{2.007500pt}%
\definecolor{currentstroke}{rgb}{0.501961,0.501961,0.501961}%
\pgfsetstrokecolor{currentstroke}%
\pgfsetdash{{7.400000pt}{3.200000pt}}{0.000000pt}%
\pgfpathmoveto{\pgfqpoint{0.842068in}{3.881055in}}%
\pgfpathlineto{\pgfqpoint{1.119846in}{3.881055in}}%
\pgfpathlineto{\pgfqpoint{1.397623in}{3.881055in}}%
\pgfusepath{stroke}%
\end{pgfscope}%
\begin{pgfscope}%
\definecolor{textcolor}{rgb}{0.000000,0.000000,0.000000}%
\pgfsetstrokecolor{textcolor}%
\pgfsetfillcolor{textcolor}%
\pgftext[x=1.619846in,y=3.783833in,left,base]{\color{textcolor}{\rmfamily\fontsize{20.000000}{24.000000}\selectfont\catcode`\^=\active\def^{\ifmmode\sp\else\^{}\fi}\catcode`\%=\active\def
\end{pgfscope}%
\end{pgfpicture}%
\makeatother%
\endgroup%

%% file: figures/evaluation/BPIC17/BPIC17_PIT_event_elapsed_norm_4layer.pgf
\begingroup%
\makeatletter%
\begin{pgfpicture}%
\pgfpathrectangle{\pgfpointorigin}{\pgfqpoint{6.159289in}{4.557174in}}%
\pgfusepath{use as bounding box, clip}%
\begin{pgfscope}%
\pgfsetbuttcap%
\pgfsetmiterjoin%
\definecolor{currentfill}{rgb}{1.000000,1.000000,1.000000}%
\pgfsetfillcolor{currentfill}%
\pgfsetlinewidth{0.000000pt}%
\definecolor{currentstroke}{rgb}{1.000000,1.000000,1.000000}%
\pgfsetstrokecolor{currentstroke}%
\pgfsetdash{}{0pt}%
\pgfpathmoveto{\pgfqpoint{0.000000in}{0.000000in}}%
\pgfpathlineto{\pgfqpoint{6.159289in}{0.000000in}}%
\pgfpathlineto{\pgfqpoint{6.159289in}{4.557174in}}%
\pgfpathlineto{\pgfqpoint{0.000000in}{4.557174in}}%
\pgfpathlineto{\pgfqpoint{0.000000in}{0.000000in}}%
\pgfpathclose%
\pgfusepath{fill}%
\end{pgfscope}%
\begin{pgfscope}%
\pgfsetbuttcap%
\pgfsetmiterjoin%
\definecolor{currentfill}{rgb}{1.000000,1.000000,1.000000}%
\pgfsetfillcolor{currentfill}%
\pgfsetlinewidth{0.000000pt}%
\definecolor{currentstroke}{rgb}{0.000000,0.000000,0.000000}%
\pgfsetstrokecolor{currentstroke}%
\pgfsetstrokeopacity{0.000000}%
\pgfsetdash{}{0pt}%
\pgfpathmoveto{\pgfqpoint{0.592068in}{0.716028in}}%
\pgfpathlineto{\pgfqpoint{5.988008in}{0.716028in}}%
\pgfpathlineto{\pgfqpoint{5.988008in}{4.233871in}}%
\pgfpathlineto{\pgfqpoint{0.592068in}{4.233871in}}%
\pgfpathlineto{\pgfqpoint{0.592068in}{0.716028in}}%
\pgfpathclose%
\pgfusepath{fill}%
\end{pgfscope}%
\begin{pgfscope}%
\pgfpathrectangle{\pgfqpoint{0.592068in}{0.716028in}}{\pgfqpoint{5.395940in}{3.517843in}}%
\pgfusepath{clip}%
\pgfsetbuttcap%
\pgfsetmiterjoin%
\definecolor{currentfill}{rgb}{0.000000,0.000000,1.000000}%
\pgfsetfillcolor{currentfill}%
\pgfsetlinewidth{0.000000pt}%
\definecolor{currentstroke}{rgb}{0.000000,0.000000,0.000000}%
\pgfsetstrokecolor{currentstroke}%
\pgfsetdash{}{0pt}%
\pgfpathmoveto{\pgfqpoint{0.592068in}{0.716028in}}%
\pgfpathlineto{\pgfqpoint{0.778135in}{0.716028in}}%
\pgfpathlineto{\pgfqpoint{0.778135in}{0.716028in}}%
\pgfpathlineto{\pgfqpoint{0.592068in}{0.716028in}}%
\pgfpathlineto{\pgfqpoint{0.592068in}{0.716028in}}%
\pgfpathclose%
\pgfusepath{fill}%
\end{pgfscope}%
\begin{pgfscope}%
\pgfpathrectangle{\pgfqpoint{0.592068in}{0.716028in}}{\pgfqpoint{5.395940in}{3.517843in}}%
\pgfusepath{clip}%
\pgfsetbuttcap%
\pgfsetmiterjoin%
\definecolor{currentfill}{rgb}{0.000000,0.000000,1.000000}%
\pgfsetfillcolor{currentfill}%
\pgfsetlinewidth{0.000000pt}%
\definecolor{currentstroke}{rgb}{0.000000,0.000000,0.000000}%
\pgfsetstrokecolor{currentstroke}%
\pgfsetdash{}{0pt}%
\pgfpathmoveto{\pgfqpoint{0.778135in}{0.716028in}}%
\pgfpathlineto{\pgfqpoint{0.964202in}{0.716028in}}%
\pgfpathlineto{\pgfqpoint{0.964202in}{0.716028in}}%
\pgfpathlineto{\pgfqpoint{0.778135in}{0.716028in}}%
\pgfpathlineto{\pgfqpoint{0.778135in}{0.716028in}}%
\pgfpathclose%
\pgfusepath{fill}%
\end{pgfscope}%
\begin{pgfscope}%
\pgfpathrectangle{\pgfqpoint{0.592068in}{0.716028in}}{\pgfqpoint{5.395940in}{3.517843in}}%
\pgfusepath{clip}%
\pgfsetbuttcap%
\pgfsetmiterjoin%
\definecolor{currentfill}{rgb}{0.000000,0.000000,1.000000}%
\pgfsetfillcolor{currentfill}%
\pgfsetlinewidth{0.000000pt}%
\definecolor{currentstroke}{rgb}{0.000000,0.000000,0.000000}%
\pgfsetstrokecolor{currentstroke}%
\pgfsetdash{}{0pt}%
\pgfpathmoveto{\pgfqpoint{0.964202in}{0.716028in}}%
\pgfpathlineto{\pgfqpoint{1.150269in}{0.716028in}}%
\pgfpathlineto{\pgfqpoint{1.150269in}{0.716028in}}%
\pgfpathlineto{\pgfqpoint{0.964202in}{0.716028in}}%
\pgfpathlineto{\pgfqpoint{0.964202in}{0.716028in}}%
\pgfpathclose%
\pgfusepath{fill}%
\end{pgfscope}%
\begin{pgfscope}%
\pgfpathrectangle{\pgfqpoint{0.592068in}{0.716028in}}{\pgfqpoint{5.395940in}{3.517843in}}%
\pgfusepath{clip}%
\pgfsetbuttcap%
\pgfsetmiterjoin%
\definecolor{currentfill}{rgb}{0.000000,0.000000,1.000000}%
\pgfsetfillcolor{currentfill}%
\pgfsetlinewidth{0.000000pt}%
\definecolor{currentstroke}{rgb}{0.000000,0.000000,0.000000}%
\pgfsetstrokecolor{currentstroke}%
\pgfsetdash{}{0pt}%
\pgfpathmoveto{\pgfqpoint{1.150269in}{0.716028in}}%
\pgfpathlineto{\pgfqpoint{1.336335in}{0.716028in}}%
\pgfpathlineto{\pgfqpoint{1.336335in}{0.716534in}}%
\pgfpathlineto{\pgfqpoint{1.150269in}{0.716534in}}%
\pgfpathlineto{\pgfqpoint{1.150269in}{0.716028in}}%
\pgfpathclose%
\pgfusepath{fill}%
\end{pgfscope}%
\begin{pgfscope}%
\pgfpathrectangle{\pgfqpoint{0.592068in}{0.716028in}}{\pgfqpoint{5.395940in}{3.517843in}}%
\pgfusepath{clip}%
\pgfsetbuttcap%
\pgfsetmiterjoin%
\definecolor{currentfill}{rgb}{0.000000,0.000000,1.000000}%
\pgfsetfillcolor{currentfill}%
\pgfsetlinewidth{0.000000pt}%
\definecolor{currentstroke}{rgb}{0.000000,0.000000,0.000000}%
\pgfsetstrokecolor{currentstroke}%
\pgfsetdash{}{0pt}%
\pgfpathmoveto{\pgfqpoint{1.336335in}{0.716028in}}%
\pgfpathlineto{\pgfqpoint{1.522402in}{0.716028in}}%
\pgfpathlineto{\pgfqpoint{1.522402in}{0.716534in}}%
\pgfpathlineto{\pgfqpoint{1.336335in}{0.716534in}}%
\pgfpathlineto{\pgfqpoint{1.336335in}{0.716028in}}%
\pgfpathclose%
\pgfusepath{fill}%
\end{pgfscope}%
\begin{pgfscope}%
\pgfpathrectangle{\pgfqpoint{0.592068in}{0.716028in}}{\pgfqpoint{5.395940in}{3.517843in}}%
\pgfusepath{clip}%
\pgfsetbuttcap%
\pgfsetmiterjoin%
\definecolor{currentfill}{rgb}{0.000000,0.000000,1.000000}%
\pgfsetfillcolor{currentfill}%
\pgfsetlinewidth{0.000000pt}%
\definecolor{currentstroke}{rgb}{0.000000,0.000000,0.000000}%
\pgfsetstrokecolor{currentstroke}%
\pgfsetdash{}{0pt}%
\pgfpathmoveto{\pgfqpoint{1.522402in}{0.716028in}}%
\pgfpathlineto{\pgfqpoint{1.708469in}{0.716028in}}%
\pgfpathlineto{\pgfqpoint{1.708469in}{0.723110in}}%
\pgfpathlineto{\pgfqpoint{1.522402in}{0.723110in}}%
\pgfpathlineto{\pgfqpoint{1.522402in}{0.716028in}}%
\pgfpathclose%
\pgfusepath{fill}%
\end{pgfscope}%
\begin{pgfscope}%
\pgfpathrectangle{\pgfqpoint{0.592068in}{0.716028in}}{\pgfqpoint{5.395940in}{3.517843in}}%
\pgfusepath{clip}%
\pgfsetbuttcap%
\pgfsetmiterjoin%
\definecolor{currentfill}{rgb}{0.000000,0.000000,1.000000}%
\pgfsetfillcolor{currentfill}%
\pgfsetlinewidth{0.000000pt}%
\definecolor{currentstroke}{rgb}{0.000000,0.000000,0.000000}%
\pgfsetstrokecolor{currentstroke}%
\pgfsetdash{}{0pt}%
\pgfpathmoveto{\pgfqpoint{1.708469in}{0.716028in}}%
\pgfpathlineto{\pgfqpoint{1.894536in}{0.716028in}}%
\pgfpathlineto{\pgfqpoint{1.894536in}{0.818882in}}%
\pgfpathlineto{\pgfqpoint{1.708469in}{0.818882in}}%
\pgfpathlineto{\pgfqpoint{1.708469in}{0.716028in}}%
\pgfpathclose%
\pgfusepath{fill}%
\end{pgfscope}%
\begin{pgfscope}%
\pgfpathrectangle{\pgfqpoint{0.592068in}{0.716028in}}{\pgfqpoint{5.395940in}{3.517843in}}%
\pgfusepath{clip}%
\pgfsetbuttcap%
\pgfsetmiterjoin%
\definecolor{currentfill}{rgb}{0.000000,0.000000,1.000000}%
\pgfsetfillcolor{currentfill}%
\pgfsetlinewidth{0.000000pt}%
\definecolor{currentstroke}{rgb}{0.000000,0.000000,0.000000}%
\pgfsetstrokecolor{currentstroke}%
\pgfsetdash{}{0pt}%
\pgfpathmoveto{\pgfqpoint{1.894536in}{0.716028in}}%
\pgfpathlineto{\pgfqpoint{2.080603in}{0.716028in}}%
\pgfpathlineto{\pgfqpoint{2.080603in}{1.341074in}}%
\pgfpathlineto{\pgfqpoint{1.894536in}{1.341074in}}%
\pgfpathlineto{\pgfqpoint{1.894536in}{0.716028in}}%
\pgfpathclose%
\pgfusepath{fill}%
\end{pgfscope}%
\begin{pgfscope}%
\pgfpathrectangle{\pgfqpoint{0.592068in}{0.716028in}}{\pgfqpoint{5.395940in}{3.517843in}}%
\pgfusepath{clip}%
\pgfsetbuttcap%
\pgfsetmiterjoin%
\definecolor{currentfill}{rgb}{0.000000,0.000000,1.000000}%
\pgfsetfillcolor{currentfill}%
\pgfsetlinewidth{0.000000pt}%
\definecolor{currentstroke}{rgb}{0.000000,0.000000,0.000000}%
\pgfsetstrokecolor{currentstroke}%
\pgfsetdash{}{0pt}%
\pgfpathmoveto{\pgfqpoint{2.080603in}{0.716028in}}%
\pgfpathlineto{\pgfqpoint{2.266670in}{0.716028in}}%
\pgfpathlineto{\pgfqpoint{2.266670in}{1.870179in}}%
\pgfpathlineto{\pgfqpoint{2.080603in}{1.870179in}}%
\pgfpathlineto{\pgfqpoint{2.080603in}{0.716028in}}%
\pgfpathclose%
\pgfusepath{fill}%
\end{pgfscope}%
\begin{pgfscope}%
\pgfpathrectangle{\pgfqpoint{0.592068in}{0.716028in}}{\pgfqpoint{5.395940in}{3.517843in}}%
\pgfusepath{clip}%
\pgfsetbuttcap%
\pgfsetmiterjoin%
\definecolor{currentfill}{rgb}{0.000000,0.000000,1.000000}%
\pgfsetfillcolor{currentfill}%
\pgfsetlinewidth{0.000000pt}%
\definecolor{currentstroke}{rgb}{0.000000,0.000000,0.000000}%
\pgfsetstrokecolor{currentstroke}%
\pgfsetdash{}{0pt}%
\pgfpathmoveto{\pgfqpoint{2.266670in}{0.716028in}}%
\pgfpathlineto{\pgfqpoint{2.452737in}{0.716028in}}%
\pgfpathlineto{\pgfqpoint{2.452737in}{1.708143in}}%
\pgfpathlineto{\pgfqpoint{2.266670in}{1.708143in}}%
\pgfpathlineto{\pgfqpoint{2.266670in}{0.716028in}}%
\pgfpathclose%
\pgfusepath{fill}%
\end{pgfscope}%
\begin{pgfscope}%
\pgfpathrectangle{\pgfqpoint{0.592068in}{0.716028in}}{\pgfqpoint{5.395940in}{3.517843in}}%
\pgfusepath{clip}%
\pgfsetbuttcap%
\pgfsetmiterjoin%
\definecolor{currentfill}{rgb}{0.000000,0.000000,1.000000}%
\pgfsetfillcolor{currentfill}%
\pgfsetlinewidth{0.000000pt}%
\definecolor{currentstroke}{rgb}{0.000000,0.000000,0.000000}%
\pgfsetstrokecolor{currentstroke}%
\pgfsetdash{}{0pt}%
\pgfpathmoveto{\pgfqpoint{2.452737in}{0.716028in}}%
\pgfpathlineto{\pgfqpoint{2.638804in}{0.716028in}}%
\pgfpathlineto{\pgfqpoint{2.638804in}{1.673746in}}%
\pgfpathlineto{\pgfqpoint{2.452737in}{1.673746in}}%
\pgfpathlineto{\pgfqpoint{2.452737in}{0.716028in}}%
\pgfpathclose%
\pgfusepath{fill}%
\end{pgfscope}%
\begin{pgfscope}%
\pgfpathrectangle{\pgfqpoint{0.592068in}{0.716028in}}{\pgfqpoint{5.395940in}{3.517843in}}%
\pgfusepath{clip}%
\pgfsetbuttcap%
\pgfsetmiterjoin%
\definecolor{currentfill}{rgb}{0.000000,0.000000,1.000000}%
\pgfsetfillcolor{currentfill}%
\pgfsetlinewidth{0.000000pt}%
\definecolor{currentstroke}{rgb}{0.000000,0.000000,0.000000}%
\pgfsetstrokecolor{currentstroke}%
\pgfsetdash{}{0pt}%
\pgfpathmoveto{\pgfqpoint{2.638804in}{0.716028in}}%
\pgfpathlineto{\pgfqpoint{2.824871in}{0.716028in}}%
\pgfpathlineto{\pgfqpoint{2.824871in}{1.649297in}}%
\pgfpathlineto{\pgfqpoint{2.638804in}{1.649297in}}%
\pgfpathlineto{\pgfqpoint{2.638804in}{0.716028in}}%
\pgfpathclose%
\pgfusepath{fill}%
\end{pgfscope}%
\begin{pgfscope}%
\pgfpathrectangle{\pgfqpoint{0.592068in}{0.716028in}}{\pgfqpoint{5.395940in}{3.517843in}}%
\pgfusepath{clip}%
\pgfsetbuttcap%
\pgfsetmiterjoin%
\definecolor{currentfill}{rgb}{0.000000,0.000000,1.000000}%
\pgfsetfillcolor{currentfill}%
\pgfsetlinewidth{0.000000pt}%
\definecolor{currentstroke}{rgb}{0.000000,0.000000,0.000000}%
\pgfsetstrokecolor{currentstroke}%
\pgfsetdash{}{0pt}%
\pgfpathmoveto{\pgfqpoint{2.824871in}{0.716028in}}%
\pgfpathlineto{\pgfqpoint{3.010937in}{0.716028in}}%
\pgfpathlineto{\pgfqpoint{3.010937in}{1.659245in}}%
\pgfpathlineto{\pgfqpoint{2.824871in}{1.659245in}}%
\pgfpathlineto{\pgfqpoint{2.824871in}{0.716028in}}%
\pgfpathclose%
\pgfusepath{fill}%
\end{pgfscope}%
\begin{pgfscope}%
\pgfpathrectangle{\pgfqpoint{0.592068in}{0.716028in}}{\pgfqpoint{5.395940in}{3.517843in}}%
\pgfusepath{clip}%
\pgfsetbuttcap%
\pgfsetmiterjoin%
\definecolor{currentfill}{rgb}{0.000000,0.000000,1.000000}%
\pgfsetfillcolor{currentfill}%
\pgfsetlinewidth{0.000000pt}%
\definecolor{currentstroke}{rgb}{0.000000,0.000000,0.000000}%
\pgfsetstrokecolor{currentstroke}%
\pgfsetdash{}{0pt}%
\pgfpathmoveto{\pgfqpoint{3.010937in}{0.716028in}}%
\pgfpathlineto{\pgfqpoint{3.197004in}{0.716028in}}%
\pgfpathlineto{\pgfqpoint{3.197004in}{1.417793in}}%
\pgfpathlineto{\pgfqpoint{3.010937in}{1.417793in}}%
\pgfpathlineto{\pgfqpoint{3.010937in}{0.716028in}}%
\pgfpathclose%
\pgfusepath{fill}%
\end{pgfscope}%
\begin{pgfscope}%
\pgfpathrectangle{\pgfqpoint{0.592068in}{0.716028in}}{\pgfqpoint{5.395940in}{3.517843in}}%
\pgfusepath{clip}%
\pgfsetbuttcap%
\pgfsetmiterjoin%
\definecolor{currentfill}{rgb}{0.000000,0.000000,1.000000}%
\pgfsetfillcolor{currentfill}%
\pgfsetlinewidth{0.000000pt}%
\definecolor{currentstroke}{rgb}{0.000000,0.000000,0.000000}%
\pgfsetstrokecolor{currentstroke}%
\pgfsetdash{}{0pt}%
\pgfpathmoveto{\pgfqpoint{3.197004in}{0.716028in}}%
\pgfpathlineto{\pgfqpoint{3.383071in}{0.716028in}}%
\pgfpathlineto{\pgfqpoint{3.383071in}{1.362656in}}%
\pgfpathlineto{\pgfqpoint{3.197004in}{1.362656in}}%
\pgfpathlineto{\pgfqpoint{3.197004in}{0.716028in}}%
\pgfpathclose%
\pgfusepath{fill}%
\end{pgfscope}%
\begin{pgfscope}%
\pgfpathrectangle{\pgfqpoint{0.592068in}{0.716028in}}{\pgfqpoint{5.395940in}{3.517843in}}%
\pgfusepath{clip}%
\pgfsetbuttcap%
\pgfsetmiterjoin%
\definecolor{currentfill}{rgb}{0.000000,0.000000,1.000000}%
\pgfsetfillcolor{currentfill}%
\pgfsetlinewidth{0.000000pt}%
\definecolor{currentstroke}{rgb}{0.000000,0.000000,0.000000}%
\pgfsetstrokecolor{currentstroke}%
\pgfsetdash{}{0pt}%
\pgfpathmoveto{\pgfqpoint{3.383071in}{0.716028in}}%
\pgfpathlineto{\pgfqpoint{3.569138in}{0.716028in}}%
\pgfpathlineto{\pgfqpoint{3.569138in}{1.465341in}}%
\pgfpathlineto{\pgfqpoint{3.383071in}{1.465341in}}%
\pgfpathlineto{\pgfqpoint{3.383071in}{0.716028in}}%
\pgfpathclose%
\pgfusepath{fill}%
\end{pgfscope}%
\begin{pgfscope}%
\pgfpathrectangle{\pgfqpoint{0.592068in}{0.716028in}}{\pgfqpoint{5.395940in}{3.517843in}}%
\pgfusepath{clip}%
\pgfsetbuttcap%
\pgfsetmiterjoin%
\definecolor{currentfill}{rgb}{0.000000,0.000000,1.000000}%
\pgfsetfillcolor{currentfill}%
\pgfsetlinewidth{0.000000pt}%
\definecolor{currentstroke}{rgb}{0.000000,0.000000,0.000000}%
\pgfsetstrokecolor{currentstroke}%
\pgfsetdash{}{0pt}%
\pgfpathmoveto{\pgfqpoint{3.569138in}{0.716028in}}%
\pgfpathlineto{\pgfqpoint{3.755205in}{0.716028in}}%
\pgfpathlineto{\pgfqpoint{3.755205in}{1.536496in}}%
\pgfpathlineto{\pgfqpoint{3.569138in}{1.536496in}}%
\pgfpathlineto{\pgfqpoint{3.569138in}{0.716028in}}%
\pgfpathclose%
\pgfusepath{fill}%
\end{pgfscope}%
\begin{pgfscope}%
\pgfpathrectangle{\pgfqpoint{0.592068in}{0.716028in}}{\pgfqpoint{5.395940in}{3.517843in}}%
\pgfusepath{clip}%
\pgfsetbuttcap%
\pgfsetmiterjoin%
\definecolor{currentfill}{rgb}{0.000000,0.000000,1.000000}%
\pgfsetfillcolor{currentfill}%
\pgfsetlinewidth{0.000000pt}%
\definecolor{currentstroke}{rgb}{0.000000,0.000000,0.000000}%
\pgfsetstrokecolor{currentstroke}%
\pgfsetdash{}{0pt}%
\pgfpathmoveto{\pgfqpoint{3.755205in}{0.716028in}}%
\pgfpathlineto{\pgfqpoint{3.941272in}{0.716028in}}%
\pgfpathlineto{\pgfqpoint{3.941272in}{1.589440in}}%
\pgfpathlineto{\pgfqpoint{3.755205in}{1.589440in}}%
\pgfpathlineto{\pgfqpoint{3.755205in}{0.716028in}}%
\pgfpathclose%
\pgfusepath{fill}%
\end{pgfscope}%
\begin{pgfscope}%
\pgfpathrectangle{\pgfqpoint{0.592068in}{0.716028in}}{\pgfqpoint{5.395940in}{3.517843in}}%
\pgfusepath{clip}%
\pgfsetbuttcap%
\pgfsetmiterjoin%
\definecolor{currentfill}{rgb}{0.000000,0.000000,1.000000}%
\pgfsetfillcolor{currentfill}%
\pgfsetlinewidth{0.000000pt}%
\definecolor{currentstroke}{rgb}{0.000000,0.000000,0.000000}%
\pgfsetstrokecolor{currentstroke}%
\pgfsetdash{}{0pt}%
\pgfpathmoveto{\pgfqpoint{3.941272in}{0.716028in}}%
\pgfpathlineto{\pgfqpoint{4.127339in}{0.716028in}}%
\pgfpathlineto{\pgfqpoint{4.127339in}{1.741528in}}%
\pgfpathlineto{\pgfqpoint{3.941272in}{1.741528in}}%
\pgfpathlineto{\pgfqpoint{3.941272in}{0.716028in}}%
\pgfpathclose%
\pgfusepath{fill}%
\end{pgfscope}%
\begin{pgfscope}%
\pgfpathrectangle{\pgfqpoint{0.592068in}{0.716028in}}{\pgfqpoint{5.395940in}{3.517843in}}%
\pgfusepath{clip}%
\pgfsetbuttcap%
\pgfsetmiterjoin%
\definecolor{currentfill}{rgb}{0.000000,0.000000,1.000000}%
\pgfsetfillcolor{currentfill}%
\pgfsetlinewidth{0.000000pt}%
\definecolor{currentstroke}{rgb}{0.000000,0.000000,0.000000}%
\pgfsetstrokecolor{currentstroke}%
\pgfsetdash{}{0pt}%
\pgfpathmoveto{\pgfqpoint{4.127339in}{0.716028in}}%
\pgfpathlineto{\pgfqpoint{4.313406in}{0.716028in}}%
\pgfpathlineto{\pgfqpoint{4.313406in}{1.852307in}}%
\pgfpathlineto{\pgfqpoint{4.127339in}{1.852307in}}%
\pgfpathlineto{\pgfqpoint{4.127339in}{0.716028in}}%
\pgfpathclose%
\pgfusepath{fill}%
\end{pgfscope}%
\begin{pgfscope}%
\pgfpathrectangle{\pgfqpoint{0.592068in}{0.716028in}}{\pgfqpoint{5.395940in}{3.517843in}}%
\pgfusepath{clip}%
\pgfsetbuttcap%
\pgfsetmiterjoin%
\definecolor{currentfill}{rgb}{0.000000,0.000000,1.000000}%
\pgfsetfillcolor{currentfill}%
\pgfsetlinewidth{0.000000pt}%
\definecolor{currentstroke}{rgb}{0.000000,0.000000,0.000000}%
\pgfsetstrokecolor{currentstroke}%
\pgfsetdash{}{0pt}%
\pgfpathmoveto{\pgfqpoint{4.313406in}{0.716028in}}%
\pgfpathlineto{\pgfqpoint{4.499473in}{0.716028in}}%
\pgfpathlineto{\pgfqpoint{4.499473in}{2.033396in}}%
\pgfpathlineto{\pgfqpoint{4.313406in}{2.033396in}}%
\pgfpathlineto{\pgfqpoint{4.313406in}{0.716028in}}%
\pgfpathclose%
\pgfusepath{fill}%
\end{pgfscope}%
\begin{pgfscope}%
\pgfpathrectangle{\pgfqpoint{0.592068in}{0.716028in}}{\pgfqpoint{5.395940in}{3.517843in}}%
\pgfusepath{clip}%
\pgfsetbuttcap%
\pgfsetmiterjoin%
\definecolor{currentfill}{rgb}{0.000000,0.000000,1.000000}%
\pgfsetfillcolor{currentfill}%
\pgfsetlinewidth{0.000000pt}%
\definecolor{currentstroke}{rgb}{0.000000,0.000000,0.000000}%
\pgfsetstrokecolor{currentstroke}%
\pgfsetdash{}{0pt}%
\pgfpathmoveto{\pgfqpoint{4.499473in}{0.716028in}}%
\pgfpathlineto{\pgfqpoint{4.685539in}{0.716028in}}%
\pgfpathlineto{\pgfqpoint{4.685539in}{2.090556in}}%
\pgfpathlineto{\pgfqpoint{4.499473in}{2.090556in}}%
\pgfpathlineto{\pgfqpoint{4.499473in}{0.716028in}}%
\pgfpathclose%
\pgfusepath{fill}%
\end{pgfscope}%
\begin{pgfscope}%
\pgfpathrectangle{\pgfqpoint{0.592068in}{0.716028in}}{\pgfqpoint{5.395940in}{3.517843in}}%
\pgfusepath{clip}%
\pgfsetbuttcap%
\pgfsetmiterjoin%
\definecolor{currentfill}{rgb}{0.000000,0.000000,1.000000}%
\pgfsetfillcolor{currentfill}%
\pgfsetlinewidth{0.000000pt}%
\definecolor{currentstroke}{rgb}{0.000000,0.000000,0.000000}%
\pgfsetstrokecolor{currentstroke}%
\pgfsetdash{}{0pt}%
\pgfpathmoveto{\pgfqpoint{4.685539in}{0.716028in}}%
\pgfpathlineto{\pgfqpoint{4.871606in}{0.716028in}}%
\pgfpathlineto{\pgfqpoint{4.871606in}{2.205718in}}%
\pgfpathlineto{\pgfqpoint{4.685539in}{2.205718in}}%
\pgfpathlineto{\pgfqpoint{4.685539in}{0.716028in}}%
\pgfpathclose%
\pgfusepath{fill}%
\end{pgfscope}%
\begin{pgfscope}%
\pgfpathrectangle{\pgfqpoint{0.592068in}{0.716028in}}{\pgfqpoint{5.395940in}{3.517843in}}%
\pgfusepath{clip}%
\pgfsetbuttcap%
\pgfsetmiterjoin%
\definecolor{currentfill}{rgb}{0.000000,0.000000,1.000000}%
\pgfsetfillcolor{currentfill}%
\pgfsetlinewidth{0.000000pt}%
\definecolor{currentstroke}{rgb}{0.000000,0.000000,0.000000}%
\pgfsetstrokecolor{currentstroke}%
\pgfsetdash{}{0pt}%
\pgfpathmoveto{\pgfqpoint{4.871606in}{0.716028in}}%
\pgfpathlineto{\pgfqpoint{5.057673in}{0.716028in}}%
\pgfpathlineto{\pgfqpoint{5.057673in}{2.271140in}}%
\pgfpathlineto{\pgfqpoint{4.871606in}{2.271140in}}%
\pgfpathlineto{\pgfqpoint{4.871606in}{0.716028in}}%
\pgfpathclose%
\pgfusepath{fill}%
\end{pgfscope}%
\begin{pgfscope}%
\pgfpathrectangle{\pgfqpoint{0.592068in}{0.716028in}}{\pgfqpoint{5.395940in}{3.517843in}}%
\pgfusepath{clip}%
\pgfsetbuttcap%
\pgfsetmiterjoin%
\definecolor{currentfill}{rgb}{0.000000,0.000000,1.000000}%
\pgfsetfillcolor{currentfill}%
\pgfsetlinewidth{0.000000pt}%
\definecolor{currentstroke}{rgb}{0.000000,0.000000,0.000000}%
\pgfsetstrokecolor{currentstroke}%
\pgfsetdash{}{0pt}%
\pgfpathmoveto{\pgfqpoint{5.057673in}{0.716028in}}%
\pgfpathlineto{\pgfqpoint{5.243740in}{0.716028in}}%
\pgfpathlineto{\pgfqpoint{5.243740in}{2.359324in}}%
\pgfpathlineto{\pgfqpoint{5.057673in}{2.359324in}}%
\pgfpathlineto{\pgfqpoint{5.057673in}{0.716028in}}%
\pgfpathclose%
\pgfusepath{fill}%
\end{pgfscope}%
\begin{pgfscope}%
\pgfpathrectangle{\pgfqpoint{0.592068in}{0.716028in}}{\pgfqpoint{5.395940in}{3.517843in}}%
\pgfusepath{clip}%
\pgfsetbuttcap%
\pgfsetmiterjoin%
\definecolor{currentfill}{rgb}{0.000000,0.000000,1.000000}%
\pgfsetfillcolor{currentfill}%
\pgfsetlinewidth{0.000000pt}%
\definecolor{currentstroke}{rgb}{0.000000,0.000000,0.000000}%
\pgfsetstrokecolor{currentstroke}%
\pgfsetdash{}{0pt}%
\pgfpathmoveto{\pgfqpoint{5.243740in}{0.716028in}}%
\pgfpathlineto{\pgfqpoint{5.429807in}{0.716028in}}%
\pgfpathlineto{\pgfqpoint{5.429807in}{2.405186in}}%
\pgfpathlineto{\pgfqpoint{5.243740in}{2.405186in}}%
\pgfpathlineto{\pgfqpoint{5.243740in}{0.716028in}}%
\pgfpathclose%
\pgfusepath{fill}%
\end{pgfscope}%
\begin{pgfscope}%
\pgfpathrectangle{\pgfqpoint{0.592068in}{0.716028in}}{\pgfqpoint{5.395940in}{3.517843in}}%
\pgfusepath{clip}%
\pgfsetbuttcap%
\pgfsetmiterjoin%
\definecolor{currentfill}{rgb}{0.000000,0.000000,1.000000}%
\pgfsetfillcolor{currentfill}%
\pgfsetlinewidth{0.000000pt}%
\definecolor{currentstroke}{rgb}{0.000000,0.000000,0.000000}%
\pgfsetstrokecolor{currentstroke}%
\pgfsetdash{}{0pt}%
\pgfpathmoveto{\pgfqpoint{5.429807in}{0.716028in}}%
\pgfpathlineto{\pgfqpoint{5.615874in}{0.716028in}}%
\pgfpathlineto{\pgfqpoint{5.615874in}{2.724369in}}%
\pgfpathlineto{\pgfqpoint{5.429807in}{2.724369in}}%
\pgfpathlineto{\pgfqpoint{5.429807in}{0.716028in}}%
\pgfpathclose%
\pgfusepath{fill}%
\end{pgfscope}%
\begin{pgfscope}%
\pgfpathrectangle{\pgfqpoint{0.592068in}{0.716028in}}{\pgfqpoint{5.395940in}{3.517843in}}%
\pgfusepath{clip}%
\pgfsetbuttcap%
\pgfsetmiterjoin%
\definecolor{currentfill}{rgb}{0.000000,0.000000,1.000000}%
\pgfsetfillcolor{currentfill}%
\pgfsetlinewidth{0.000000pt}%
\definecolor{currentstroke}{rgb}{0.000000,0.000000,0.000000}%
\pgfsetstrokecolor{currentstroke}%
\pgfsetdash{}{0pt}%
\pgfpathmoveto{\pgfqpoint{5.615874in}{0.716028in}}%
\pgfpathlineto{\pgfqpoint{5.801941in}{0.716028in}}%
\pgfpathlineto{\pgfqpoint{5.801941in}{4.066355in}}%
\pgfpathlineto{\pgfqpoint{5.615874in}{4.066355in}}%
\pgfpathlineto{\pgfqpoint{5.615874in}{0.716028in}}%
\pgfpathclose%
\pgfusepath{fill}%
\end{pgfscope}%
\begin{pgfscope}%
\pgfpathrectangle{\pgfqpoint{0.592068in}{0.716028in}}{\pgfqpoint{5.395940in}{3.517843in}}%
\pgfusepath{clip}%
\pgfsetbuttcap%
\pgfsetmiterjoin%
\definecolor{currentfill}{rgb}{0.000000,0.000000,1.000000}%
\pgfsetfillcolor{currentfill}%
\pgfsetlinewidth{0.000000pt}%
\definecolor{currentstroke}{rgb}{0.000000,0.000000,0.000000}%
\pgfsetstrokecolor{currentstroke}%
\pgfsetdash{}{0pt}%
\pgfpathmoveto{\pgfqpoint{5.801941in}{0.716028in}}%
\pgfpathlineto{\pgfqpoint{5.988008in}{0.716028in}}%
\pgfpathlineto{\pgfqpoint{5.988008in}{3.967885in}}%
\pgfpathlineto{\pgfqpoint{5.801941in}{3.967885in}}%
\pgfpathlineto{\pgfqpoint{5.801941in}{0.716028in}}%
\pgfpathclose%
\pgfusepath{fill}%
\end{pgfscope}%
\begin{pgfscope}%
\pgfsetbuttcap%
\pgfsetroundjoin%
\definecolor{currentfill}{rgb}{0.000000,0.000000,0.000000}%
\pgfsetfillcolor{currentfill}%
\pgfsetlinewidth{0.200750pt}%
\definecolor{currentstroke}{rgb}{0.000000,0.000000,0.000000}%
\pgfsetstrokecolor{currentstroke}%
\pgfsetdash{}{0pt}%
\pgfsys@defobject{currentmarker}{\pgfqpoint{0.000000in}{-0.048611in}}{\pgfqpoint{0.000000in}{0.000000in}}{%
\pgfpathmoveto{\pgfqpoint{0.000000in}{0.000000in}}%
\pgfpathlineto{\pgfqpoint{0.000000in}{-0.048611in}}%
\pgfusepath{stroke,fill}%
}%
\begin{pgfscope}%
\pgfsys@transformshift{0.592068in}{0.716028in}%
\pgfsys@useobject{currentmarker}{}%
\end{pgfscope}%
\end{pgfscope}%
\begin{pgfscope}%
\definecolor{textcolor}{rgb}{0.000000,0.000000,0.000000}%
\pgfsetstrokecolor{textcolor}%
\pgfsetfillcolor{textcolor}%
\pgftext[x=0.592068in,y=0.618806in,,top]{\color{textcolor}{\rmfamily\fontsize{20.000000}{24.000000}\selectfont\catcode`\^=\active\def^{\ifmmode\sp\else\^{}\fi}\catcode`\%=\active\def
\end{pgfscope}%
\begin{pgfscope}%
\pgfsetbuttcap%
\pgfsetroundjoin%
\definecolor{currentfill}{rgb}{0.000000,0.000000,0.000000}%
\pgfsetfillcolor{currentfill}%
\pgfsetlinewidth{0.200750pt}%
\definecolor{currentstroke}{rgb}{0.000000,0.000000,0.000000}%
\pgfsetstrokecolor{currentstroke}%
\pgfsetdash{}{0pt}%
\pgfsys@defobject{currentmarker}{\pgfqpoint{0.000000in}{-0.048611in}}{\pgfqpoint{0.000000in}{0.000000in}}{%
\pgfpathmoveto{\pgfqpoint{0.000000in}{0.000000in}}%
\pgfpathlineto{\pgfqpoint{0.000000in}{-0.048611in}}%
\pgfusepath{stroke,fill}%
}%
\begin{pgfscope}%
\pgfsys@transformshift{1.671256in}{0.716028in}%
\pgfsys@useobject{currentmarker}{}%
\end{pgfscope}%
\end{pgfscope}%
\begin{pgfscope}%
\definecolor{textcolor}{rgb}{0.000000,0.000000,0.000000}%
\pgfsetstrokecolor{textcolor}%
\pgfsetfillcolor{textcolor}%
\pgftext[x=1.671256in,y=0.618806in,,top]{\color{textcolor}{\rmfamily\fontsize{20.000000}{24.000000}\selectfont\catcode`\^=\active\def^{\ifmmode\sp\else\^{}\fi}\catcode`\%=\active\def
\end{pgfscope}%
\begin{pgfscope}%
\pgfsetbuttcap%
\pgfsetroundjoin%
\definecolor{currentfill}{rgb}{0.000000,0.000000,0.000000}%
\pgfsetfillcolor{currentfill}%
\pgfsetlinewidth{0.200750pt}%
\definecolor{currentstroke}{rgb}{0.000000,0.000000,0.000000}%
\pgfsetstrokecolor{currentstroke}%
\pgfsetdash{}{0pt}%
\pgfsys@defobject{currentmarker}{\pgfqpoint{0.000000in}{-0.048611in}}{\pgfqpoint{0.000000in}{0.000000in}}{%
\pgfpathmoveto{\pgfqpoint{0.000000in}{0.000000in}}%
\pgfpathlineto{\pgfqpoint{0.000000in}{-0.048611in}}%
\pgfusepath{stroke,fill}%
}%
\begin{pgfscope}%
\pgfsys@transformshift{2.750444in}{0.716028in}%
\pgfsys@useobject{currentmarker}{}%
\end{pgfscope}%
\end{pgfscope}%
\begin{pgfscope}%
\definecolor{textcolor}{rgb}{0.000000,0.000000,0.000000}%
\pgfsetstrokecolor{textcolor}%
\pgfsetfillcolor{textcolor}%
\pgftext[x=2.750444in,y=0.618806in,,top]{\color{textcolor}{\rmfamily\fontsize{20.000000}{24.000000}\selectfont\catcode`\^=\active\def^{\ifmmode\sp\else\^{}\fi}\catcode`\%=\active\def
\end{pgfscope}%
\begin{pgfscope}%
\pgfsetbuttcap%
\pgfsetroundjoin%
\definecolor{currentfill}{rgb}{0.000000,0.000000,0.000000}%
\pgfsetfillcolor{currentfill}%
\pgfsetlinewidth{0.200750pt}%
\definecolor{currentstroke}{rgb}{0.000000,0.000000,0.000000}%
\pgfsetstrokecolor{currentstroke}%
\pgfsetdash{}{0pt}%
\pgfsys@defobject{currentmarker}{\pgfqpoint{0.000000in}{-0.048611in}}{\pgfqpoint{0.000000in}{0.000000in}}{%
\pgfpathmoveto{\pgfqpoint{0.000000in}{0.000000in}}%
\pgfpathlineto{\pgfqpoint{0.000000in}{-0.048611in}}%
\pgfusepath{stroke,fill}%
}%
\begin{pgfscope}%
\pgfsys@transformshift{3.829632in}{0.716028in}%
\pgfsys@useobject{currentmarker}{}%
\end{pgfscope}%
\end{pgfscope}%
\begin{pgfscope}%
\definecolor{textcolor}{rgb}{0.000000,0.000000,0.000000}%
\pgfsetstrokecolor{textcolor}%
\pgfsetfillcolor{textcolor}%
\pgftext[x=3.829632in,y=0.618806in,,top]{\color{textcolor}{\rmfamily\fontsize{20.000000}{24.000000}\selectfont\catcode`\^=\active\def^{\ifmmode\sp\else\^{}\fi}\catcode`\%=\active\def
\end{pgfscope}%
\begin{pgfscope}%
\pgfsetbuttcap%
\pgfsetroundjoin%
\definecolor{currentfill}{rgb}{0.000000,0.000000,0.000000}%
\pgfsetfillcolor{currentfill}%
\pgfsetlinewidth{0.200750pt}%
\definecolor{currentstroke}{rgb}{0.000000,0.000000,0.000000}%
\pgfsetstrokecolor{currentstroke}%
\pgfsetdash{}{0pt}%
\pgfsys@defobject{currentmarker}{\pgfqpoint{0.000000in}{-0.048611in}}{\pgfqpoint{0.000000in}{0.000000in}}{%
\pgfpathmoveto{\pgfqpoint{0.000000in}{0.000000in}}%
\pgfpathlineto{\pgfqpoint{0.000000in}{-0.048611in}}%
\pgfusepath{stroke,fill}%
}%
\begin{pgfscope}%
\pgfsys@transformshift{4.908820in}{0.716028in}%
\pgfsys@useobject{currentmarker}{}%
\end{pgfscope}%
\end{pgfscope}%
\begin{pgfscope}%
\definecolor{textcolor}{rgb}{0.000000,0.000000,0.000000}%
\pgfsetstrokecolor{textcolor}%
\pgfsetfillcolor{textcolor}%
\pgftext[x=4.908820in,y=0.618806in,,top]{\color{textcolor}{\rmfamily\fontsize{20.000000}{24.000000}\selectfont\catcode`\^=\active\def^{\ifmmode\sp\else\^{}\fi}\catcode`\%=\active\def
\end{pgfscope}%
\begin{pgfscope}%
\pgfsetbuttcap%
\pgfsetroundjoin%
\definecolor{currentfill}{rgb}{0.000000,0.000000,0.000000}%
\pgfsetfillcolor{currentfill}%
\pgfsetlinewidth{0.200750pt}%
\definecolor{currentstroke}{rgb}{0.000000,0.000000,0.000000}%
\pgfsetstrokecolor{currentstroke}%
\pgfsetdash{}{0pt}%
\pgfsys@defobject{currentmarker}{\pgfqpoint{0.000000in}{-0.048611in}}{\pgfqpoint{0.000000in}{0.000000in}}{%
\pgfpathmoveto{\pgfqpoint{0.000000in}{0.000000in}}%
\pgfpathlineto{\pgfqpoint{0.000000in}{-0.048611in}}%
\pgfusepath{stroke,fill}%
}%
\begin{pgfscope}%
\pgfsys@transformshift{5.988008in}{0.716028in}%
\pgfsys@useobject{currentmarker}{}%
\end{pgfscope}%
\end{pgfscope}%
\begin{pgfscope}%
\definecolor{textcolor}{rgb}{0.000000,0.000000,0.000000}%
\pgfsetstrokecolor{textcolor}%
\pgfsetfillcolor{textcolor}%
\pgftext[x=5.988008in,y=0.618806in,,top]{\color{textcolor}{\rmfamily\fontsize{20.000000}{24.000000}\selectfont\catcode`\^=\active\def^{\ifmmode\sp\else\^{}\fi}\catcode`\%=\active\def
\end{pgfscope}%
\begin{pgfscope}%
\definecolor{textcolor}{rgb}{0.000000,0.000000,0.000000}%
\pgfsetstrokecolor{textcolor}%
\pgfsetfillcolor{textcolor}%
\pgftext[x=3.290038in,y=0.307183in,,top]{\color{textcolor}{\rmfamily\fontsize{26.000000}{31.200000}\selectfont\catcode`\^=\active\def^{\ifmmode\sp\else\^{}\fi}\catcode`\%=\active\def
\end{pgfscope}%
\begin{pgfscope}%
\pgfsetbuttcap%
\pgfsetroundjoin%
\definecolor{currentfill}{rgb}{0.000000,0.000000,0.000000}%
\pgfsetfillcolor{currentfill}%
\pgfsetlinewidth{0.200750pt}%
\definecolor{currentstroke}{rgb}{0.000000,0.000000,0.000000}%
\pgfsetstrokecolor{currentstroke}%
\pgfsetdash{}{0pt}%
\pgfsys@defobject{currentmarker}{\pgfqpoint{-0.048611in}{0.000000in}}{\pgfqpoint{-0.000000in}{0.000000in}}{%
\pgfpathmoveto{\pgfqpoint{-0.000000in}{0.000000in}}%
\pgfpathlineto{\pgfqpoint{-0.048611in}{0.000000in}}%
\pgfusepath{stroke,fill}%
}%
\begin{pgfscope}%
\pgfsys@transformshift{0.592068in}{0.716028in}%
\pgfsys@useobject{currentmarker}{}%
\end{pgfscope}%
\end{pgfscope}%
\begin{pgfscope}%
\definecolor{textcolor}{rgb}{0.000000,0.000000,0.000000}%
\pgfsetstrokecolor{textcolor}%
\pgfsetfillcolor{textcolor}%
\pgftext[x=0.362738in, y=0.616009in, left, base]{\color{textcolor}{\rmfamily\fontsize{20.000000}{24.000000}\selectfont\catcode`\^=\active\def^{\ifmmode\sp\else\^{}\fi}\catcode`\%=\active\def
\end{pgfscope}%
\begin{pgfscope}%
\pgfsetbuttcap%
\pgfsetroundjoin%
\definecolor{currentfill}{rgb}{0.000000,0.000000,0.000000}%
\pgfsetfillcolor{currentfill}%
\pgfsetlinewidth{0.200750pt}%
\definecolor{currentstroke}{rgb}{0.000000,0.000000,0.000000}%
\pgfsetstrokecolor{currentstroke}%
\pgfsetdash{}{0pt}%
\pgfsys@defobject{currentmarker}{\pgfqpoint{-0.048611in}{0.000000in}}{\pgfqpoint{-0.000000in}{0.000000in}}{%
\pgfpathmoveto{\pgfqpoint{-0.000000in}{0.000000in}}%
\pgfpathlineto{\pgfqpoint{-0.048611in}{0.000000in}}%
\pgfusepath{stroke,fill}%
}%
\begin{pgfscope}%
\pgfsys@transformshift{0.592068in}{1.728080in}%
\pgfsys@useobject{currentmarker}{}%
\end{pgfscope}%
\end{pgfscope}%
\begin{pgfscope}%
\definecolor{textcolor}{rgb}{0.000000,0.000000,0.000000}%
\pgfsetstrokecolor{textcolor}%
\pgfsetfillcolor{textcolor}%
\pgftext[x=0.362738in, y=1.628061in, left, base]{\color{textcolor}{\rmfamily\fontsize{20.000000}{24.000000}\selectfont\catcode`\^=\active\def^{\ifmmode\sp\else\^{}\fi}\catcode`\%=\active\def
\end{pgfscope}%
\begin{pgfscope}%
\pgfsetbuttcap%
\pgfsetroundjoin%
\definecolor{currentfill}{rgb}{0.000000,0.000000,0.000000}%
\pgfsetfillcolor{currentfill}%
\pgfsetlinewidth{0.200750pt}%
\definecolor{currentstroke}{rgb}{0.000000,0.000000,0.000000}%
\pgfsetstrokecolor{currentstroke}%
\pgfsetdash{}{0pt}%
\pgfsys@defobject{currentmarker}{\pgfqpoint{-0.048611in}{0.000000in}}{\pgfqpoint{-0.000000in}{0.000000in}}{%
\pgfpathmoveto{\pgfqpoint{-0.000000in}{0.000000in}}%
\pgfpathlineto{\pgfqpoint{-0.048611in}{0.000000in}}%
\pgfusepath{stroke,fill}%
}%
\begin{pgfscope}%
\pgfsys@transformshift{0.592068in}{2.740132in}%
\pgfsys@useobject{currentmarker}{}%
\end{pgfscope}%
\end{pgfscope}%
\begin{pgfscope}%
\definecolor{textcolor}{rgb}{0.000000,0.000000,0.000000}%
\pgfsetstrokecolor{textcolor}%
\pgfsetfillcolor{textcolor}%
\pgftext[x=0.362738in, y=2.640112in, left, base]{\color{textcolor}{\rmfamily\fontsize{20.000000}{24.000000}\selectfont\catcode`\^=\active\def^{\ifmmode\sp\else\^{}\fi}\catcode`\%=\active\def
\end{pgfscope}%
\begin{pgfscope}%
\pgfsetbuttcap%
\pgfsetroundjoin%
\definecolor{currentfill}{rgb}{0.000000,0.000000,0.000000}%
\pgfsetfillcolor{currentfill}%
\pgfsetlinewidth{0.200750pt}%
\definecolor{currentstroke}{rgb}{0.000000,0.000000,0.000000}%
\pgfsetstrokecolor{currentstroke}%
\pgfsetdash{}{0pt}%
\pgfsys@defobject{currentmarker}{\pgfqpoint{-0.048611in}{0.000000in}}{\pgfqpoint{-0.000000in}{0.000000in}}{%
\pgfpathmoveto{\pgfqpoint{-0.000000in}{0.000000in}}%
\pgfpathlineto{\pgfqpoint{-0.048611in}{0.000000in}}%
\pgfusepath{stroke,fill}%
}%
\begin{pgfscope}%
\pgfsys@transformshift{0.592068in}{3.752184in}%
\pgfsys@useobject{currentmarker}{}%
\end{pgfscope}%
\end{pgfscope}%
\begin{pgfscope}%
\definecolor{textcolor}{rgb}{0.000000,0.000000,0.000000}%
\pgfsetstrokecolor{textcolor}%
\pgfsetfillcolor{textcolor}%
\pgftext[x=0.362738in, y=3.652164in, left, base]{\color{textcolor}{\rmfamily\fontsize{20.000000}{24.000000}\selectfont\catcode`\^=\active\def^{\ifmmode\sp\else\^{}\fi}\catcode`\%=\active\def
\end{pgfscope}%
\begin{pgfscope}%
\definecolor{textcolor}{rgb}{0.000000,0.000000,0.000000}%
\pgfsetstrokecolor{textcolor}%
\pgfsetfillcolor{textcolor}%
\pgftext[x=0.307183in,y=2.474950in,,bottom,rotate=90.000000]{\color{textcolor}{\rmfamily\fontsize{26.000000}{31.200000}\selectfont\catcode`\^=\active\def^{\ifmmode\sp\else\^{}\fi}\catcode`\%=\active\def
\end{pgfscope}%
\begin{pgfscope}%
\pgfpathrectangle{\pgfqpoint{0.592068in}{0.716028in}}{\pgfqpoint{5.395940in}{3.517843in}}%
\pgfusepath{clip}%
\pgfsetbuttcap%
\pgfsetroundjoin%
\pgfsetlinewidth{2.007500pt}%
\definecolor{currentstroke}{rgb}{0.501961,0.501961,0.501961}%
\pgfsetstrokecolor{currentstroke}%
\pgfsetdash{{7.400000pt}{3.200000pt}}{0.000000pt}%
\pgfpathmoveto{\pgfqpoint{0.592068in}{1.728080in}}%
\pgfpathlineto{\pgfqpoint{5.988008in}{1.728080in}}%
\pgfusepath{stroke}%
\end{pgfscope}%
\begin{pgfscope}%
\pgfsetrectcap%
\pgfsetmiterjoin%
\pgfsetlinewidth{0.803000pt}%
\definecolor{currentstroke}{rgb}{0.000000,0.000000,0.000000}%
\pgfsetstrokecolor{currentstroke}%
\pgfsetdash{}{0pt}%
\pgfpathmoveto{\pgfqpoint{0.592068in}{0.716028in}}%
\pgfpathlineto{\pgfqpoint{0.592068in}{4.233871in}}%
\pgfusepath{stroke}%
\end{pgfscope}%
\begin{pgfscope}%
\pgfsetrectcap%
\pgfsetmiterjoin%
\pgfsetlinewidth{0.803000pt}%
\definecolor{currentstroke}{rgb}{0.000000,0.000000,0.000000}%
\pgfsetstrokecolor{currentstroke}%
\pgfsetdash{}{0pt}%
\pgfpathmoveto{\pgfqpoint{5.988008in}{0.716028in}}%
\pgfpathlineto{\pgfqpoint{5.988008in}{4.233871in}}%
\pgfusepath{stroke}%
\end{pgfscope}%
\begin{pgfscope}%
\pgfsetrectcap%
\pgfsetmiterjoin%
\pgfsetlinewidth{0.803000pt}%
\definecolor{currentstroke}{rgb}{0.000000,0.000000,0.000000}%
\pgfsetstrokecolor{currentstroke}%
\pgfsetdash{}{0pt}%
\pgfpathmoveto{\pgfqpoint{0.592068in}{0.716028in}}%
\pgfpathlineto{\pgfqpoint{5.988008in}{0.716028in}}%
\pgfusepath{stroke}%
\end{pgfscope}%
\begin{pgfscope}%
\pgfsetrectcap%
\pgfsetmiterjoin%
\pgfsetlinewidth{0.803000pt}%
\definecolor{currentstroke}{rgb}{0.000000,0.000000,0.000000}%
\pgfsetstrokecolor{currentstroke}%
\pgfsetdash{}{0pt}%
\pgfpathmoveto{\pgfqpoint{0.592068in}{4.233871in}}%
\pgfpathlineto{\pgfqpoint{5.988008in}{4.233871in}}%
\pgfusepath{stroke}%
\end{pgfscope}%
\begin{pgfscope}%
\definecolor{textcolor}{rgb}{0.000000,0.000000,0.000000}%
\pgfsetstrokecolor{textcolor}%
\pgfsetfillcolor{textcolor}%
\pgftext[x=3.290038in,y=4.317204in,,base]{\color{textcolor}{\rmfamily\fontsize{25.000000}{30.000000}\selectfont\catcode`\^=\active\def^{\ifmmode\sp\else\^{}\fi}\catcode`\%=\active\def
\end{pgfscope}%
\begin{pgfscope}%
\pgfsetbuttcap%
\pgfsetmiterjoin%
\definecolor{currentfill}{rgb}{1.000000,1.000000,1.000000}%
\pgfsetfillcolor{currentfill}%
\pgfsetfillopacity{0.800000}%
\pgfsetlinewidth{1.003750pt}%
\definecolor{currentstroke}{rgb}{0.800000,0.800000,0.800000}%
\pgfsetstrokecolor{currentstroke}%
\pgfsetstrokeopacity{0.800000}%
\pgfsetdash{}{0pt}%
\pgfpathmoveto{\pgfqpoint{0.786512in}{3.616692in}}%
\pgfpathlineto{\pgfqpoint{2.328414in}{3.616692in}}%
\pgfpathquadraticcurveto{\pgfqpoint{2.383970in}{3.616692in}}{\pgfqpoint{2.383970in}{3.672248in}}%
\pgfpathlineto{\pgfqpoint{2.383970in}{4.039427in}}%
\pgfpathquadraticcurveto{\pgfqpoint{2.383970in}{4.094982in}}{\pgfqpoint{2.328414in}{4.094982in}}%
\pgfpathlineto{\pgfqpoint{0.786512in}{4.094982in}}%
\pgfpathquadraticcurveto{\pgfqpoint{0.730957in}{4.094982in}}{\pgfqpoint{0.730957in}{4.039427in}}%
\pgfpathlineto{\pgfqpoint{0.730957in}{3.672248in}}%
\pgfpathquadraticcurveto{\pgfqpoint{0.730957in}{3.616692in}}{\pgfqpoint{0.786512in}{3.616692in}}%
\pgfpathlineto{\pgfqpoint{0.786512in}{3.616692in}}%
\pgfpathclose%
\pgfusepath{stroke,fill}%
\end{pgfscope}%
\begin{pgfscope}%
\pgfsetbuttcap%
\pgfsetroundjoin%
\pgfsetlinewidth{2.007500pt}%
\definecolor{currentstroke}{rgb}{0.501961,0.501961,0.501961}%
\pgfsetstrokecolor{currentstroke}%
\pgfsetdash{{7.400000pt}{3.200000pt}}{0.000000pt}%
\pgfpathmoveto{\pgfqpoint{0.842068in}{3.881055in}}%
\pgfpathlineto{\pgfqpoint{1.119846in}{3.881055in}}%
\pgfpathlineto{\pgfqpoint{1.397623in}{3.881055in}}%
\pgfusepath{stroke}%
\end{pgfscope}%
\begin{pgfscope}%
\definecolor{textcolor}{rgb}{0.000000,0.000000,0.000000}%
\pgfsetstrokecolor{textcolor}%
\pgfsetfillcolor{textcolor}%
\pgftext[x=1.619846in,y=3.783833in,left,base]{\color{textcolor}{\rmfamily\fontsize{20.000000}{24.000000}\selectfont\catcode`\^=\active\def^{\ifmmode\sp\else\^{}\fi}\catcode`\%=\active\def
\end{pgfscope}%
\end{pgfpicture}%
\makeatother%
\endgroup%

%% file: figures/evaluation/BPIC17/BPIC17_PIT_remaining_time_norm_4layer.pgf
\begingroup%
\makeatletter%
\begin{pgfpicture}%
\pgfpathrectangle{\pgfpointorigin}{\pgfqpoint{6.159289in}{4.557174in}}%
\pgfusepath{use as bounding box, clip}%
\begin{pgfscope}%
\pgfsetbuttcap%
\pgfsetmiterjoin%
\definecolor{currentfill}{rgb}{1.000000,1.000000,1.000000}%
\pgfsetfillcolor{currentfill}%
\pgfsetlinewidth{0.000000pt}%
\definecolor{currentstroke}{rgb}{1.000000,1.000000,1.000000}%
\pgfsetstrokecolor{currentstroke}%
\pgfsetdash{}{0pt}%
\pgfpathmoveto{\pgfqpoint{0.000000in}{0.000000in}}%
\pgfpathlineto{\pgfqpoint{6.159289in}{0.000000in}}%
\pgfpathlineto{\pgfqpoint{6.159289in}{4.557174in}}%
\pgfpathlineto{\pgfqpoint{0.000000in}{4.557174in}}%
\pgfpathlineto{\pgfqpoint{0.000000in}{0.000000in}}%
\pgfpathclose%
\pgfusepath{fill}%
\end{pgfscope}%
\begin{pgfscope}%
\pgfsetbuttcap%
\pgfsetmiterjoin%
\definecolor{currentfill}{rgb}{1.000000,1.000000,1.000000}%
\pgfsetfillcolor{currentfill}%
\pgfsetlinewidth{0.000000pt}%
\definecolor{currentstroke}{rgb}{0.000000,0.000000,0.000000}%
\pgfsetstrokecolor{currentstroke}%
\pgfsetstrokeopacity{0.000000}%
\pgfsetdash{}{0pt}%
\pgfpathmoveto{\pgfqpoint{0.592068in}{0.716028in}}%
\pgfpathlineto{\pgfqpoint{5.988008in}{0.716028in}}%
\pgfpathlineto{\pgfqpoint{5.988008in}{4.233871in}}%
\pgfpathlineto{\pgfqpoint{0.592068in}{4.233871in}}%
\pgfpathlineto{\pgfqpoint{0.592068in}{0.716028in}}%
\pgfpathclose%
\pgfusepath{fill}%
\end{pgfscope}%
\begin{pgfscope}%
\pgfpathrectangle{\pgfqpoint{0.592068in}{0.716028in}}{\pgfqpoint{5.395940in}{3.517843in}}%
\pgfusepath{clip}%
\pgfsetbuttcap%
\pgfsetmiterjoin%
\definecolor{currentfill}{rgb}{0.000000,0.000000,1.000000}%
\pgfsetfillcolor{currentfill}%
\pgfsetlinewidth{0.000000pt}%
\definecolor{currentstroke}{rgb}{0.000000,0.000000,0.000000}%
\pgfsetstrokecolor{currentstroke}%
\pgfsetdash{}{0pt}%
\pgfpathmoveto{\pgfqpoint{0.592068in}{0.716028in}}%
\pgfpathlineto{\pgfqpoint{0.778135in}{0.716028in}}%
\pgfpathlineto{\pgfqpoint{0.778135in}{4.066355in}}%
\pgfpathlineto{\pgfqpoint{0.592068in}{4.066355in}}%
\pgfpathlineto{\pgfqpoint{0.592068in}{0.716028in}}%
\pgfpathclose%
\pgfusepath{fill}%
\end{pgfscope}%
\begin{pgfscope}%
\pgfpathrectangle{\pgfqpoint{0.592068in}{0.716028in}}{\pgfqpoint{5.395940in}{3.517843in}}%
\pgfusepath{clip}%
\pgfsetbuttcap%
\pgfsetmiterjoin%
\definecolor{currentfill}{rgb}{0.000000,0.000000,1.000000}%
\pgfsetfillcolor{currentfill}%
\pgfsetlinewidth{0.000000pt}%
\definecolor{currentstroke}{rgb}{0.000000,0.000000,0.000000}%
\pgfsetstrokecolor{currentstroke}%
\pgfsetdash{}{0pt}%
\pgfpathmoveto{\pgfqpoint{0.778135in}{0.716028in}}%
\pgfpathlineto{\pgfqpoint{0.964202in}{0.716028in}}%
\pgfpathlineto{\pgfqpoint{0.964202in}{3.313560in}}%
\pgfpathlineto{\pgfqpoint{0.778135in}{3.313560in}}%
\pgfpathlineto{\pgfqpoint{0.778135in}{0.716028in}}%
\pgfpathclose%
\pgfusepath{fill}%
\end{pgfscope}%
\begin{pgfscope}%
\pgfpathrectangle{\pgfqpoint{0.592068in}{0.716028in}}{\pgfqpoint{5.395940in}{3.517843in}}%
\pgfusepath{clip}%
\pgfsetbuttcap%
\pgfsetmiterjoin%
\definecolor{currentfill}{rgb}{0.000000,0.000000,1.000000}%
\pgfsetfillcolor{currentfill}%
\pgfsetlinewidth{0.000000pt}%
\definecolor{currentstroke}{rgb}{0.000000,0.000000,0.000000}%
\pgfsetstrokecolor{currentstroke}%
\pgfsetdash{}{0pt}%
\pgfpathmoveto{\pgfqpoint{0.964202in}{0.716028in}}%
\pgfpathlineto{\pgfqpoint{1.150269in}{0.716028in}}%
\pgfpathlineto{\pgfqpoint{1.150269in}{2.603205in}}%
\pgfpathlineto{\pgfqpoint{0.964202in}{2.603205in}}%
\pgfpathlineto{\pgfqpoint{0.964202in}{0.716028in}}%
\pgfpathclose%
\pgfusepath{fill}%
\end{pgfscope}%
\begin{pgfscope}%
\pgfpathrectangle{\pgfqpoint{0.592068in}{0.716028in}}{\pgfqpoint{5.395940in}{3.517843in}}%
\pgfusepath{clip}%
\pgfsetbuttcap%
\pgfsetmiterjoin%
\definecolor{currentfill}{rgb}{0.000000,0.000000,1.000000}%
\pgfsetfillcolor{currentfill}%
\pgfsetlinewidth{0.000000pt}%
\definecolor{currentstroke}{rgb}{0.000000,0.000000,0.000000}%
\pgfsetstrokecolor{currentstroke}%
\pgfsetdash{}{0pt}%
\pgfpathmoveto{\pgfqpoint{1.150269in}{0.716028in}}%
\pgfpathlineto{\pgfqpoint{1.336335in}{0.716028in}}%
\pgfpathlineto{\pgfqpoint{1.336335in}{2.263145in}}%
\pgfpathlineto{\pgfqpoint{1.150269in}{2.263145in}}%
\pgfpathlineto{\pgfqpoint{1.150269in}{0.716028in}}%
\pgfpathclose%
\pgfusepath{fill}%
\end{pgfscope}%
\begin{pgfscope}%
\pgfpathrectangle{\pgfqpoint{0.592068in}{0.716028in}}{\pgfqpoint{5.395940in}{3.517843in}}%
\pgfusepath{clip}%
\pgfsetbuttcap%
\pgfsetmiterjoin%
\definecolor{currentfill}{rgb}{0.000000,0.000000,1.000000}%
\pgfsetfillcolor{currentfill}%
\pgfsetlinewidth{0.000000pt}%
\definecolor{currentstroke}{rgb}{0.000000,0.000000,0.000000}%
\pgfsetstrokecolor{currentstroke}%
\pgfsetdash{}{0pt}%
\pgfpathmoveto{\pgfqpoint{1.336335in}{0.716028in}}%
\pgfpathlineto{\pgfqpoint{1.522402in}{0.716028in}}%
\pgfpathlineto{\pgfqpoint{1.522402in}{2.197028in}}%
\pgfpathlineto{\pgfqpoint{1.336335in}{2.197028in}}%
\pgfpathlineto{\pgfqpoint{1.336335in}{0.716028in}}%
\pgfpathclose%
\pgfusepath{fill}%
\end{pgfscope}%
\begin{pgfscope}%
\pgfpathrectangle{\pgfqpoint{0.592068in}{0.716028in}}{\pgfqpoint{5.395940in}{3.517843in}}%
\pgfusepath{clip}%
\pgfsetbuttcap%
\pgfsetmiterjoin%
\definecolor{currentfill}{rgb}{0.000000,0.000000,1.000000}%
\pgfsetfillcolor{currentfill}%
\pgfsetlinewidth{0.000000pt}%
\definecolor{currentstroke}{rgb}{0.000000,0.000000,0.000000}%
\pgfsetstrokecolor{currentstroke}%
\pgfsetdash{}{0pt}%
\pgfpathmoveto{\pgfqpoint{1.522402in}{0.716028in}}%
\pgfpathlineto{\pgfqpoint{1.708469in}{0.716028in}}%
\pgfpathlineto{\pgfqpoint{1.708469in}{2.108507in}}%
\pgfpathlineto{\pgfqpoint{1.522402in}{2.108507in}}%
\pgfpathlineto{\pgfqpoint{1.522402in}{0.716028in}}%
\pgfpathclose%
\pgfusepath{fill}%
\end{pgfscope}%
\begin{pgfscope}%
\pgfpathrectangle{\pgfqpoint{0.592068in}{0.716028in}}{\pgfqpoint{5.395940in}{3.517843in}}%
\pgfusepath{clip}%
\pgfsetbuttcap%
\pgfsetmiterjoin%
\definecolor{currentfill}{rgb}{0.000000,0.000000,1.000000}%
\pgfsetfillcolor{currentfill}%
\pgfsetlinewidth{0.000000pt}%
\definecolor{currentstroke}{rgb}{0.000000,0.000000,0.000000}%
\pgfsetstrokecolor{currentstroke}%
\pgfsetdash{}{0pt}%
\pgfpathmoveto{\pgfqpoint{1.708469in}{0.716028in}}%
\pgfpathlineto{\pgfqpoint{1.894536in}{0.716028in}}%
\pgfpathlineto{\pgfqpoint{1.894536in}{2.040567in}}%
\pgfpathlineto{\pgfqpoint{1.708469in}{2.040567in}}%
\pgfpathlineto{\pgfqpoint{1.708469in}{0.716028in}}%
\pgfpathclose%
\pgfusepath{fill}%
\end{pgfscope}%
\begin{pgfscope}%
\pgfpathrectangle{\pgfqpoint{0.592068in}{0.716028in}}{\pgfqpoint{5.395940in}{3.517843in}}%
\pgfusepath{clip}%
\pgfsetbuttcap%
\pgfsetmiterjoin%
\definecolor{currentfill}{rgb}{0.000000,0.000000,1.000000}%
\pgfsetfillcolor{currentfill}%
\pgfsetlinewidth{0.000000pt}%
\definecolor{currentstroke}{rgb}{0.000000,0.000000,0.000000}%
\pgfsetstrokecolor{currentstroke}%
\pgfsetdash{}{0pt}%
\pgfpathmoveto{\pgfqpoint{1.894536in}{0.716028in}}%
\pgfpathlineto{\pgfqpoint{2.080603in}{0.716028in}}%
\pgfpathlineto{\pgfqpoint{2.080603in}{1.919078in}}%
\pgfpathlineto{\pgfqpoint{1.894536in}{1.919078in}}%
\pgfpathlineto{\pgfqpoint{1.894536in}{0.716028in}}%
\pgfpathclose%
\pgfusepath{fill}%
\end{pgfscope}%
\begin{pgfscope}%
\pgfpathrectangle{\pgfqpoint{0.592068in}{0.716028in}}{\pgfqpoint{5.395940in}{3.517843in}}%
\pgfusepath{clip}%
\pgfsetbuttcap%
\pgfsetmiterjoin%
\definecolor{currentfill}{rgb}{0.000000,0.000000,1.000000}%
\pgfsetfillcolor{currentfill}%
\pgfsetlinewidth{0.000000pt}%
\definecolor{currentstroke}{rgb}{0.000000,0.000000,0.000000}%
\pgfsetstrokecolor{currentstroke}%
\pgfsetdash{}{0pt}%
\pgfpathmoveto{\pgfqpoint{2.080603in}{0.716028in}}%
\pgfpathlineto{\pgfqpoint{2.266670in}{0.716028in}}%
\pgfpathlineto{\pgfqpoint{2.266670in}{1.905964in}}%
\pgfpathlineto{\pgfqpoint{2.080603in}{1.905964in}}%
\pgfpathlineto{\pgfqpoint{2.080603in}{0.716028in}}%
\pgfpathclose%
\pgfusepath{fill}%
\end{pgfscope}%
\begin{pgfscope}%
\pgfpathrectangle{\pgfqpoint{0.592068in}{0.716028in}}{\pgfqpoint{5.395940in}{3.517843in}}%
\pgfusepath{clip}%
\pgfsetbuttcap%
\pgfsetmiterjoin%
\definecolor{currentfill}{rgb}{0.000000,0.000000,1.000000}%
\pgfsetfillcolor{currentfill}%
\pgfsetlinewidth{0.000000pt}%
\definecolor{currentstroke}{rgb}{0.000000,0.000000,0.000000}%
\pgfsetstrokecolor{currentstroke}%
\pgfsetdash{}{0pt}%
\pgfpathmoveto{\pgfqpoint{2.266670in}{0.716028in}}%
\pgfpathlineto{\pgfqpoint{2.452737in}{0.716028in}}%
\pgfpathlineto{\pgfqpoint{2.452737in}{1.858789in}}%
\pgfpathlineto{\pgfqpoint{2.266670in}{1.858789in}}%
\pgfpathlineto{\pgfqpoint{2.266670in}{0.716028in}}%
\pgfpathclose%
\pgfusepath{fill}%
\end{pgfscope}%
\begin{pgfscope}%
\pgfpathrectangle{\pgfqpoint{0.592068in}{0.716028in}}{\pgfqpoint{5.395940in}{3.517843in}}%
\pgfusepath{clip}%
\pgfsetbuttcap%
\pgfsetmiterjoin%
\definecolor{currentfill}{rgb}{0.000000,0.000000,1.000000}%
\pgfsetfillcolor{currentfill}%
\pgfsetlinewidth{0.000000pt}%
\definecolor{currentstroke}{rgb}{0.000000,0.000000,0.000000}%
\pgfsetstrokecolor{currentstroke}%
\pgfsetdash{}{0pt}%
\pgfpathmoveto{\pgfqpoint{2.452737in}{0.716028in}}%
\pgfpathlineto{\pgfqpoint{2.638804in}{0.716028in}}%
\pgfpathlineto{\pgfqpoint{2.638804in}{1.850593in}}%
\pgfpathlineto{\pgfqpoint{2.452737in}{1.850593in}}%
\pgfpathlineto{\pgfqpoint{2.452737in}{0.716028in}}%
\pgfpathclose%
\pgfusepath{fill}%
\end{pgfscope}%
\begin{pgfscope}%
\pgfpathrectangle{\pgfqpoint{0.592068in}{0.716028in}}{\pgfqpoint{5.395940in}{3.517843in}}%
\pgfusepath{clip}%
\pgfsetbuttcap%
\pgfsetmiterjoin%
\definecolor{currentfill}{rgb}{0.000000,0.000000,1.000000}%
\pgfsetfillcolor{currentfill}%
\pgfsetlinewidth{0.000000pt}%
\definecolor{currentstroke}{rgb}{0.000000,0.000000,0.000000}%
\pgfsetstrokecolor{currentstroke}%
\pgfsetdash{}{0pt}%
\pgfpathmoveto{\pgfqpoint{2.638804in}{0.716028in}}%
\pgfpathlineto{\pgfqpoint{2.824871in}{0.716028in}}%
\pgfpathlineto{\pgfqpoint{2.824871in}{1.812707in}}%
\pgfpathlineto{\pgfqpoint{2.638804in}{1.812707in}}%
\pgfpathlineto{\pgfqpoint{2.638804in}{0.716028in}}%
\pgfpathclose%
\pgfusepath{fill}%
\end{pgfscope}%
\begin{pgfscope}%
\pgfpathrectangle{\pgfqpoint{0.592068in}{0.716028in}}{\pgfqpoint{5.395940in}{3.517843in}}%
\pgfusepath{clip}%
\pgfsetbuttcap%
\pgfsetmiterjoin%
\definecolor{currentfill}{rgb}{0.000000,0.000000,1.000000}%
\pgfsetfillcolor{currentfill}%
\pgfsetlinewidth{0.000000pt}%
\definecolor{currentstroke}{rgb}{0.000000,0.000000,0.000000}%
\pgfsetstrokecolor{currentstroke}%
\pgfsetdash{}{0pt}%
\pgfpathmoveto{\pgfqpoint{2.824871in}{0.716028in}}%
\pgfpathlineto{\pgfqpoint{3.010937in}{0.716028in}}%
\pgfpathlineto{\pgfqpoint{3.010937in}{1.820175in}}%
\pgfpathlineto{\pgfqpoint{2.824871in}{1.820175in}}%
\pgfpathlineto{\pgfqpoint{2.824871in}{0.716028in}}%
\pgfpathclose%
\pgfusepath{fill}%
\end{pgfscope}%
\begin{pgfscope}%
\pgfpathrectangle{\pgfqpoint{0.592068in}{0.716028in}}{\pgfqpoint{5.395940in}{3.517843in}}%
\pgfusepath{clip}%
\pgfsetbuttcap%
\pgfsetmiterjoin%
\definecolor{currentfill}{rgb}{0.000000,0.000000,1.000000}%
\pgfsetfillcolor{currentfill}%
\pgfsetlinewidth{0.000000pt}%
\definecolor{currentstroke}{rgb}{0.000000,0.000000,0.000000}%
\pgfsetstrokecolor{currentstroke}%
\pgfsetdash{}{0pt}%
\pgfpathmoveto{\pgfqpoint{3.010937in}{0.716028in}}%
\pgfpathlineto{\pgfqpoint{3.197004in}{0.716028in}}%
\pgfpathlineto{\pgfqpoint{3.197004in}{1.828554in}}%
\pgfpathlineto{\pgfqpoint{3.010937in}{1.828554in}}%
\pgfpathlineto{\pgfqpoint{3.010937in}{0.716028in}}%
\pgfpathclose%
\pgfusepath{fill}%
\end{pgfscope}%
\begin{pgfscope}%
\pgfpathrectangle{\pgfqpoint{0.592068in}{0.716028in}}{\pgfqpoint{5.395940in}{3.517843in}}%
\pgfusepath{clip}%
\pgfsetbuttcap%
\pgfsetmiterjoin%
\definecolor{currentfill}{rgb}{0.000000,0.000000,1.000000}%
\pgfsetfillcolor{currentfill}%
\pgfsetlinewidth{0.000000pt}%
\definecolor{currentstroke}{rgb}{0.000000,0.000000,0.000000}%
\pgfsetstrokecolor{currentstroke}%
\pgfsetdash{}{0pt}%
\pgfpathmoveto{\pgfqpoint{3.197004in}{0.716028in}}%
\pgfpathlineto{\pgfqpoint{3.383071in}{0.716028in}}%
\pgfpathlineto{\pgfqpoint{3.383071in}{1.935107in}}%
\pgfpathlineto{\pgfqpoint{3.197004in}{1.935107in}}%
\pgfpathlineto{\pgfqpoint{3.197004in}{0.716028in}}%
\pgfpathclose%
\pgfusepath{fill}%
\end{pgfscope}%
\begin{pgfscope}%
\pgfpathrectangle{\pgfqpoint{0.592068in}{0.716028in}}{\pgfqpoint{5.395940in}{3.517843in}}%
\pgfusepath{clip}%
\pgfsetbuttcap%
\pgfsetmiterjoin%
\definecolor{currentfill}{rgb}{0.000000,0.000000,1.000000}%
\pgfsetfillcolor{currentfill}%
\pgfsetlinewidth{0.000000pt}%
\definecolor{currentstroke}{rgb}{0.000000,0.000000,0.000000}%
\pgfsetstrokecolor{currentstroke}%
\pgfsetdash{}{0pt}%
\pgfpathmoveto{\pgfqpoint{3.383071in}{0.716028in}}%
\pgfpathlineto{\pgfqpoint{3.569138in}{0.716028in}}%
\pgfpathlineto{\pgfqpoint{3.569138in}{1.812161in}}%
\pgfpathlineto{\pgfqpoint{3.383071in}{1.812161in}}%
\pgfpathlineto{\pgfqpoint{3.383071in}{0.716028in}}%
\pgfpathclose%
\pgfusepath{fill}%
\end{pgfscope}%
\begin{pgfscope}%
\pgfpathrectangle{\pgfqpoint{0.592068in}{0.716028in}}{\pgfqpoint{5.395940in}{3.517843in}}%
\pgfusepath{clip}%
\pgfsetbuttcap%
\pgfsetmiterjoin%
\definecolor{currentfill}{rgb}{0.000000,0.000000,1.000000}%
\pgfsetfillcolor{currentfill}%
\pgfsetlinewidth{0.000000pt}%
\definecolor{currentstroke}{rgb}{0.000000,0.000000,0.000000}%
\pgfsetstrokecolor{currentstroke}%
\pgfsetdash{}{0pt}%
\pgfpathmoveto{\pgfqpoint{3.569138in}{0.716028in}}%
\pgfpathlineto{\pgfqpoint{3.755205in}{0.716028in}}%
\pgfpathlineto{\pgfqpoint{3.755205in}{1.694679in}}%
\pgfpathlineto{\pgfqpoint{3.569138in}{1.694679in}}%
\pgfpathlineto{\pgfqpoint{3.569138in}{0.716028in}}%
\pgfpathclose%
\pgfusepath{fill}%
\end{pgfscope}%
\begin{pgfscope}%
\pgfpathrectangle{\pgfqpoint{0.592068in}{0.716028in}}{\pgfqpoint{5.395940in}{3.517843in}}%
\pgfusepath{clip}%
\pgfsetbuttcap%
\pgfsetmiterjoin%
\definecolor{currentfill}{rgb}{0.000000,0.000000,1.000000}%
\pgfsetfillcolor{currentfill}%
\pgfsetlinewidth{0.000000pt}%
\definecolor{currentstroke}{rgb}{0.000000,0.000000,0.000000}%
\pgfsetstrokecolor{currentstroke}%
\pgfsetdash{}{0pt}%
\pgfpathmoveto{\pgfqpoint{3.755205in}{0.716028in}}%
\pgfpathlineto{\pgfqpoint{3.941272in}{0.716028in}}%
\pgfpathlineto{\pgfqpoint{3.941272in}{1.626558in}}%
\pgfpathlineto{\pgfqpoint{3.755205in}{1.626558in}}%
\pgfpathlineto{\pgfqpoint{3.755205in}{0.716028in}}%
\pgfpathclose%
\pgfusepath{fill}%
\end{pgfscope}%
\begin{pgfscope}%
\pgfpathrectangle{\pgfqpoint{0.592068in}{0.716028in}}{\pgfqpoint{5.395940in}{3.517843in}}%
\pgfusepath{clip}%
\pgfsetbuttcap%
\pgfsetmiterjoin%
\definecolor{currentfill}{rgb}{0.000000,0.000000,1.000000}%
\pgfsetfillcolor{currentfill}%
\pgfsetlinewidth{0.000000pt}%
\definecolor{currentstroke}{rgb}{0.000000,0.000000,0.000000}%
\pgfsetstrokecolor{currentstroke}%
\pgfsetdash{}{0pt}%
\pgfpathmoveto{\pgfqpoint{3.941272in}{0.716028in}}%
\pgfpathlineto{\pgfqpoint{4.127339in}{0.716028in}}%
\pgfpathlineto{\pgfqpoint{4.127339in}{1.794311in}}%
\pgfpathlineto{\pgfqpoint{3.941272in}{1.794311in}}%
\pgfpathlineto{\pgfqpoint{3.941272in}{0.716028in}}%
\pgfpathclose%
\pgfusepath{fill}%
\end{pgfscope}%
\begin{pgfscope}%
\pgfpathrectangle{\pgfqpoint{0.592068in}{0.716028in}}{\pgfqpoint{5.395940in}{3.517843in}}%
\pgfusepath{clip}%
\pgfsetbuttcap%
\pgfsetmiterjoin%
\definecolor{currentfill}{rgb}{0.000000,0.000000,1.000000}%
\pgfsetfillcolor{currentfill}%
\pgfsetlinewidth{0.000000pt}%
\definecolor{currentstroke}{rgb}{0.000000,0.000000,0.000000}%
\pgfsetstrokecolor{currentstroke}%
\pgfsetdash{}{0pt}%
\pgfpathmoveto{\pgfqpoint{4.127339in}{0.716028in}}%
\pgfpathlineto{\pgfqpoint{4.313406in}{0.716028in}}%
\pgfpathlineto{\pgfqpoint{4.313406in}{1.973175in}}%
\pgfpathlineto{\pgfqpoint{4.127339in}{1.973175in}}%
\pgfpathlineto{\pgfqpoint{4.127339in}{0.716028in}}%
\pgfpathclose%
\pgfusepath{fill}%
\end{pgfscope}%
\begin{pgfscope}%
\pgfpathrectangle{\pgfqpoint{0.592068in}{0.716028in}}{\pgfqpoint{5.395940in}{3.517843in}}%
\pgfusepath{clip}%
\pgfsetbuttcap%
\pgfsetmiterjoin%
\definecolor{currentfill}{rgb}{0.000000,0.000000,1.000000}%
\pgfsetfillcolor{currentfill}%
\pgfsetlinewidth{0.000000pt}%
\definecolor{currentstroke}{rgb}{0.000000,0.000000,0.000000}%
\pgfsetstrokecolor{currentstroke}%
\pgfsetdash{}{0pt}%
\pgfpathmoveto{\pgfqpoint{4.313406in}{0.716028in}}%
\pgfpathlineto{\pgfqpoint{4.499473in}{0.716028in}}%
\pgfpathlineto{\pgfqpoint{4.499473in}{2.120164in}}%
\pgfpathlineto{\pgfqpoint{4.313406in}{2.120164in}}%
\pgfpathlineto{\pgfqpoint{4.313406in}{0.716028in}}%
\pgfpathclose%
\pgfusepath{fill}%
\end{pgfscope}%
\begin{pgfscope}%
\pgfpathrectangle{\pgfqpoint{0.592068in}{0.716028in}}{\pgfqpoint{5.395940in}{3.517843in}}%
\pgfusepath{clip}%
\pgfsetbuttcap%
\pgfsetmiterjoin%
\definecolor{currentfill}{rgb}{0.000000,0.000000,1.000000}%
\pgfsetfillcolor{currentfill}%
\pgfsetlinewidth{0.000000pt}%
\definecolor{currentstroke}{rgb}{0.000000,0.000000,0.000000}%
\pgfsetstrokecolor{currentstroke}%
\pgfsetdash{}{0pt}%
\pgfpathmoveto{\pgfqpoint{4.499473in}{0.716028in}}%
\pgfpathlineto{\pgfqpoint{4.685539in}{0.716028in}}%
\pgfpathlineto{\pgfqpoint{4.685539in}{1.375019in}}%
\pgfpathlineto{\pgfqpoint{4.499473in}{1.375019in}}%
\pgfpathlineto{\pgfqpoint{4.499473in}{0.716028in}}%
\pgfpathclose%
\pgfusepath{fill}%
\end{pgfscope}%
\begin{pgfscope}%
\pgfpathrectangle{\pgfqpoint{0.592068in}{0.716028in}}{\pgfqpoint{5.395940in}{3.517843in}}%
\pgfusepath{clip}%
\pgfsetbuttcap%
\pgfsetmiterjoin%
\definecolor{currentfill}{rgb}{0.000000,0.000000,1.000000}%
\pgfsetfillcolor{currentfill}%
\pgfsetlinewidth{0.000000pt}%
\definecolor{currentstroke}{rgb}{0.000000,0.000000,0.000000}%
\pgfsetstrokecolor{currentstroke}%
\pgfsetdash{}{0pt}%
\pgfpathmoveto{\pgfqpoint{4.685539in}{0.716028in}}%
\pgfpathlineto{\pgfqpoint{4.871606in}{0.716028in}}%
\pgfpathlineto{\pgfqpoint{4.871606in}{1.196155in}}%
\pgfpathlineto{\pgfqpoint{4.685539in}{1.196155in}}%
\pgfpathlineto{\pgfqpoint{4.685539in}{0.716028in}}%
\pgfpathclose%
\pgfusepath{fill}%
\end{pgfscope}%
\begin{pgfscope}%
\pgfpathrectangle{\pgfqpoint{0.592068in}{0.716028in}}{\pgfqpoint{5.395940in}{3.517843in}}%
\pgfusepath{clip}%
\pgfsetbuttcap%
\pgfsetmiterjoin%
\definecolor{currentfill}{rgb}{0.000000,0.000000,1.000000}%
\pgfsetfillcolor{currentfill}%
\pgfsetlinewidth{0.000000pt}%
\definecolor{currentstroke}{rgb}{0.000000,0.000000,0.000000}%
\pgfsetstrokecolor{currentstroke}%
\pgfsetdash{}{0pt}%
\pgfpathmoveto{\pgfqpoint{4.871606in}{0.716028in}}%
\pgfpathlineto{\pgfqpoint{5.057673in}{0.716028in}}%
\pgfpathlineto{\pgfqpoint{5.057673in}{1.061552in}}%
\pgfpathlineto{\pgfqpoint{4.871606in}{1.061552in}}%
\pgfpathlineto{\pgfqpoint{4.871606in}{0.716028in}}%
\pgfpathclose%
\pgfusepath{fill}%
\end{pgfscope}%
\begin{pgfscope}%
\pgfpathrectangle{\pgfqpoint{0.592068in}{0.716028in}}{\pgfqpoint{5.395940in}{3.517843in}}%
\pgfusepath{clip}%
\pgfsetbuttcap%
\pgfsetmiterjoin%
\definecolor{currentfill}{rgb}{0.000000,0.000000,1.000000}%
\pgfsetfillcolor{currentfill}%
\pgfsetlinewidth{0.000000pt}%
\definecolor{currentstroke}{rgb}{0.000000,0.000000,0.000000}%
\pgfsetstrokecolor{currentstroke}%
\pgfsetdash{}{0pt}%
\pgfpathmoveto{\pgfqpoint{5.057673in}{0.716028in}}%
\pgfpathlineto{\pgfqpoint{5.243740in}{0.716028in}}%
\pgfpathlineto{\pgfqpoint{5.243740in}{0.953542in}}%
\pgfpathlineto{\pgfqpoint{5.057673in}{0.953542in}}%
\pgfpathlineto{\pgfqpoint{5.057673in}{0.716028in}}%
\pgfpathclose%
\pgfusepath{fill}%
\end{pgfscope}%
\begin{pgfscope}%
\pgfpathrectangle{\pgfqpoint{0.592068in}{0.716028in}}{\pgfqpoint{5.395940in}{3.517843in}}%
\pgfusepath{clip}%
\pgfsetbuttcap%
\pgfsetmiterjoin%
\definecolor{currentfill}{rgb}{0.000000,0.000000,1.000000}%
\pgfsetfillcolor{currentfill}%
\pgfsetlinewidth{0.000000pt}%
\definecolor{currentstroke}{rgb}{0.000000,0.000000,0.000000}%
\pgfsetstrokecolor{currentstroke}%
\pgfsetdash{}{0pt}%
\pgfpathmoveto{\pgfqpoint{5.243740in}{0.716028in}}%
\pgfpathlineto{\pgfqpoint{5.429807in}{0.716028in}}%
\pgfpathlineto{\pgfqpoint{5.429807in}{0.891067in}}%
\pgfpathlineto{\pgfqpoint{5.243740in}{0.891067in}}%
\pgfpathlineto{\pgfqpoint{5.243740in}{0.716028in}}%
\pgfpathclose%
\pgfusepath{fill}%
\end{pgfscope}%
\begin{pgfscope}%
\pgfpathrectangle{\pgfqpoint{0.592068in}{0.716028in}}{\pgfqpoint{5.395940in}{3.517843in}}%
\pgfusepath{clip}%
\pgfsetbuttcap%
\pgfsetmiterjoin%
\definecolor{currentfill}{rgb}{0.000000,0.000000,1.000000}%
\pgfsetfillcolor{currentfill}%
\pgfsetlinewidth{0.000000pt}%
\definecolor{currentstroke}{rgb}{0.000000,0.000000,0.000000}%
\pgfsetstrokecolor{currentstroke}%
\pgfsetdash{}{0pt}%
\pgfpathmoveto{\pgfqpoint{5.429807in}{0.716028in}}%
\pgfpathlineto{\pgfqpoint{5.615874in}{0.716028in}}%
\pgfpathlineto{\pgfqpoint{5.615874in}{0.832053in}}%
\pgfpathlineto{\pgfqpoint{5.429807in}{0.832053in}}%
\pgfpathlineto{\pgfqpoint{5.429807in}{0.716028in}}%
\pgfpathclose%
\pgfusepath{fill}%
\end{pgfscope}%
\begin{pgfscope}%
\pgfpathrectangle{\pgfqpoint{0.592068in}{0.716028in}}{\pgfqpoint{5.395940in}{3.517843in}}%
\pgfusepath{clip}%
\pgfsetbuttcap%
\pgfsetmiterjoin%
\definecolor{currentfill}{rgb}{0.000000,0.000000,1.000000}%
\pgfsetfillcolor{currentfill}%
\pgfsetlinewidth{0.000000pt}%
\definecolor{currentstroke}{rgb}{0.000000,0.000000,0.000000}%
\pgfsetstrokecolor{currentstroke}%
\pgfsetdash{}{0pt}%
\pgfpathmoveto{\pgfqpoint{5.615874in}{0.716028in}}%
\pgfpathlineto{\pgfqpoint{5.801941in}{0.716028in}}%
\pgfpathlineto{\pgfqpoint{5.801941in}{0.803092in}}%
\pgfpathlineto{\pgfqpoint{5.615874in}{0.803092in}}%
\pgfpathlineto{\pgfqpoint{5.615874in}{0.716028in}}%
\pgfpathclose%
\pgfusepath{fill}%
\end{pgfscope}%
\begin{pgfscope}%
\pgfpathrectangle{\pgfqpoint{0.592068in}{0.716028in}}{\pgfqpoint{5.395940in}{3.517843in}}%
\pgfusepath{clip}%
\pgfsetbuttcap%
\pgfsetmiterjoin%
\definecolor{currentfill}{rgb}{0.000000,0.000000,1.000000}%
\pgfsetfillcolor{currentfill}%
\pgfsetlinewidth{0.000000pt}%
\definecolor{currentstroke}{rgb}{0.000000,0.000000,0.000000}%
\pgfsetstrokecolor{currentstroke}%
\pgfsetdash{}{0pt}%
\pgfpathmoveto{\pgfqpoint{5.801941in}{0.716028in}}%
\pgfpathlineto{\pgfqpoint{5.988008in}{0.716028in}}%
\pgfpathlineto{\pgfqpoint{5.988008in}{0.812564in}}%
\pgfpathlineto{\pgfqpoint{5.801941in}{0.812564in}}%
\pgfpathlineto{\pgfqpoint{5.801941in}{0.716028in}}%
\pgfpathclose%
\pgfusepath{fill}%
\end{pgfscope}%
\begin{pgfscope}%
\pgfsetbuttcap%
\pgfsetroundjoin%
\definecolor{currentfill}{rgb}{0.000000,0.000000,0.000000}%
\pgfsetfillcolor{currentfill}%
\pgfsetlinewidth{0.200750pt}%
\definecolor{currentstroke}{rgb}{0.000000,0.000000,0.000000}%
\pgfsetstrokecolor{currentstroke}%
\pgfsetdash{}{0pt}%
\pgfsys@defobject{currentmarker}{\pgfqpoint{0.000000in}{-0.048611in}}{\pgfqpoint{0.000000in}{0.000000in}}{%
\pgfpathmoveto{\pgfqpoint{0.000000in}{0.000000in}}%
\pgfpathlineto{\pgfqpoint{0.000000in}{-0.048611in}}%
\pgfusepath{stroke,fill}%
}%
\begin{pgfscope}%
\pgfsys@transformshift{0.592068in}{0.716028in}%
\pgfsys@useobject{currentmarker}{}%
\end{pgfscope}%
\end{pgfscope}%
\begin{pgfscope}%
\definecolor{textcolor}{rgb}{0.000000,0.000000,0.000000}%
\pgfsetstrokecolor{textcolor}%
\pgfsetfillcolor{textcolor}%
\pgftext[x=0.592068in,y=0.618806in,,top]{\color{textcolor}{\rmfamily\fontsize{20.000000}{24.000000}\selectfont\catcode`\^=\active\def^{\ifmmode\sp\else\^{}\fi}\catcode`\%=\active\def
\end{pgfscope}%
\begin{pgfscope}%
\pgfsetbuttcap%
\pgfsetroundjoin%
\definecolor{currentfill}{rgb}{0.000000,0.000000,0.000000}%
\pgfsetfillcolor{currentfill}%
\pgfsetlinewidth{0.200750pt}%
\definecolor{currentstroke}{rgb}{0.000000,0.000000,0.000000}%
\pgfsetstrokecolor{currentstroke}%
\pgfsetdash{}{0pt}%
\pgfsys@defobject{currentmarker}{\pgfqpoint{0.000000in}{-0.048611in}}{\pgfqpoint{0.000000in}{0.000000in}}{%
\pgfpathmoveto{\pgfqpoint{0.000000in}{0.000000in}}%
\pgfpathlineto{\pgfqpoint{0.000000in}{-0.048611in}}%
\pgfusepath{stroke,fill}%
}%
\begin{pgfscope}%
\pgfsys@transformshift{1.671256in}{0.716028in}%
\pgfsys@useobject{currentmarker}{}%
\end{pgfscope}%
\end{pgfscope}%
\begin{pgfscope}%
\definecolor{textcolor}{rgb}{0.000000,0.000000,0.000000}%
\pgfsetstrokecolor{textcolor}%
\pgfsetfillcolor{textcolor}%
\pgftext[x=1.671256in,y=0.618806in,,top]{\color{textcolor}{\rmfamily\fontsize{20.000000}{24.000000}\selectfont\catcode`\^=\active\def^{\ifmmode\sp\else\^{}\fi}\catcode`\%=\active\def
\end{pgfscope}%
\begin{pgfscope}%
\pgfsetbuttcap%
\pgfsetroundjoin%
\definecolor{currentfill}{rgb}{0.000000,0.000000,0.000000}%
\pgfsetfillcolor{currentfill}%
\pgfsetlinewidth{0.200750pt}%
\definecolor{currentstroke}{rgb}{0.000000,0.000000,0.000000}%
\pgfsetstrokecolor{currentstroke}%
\pgfsetdash{}{0pt}%
\pgfsys@defobject{currentmarker}{\pgfqpoint{0.000000in}{-0.048611in}}{\pgfqpoint{0.000000in}{0.000000in}}{%
\pgfpathmoveto{\pgfqpoint{0.000000in}{0.000000in}}%
\pgfpathlineto{\pgfqpoint{0.000000in}{-0.048611in}}%
\pgfusepath{stroke,fill}%
}%
\begin{pgfscope}%
\pgfsys@transformshift{2.750444in}{0.716028in}%
\pgfsys@useobject{currentmarker}{}%
\end{pgfscope}%
\end{pgfscope}%
\begin{pgfscope}%
\definecolor{textcolor}{rgb}{0.000000,0.000000,0.000000}%
\pgfsetstrokecolor{textcolor}%
\pgfsetfillcolor{textcolor}%
\pgftext[x=2.750444in,y=0.618806in,,top]{\color{textcolor}{\rmfamily\fontsize{20.000000}{24.000000}\selectfont\catcode`\^=\active\def^{\ifmmode\sp\else\^{}\fi}\catcode`\%=\active\def
\end{pgfscope}%
\begin{pgfscope}%
\pgfsetbuttcap%
\pgfsetroundjoin%
\definecolor{currentfill}{rgb}{0.000000,0.000000,0.000000}%
\pgfsetfillcolor{currentfill}%
\pgfsetlinewidth{0.200750pt}%
\definecolor{currentstroke}{rgb}{0.000000,0.000000,0.000000}%
\pgfsetstrokecolor{currentstroke}%
\pgfsetdash{}{0pt}%
\pgfsys@defobject{currentmarker}{\pgfqpoint{0.000000in}{-0.048611in}}{\pgfqpoint{0.000000in}{0.000000in}}{%
\pgfpathmoveto{\pgfqpoint{0.000000in}{0.000000in}}%
\pgfpathlineto{\pgfqpoint{0.000000in}{-0.048611in}}%
\pgfusepath{stroke,fill}%
}%
\begin{pgfscope}%
\pgfsys@transformshift{3.829632in}{0.716028in}%
\pgfsys@useobject{currentmarker}{}%
\end{pgfscope}%
\end{pgfscope}%
\begin{pgfscope}%
\definecolor{textcolor}{rgb}{0.000000,0.000000,0.000000}%
\pgfsetstrokecolor{textcolor}%
\pgfsetfillcolor{textcolor}%
\pgftext[x=3.829632in,y=0.618806in,,top]{\color{textcolor}{\rmfamily\fontsize{20.000000}{24.000000}\selectfont\catcode`\^=\active\def^{\ifmmode\sp\else\^{}\fi}\catcode`\%=\active\def
\end{pgfscope}%
\begin{pgfscope}%
\pgfsetbuttcap%
\pgfsetroundjoin%
\definecolor{currentfill}{rgb}{0.000000,0.000000,0.000000}%
\pgfsetfillcolor{currentfill}%
\pgfsetlinewidth{0.200750pt}%
\definecolor{currentstroke}{rgb}{0.000000,0.000000,0.000000}%
\pgfsetstrokecolor{currentstroke}%
\pgfsetdash{}{0pt}%
\pgfsys@defobject{currentmarker}{\pgfqpoint{0.000000in}{-0.048611in}}{\pgfqpoint{0.000000in}{0.000000in}}{%
\pgfpathmoveto{\pgfqpoint{0.000000in}{0.000000in}}%
\pgfpathlineto{\pgfqpoint{0.000000in}{-0.048611in}}%
\pgfusepath{stroke,fill}%
}%
\begin{pgfscope}%
\pgfsys@transformshift{4.908820in}{0.716028in}%
\pgfsys@useobject{currentmarker}{}%
\end{pgfscope}%
\end{pgfscope}%
\begin{pgfscope}%
\definecolor{textcolor}{rgb}{0.000000,0.000000,0.000000}%
\pgfsetstrokecolor{textcolor}%
\pgfsetfillcolor{textcolor}%
\pgftext[x=4.908820in,y=0.618806in,,top]{\color{textcolor}{\rmfamily\fontsize{20.000000}{24.000000}\selectfont\catcode`\^=\active\def^{\ifmmode\sp\else\^{}\fi}\catcode`\%=\active\def
\end{pgfscope}%
\begin{pgfscope}%
\pgfsetbuttcap%
\pgfsetroundjoin%
\definecolor{currentfill}{rgb}{0.000000,0.000000,0.000000}%
\pgfsetfillcolor{currentfill}%
\pgfsetlinewidth{0.200750pt}%
\definecolor{currentstroke}{rgb}{0.000000,0.000000,0.000000}%
\pgfsetstrokecolor{currentstroke}%
\pgfsetdash{}{0pt}%
\pgfsys@defobject{currentmarker}{\pgfqpoint{0.000000in}{-0.048611in}}{\pgfqpoint{0.000000in}{0.000000in}}{%
\pgfpathmoveto{\pgfqpoint{0.000000in}{0.000000in}}%
\pgfpathlineto{\pgfqpoint{0.000000in}{-0.048611in}}%
\pgfusepath{stroke,fill}%
}%
\begin{pgfscope}%
\pgfsys@transformshift{5.988008in}{0.716028in}%
\pgfsys@useobject{currentmarker}{}%
\end{pgfscope}%
\end{pgfscope}%
\begin{pgfscope}%
\definecolor{textcolor}{rgb}{0.000000,0.000000,0.000000}%
\pgfsetstrokecolor{textcolor}%
\pgfsetfillcolor{textcolor}%
\pgftext[x=5.988008in,y=0.618806in,,top]{\color{textcolor}{\rmfamily\fontsize{20.000000}{24.000000}\selectfont\catcode`\^=\active\def^{\ifmmode\sp\else\^{}\fi}\catcode`\%=\active\def
\end{pgfscope}%
\begin{pgfscope}%
\definecolor{textcolor}{rgb}{0.000000,0.000000,0.000000}%
\pgfsetstrokecolor{textcolor}%
\pgfsetfillcolor{textcolor}%
\pgftext[x=3.290038in,y=0.307183in,,top]{\color{textcolor}{\rmfamily\fontsize{26.000000}{31.200000}\selectfont\catcode`\^=\active\def^{\ifmmode\sp\else\^{}\fi}\catcode`\%=\active\def
\end{pgfscope}%
\begin{pgfscope}%
\pgfsetbuttcap%
\pgfsetroundjoin%
\definecolor{currentfill}{rgb}{0.000000,0.000000,0.000000}%
\pgfsetfillcolor{currentfill}%
\pgfsetlinewidth{0.200750pt}%
\definecolor{currentstroke}{rgb}{0.000000,0.000000,0.000000}%
\pgfsetstrokecolor{currentstroke}%
\pgfsetdash{}{0pt}%
\pgfsys@defobject{currentmarker}{\pgfqpoint{-0.048611in}{0.000000in}}{\pgfqpoint{-0.000000in}{0.000000in}}{%
\pgfpathmoveto{\pgfqpoint{-0.000000in}{0.000000in}}%
\pgfpathlineto{\pgfqpoint{-0.048611in}{0.000000in}}%
\pgfusepath{stroke,fill}%
}%
\begin{pgfscope}%
\pgfsys@transformshift{0.592068in}{0.716028in}%
\pgfsys@useobject{currentmarker}{}%
\end{pgfscope}%
\end{pgfscope}%
\begin{pgfscope}%
\definecolor{textcolor}{rgb}{0.000000,0.000000,0.000000}%
\pgfsetstrokecolor{textcolor}%
\pgfsetfillcolor{textcolor}%
\pgftext[x=0.362738in, y=0.616009in, left, base]{\color{textcolor}{\rmfamily\fontsize{20.000000}{24.000000}\selectfont\catcode`\^=\active\def^{\ifmmode\sp\else\^{}\fi}\catcode`\%=\active\def
\end{pgfscope}%
\begin{pgfscope}%
\pgfsetbuttcap%
\pgfsetroundjoin%
\definecolor{currentfill}{rgb}{0.000000,0.000000,0.000000}%
\pgfsetfillcolor{currentfill}%
\pgfsetlinewidth{0.200750pt}%
\definecolor{currentstroke}{rgb}{0.000000,0.000000,0.000000}%
\pgfsetstrokecolor{currentstroke}%
\pgfsetdash{}{0pt}%
\pgfsys@defobject{currentmarker}{\pgfqpoint{-0.048611in}{0.000000in}}{\pgfqpoint{-0.000000in}{0.000000in}}{%
\pgfpathmoveto{\pgfqpoint{-0.000000in}{0.000000in}}%
\pgfpathlineto{\pgfqpoint{-0.048611in}{0.000000in}}%
\pgfusepath{stroke,fill}%
}%
\begin{pgfscope}%
\pgfsys@transformshift{0.592068in}{1.809291in}%
\pgfsys@useobject{currentmarker}{}%
\end{pgfscope}%
\end{pgfscope}%
\begin{pgfscope}%
\definecolor{textcolor}{rgb}{0.000000,0.000000,0.000000}%
\pgfsetstrokecolor{textcolor}%
\pgfsetfillcolor{textcolor}%
\pgftext[x=0.362738in, y=1.709271in, left, base]{\color{textcolor}{\rmfamily\fontsize{20.000000}{24.000000}\selectfont\catcode`\^=\active\def^{\ifmmode\sp\else\^{}\fi}\catcode`\%=\active\def
\end{pgfscope}%
\begin{pgfscope}%
\pgfsetbuttcap%
\pgfsetroundjoin%
\definecolor{currentfill}{rgb}{0.000000,0.000000,0.000000}%
\pgfsetfillcolor{currentfill}%
\pgfsetlinewidth{0.200750pt}%
\definecolor{currentstroke}{rgb}{0.000000,0.000000,0.000000}%
\pgfsetstrokecolor{currentstroke}%
\pgfsetdash{}{0pt}%
\pgfsys@defobject{currentmarker}{\pgfqpoint{-0.048611in}{0.000000in}}{\pgfqpoint{-0.000000in}{0.000000in}}{%
\pgfpathmoveto{\pgfqpoint{-0.000000in}{0.000000in}}%
\pgfpathlineto{\pgfqpoint{-0.048611in}{0.000000in}}%
\pgfusepath{stroke,fill}%
}%
\begin{pgfscope}%
\pgfsys@transformshift{0.592068in}{2.902553in}%
\pgfsys@useobject{currentmarker}{}%
\end{pgfscope}%
\end{pgfscope}%
\begin{pgfscope}%
\definecolor{textcolor}{rgb}{0.000000,0.000000,0.000000}%
\pgfsetstrokecolor{textcolor}%
\pgfsetfillcolor{textcolor}%
\pgftext[x=0.362738in, y=2.802534in, left, base]{\color{textcolor}{\rmfamily\fontsize{20.000000}{24.000000}\selectfont\catcode`\^=\active\def^{\ifmmode\sp\else\^{}\fi}\catcode`\%=\active\def
\end{pgfscope}%
\begin{pgfscope}%
\pgfsetbuttcap%
\pgfsetroundjoin%
\definecolor{currentfill}{rgb}{0.000000,0.000000,0.000000}%
\pgfsetfillcolor{currentfill}%
\pgfsetlinewidth{0.200750pt}%
\definecolor{currentstroke}{rgb}{0.000000,0.000000,0.000000}%
\pgfsetstrokecolor{currentstroke}%
\pgfsetdash{}{0pt}%
\pgfsys@defobject{currentmarker}{\pgfqpoint{-0.048611in}{0.000000in}}{\pgfqpoint{-0.000000in}{0.000000in}}{%
\pgfpathmoveto{\pgfqpoint{-0.000000in}{0.000000in}}%
\pgfpathlineto{\pgfqpoint{-0.048611in}{0.000000in}}%
\pgfusepath{stroke,fill}%
}%
\begin{pgfscope}%
\pgfsys@transformshift{0.592068in}{3.995815in}%
\pgfsys@useobject{currentmarker}{}%
\end{pgfscope}%
\end{pgfscope}%
\begin{pgfscope}%
\definecolor{textcolor}{rgb}{0.000000,0.000000,0.000000}%
\pgfsetstrokecolor{textcolor}%
\pgfsetfillcolor{textcolor}%
\pgftext[x=0.362738in, y=3.895796in, left, base]{\color{textcolor}{\rmfamily\fontsize{20.000000}{24.000000}\selectfont\catcode`\^=\active\def^{\ifmmode\sp\else\^{}\fi}\catcode`\%=\active\def
\end{pgfscope}%
\begin{pgfscope}%
\definecolor{textcolor}{rgb}{0.000000,0.000000,0.000000}%
\pgfsetstrokecolor{textcolor}%
\pgfsetfillcolor{textcolor}%
\pgftext[x=0.307183in,y=2.474950in,,bottom,rotate=90.000000]{\color{textcolor}{\rmfamily\fontsize{26.000000}{31.200000}\selectfont\catcode`\^=\active\def^{\ifmmode\sp\else\^{}\fi}\catcode`\%=\active\def
\end{pgfscope}%
\begin{pgfscope}%
\pgfpathrectangle{\pgfqpoint{0.592068in}{0.716028in}}{\pgfqpoint{5.395940in}{3.517843in}}%
\pgfusepath{clip}%
\pgfsetbuttcap%
\pgfsetroundjoin%
\pgfsetlinewidth{2.007500pt}%
\definecolor{currentstroke}{rgb}{0.501961,0.501961,0.501961}%
\pgfsetstrokecolor{currentstroke}%
\pgfsetdash{{7.400000pt}{3.200000pt}}{0.000000pt}%
\pgfpathmoveto{\pgfqpoint{0.592068in}{1.809291in}}%
\pgfpathlineto{\pgfqpoint{5.988008in}{1.809291in}}%
\pgfusepath{stroke}%
\end{pgfscope}%
\begin{pgfscope}%
\pgfsetrectcap%
\pgfsetmiterjoin%
\pgfsetlinewidth{0.803000pt}%
\definecolor{currentstroke}{rgb}{0.000000,0.000000,0.000000}%
\pgfsetstrokecolor{currentstroke}%
\pgfsetdash{}{0pt}%
\pgfpathmoveto{\pgfqpoint{0.592068in}{0.716028in}}%
\pgfpathlineto{\pgfqpoint{0.592068in}{4.233871in}}%
\pgfusepath{stroke}%
\end{pgfscope}%
\begin{pgfscope}%
\pgfsetrectcap%
\pgfsetmiterjoin%
\pgfsetlinewidth{0.803000pt}%
\definecolor{currentstroke}{rgb}{0.000000,0.000000,0.000000}%
\pgfsetstrokecolor{currentstroke}%
\pgfsetdash{}{0pt}%
\pgfpathmoveto{\pgfqpoint{5.988008in}{0.716028in}}%
\pgfpathlineto{\pgfqpoint{5.988008in}{4.233871in}}%
\pgfusepath{stroke}%
\end{pgfscope}%
\begin{pgfscope}%
\pgfsetrectcap%
\pgfsetmiterjoin%
\pgfsetlinewidth{0.803000pt}%
\definecolor{currentstroke}{rgb}{0.000000,0.000000,0.000000}%
\pgfsetstrokecolor{currentstroke}%
\pgfsetdash{}{0pt}%
\pgfpathmoveto{\pgfqpoint{0.592068in}{0.716028in}}%
\pgfpathlineto{\pgfqpoint{5.988008in}{0.716028in}}%
\pgfusepath{stroke}%
\end{pgfscope}%
\begin{pgfscope}%
\pgfsetrectcap%
\pgfsetmiterjoin%
\pgfsetlinewidth{0.803000pt}%
\definecolor{currentstroke}{rgb}{0.000000,0.000000,0.000000}%
\pgfsetstrokecolor{currentstroke}%
\pgfsetdash{}{0pt}%
\pgfpathmoveto{\pgfqpoint{0.592068in}{4.233871in}}%
\pgfpathlineto{\pgfqpoint{5.988008in}{4.233871in}}%
\pgfusepath{stroke}%
\end{pgfscope}%
\begin{pgfscope}%
\definecolor{textcolor}{rgb}{0.000000,0.000000,0.000000}%
\pgfsetstrokecolor{textcolor}%
\pgfsetfillcolor{textcolor}%
\pgftext[x=3.290038in,y=4.317204in,,base]{\color{textcolor}{\rmfamily\fontsize{25.000000}{30.000000}\selectfont\catcode`\^=\active\def^{\ifmmode\sp\else\^{}\fi}\catcode`\%=\active\def
\end{pgfscope}%
\begin{pgfscope}%
\pgfsetbuttcap%
\pgfsetmiterjoin%
\definecolor{currentfill}{rgb}{1.000000,1.000000,1.000000}%
\pgfsetfillcolor{currentfill}%
\pgfsetfillopacity{0.800000}%
\pgfsetlinewidth{1.003750pt}%
\definecolor{currentstroke}{rgb}{0.800000,0.800000,0.800000}%
\pgfsetstrokecolor{currentstroke}%
\pgfsetstrokeopacity{0.800000}%
\pgfsetdash{}{0pt}%
\pgfpathmoveto{\pgfqpoint{4.251661in}{3.616692in}}%
\pgfpathlineto{\pgfqpoint{5.793563in}{3.616692in}}%
\pgfpathquadraticcurveto{\pgfqpoint{5.849119in}{3.616692in}}{\pgfqpoint{5.849119in}{3.672248in}}%
\pgfpathlineto{\pgfqpoint{5.849119in}{4.039427in}}%
\pgfpathquadraticcurveto{\pgfqpoint{5.849119in}{4.094982in}}{\pgfqpoint{5.793563in}{4.094982in}}%
\pgfpathlineto{\pgfqpoint{4.251661in}{4.094982in}}%
\pgfpathquadraticcurveto{\pgfqpoint{4.196106in}{4.094982in}}{\pgfqpoint{4.196106in}{4.039427in}}%
\pgfpathlineto{\pgfqpoint{4.196106in}{3.672248in}}%
\pgfpathquadraticcurveto{\pgfqpoint{4.196106in}{3.616692in}}{\pgfqpoint{4.251661in}{3.616692in}}%
\pgfpathlineto{\pgfqpoint{4.251661in}{3.616692in}}%
\pgfpathclose%
\pgfusepath{stroke,fill}%
\end{pgfscope}%
\begin{pgfscope}%
\pgfsetbuttcap%
\pgfsetroundjoin%
\pgfsetlinewidth{2.007500pt}%
\definecolor{currentstroke}{rgb}{0.501961,0.501961,0.501961}%
\pgfsetstrokecolor{currentstroke}%
\pgfsetdash{{7.400000pt}{3.200000pt}}{0.000000pt}%
\pgfpathmoveto{\pgfqpoint{4.307217in}{3.881055in}}%
\pgfpathlineto{\pgfqpoint{4.584995in}{3.881055in}}%
\pgfpathlineto{\pgfqpoint{4.862772in}{3.881055in}}%
\pgfusepath{stroke}%
\end{pgfscope}%
\begin{pgfscope}%
\definecolor{textcolor}{rgb}{0.000000,0.000000,0.000000}%
\pgfsetstrokecolor{textcolor}%
\pgfsetfillcolor{textcolor}%
\pgftext[x=5.084995in,y=3.783833in,left,base]{\color{textcolor}{\rmfamily\fontsize{20.000000}{24.000000}\selectfont\catcode`\^=\active\def^{\ifmmode\sp\else\^{}\fi}\catcode`\%=\active\def
\end{pgfscope}%
\end{pgfpicture}%
\makeatother%
\endgroup%

%% file: figures/evaluation/PCR/PCR_PIT_event_elapsed_norm_4layer.pgf
\begingroup%
\makeatletter%
\begin{pgfpicture}%
\pgfpathrectangle{\pgfpointorigin}{\pgfqpoint{6.159289in}{4.557174in}}%
\pgfusepath{use as bounding box, clip}%
\begin{pgfscope}%
\pgfsetbuttcap%
\pgfsetmiterjoin%
\definecolor{currentfill}{rgb}{1.000000,1.000000,1.000000}%
\pgfsetfillcolor{currentfill}%
\pgfsetlinewidth{0.000000pt}%
\definecolor{currentstroke}{rgb}{1.000000,1.000000,1.000000}%
\pgfsetstrokecolor{currentstroke}%
\pgfsetdash{}{0pt}%
\pgfpathmoveto{\pgfqpoint{0.000000in}{0.000000in}}%
\pgfpathlineto{\pgfqpoint{6.159289in}{0.000000in}}%
\pgfpathlineto{\pgfqpoint{6.159289in}{4.557174in}}%
\pgfpathlineto{\pgfqpoint{0.000000in}{4.557174in}}%
\pgfpathlineto{\pgfqpoint{0.000000in}{0.000000in}}%
\pgfpathclose%
\pgfusepath{fill}%
\end{pgfscope}%
\begin{pgfscope}%
\pgfsetbuttcap%
\pgfsetmiterjoin%
\definecolor{currentfill}{rgb}{1.000000,1.000000,1.000000}%
\pgfsetfillcolor{currentfill}%
\pgfsetlinewidth{0.000000pt}%
\definecolor{currentstroke}{rgb}{0.000000,0.000000,0.000000}%
\pgfsetstrokecolor{currentstroke}%
\pgfsetstrokeopacity{0.000000}%
\pgfsetdash{}{0pt}%
\pgfpathmoveto{\pgfqpoint{0.592068in}{0.716028in}}%
\pgfpathlineto{\pgfqpoint{5.988008in}{0.716028in}}%
\pgfpathlineto{\pgfqpoint{5.988008in}{4.233871in}}%
\pgfpathlineto{\pgfqpoint{0.592068in}{4.233871in}}%
\pgfpathlineto{\pgfqpoint{0.592068in}{0.716028in}}%
\pgfpathclose%
\pgfusepath{fill}%
\end{pgfscope}%
\begin{pgfscope}%
\pgfpathrectangle{\pgfqpoint{0.592068in}{0.716028in}}{\pgfqpoint{5.395940in}{3.517843in}}%
\pgfusepath{clip}%
\pgfsetbuttcap%
\pgfsetmiterjoin%
\definecolor{currentfill}{rgb}{0.000000,0.000000,1.000000}%
\pgfsetfillcolor{currentfill}%
\pgfsetlinewidth{0.000000pt}%
\definecolor{currentstroke}{rgb}{0.000000,0.000000,0.000000}%
\pgfsetstrokecolor{currentstroke}%
\pgfsetdash{}{0pt}%
\pgfpathmoveto{\pgfqpoint{0.592068in}{0.716028in}}%
\pgfpathlineto{\pgfqpoint{0.778135in}{0.716028in}}%
\pgfpathlineto{\pgfqpoint{0.778135in}{3.744710in}}%
\pgfpathlineto{\pgfqpoint{0.592068in}{3.744710in}}%
\pgfpathlineto{\pgfqpoint{0.592068in}{0.716028in}}%
\pgfpathclose%
\pgfusepath{fill}%
\end{pgfscope}%
\begin{pgfscope}%
\pgfpathrectangle{\pgfqpoint{0.592068in}{0.716028in}}{\pgfqpoint{5.395940in}{3.517843in}}%
\pgfusepath{clip}%
\pgfsetbuttcap%
\pgfsetmiterjoin%
\definecolor{currentfill}{rgb}{0.000000,0.000000,1.000000}%
\pgfsetfillcolor{currentfill}%
\pgfsetlinewidth{0.000000pt}%
\definecolor{currentstroke}{rgb}{0.000000,0.000000,0.000000}%
\pgfsetstrokecolor{currentstroke}%
\pgfsetdash{}{0pt}%
\pgfpathmoveto{\pgfqpoint{0.778135in}{0.716028in}}%
\pgfpathlineto{\pgfqpoint{0.964202in}{0.716028in}}%
\pgfpathlineto{\pgfqpoint{0.964202in}{4.066355in}}%
\pgfpathlineto{\pgfqpoint{0.778135in}{4.066355in}}%
\pgfpathlineto{\pgfqpoint{0.778135in}{0.716028in}}%
\pgfpathclose%
\pgfusepath{fill}%
\end{pgfscope}%
\begin{pgfscope}%
\pgfpathrectangle{\pgfqpoint{0.592068in}{0.716028in}}{\pgfqpoint{5.395940in}{3.517843in}}%
\pgfusepath{clip}%
\pgfsetbuttcap%
\pgfsetmiterjoin%
\definecolor{currentfill}{rgb}{0.000000,0.000000,1.000000}%
\pgfsetfillcolor{currentfill}%
\pgfsetlinewidth{0.000000pt}%
\definecolor{currentstroke}{rgb}{0.000000,0.000000,0.000000}%
\pgfsetstrokecolor{currentstroke}%
\pgfsetdash{}{0pt}%
\pgfpathmoveto{\pgfqpoint{0.964202in}{0.716028in}}%
\pgfpathlineto{\pgfqpoint{1.150269in}{0.716028in}}%
\pgfpathlineto{\pgfqpoint{1.150269in}{1.640144in}}%
\pgfpathlineto{\pgfqpoint{0.964202in}{1.640144in}}%
\pgfpathlineto{\pgfqpoint{0.964202in}{0.716028in}}%
\pgfpathclose%
\pgfusepath{fill}%
\end{pgfscope}%
\begin{pgfscope}%
\pgfpathrectangle{\pgfqpoint{0.592068in}{0.716028in}}{\pgfqpoint{5.395940in}{3.517843in}}%
\pgfusepath{clip}%
\pgfsetbuttcap%
\pgfsetmiterjoin%
\definecolor{currentfill}{rgb}{0.000000,0.000000,1.000000}%
\pgfsetfillcolor{currentfill}%
\pgfsetlinewidth{0.000000pt}%
\definecolor{currentstroke}{rgb}{0.000000,0.000000,0.000000}%
\pgfsetstrokecolor{currentstroke}%
\pgfsetdash{}{0pt}%
\pgfpathmoveto{\pgfqpoint{1.150269in}{0.716028in}}%
\pgfpathlineto{\pgfqpoint{1.336335in}{0.716028in}}%
\pgfpathlineto{\pgfqpoint{1.336335in}{1.276049in}}%
\pgfpathlineto{\pgfqpoint{1.150269in}{1.276049in}}%
\pgfpathlineto{\pgfqpoint{1.150269in}{0.716028in}}%
\pgfpathclose%
\pgfusepath{fill}%
\end{pgfscope}%
\begin{pgfscope}%
\pgfpathrectangle{\pgfqpoint{0.592068in}{0.716028in}}{\pgfqpoint{5.395940in}{3.517843in}}%
\pgfusepath{clip}%
\pgfsetbuttcap%
\pgfsetmiterjoin%
\definecolor{currentfill}{rgb}{0.000000,0.000000,1.000000}%
\pgfsetfillcolor{currentfill}%
\pgfsetlinewidth{0.000000pt}%
\definecolor{currentstroke}{rgb}{0.000000,0.000000,0.000000}%
\pgfsetstrokecolor{currentstroke}%
\pgfsetdash{}{0pt}%
\pgfpathmoveto{\pgfqpoint{1.336335in}{0.716028in}}%
\pgfpathlineto{\pgfqpoint{1.522402in}{0.716028in}}%
\pgfpathlineto{\pgfqpoint{1.522402in}{1.364215in}}%
\pgfpathlineto{\pgfqpoint{1.336335in}{1.364215in}}%
\pgfpathlineto{\pgfqpoint{1.336335in}{0.716028in}}%
\pgfpathclose%
\pgfusepath{fill}%
\end{pgfscope}%
\begin{pgfscope}%
\pgfpathrectangle{\pgfqpoint{0.592068in}{0.716028in}}{\pgfqpoint{5.395940in}{3.517843in}}%
\pgfusepath{clip}%
\pgfsetbuttcap%
\pgfsetmiterjoin%
\definecolor{currentfill}{rgb}{0.000000,0.000000,1.000000}%
\pgfsetfillcolor{currentfill}%
\pgfsetlinewidth{0.000000pt}%
\definecolor{currentstroke}{rgb}{0.000000,0.000000,0.000000}%
\pgfsetstrokecolor{currentstroke}%
\pgfsetdash{}{0pt}%
\pgfpathmoveto{\pgfqpoint{1.522402in}{0.716028in}}%
\pgfpathlineto{\pgfqpoint{1.708469in}{0.716028in}}%
\pgfpathlineto{\pgfqpoint{1.708469in}{1.207475in}}%
\pgfpathlineto{\pgfqpoint{1.522402in}{1.207475in}}%
\pgfpathlineto{\pgfqpoint{1.522402in}{0.716028in}}%
\pgfpathclose%
\pgfusepath{fill}%
\end{pgfscope}%
\begin{pgfscope}%
\pgfpathrectangle{\pgfqpoint{0.592068in}{0.716028in}}{\pgfqpoint{5.395940in}{3.517843in}}%
\pgfusepath{clip}%
\pgfsetbuttcap%
\pgfsetmiterjoin%
\definecolor{currentfill}{rgb}{0.000000,0.000000,1.000000}%
\pgfsetfillcolor{currentfill}%
\pgfsetlinewidth{0.000000pt}%
\definecolor{currentstroke}{rgb}{0.000000,0.000000,0.000000}%
\pgfsetstrokecolor{currentstroke}%
\pgfsetdash{}{0pt}%
\pgfpathmoveto{\pgfqpoint{1.708469in}{0.716028in}}%
\pgfpathlineto{\pgfqpoint{1.894536in}{0.716028in}}%
\pgfpathlineto{\pgfqpoint{1.894536in}{1.276049in}}%
\pgfpathlineto{\pgfqpoint{1.708469in}{1.276049in}}%
\pgfpathlineto{\pgfqpoint{1.708469in}{0.716028in}}%
\pgfpathclose%
\pgfusepath{fill}%
\end{pgfscope}%
\begin{pgfscope}%
\pgfpathrectangle{\pgfqpoint{0.592068in}{0.716028in}}{\pgfqpoint{5.395940in}{3.517843in}}%
\pgfusepath{clip}%
\pgfsetbuttcap%
\pgfsetmiterjoin%
\definecolor{currentfill}{rgb}{0.000000,0.000000,1.000000}%
\pgfsetfillcolor{currentfill}%
\pgfsetlinewidth{0.000000pt}%
\definecolor{currentstroke}{rgb}{0.000000,0.000000,0.000000}%
\pgfsetstrokecolor{currentstroke}%
\pgfsetdash{}{0pt}%
\pgfpathmoveto{\pgfqpoint{1.894536in}{0.716028in}}%
\pgfpathlineto{\pgfqpoint{2.080603in}{0.716028in}}%
\pgfpathlineto{\pgfqpoint{2.080603in}{1.488301in}}%
\pgfpathlineto{\pgfqpoint{1.894536in}{1.488301in}}%
\pgfpathlineto{\pgfqpoint{1.894536in}{0.716028in}}%
\pgfpathclose%
\pgfusepath{fill}%
\end{pgfscope}%
\begin{pgfscope}%
\pgfpathrectangle{\pgfqpoint{0.592068in}{0.716028in}}{\pgfqpoint{5.395940in}{3.517843in}}%
\pgfusepath{clip}%
\pgfsetbuttcap%
\pgfsetmiterjoin%
\definecolor{currentfill}{rgb}{0.000000,0.000000,1.000000}%
\pgfsetfillcolor{currentfill}%
\pgfsetlinewidth{0.000000pt}%
\definecolor{currentstroke}{rgb}{0.000000,0.000000,0.000000}%
\pgfsetstrokecolor{currentstroke}%
\pgfsetdash{}{0pt}%
\pgfpathmoveto{\pgfqpoint{2.080603in}{0.716028in}}%
\pgfpathlineto{\pgfqpoint{2.266670in}{0.716028in}}%
\pgfpathlineto{\pgfqpoint{2.266670in}{2.670385in}}%
\pgfpathlineto{\pgfqpoint{2.080603in}{2.670385in}}%
\pgfpathlineto{\pgfqpoint{2.080603in}{0.716028in}}%
\pgfpathclose%
\pgfusepath{fill}%
\end{pgfscope}%
\begin{pgfscope}%
\pgfpathrectangle{\pgfqpoint{0.592068in}{0.716028in}}{\pgfqpoint{5.395940in}{3.517843in}}%
\pgfusepath{clip}%
\pgfsetbuttcap%
\pgfsetmiterjoin%
\definecolor{currentfill}{rgb}{0.000000,0.000000,1.000000}%
\pgfsetfillcolor{currentfill}%
\pgfsetlinewidth{0.000000pt}%
\definecolor{currentstroke}{rgb}{0.000000,0.000000,0.000000}%
\pgfsetstrokecolor{currentstroke}%
\pgfsetdash{}{0pt}%
\pgfpathmoveto{\pgfqpoint{2.266670in}{0.716028in}}%
\pgfpathlineto{\pgfqpoint{2.452737in}{0.716028in}}%
\pgfpathlineto{\pgfqpoint{2.452737in}{3.119381in}}%
\pgfpathlineto{\pgfqpoint{2.266670in}{3.119381in}}%
\pgfpathlineto{\pgfqpoint{2.266670in}{0.716028in}}%
\pgfpathclose%
\pgfusepath{fill}%
\end{pgfscope}%
\begin{pgfscope}%
\pgfpathrectangle{\pgfqpoint{0.592068in}{0.716028in}}{\pgfqpoint{5.395940in}{3.517843in}}%
\pgfusepath{clip}%
\pgfsetbuttcap%
\pgfsetmiterjoin%
\definecolor{currentfill}{rgb}{0.000000,0.000000,1.000000}%
\pgfsetfillcolor{currentfill}%
\pgfsetlinewidth{0.000000pt}%
\definecolor{currentstroke}{rgb}{0.000000,0.000000,0.000000}%
\pgfsetstrokecolor{currentstroke}%
\pgfsetdash{}{0pt}%
\pgfpathmoveto{\pgfqpoint{2.452737in}{0.716028in}}%
\pgfpathlineto{\pgfqpoint{2.638804in}{0.716028in}}%
\pgfpathlineto{\pgfqpoint{2.638804in}{2.933252in}}%
\pgfpathlineto{\pgfqpoint{2.452737in}{2.933252in}}%
\pgfpathlineto{\pgfqpoint{2.452737in}{0.716028in}}%
\pgfpathclose%
\pgfusepath{fill}%
\end{pgfscope}%
\begin{pgfscope}%
\pgfpathrectangle{\pgfqpoint{0.592068in}{0.716028in}}{\pgfqpoint{5.395940in}{3.517843in}}%
\pgfusepath{clip}%
\pgfsetbuttcap%
\pgfsetmiterjoin%
\definecolor{currentfill}{rgb}{0.000000,0.000000,1.000000}%
\pgfsetfillcolor{currentfill}%
\pgfsetlinewidth{0.000000pt}%
\definecolor{currentstroke}{rgb}{0.000000,0.000000,0.000000}%
\pgfsetstrokecolor{currentstroke}%
\pgfsetdash{}{0pt}%
\pgfpathmoveto{\pgfqpoint{2.638804in}{0.716028in}}%
\pgfpathlineto{\pgfqpoint{2.824871in}{0.716028in}}%
\pgfpathlineto{\pgfqpoint{2.824871in}{3.295714in}}%
\pgfpathlineto{\pgfqpoint{2.638804in}{3.295714in}}%
\pgfpathlineto{\pgfqpoint{2.638804in}{0.716028in}}%
\pgfpathclose%
\pgfusepath{fill}%
\end{pgfscope}%
\begin{pgfscope}%
\pgfpathrectangle{\pgfqpoint{0.592068in}{0.716028in}}{\pgfqpoint{5.395940in}{3.517843in}}%
\pgfusepath{clip}%
\pgfsetbuttcap%
\pgfsetmiterjoin%
\definecolor{currentfill}{rgb}{0.000000,0.000000,1.000000}%
\pgfsetfillcolor{currentfill}%
\pgfsetlinewidth{0.000000pt}%
\definecolor{currentstroke}{rgb}{0.000000,0.000000,0.000000}%
\pgfsetstrokecolor{currentstroke}%
\pgfsetdash{}{0pt}%
\pgfpathmoveto{\pgfqpoint{2.824871in}{0.716028in}}%
\pgfpathlineto{\pgfqpoint{3.010937in}{0.716028in}}%
\pgfpathlineto{\pgfqpoint{3.010937in}{3.021419in}}%
\pgfpathlineto{\pgfqpoint{2.824871in}{3.021419in}}%
\pgfpathlineto{\pgfqpoint{2.824871in}{0.716028in}}%
\pgfpathclose%
\pgfusepath{fill}%
\end{pgfscope}%
\begin{pgfscope}%
\pgfpathrectangle{\pgfqpoint{0.592068in}{0.716028in}}{\pgfqpoint{5.395940in}{3.517843in}}%
\pgfusepath{clip}%
\pgfsetbuttcap%
\pgfsetmiterjoin%
\definecolor{currentfill}{rgb}{0.000000,0.000000,1.000000}%
\pgfsetfillcolor{currentfill}%
\pgfsetlinewidth{0.000000pt}%
\definecolor{currentstroke}{rgb}{0.000000,0.000000,0.000000}%
\pgfsetstrokecolor{currentstroke}%
\pgfsetdash{}{0pt}%
\pgfpathmoveto{\pgfqpoint{3.010937in}{0.716028in}}%
\pgfpathlineto{\pgfqpoint{3.197004in}{0.716028in}}%
\pgfpathlineto{\pgfqpoint{3.197004in}{2.054853in}}%
\pgfpathlineto{\pgfqpoint{3.010937in}{2.054853in}}%
\pgfpathlineto{\pgfqpoint{3.010937in}{0.716028in}}%
\pgfpathclose%
\pgfusepath{fill}%
\end{pgfscope}%
\begin{pgfscope}%
\pgfpathrectangle{\pgfqpoint{0.592068in}{0.716028in}}{\pgfqpoint{5.395940in}{3.517843in}}%
\pgfusepath{clip}%
\pgfsetbuttcap%
\pgfsetmiterjoin%
\definecolor{currentfill}{rgb}{0.000000,0.000000,1.000000}%
\pgfsetfillcolor{currentfill}%
\pgfsetlinewidth{0.000000pt}%
\definecolor{currentstroke}{rgb}{0.000000,0.000000,0.000000}%
\pgfsetstrokecolor{currentstroke}%
\pgfsetdash{}{0pt}%
\pgfpathmoveto{\pgfqpoint{3.197004in}{0.716028in}}%
\pgfpathlineto{\pgfqpoint{3.383071in}{0.716028in}}%
\pgfpathlineto{\pgfqpoint{3.383071in}{1.594428in}}%
\pgfpathlineto{\pgfqpoint{3.197004in}{1.594428in}}%
\pgfpathlineto{\pgfqpoint{3.197004in}{0.716028in}}%
\pgfpathclose%
\pgfusepath{fill}%
\end{pgfscope}%
\begin{pgfscope}%
\pgfpathrectangle{\pgfqpoint{0.592068in}{0.716028in}}{\pgfqpoint{5.395940in}{3.517843in}}%
\pgfusepath{clip}%
\pgfsetbuttcap%
\pgfsetmiterjoin%
\definecolor{currentfill}{rgb}{0.000000,0.000000,1.000000}%
\pgfsetfillcolor{currentfill}%
\pgfsetlinewidth{0.000000pt}%
\definecolor{currentstroke}{rgb}{0.000000,0.000000,0.000000}%
\pgfsetstrokecolor{currentstroke}%
\pgfsetdash{}{0pt}%
\pgfpathmoveto{\pgfqpoint{3.383071in}{0.716028in}}%
\pgfpathlineto{\pgfqpoint{3.569138in}{0.716028in}}%
\pgfpathlineto{\pgfqpoint{3.569138in}{1.153595in}}%
\pgfpathlineto{\pgfqpoint{3.383071in}{1.153595in}}%
\pgfpathlineto{\pgfqpoint{3.383071in}{0.716028in}}%
\pgfpathclose%
\pgfusepath{fill}%
\end{pgfscope}%
\begin{pgfscope}%
\pgfpathrectangle{\pgfqpoint{0.592068in}{0.716028in}}{\pgfqpoint{5.395940in}{3.517843in}}%
\pgfusepath{clip}%
\pgfsetbuttcap%
\pgfsetmiterjoin%
\definecolor{currentfill}{rgb}{0.000000,0.000000,1.000000}%
\pgfsetfillcolor{currentfill}%
\pgfsetlinewidth{0.000000pt}%
\definecolor{currentstroke}{rgb}{0.000000,0.000000,0.000000}%
\pgfsetstrokecolor{currentstroke}%
\pgfsetdash{}{0pt}%
\pgfpathmoveto{\pgfqpoint{3.569138in}{0.716028in}}%
\pgfpathlineto{\pgfqpoint{3.755205in}{0.716028in}}%
\pgfpathlineto{\pgfqpoint{3.755205in}{0.933179in}}%
\pgfpathlineto{\pgfqpoint{3.569138in}{0.933179in}}%
\pgfpathlineto{\pgfqpoint{3.569138in}{0.716028in}}%
\pgfpathclose%
\pgfusepath{fill}%
\end{pgfscope}%
\begin{pgfscope}%
\pgfpathrectangle{\pgfqpoint{0.592068in}{0.716028in}}{\pgfqpoint{5.395940in}{3.517843in}}%
\pgfusepath{clip}%
\pgfsetbuttcap%
\pgfsetmiterjoin%
\definecolor{currentfill}{rgb}{0.000000,0.000000,1.000000}%
\pgfsetfillcolor{currentfill}%
\pgfsetlinewidth{0.000000pt}%
\definecolor{currentstroke}{rgb}{0.000000,0.000000,0.000000}%
\pgfsetstrokecolor{currentstroke}%
\pgfsetdash{}{0pt}%
\pgfpathmoveto{\pgfqpoint{3.755205in}{0.716028in}}%
\pgfpathlineto{\pgfqpoint{3.941272in}{0.716028in}}%
\pgfpathlineto{\pgfqpoint{3.941272in}{0.928281in}}%
\pgfpathlineto{\pgfqpoint{3.755205in}{0.928281in}}%
\pgfpathlineto{\pgfqpoint{3.755205in}{0.716028in}}%
\pgfpathclose%
\pgfusepath{fill}%
\end{pgfscope}%
\begin{pgfscope}%
\pgfpathrectangle{\pgfqpoint{0.592068in}{0.716028in}}{\pgfqpoint{5.395940in}{3.517843in}}%
\pgfusepath{clip}%
\pgfsetbuttcap%
\pgfsetmiterjoin%
\definecolor{currentfill}{rgb}{0.000000,0.000000,1.000000}%
\pgfsetfillcolor{currentfill}%
\pgfsetlinewidth{0.000000pt}%
\definecolor{currentstroke}{rgb}{0.000000,0.000000,0.000000}%
\pgfsetstrokecolor{currentstroke}%
\pgfsetdash{}{0pt}%
\pgfpathmoveto{\pgfqpoint{3.941272in}{0.716028in}}%
\pgfpathlineto{\pgfqpoint{4.127339in}{0.716028in}}%
\pgfpathlineto{\pgfqpoint{4.127339in}{0.874401in}}%
\pgfpathlineto{\pgfqpoint{3.941272in}{0.874401in}}%
\pgfpathlineto{\pgfqpoint{3.941272in}{0.716028in}}%
\pgfpathclose%
\pgfusepath{fill}%
\end{pgfscope}%
\begin{pgfscope}%
\pgfpathrectangle{\pgfqpoint{0.592068in}{0.716028in}}{\pgfqpoint{5.395940in}{3.517843in}}%
\pgfusepath{clip}%
\pgfsetbuttcap%
\pgfsetmiterjoin%
\definecolor{currentfill}{rgb}{0.000000,0.000000,1.000000}%
\pgfsetfillcolor{currentfill}%
\pgfsetlinewidth{0.000000pt}%
\definecolor{currentstroke}{rgb}{0.000000,0.000000,0.000000}%
\pgfsetstrokecolor{currentstroke}%
\pgfsetdash{}{0pt}%
\pgfpathmoveto{\pgfqpoint{4.127339in}{0.716028in}}%
\pgfpathlineto{\pgfqpoint{4.313406in}{0.716028in}}%
\pgfpathlineto{\pgfqpoint{4.313406in}{0.830318in}}%
\pgfpathlineto{\pgfqpoint{4.127339in}{0.830318in}}%
\pgfpathlineto{\pgfqpoint{4.127339in}{0.716028in}}%
\pgfpathclose%
\pgfusepath{fill}%
\end{pgfscope}%
\begin{pgfscope}%
\pgfpathrectangle{\pgfqpoint{0.592068in}{0.716028in}}{\pgfqpoint{5.395940in}{3.517843in}}%
\pgfusepath{clip}%
\pgfsetbuttcap%
\pgfsetmiterjoin%
\definecolor{currentfill}{rgb}{0.000000,0.000000,1.000000}%
\pgfsetfillcolor{currentfill}%
\pgfsetlinewidth{0.000000pt}%
\definecolor{currentstroke}{rgb}{0.000000,0.000000,0.000000}%
\pgfsetstrokecolor{currentstroke}%
\pgfsetdash{}{0pt}%
\pgfpathmoveto{\pgfqpoint{4.313406in}{0.716028in}}%
\pgfpathlineto{\pgfqpoint{4.499473in}{0.716028in}}%
\pgfpathlineto{\pgfqpoint{4.499473in}{0.833584in}}%
\pgfpathlineto{\pgfqpoint{4.313406in}{0.833584in}}%
\pgfpathlineto{\pgfqpoint{4.313406in}{0.716028in}}%
\pgfpathclose%
\pgfusepath{fill}%
\end{pgfscope}%
\begin{pgfscope}%
\pgfpathrectangle{\pgfqpoint{0.592068in}{0.716028in}}{\pgfqpoint{5.395940in}{3.517843in}}%
\pgfusepath{clip}%
\pgfsetbuttcap%
\pgfsetmiterjoin%
\definecolor{currentfill}{rgb}{0.000000,0.000000,1.000000}%
\pgfsetfillcolor{currentfill}%
\pgfsetlinewidth{0.000000pt}%
\definecolor{currentstroke}{rgb}{0.000000,0.000000,0.000000}%
\pgfsetstrokecolor{currentstroke}%
\pgfsetdash{}{0pt}%
\pgfpathmoveto{\pgfqpoint{4.499473in}{0.716028in}}%
\pgfpathlineto{\pgfqpoint{4.685539in}{0.716028in}}%
\pgfpathlineto{\pgfqpoint{4.685539in}{0.815624in}}%
\pgfpathlineto{\pgfqpoint{4.499473in}{0.815624in}}%
\pgfpathlineto{\pgfqpoint{4.499473in}{0.716028in}}%
\pgfpathclose%
\pgfusepath{fill}%
\end{pgfscope}%
\begin{pgfscope}%
\pgfpathrectangle{\pgfqpoint{0.592068in}{0.716028in}}{\pgfqpoint{5.395940in}{3.517843in}}%
\pgfusepath{clip}%
\pgfsetbuttcap%
\pgfsetmiterjoin%
\definecolor{currentfill}{rgb}{0.000000,0.000000,1.000000}%
\pgfsetfillcolor{currentfill}%
\pgfsetlinewidth{0.000000pt}%
\definecolor{currentstroke}{rgb}{0.000000,0.000000,0.000000}%
\pgfsetstrokecolor{currentstroke}%
\pgfsetdash{}{0pt}%
\pgfpathmoveto{\pgfqpoint{4.685539in}{0.716028in}}%
\pgfpathlineto{\pgfqpoint{4.871606in}{0.716028in}}%
\pgfpathlineto{\pgfqpoint{4.871606in}{0.947873in}}%
\pgfpathlineto{\pgfqpoint{4.685539in}{0.947873in}}%
\pgfpathlineto{\pgfqpoint{4.685539in}{0.716028in}}%
\pgfpathclose%
\pgfusepath{fill}%
\end{pgfscope}%
\begin{pgfscope}%
\pgfpathrectangle{\pgfqpoint{0.592068in}{0.716028in}}{\pgfqpoint{5.395940in}{3.517843in}}%
\pgfusepath{clip}%
\pgfsetbuttcap%
\pgfsetmiterjoin%
\definecolor{currentfill}{rgb}{0.000000,0.000000,1.000000}%
\pgfsetfillcolor{currentfill}%
\pgfsetlinewidth{0.000000pt}%
\definecolor{currentstroke}{rgb}{0.000000,0.000000,0.000000}%
\pgfsetstrokecolor{currentstroke}%
\pgfsetdash{}{0pt}%
\pgfpathmoveto{\pgfqpoint{4.871606in}{0.716028in}}%
\pgfpathlineto{\pgfqpoint{5.057673in}{0.716028in}}%
\pgfpathlineto{\pgfqpoint{5.057673in}{1.047469in}}%
\pgfpathlineto{\pgfqpoint{4.871606in}{1.047469in}}%
\pgfpathlineto{\pgfqpoint{4.871606in}{0.716028in}}%
\pgfpathclose%
\pgfusepath{fill}%
\end{pgfscope}%
\begin{pgfscope}%
\pgfpathrectangle{\pgfqpoint{0.592068in}{0.716028in}}{\pgfqpoint{5.395940in}{3.517843in}}%
\pgfusepath{clip}%
\pgfsetbuttcap%
\pgfsetmiterjoin%
\definecolor{currentfill}{rgb}{0.000000,0.000000,1.000000}%
\pgfsetfillcolor{currentfill}%
\pgfsetlinewidth{0.000000pt}%
\definecolor{currentstroke}{rgb}{0.000000,0.000000,0.000000}%
\pgfsetstrokecolor{currentstroke}%
\pgfsetdash{}{0pt}%
\pgfpathmoveto{\pgfqpoint{5.057673in}{0.716028in}}%
\pgfpathlineto{\pgfqpoint{5.243740in}{0.716028in}}%
\pgfpathlineto{\pgfqpoint{5.243740in}{1.165024in}}%
\pgfpathlineto{\pgfqpoint{5.057673in}{1.165024in}}%
\pgfpathlineto{\pgfqpoint{5.057673in}{0.716028in}}%
\pgfpathclose%
\pgfusepath{fill}%
\end{pgfscope}%
\begin{pgfscope}%
\pgfpathrectangle{\pgfqpoint{0.592068in}{0.716028in}}{\pgfqpoint{5.395940in}{3.517843in}}%
\pgfusepath{clip}%
\pgfsetbuttcap%
\pgfsetmiterjoin%
\definecolor{currentfill}{rgb}{0.000000,0.000000,1.000000}%
\pgfsetfillcolor{currentfill}%
\pgfsetlinewidth{0.000000pt}%
\definecolor{currentstroke}{rgb}{0.000000,0.000000,0.000000}%
\pgfsetstrokecolor{currentstroke}%
\pgfsetdash{}{0pt}%
\pgfpathmoveto{\pgfqpoint{5.243740in}{0.716028in}}%
\pgfpathlineto{\pgfqpoint{5.429807in}{0.716028in}}%
\pgfpathlineto{\pgfqpoint{5.429807in}{1.194413in}}%
\pgfpathlineto{\pgfqpoint{5.243740in}{1.194413in}}%
\pgfpathlineto{\pgfqpoint{5.243740in}{0.716028in}}%
\pgfpathclose%
\pgfusepath{fill}%
\end{pgfscope}%
\begin{pgfscope}%
\pgfpathrectangle{\pgfqpoint{0.592068in}{0.716028in}}{\pgfqpoint{5.395940in}{3.517843in}}%
\pgfusepath{clip}%
\pgfsetbuttcap%
\pgfsetmiterjoin%
\definecolor{currentfill}{rgb}{0.000000,0.000000,1.000000}%
\pgfsetfillcolor{currentfill}%
\pgfsetlinewidth{0.000000pt}%
\definecolor{currentstroke}{rgb}{0.000000,0.000000,0.000000}%
\pgfsetstrokecolor{currentstroke}%
\pgfsetdash{}{0pt}%
\pgfpathmoveto{\pgfqpoint{5.429807in}{0.716028in}}%
\pgfpathlineto{\pgfqpoint{5.615874in}{0.716028in}}%
\pgfpathlineto{\pgfqpoint{5.615874in}{1.163391in}}%
\pgfpathlineto{\pgfqpoint{5.429807in}{1.163391in}}%
\pgfpathlineto{\pgfqpoint{5.429807in}{0.716028in}}%
\pgfpathclose%
\pgfusepath{fill}%
\end{pgfscope}%
\begin{pgfscope}%
\pgfpathrectangle{\pgfqpoint{0.592068in}{0.716028in}}{\pgfqpoint{5.395940in}{3.517843in}}%
\pgfusepath{clip}%
\pgfsetbuttcap%
\pgfsetmiterjoin%
\definecolor{currentfill}{rgb}{0.000000,0.000000,1.000000}%
\pgfsetfillcolor{currentfill}%
\pgfsetlinewidth{0.000000pt}%
\definecolor{currentstroke}{rgb}{0.000000,0.000000,0.000000}%
\pgfsetstrokecolor{currentstroke}%
\pgfsetdash{}{0pt}%
\pgfpathmoveto{\pgfqpoint{5.615874in}{0.716028in}}%
\pgfpathlineto{\pgfqpoint{5.801941in}{0.716028in}}%
\pgfpathlineto{\pgfqpoint{5.801941in}{1.019713in}}%
\pgfpathlineto{\pgfqpoint{5.615874in}{1.019713in}}%
\pgfpathlineto{\pgfqpoint{5.615874in}{0.716028in}}%
\pgfpathclose%
\pgfusepath{fill}%
\end{pgfscope}%
\begin{pgfscope}%
\pgfpathrectangle{\pgfqpoint{0.592068in}{0.716028in}}{\pgfqpoint{5.395940in}{3.517843in}}%
\pgfusepath{clip}%
\pgfsetbuttcap%
\pgfsetmiterjoin%
\definecolor{currentfill}{rgb}{0.000000,0.000000,1.000000}%
\pgfsetfillcolor{currentfill}%
\pgfsetlinewidth{0.000000pt}%
\definecolor{currentstroke}{rgb}{0.000000,0.000000,0.000000}%
\pgfsetstrokecolor{currentstroke}%
\pgfsetdash{}{0pt}%
\pgfpathmoveto{\pgfqpoint{5.801941in}{0.716028in}}%
\pgfpathlineto{\pgfqpoint{5.988008in}{0.716028in}}%
\pgfpathlineto{\pgfqpoint{5.988008in}{2.913660in}}%
\pgfpathlineto{\pgfqpoint{5.801941in}{2.913660in}}%
\pgfpathlineto{\pgfqpoint{5.801941in}{0.716028in}}%
\pgfpathclose%
\pgfusepath{fill}%
\end{pgfscope}%
\begin{pgfscope}%
\pgfsetbuttcap%
\pgfsetroundjoin%
\definecolor{currentfill}{rgb}{0.000000,0.000000,0.000000}%
\pgfsetfillcolor{currentfill}%
\pgfsetlinewidth{0.200750pt}%
\definecolor{currentstroke}{rgb}{0.000000,0.000000,0.000000}%
\pgfsetstrokecolor{currentstroke}%
\pgfsetdash{}{0pt}%
\pgfsys@defobject{currentmarker}{\pgfqpoint{0.000000in}{-0.048611in}}{\pgfqpoint{0.000000in}{0.000000in}}{%
\pgfpathmoveto{\pgfqpoint{0.000000in}{0.000000in}}%
\pgfpathlineto{\pgfqpoint{0.000000in}{-0.048611in}}%
\pgfusepath{stroke,fill}%
}%
\begin{pgfscope}%
\pgfsys@transformshift{0.592068in}{0.716028in}%
\pgfsys@useobject{currentmarker}{}%
\end{pgfscope}%
\end{pgfscope}%
\begin{pgfscope}%
\definecolor{textcolor}{rgb}{0.000000,0.000000,0.000000}%
\pgfsetstrokecolor{textcolor}%
\pgfsetfillcolor{textcolor}%
\pgftext[x=0.592068in,y=0.618806in,,top]{\color{textcolor}{\rmfamily\fontsize{20.000000}{24.000000}\selectfont\catcode`\^=\active\def^{\ifmmode\sp\else\^{}\fi}\catcode`\%=\active\def
\end{pgfscope}%
\begin{pgfscope}%
\pgfsetbuttcap%
\pgfsetroundjoin%
\definecolor{currentfill}{rgb}{0.000000,0.000000,0.000000}%
\pgfsetfillcolor{currentfill}%
\pgfsetlinewidth{0.200750pt}%
\definecolor{currentstroke}{rgb}{0.000000,0.000000,0.000000}%
\pgfsetstrokecolor{currentstroke}%
\pgfsetdash{}{0pt}%
\pgfsys@defobject{currentmarker}{\pgfqpoint{0.000000in}{-0.048611in}}{\pgfqpoint{0.000000in}{0.000000in}}{%
\pgfpathmoveto{\pgfqpoint{0.000000in}{0.000000in}}%
\pgfpathlineto{\pgfqpoint{0.000000in}{-0.048611in}}%
\pgfusepath{stroke,fill}%
}%
\begin{pgfscope}%
\pgfsys@transformshift{1.671256in}{0.716028in}%
\pgfsys@useobject{currentmarker}{}%
\end{pgfscope}%
\end{pgfscope}%
\begin{pgfscope}%
\definecolor{textcolor}{rgb}{0.000000,0.000000,0.000000}%
\pgfsetstrokecolor{textcolor}%
\pgfsetfillcolor{textcolor}%
\pgftext[x=1.671256in,y=0.618806in,,top]{\color{textcolor}{\rmfamily\fontsize{20.000000}{24.000000}\selectfont\catcode`\^=\active\def^{\ifmmode\sp\else\^{}\fi}\catcode`\%=\active\def
\end{pgfscope}%
\begin{pgfscope}%
\pgfsetbuttcap%
\pgfsetroundjoin%
\definecolor{currentfill}{rgb}{0.000000,0.000000,0.000000}%
\pgfsetfillcolor{currentfill}%
\pgfsetlinewidth{0.200750pt}%
\definecolor{currentstroke}{rgb}{0.000000,0.000000,0.000000}%
\pgfsetstrokecolor{currentstroke}%
\pgfsetdash{}{0pt}%
\pgfsys@defobject{currentmarker}{\pgfqpoint{0.000000in}{-0.048611in}}{\pgfqpoint{0.000000in}{0.000000in}}{%
\pgfpathmoveto{\pgfqpoint{0.000000in}{0.000000in}}%
\pgfpathlineto{\pgfqpoint{0.000000in}{-0.048611in}}%
\pgfusepath{stroke,fill}%
}%
\begin{pgfscope}%
\pgfsys@transformshift{2.750444in}{0.716028in}%
\pgfsys@useobject{currentmarker}{}%
\end{pgfscope}%
\end{pgfscope}%
\begin{pgfscope}%
\definecolor{textcolor}{rgb}{0.000000,0.000000,0.000000}%
\pgfsetstrokecolor{textcolor}%
\pgfsetfillcolor{textcolor}%
\pgftext[x=2.750444in,y=0.618806in,,top]{\color{textcolor}{\rmfamily\fontsize{20.000000}{24.000000}\selectfont\catcode`\^=\active\def^{\ifmmode\sp\else\^{}\fi}\catcode`\%=\active\def
\end{pgfscope}%
\begin{pgfscope}%
\pgfsetbuttcap%
\pgfsetroundjoin%
\definecolor{currentfill}{rgb}{0.000000,0.000000,0.000000}%
\pgfsetfillcolor{currentfill}%
\pgfsetlinewidth{0.200750pt}%
\definecolor{currentstroke}{rgb}{0.000000,0.000000,0.000000}%
\pgfsetstrokecolor{currentstroke}%
\pgfsetdash{}{0pt}%
\pgfsys@defobject{currentmarker}{\pgfqpoint{0.000000in}{-0.048611in}}{\pgfqpoint{0.000000in}{0.000000in}}{%
\pgfpathmoveto{\pgfqpoint{0.000000in}{0.000000in}}%
\pgfpathlineto{\pgfqpoint{0.000000in}{-0.048611in}}%
\pgfusepath{stroke,fill}%
}%
\begin{pgfscope}%
\pgfsys@transformshift{3.829632in}{0.716028in}%
\pgfsys@useobject{currentmarker}{}%
\end{pgfscope}%
\end{pgfscope}%
\begin{pgfscope}%
\definecolor{textcolor}{rgb}{0.000000,0.000000,0.000000}%
\pgfsetstrokecolor{textcolor}%
\pgfsetfillcolor{textcolor}%
\pgftext[x=3.829632in,y=0.618806in,,top]{\color{textcolor}{\rmfamily\fontsize{20.000000}{24.000000}\selectfont\catcode`\^=\active\def^{\ifmmode\sp\else\^{}\fi}\catcode`\%=\active\def
\end{pgfscope}%
\begin{pgfscope}%
\pgfsetbuttcap%
\pgfsetroundjoin%
\definecolor{currentfill}{rgb}{0.000000,0.000000,0.000000}%
\pgfsetfillcolor{currentfill}%
\pgfsetlinewidth{0.200750pt}%
\definecolor{currentstroke}{rgb}{0.000000,0.000000,0.000000}%
\pgfsetstrokecolor{currentstroke}%
\pgfsetdash{}{0pt}%
\pgfsys@defobject{currentmarker}{\pgfqpoint{0.000000in}{-0.048611in}}{\pgfqpoint{0.000000in}{0.000000in}}{%
\pgfpathmoveto{\pgfqpoint{0.000000in}{0.000000in}}%
\pgfpathlineto{\pgfqpoint{0.000000in}{-0.048611in}}%
\pgfusepath{stroke,fill}%
}%
\begin{pgfscope}%
\pgfsys@transformshift{4.908820in}{0.716028in}%
\pgfsys@useobject{currentmarker}{}%
\end{pgfscope}%
\end{pgfscope}%
\begin{pgfscope}%
\definecolor{textcolor}{rgb}{0.000000,0.000000,0.000000}%
\pgfsetstrokecolor{textcolor}%
\pgfsetfillcolor{textcolor}%
\pgftext[x=4.908820in,y=0.618806in,,top]{\color{textcolor}{\rmfamily\fontsize{20.000000}{24.000000}\selectfont\catcode`\^=\active\def^{\ifmmode\sp\else\^{}\fi}\catcode`\%=\active\def
\end{pgfscope}%
\begin{pgfscope}%
\pgfsetbuttcap%
\pgfsetroundjoin%
\definecolor{currentfill}{rgb}{0.000000,0.000000,0.000000}%
\pgfsetfillcolor{currentfill}%
\pgfsetlinewidth{0.200750pt}%
\definecolor{currentstroke}{rgb}{0.000000,0.000000,0.000000}%
\pgfsetstrokecolor{currentstroke}%
\pgfsetdash{}{0pt}%
\pgfsys@defobject{currentmarker}{\pgfqpoint{0.000000in}{-0.048611in}}{\pgfqpoint{0.000000in}{0.000000in}}{%
\pgfpathmoveto{\pgfqpoint{0.000000in}{0.000000in}}%
\pgfpathlineto{\pgfqpoint{0.000000in}{-0.048611in}}%
\pgfusepath{stroke,fill}%
}%
\begin{pgfscope}%
\pgfsys@transformshift{5.988008in}{0.716028in}%
\pgfsys@useobject{currentmarker}{}%
\end{pgfscope}%
\end{pgfscope}%
\begin{pgfscope}%
\definecolor{textcolor}{rgb}{0.000000,0.000000,0.000000}%
\pgfsetstrokecolor{textcolor}%
\pgfsetfillcolor{textcolor}%
\pgftext[x=5.988008in,y=0.618806in,,top]{\color{textcolor}{\rmfamily\fontsize{20.000000}{24.000000}\selectfont\catcode`\^=\active\def^{\ifmmode\sp\else\^{}\fi}\catcode`\%=\active\def
\end{pgfscope}%
\begin{pgfscope}%
\definecolor{textcolor}{rgb}{0.000000,0.000000,0.000000}%
\pgfsetstrokecolor{textcolor}%
\pgfsetfillcolor{textcolor}%
\pgftext[x=3.290038in,y=0.307183in,,top]{\color{textcolor}{\rmfamily\fontsize{26.000000}{31.200000}\selectfont\catcode`\^=\active\def^{\ifmmode\sp\else\^{}\fi}\catcode`\%=\active\def
\end{pgfscope}%
\begin{pgfscope}%
\pgfsetbuttcap%
\pgfsetroundjoin%
\definecolor{currentfill}{rgb}{0.000000,0.000000,0.000000}%
\pgfsetfillcolor{currentfill}%
\pgfsetlinewidth{0.200750pt}%
\definecolor{currentstroke}{rgb}{0.000000,0.000000,0.000000}%
\pgfsetstrokecolor{currentstroke}%
\pgfsetdash{}{0pt}%
\pgfsys@defobject{currentmarker}{\pgfqpoint{-0.048611in}{0.000000in}}{\pgfqpoint{-0.000000in}{0.000000in}}{%
\pgfpathmoveto{\pgfqpoint{-0.000000in}{0.000000in}}%
\pgfpathlineto{\pgfqpoint{-0.048611in}{0.000000in}}%
\pgfusepath{stroke,fill}%
}%
\begin{pgfscope}%
\pgfsys@transformshift{0.592068in}{0.716028in}%
\pgfsys@useobject{currentmarker}{}%
\end{pgfscope}%
\end{pgfscope}%
\begin{pgfscope}%
\definecolor{textcolor}{rgb}{0.000000,0.000000,0.000000}%
\pgfsetstrokecolor{textcolor}%
\pgfsetfillcolor{textcolor}%
\pgftext[x=0.362738in, y=0.616009in, left, base]{\color{textcolor}{\rmfamily\fontsize{20.000000}{24.000000}\selectfont\catcode`\^=\active\def^{\ifmmode\sp\else\^{}\fi}\catcode`\%=\active\def
\end{pgfscope}%
\begin{pgfscope}%
\pgfsetbuttcap%
\pgfsetroundjoin%
\definecolor{currentfill}{rgb}{0.000000,0.000000,0.000000}%
\pgfsetfillcolor{currentfill}%
\pgfsetlinewidth{0.200750pt}%
\definecolor{currentstroke}{rgb}{0.000000,0.000000,0.000000}%
\pgfsetstrokecolor{currentstroke}%
\pgfsetdash{}{0pt}%
\pgfsys@defobject{currentmarker}{\pgfqpoint{-0.048611in}{0.000000in}}{\pgfqpoint{-0.000000in}{0.000000in}}{%
\pgfpathmoveto{\pgfqpoint{-0.000000in}{0.000000in}}%
\pgfpathlineto{\pgfqpoint{-0.048611in}{0.000000in}}%
\pgfusepath{stroke,fill}%
}%
\begin{pgfscope}%
\pgfsys@transformshift{0.592068in}{1.743905in}%
\pgfsys@useobject{currentmarker}{}%
\end{pgfscope}%
\end{pgfscope}%
\begin{pgfscope}%
\definecolor{textcolor}{rgb}{0.000000,0.000000,0.000000}%
\pgfsetstrokecolor{textcolor}%
\pgfsetfillcolor{textcolor}%
\pgftext[x=0.362738in, y=1.643886in, left, base]{\color{textcolor}{\rmfamily\fontsize{20.000000}{24.000000}\selectfont\catcode`\^=\active\def^{\ifmmode\sp\else\^{}\fi}\catcode`\%=\active\def
\end{pgfscope}%
\begin{pgfscope}%
\pgfsetbuttcap%
\pgfsetroundjoin%
\definecolor{currentfill}{rgb}{0.000000,0.000000,0.000000}%
\pgfsetfillcolor{currentfill}%
\pgfsetlinewidth{0.200750pt}%
\definecolor{currentstroke}{rgb}{0.000000,0.000000,0.000000}%
\pgfsetstrokecolor{currentstroke}%
\pgfsetdash{}{0pt}%
\pgfsys@defobject{currentmarker}{\pgfqpoint{-0.048611in}{0.000000in}}{\pgfqpoint{-0.000000in}{0.000000in}}{%
\pgfpathmoveto{\pgfqpoint{-0.000000in}{0.000000in}}%
\pgfpathlineto{\pgfqpoint{-0.048611in}{0.000000in}}%
\pgfusepath{stroke,fill}%
}%
\begin{pgfscope}%
\pgfsys@transformshift{0.592068in}{2.771782in}%
\pgfsys@useobject{currentmarker}{}%
\end{pgfscope}%
\end{pgfscope}%
\begin{pgfscope}%
\definecolor{textcolor}{rgb}{0.000000,0.000000,0.000000}%
\pgfsetstrokecolor{textcolor}%
\pgfsetfillcolor{textcolor}%
\pgftext[x=0.362738in, y=2.671763in, left, base]{\color{textcolor}{\rmfamily\fontsize{20.000000}{24.000000}\selectfont\catcode`\^=\active\def^{\ifmmode\sp\else\^{}\fi}\catcode`\%=\active\def
\end{pgfscope}%
\begin{pgfscope}%
\pgfsetbuttcap%
\pgfsetroundjoin%
\definecolor{currentfill}{rgb}{0.000000,0.000000,0.000000}%
\pgfsetfillcolor{currentfill}%
\pgfsetlinewidth{0.200750pt}%
\definecolor{currentstroke}{rgb}{0.000000,0.000000,0.000000}%
\pgfsetstrokecolor{currentstroke}%
\pgfsetdash{}{0pt}%
\pgfsys@defobject{currentmarker}{\pgfqpoint{-0.048611in}{0.000000in}}{\pgfqpoint{-0.000000in}{0.000000in}}{%
\pgfpathmoveto{\pgfqpoint{-0.000000in}{0.000000in}}%
\pgfpathlineto{\pgfqpoint{-0.048611in}{0.000000in}}%
\pgfusepath{stroke,fill}%
}%
\begin{pgfscope}%
\pgfsys@transformshift{0.592068in}{3.799660in}%
\pgfsys@useobject{currentmarker}{}%
\end{pgfscope}%
\end{pgfscope}%
\begin{pgfscope}%
\definecolor{textcolor}{rgb}{0.000000,0.000000,0.000000}%
\pgfsetstrokecolor{textcolor}%
\pgfsetfillcolor{textcolor}%
\pgftext[x=0.362738in, y=3.699640in, left, base]{\color{textcolor}{\rmfamily\fontsize{20.000000}{24.000000}\selectfont\catcode`\^=\active\def^{\ifmmode\sp\else\^{}\fi}\catcode`\%=\active\def
\end{pgfscope}%
\begin{pgfscope}%
\definecolor{textcolor}{rgb}{0.000000,0.000000,0.000000}%
\pgfsetstrokecolor{textcolor}%
\pgfsetfillcolor{textcolor}%
\pgftext[x=0.307183in,y=2.474950in,,bottom,rotate=90.000000]{\color{textcolor}{\rmfamily\fontsize{26.000000}{31.200000}\selectfont\catcode`\^=\active\def^{\ifmmode\sp\else\^{}\fi}\catcode`\%=\active\def
\end{pgfscope}%
\begin{pgfscope}%
\pgfpathrectangle{\pgfqpoint{0.592068in}{0.716028in}}{\pgfqpoint{5.395940in}{3.517843in}}%
\pgfusepath{clip}%
\pgfsetbuttcap%
\pgfsetroundjoin%
\pgfsetlinewidth{2.007500pt}%
\definecolor{currentstroke}{rgb}{0.501961,0.501961,0.501961}%
\pgfsetstrokecolor{currentstroke}%
\pgfsetdash{{7.400000pt}{3.200000pt}}{0.000000pt}%
\pgfpathmoveto{\pgfqpoint{0.592068in}{1.743905in}}%
\pgfpathlineto{\pgfqpoint{5.988008in}{1.743905in}}%
\pgfusepath{stroke}%
\end{pgfscope}%
\begin{pgfscope}%
\pgfsetrectcap%
\pgfsetmiterjoin%
\pgfsetlinewidth{0.803000pt}%
\definecolor{currentstroke}{rgb}{0.000000,0.000000,0.000000}%
\pgfsetstrokecolor{currentstroke}%
\pgfsetdash{}{0pt}%
\pgfpathmoveto{\pgfqpoint{0.592068in}{0.716028in}}%
\pgfpathlineto{\pgfqpoint{0.592068in}{4.233871in}}%
\pgfusepath{stroke}%
\end{pgfscope}%
\begin{pgfscope}%
\pgfsetrectcap%
\pgfsetmiterjoin%
\pgfsetlinewidth{0.803000pt}%
\definecolor{currentstroke}{rgb}{0.000000,0.000000,0.000000}%
\pgfsetstrokecolor{currentstroke}%
\pgfsetdash{}{0pt}%
\pgfpathmoveto{\pgfqpoint{5.988008in}{0.716028in}}%
\pgfpathlineto{\pgfqpoint{5.988008in}{4.233871in}}%
\pgfusepath{stroke}%
\end{pgfscope}%
\begin{pgfscope}%
\pgfsetrectcap%
\pgfsetmiterjoin%
\pgfsetlinewidth{0.803000pt}%
\definecolor{currentstroke}{rgb}{0.000000,0.000000,0.000000}%
\pgfsetstrokecolor{currentstroke}%
\pgfsetdash{}{0pt}%
\pgfpathmoveto{\pgfqpoint{0.592068in}{0.716028in}}%
\pgfpathlineto{\pgfqpoint{5.988008in}{0.716028in}}%
\pgfusepath{stroke}%
\end{pgfscope}%
\begin{pgfscope}%
\pgfsetrectcap%
\pgfsetmiterjoin%
\pgfsetlinewidth{0.803000pt}%
\definecolor{currentstroke}{rgb}{0.000000,0.000000,0.000000}%
\pgfsetstrokecolor{currentstroke}%
\pgfsetdash{}{0pt}%
\pgfpathmoveto{\pgfqpoint{0.592068in}{4.233871in}}%
\pgfpathlineto{\pgfqpoint{5.988008in}{4.233871in}}%
\pgfusepath{stroke}%
\end{pgfscope}%
\begin{pgfscope}%
\definecolor{textcolor}{rgb}{0.000000,0.000000,0.000000}%
\pgfsetstrokecolor{textcolor}%
\pgfsetfillcolor{textcolor}%
\pgftext[x=3.290038in,y=4.317204in,,base]{\color{textcolor}{\rmfamily\fontsize{25.000000}{30.000000}\selectfont\catcode`\^=\active\def^{\ifmmode\sp\else\^{}\fi}\catcode`\%=\active\def
\end{pgfscope}%
\begin{pgfscope}%
\pgfsetbuttcap%
\pgfsetmiterjoin%
\definecolor{currentfill}{rgb}{1.000000,1.000000,1.000000}%
\pgfsetfillcolor{currentfill}%
\pgfsetfillopacity{0.800000}%
\pgfsetlinewidth{1.003750pt}%
\definecolor{currentstroke}{rgb}{0.800000,0.800000,0.800000}%
\pgfsetstrokecolor{currentstroke}%
\pgfsetstrokeopacity{0.800000}%
\pgfsetdash{}{0pt}%
\pgfpathmoveto{\pgfqpoint{4.251661in}{3.616692in}}%
\pgfpathlineto{\pgfqpoint{5.793563in}{3.616692in}}%
\pgfpathquadraticcurveto{\pgfqpoint{5.849119in}{3.616692in}}{\pgfqpoint{5.849119in}{3.672248in}}%
\pgfpathlineto{\pgfqpoint{5.849119in}{4.039427in}}%
\pgfpathquadraticcurveto{\pgfqpoint{5.849119in}{4.094982in}}{\pgfqpoint{5.793563in}{4.094982in}}%
\pgfpathlineto{\pgfqpoint{4.251661in}{4.094982in}}%
\pgfpathquadraticcurveto{\pgfqpoint{4.196106in}{4.094982in}}{\pgfqpoint{4.196106in}{4.039427in}}%
\pgfpathlineto{\pgfqpoint{4.196106in}{3.672248in}}%
\pgfpathquadraticcurveto{\pgfqpoint{4.196106in}{3.616692in}}{\pgfqpoint{4.251661in}{3.616692in}}%
\pgfpathlineto{\pgfqpoint{4.251661in}{3.616692in}}%
\pgfpathclose%
\pgfusepath{stroke,fill}%
\end{pgfscope}%
\begin{pgfscope}%
\pgfsetbuttcap%
\pgfsetroundjoin%
\pgfsetlinewidth{2.007500pt}%
\definecolor{currentstroke}{rgb}{0.501961,0.501961,0.501961}%
\pgfsetstrokecolor{currentstroke}%
\pgfsetdash{{7.400000pt}{3.200000pt}}{0.000000pt}%
\pgfpathmoveto{\pgfqpoint{4.307217in}{3.881055in}}%
\pgfpathlineto{\pgfqpoint{4.584995in}{3.881055in}}%
\pgfpathlineto{\pgfqpoint{4.862772in}{3.881055in}}%
\pgfusepath{stroke}%
\end{pgfscope}%
\begin{pgfscope}%
\definecolor{textcolor}{rgb}{0.000000,0.000000,0.000000}%
\pgfsetstrokecolor{textcolor}%
\pgfsetfillcolor{textcolor}%
\pgftext[x=5.084995in,y=3.783833in,left,base]{\color{textcolor}{\rmfamily\fontsize{20.000000}{24.000000}\selectfont\catcode`\^=\active\def^{\ifmmode\sp\else\^{}\fi}\catcode`\%=\active\def
\end{pgfscope}%
\end{pgfpicture}%
\makeatother%
\endgroup%

%% file: figures/evaluation/PCR/PCR_PIT_remaining_time_norm_4layer.pgf
\begingroup%
\makeatletter%
\begin{pgfpicture}%
\pgfpathrectangle{\pgfpointorigin}{\pgfqpoint{6.159289in}{4.557174in}}%
\pgfusepath{use as bounding box, clip}%
\begin{pgfscope}%
\pgfsetbuttcap%
\pgfsetmiterjoin%
\definecolor{currentfill}{rgb}{1.000000,1.000000,1.000000}%
\pgfsetfillcolor{currentfill}%
\pgfsetlinewidth{0.000000pt}%
\definecolor{currentstroke}{rgb}{1.000000,1.000000,1.000000}%
\pgfsetstrokecolor{currentstroke}%
\pgfsetdash{}{0pt}%
\pgfpathmoveto{\pgfqpoint{0.000000in}{0.000000in}}%
\pgfpathlineto{\pgfqpoint{6.159289in}{0.000000in}}%
\pgfpathlineto{\pgfqpoint{6.159289in}{4.557174in}}%
\pgfpathlineto{\pgfqpoint{0.000000in}{4.557174in}}%
\pgfpathlineto{\pgfqpoint{0.000000in}{0.000000in}}%
\pgfpathclose%
\pgfusepath{fill}%
\end{pgfscope}%
\begin{pgfscope}%
\pgfsetbuttcap%
\pgfsetmiterjoin%
\definecolor{currentfill}{rgb}{1.000000,1.000000,1.000000}%
\pgfsetfillcolor{currentfill}%
\pgfsetlinewidth{0.000000pt}%
\definecolor{currentstroke}{rgb}{0.000000,0.000000,0.000000}%
\pgfsetstrokecolor{currentstroke}%
\pgfsetstrokeopacity{0.000000}%
\pgfsetdash{}{0pt}%
\pgfpathmoveto{\pgfqpoint{0.592068in}{0.716028in}}%
\pgfpathlineto{\pgfqpoint{5.988008in}{0.716028in}}%
\pgfpathlineto{\pgfqpoint{5.988008in}{4.233871in}}%
\pgfpathlineto{\pgfqpoint{0.592068in}{4.233871in}}%
\pgfpathlineto{\pgfqpoint{0.592068in}{0.716028in}}%
\pgfpathclose%
\pgfusepath{fill}%
\end{pgfscope}%
\begin{pgfscope}%
\pgfpathrectangle{\pgfqpoint{0.592068in}{0.716028in}}{\pgfqpoint{5.395940in}{3.517843in}}%
\pgfusepath{clip}%
\pgfsetbuttcap%
\pgfsetmiterjoin%
\definecolor{currentfill}{rgb}{0.000000,0.000000,1.000000}%
\pgfsetfillcolor{currentfill}%
\pgfsetlinewidth{0.000000pt}%
\definecolor{currentstroke}{rgb}{0.000000,0.000000,0.000000}%
\pgfsetstrokecolor{currentstroke}%
\pgfsetdash{}{0pt}%
\pgfpathmoveto{\pgfqpoint{0.592068in}{0.716028in}}%
\pgfpathlineto{\pgfqpoint{0.778135in}{0.716028in}}%
\pgfpathlineto{\pgfqpoint{0.778135in}{3.948268in}}%
\pgfpathlineto{\pgfqpoint{0.592068in}{3.948268in}}%
\pgfpathlineto{\pgfqpoint{0.592068in}{0.716028in}}%
\pgfpathclose%
\pgfusepath{fill}%
\end{pgfscope}%
\begin{pgfscope}%
\pgfpathrectangle{\pgfqpoint{0.592068in}{0.716028in}}{\pgfqpoint{5.395940in}{3.517843in}}%
\pgfusepath{clip}%
\pgfsetbuttcap%
\pgfsetmiterjoin%
\definecolor{currentfill}{rgb}{0.000000,0.000000,1.000000}%
\pgfsetfillcolor{currentfill}%
\pgfsetlinewidth{0.000000pt}%
\definecolor{currentstroke}{rgb}{0.000000,0.000000,0.000000}%
\pgfsetstrokecolor{currentstroke}%
\pgfsetdash{}{0pt}%
\pgfpathmoveto{\pgfqpoint{0.778135in}{0.716028in}}%
\pgfpathlineto{\pgfqpoint{0.964202in}{0.716028in}}%
\pgfpathlineto{\pgfqpoint{0.964202in}{4.066355in}}%
\pgfpathlineto{\pgfqpoint{0.778135in}{4.066355in}}%
\pgfpathlineto{\pgfqpoint{0.778135in}{0.716028in}}%
\pgfpathclose%
\pgfusepath{fill}%
\end{pgfscope}%
\begin{pgfscope}%
\pgfpathrectangle{\pgfqpoint{0.592068in}{0.716028in}}{\pgfqpoint{5.395940in}{3.517843in}}%
\pgfusepath{clip}%
\pgfsetbuttcap%
\pgfsetmiterjoin%
\definecolor{currentfill}{rgb}{0.000000,0.000000,1.000000}%
\pgfsetfillcolor{currentfill}%
\pgfsetlinewidth{0.000000pt}%
\definecolor{currentstroke}{rgb}{0.000000,0.000000,0.000000}%
\pgfsetstrokecolor{currentstroke}%
\pgfsetdash{}{0pt}%
\pgfpathmoveto{\pgfqpoint{0.964202in}{0.716028in}}%
\pgfpathlineto{\pgfqpoint{1.150269in}{0.716028in}}%
\pgfpathlineto{\pgfqpoint{1.150269in}{3.430504in}}%
\pgfpathlineto{\pgfqpoint{0.964202in}{3.430504in}}%
\pgfpathlineto{\pgfqpoint{0.964202in}{0.716028in}}%
\pgfpathclose%
\pgfusepath{fill}%
\end{pgfscope}%
\begin{pgfscope}%
\pgfpathrectangle{\pgfqpoint{0.592068in}{0.716028in}}{\pgfqpoint{5.395940in}{3.517843in}}%
\pgfusepath{clip}%
\pgfsetbuttcap%
\pgfsetmiterjoin%
\definecolor{currentfill}{rgb}{0.000000,0.000000,1.000000}%
\pgfsetfillcolor{currentfill}%
\pgfsetlinewidth{0.000000pt}%
\definecolor{currentstroke}{rgb}{0.000000,0.000000,0.000000}%
\pgfsetstrokecolor{currentstroke}%
\pgfsetdash{}{0pt}%
\pgfpathmoveto{\pgfqpoint{1.150269in}{0.716028in}}%
\pgfpathlineto{\pgfqpoint{1.336335in}{0.716028in}}%
\pgfpathlineto{\pgfqpoint{1.336335in}{3.189789in}}%
\pgfpathlineto{\pgfqpoint{1.150269in}{3.189789in}}%
\pgfpathlineto{\pgfqpoint{1.150269in}{0.716028in}}%
\pgfpathclose%
\pgfusepath{fill}%
\end{pgfscope}%
\begin{pgfscope}%
\pgfpathrectangle{\pgfqpoint{0.592068in}{0.716028in}}{\pgfqpoint{5.395940in}{3.517843in}}%
\pgfusepath{clip}%
\pgfsetbuttcap%
\pgfsetmiterjoin%
\definecolor{currentfill}{rgb}{0.000000,0.000000,1.000000}%
\pgfsetfillcolor{currentfill}%
\pgfsetlinewidth{0.000000pt}%
\definecolor{currentstroke}{rgb}{0.000000,0.000000,0.000000}%
\pgfsetstrokecolor{currentstroke}%
\pgfsetdash{}{0pt}%
\pgfpathmoveto{\pgfqpoint{1.336335in}{0.716028in}}%
\pgfpathlineto{\pgfqpoint{1.522402in}{0.716028in}}%
\pgfpathlineto{\pgfqpoint{1.522402in}{3.164053in}}%
\pgfpathlineto{\pgfqpoint{1.336335in}{3.164053in}}%
\pgfpathlineto{\pgfqpoint{1.336335in}{0.716028in}}%
\pgfpathclose%
\pgfusepath{fill}%
\end{pgfscope}%
\begin{pgfscope}%
\pgfpathrectangle{\pgfqpoint{0.592068in}{0.716028in}}{\pgfqpoint{5.395940in}{3.517843in}}%
\pgfusepath{clip}%
\pgfsetbuttcap%
\pgfsetmiterjoin%
\definecolor{currentfill}{rgb}{0.000000,0.000000,1.000000}%
\pgfsetfillcolor{currentfill}%
\pgfsetlinewidth{0.000000pt}%
\definecolor{currentstroke}{rgb}{0.000000,0.000000,0.000000}%
\pgfsetstrokecolor{currentstroke}%
\pgfsetdash{}{0pt}%
\pgfpathmoveto{\pgfqpoint{1.522402in}{0.716028in}}%
\pgfpathlineto{\pgfqpoint{1.708469in}{0.716028in}}%
\pgfpathlineto{\pgfqpoint{1.708469in}{2.681109in}}%
\pgfpathlineto{\pgfqpoint{1.522402in}{2.681109in}}%
\pgfpathlineto{\pgfqpoint{1.522402in}{0.716028in}}%
\pgfpathclose%
\pgfusepath{fill}%
\end{pgfscope}%
\begin{pgfscope}%
\pgfpathrectangle{\pgfqpoint{0.592068in}{0.716028in}}{\pgfqpoint{5.395940in}{3.517843in}}%
\pgfusepath{clip}%
\pgfsetbuttcap%
\pgfsetmiterjoin%
\definecolor{currentfill}{rgb}{0.000000,0.000000,1.000000}%
\pgfsetfillcolor{currentfill}%
\pgfsetlinewidth{0.000000pt}%
\definecolor{currentstroke}{rgb}{0.000000,0.000000,0.000000}%
\pgfsetstrokecolor{currentstroke}%
\pgfsetdash{}{0pt}%
\pgfpathmoveto{\pgfqpoint{1.708469in}{0.716028in}}%
\pgfpathlineto{\pgfqpoint{1.894536in}{0.716028in}}%
\pgfpathlineto{\pgfqpoint{1.894536in}{2.173943in}}%
\pgfpathlineto{\pgfqpoint{1.708469in}{2.173943in}}%
\pgfpathlineto{\pgfqpoint{1.708469in}{0.716028in}}%
\pgfpathclose%
\pgfusepath{fill}%
\end{pgfscope}%
\begin{pgfscope}%
\pgfpathrectangle{\pgfqpoint{0.592068in}{0.716028in}}{\pgfqpoint{5.395940in}{3.517843in}}%
\pgfusepath{clip}%
\pgfsetbuttcap%
\pgfsetmiterjoin%
\definecolor{currentfill}{rgb}{0.000000,0.000000,1.000000}%
\pgfsetfillcolor{currentfill}%
\pgfsetlinewidth{0.000000pt}%
\definecolor{currentstroke}{rgb}{0.000000,0.000000,0.000000}%
\pgfsetstrokecolor{currentstroke}%
\pgfsetdash{}{0pt}%
\pgfpathmoveto{\pgfqpoint{1.894536in}{0.716028in}}%
\pgfpathlineto{\pgfqpoint{2.080603in}{0.716028in}}%
\pgfpathlineto{\pgfqpoint{2.080603in}{1.689485in}}%
\pgfpathlineto{\pgfqpoint{1.894536in}{1.689485in}}%
\pgfpathlineto{\pgfqpoint{1.894536in}{0.716028in}}%
\pgfpathclose%
\pgfusepath{fill}%
\end{pgfscope}%
\begin{pgfscope}%
\pgfpathrectangle{\pgfqpoint{0.592068in}{0.716028in}}{\pgfqpoint{5.395940in}{3.517843in}}%
\pgfusepath{clip}%
\pgfsetbuttcap%
\pgfsetmiterjoin%
\definecolor{currentfill}{rgb}{0.000000,0.000000,1.000000}%
\pgfsetfillcolor{currentfill}%
\pgfsetlinewidth{0.000000pt}%
\definecolor{currentstroke}{rgb}{0.000000,0.000000,0.000000}%
\pgfsetstrokecolor{currentstroke}%
\pgfsetdash{}{0pt}%
\pgfpathmoveto{\pgfqpoint{2.080603in}{0.716028in}}%
\pgfpathlineto{\pgfqpoint{2.266670in}{0.716028in}}%
\pgfpathlineto{\pgfqpoint{2.266670in}{1.383671in}}%
\pgfpathlineto{\pgfqpoint{2.080603in}{1.383671in}}%
\pgfpathlineto{\pgfqpoint{2.080603in}{0.716028in}}%
\pgfpathclose%
\pgfusepath{fill}%
\end{pgfscope}%
\begin{pgfscope}%
\pgfpathrectangle{\pgfqpoint{0.592068in}{0.716028in}}{\pgfqpoint{5.395940in}{3.517843in}}%
\pgfusepath{clip}%
\pgfsetbuttcap%
\pgfsetmiterjoin%
\definecolor{currentfill}{rgb}{0.000000,0.000000,1.000000}%
\pgfsetfillcolor{currentfill}%
\pgfsetlinewidth{0.000000pt}%
\definecolor{currentstroke}{rgb}{0.000000,0.000000,0.000000}%
\pgfsetstrokecolor{currentstroke}%
\pgfsetdash{}{0pt}%
\pgfpathmoveto{\pgfqpoint{2.266670in}{0.716028in}}%
\pgfpathlineto{\pgfqpoint{2.452737in}{0.716028in}}%
\pgfpathlineto{\pgfqpoint{2.452737in}{1.357934in}}%
\pgfpathlineto{\pgfqpoint{2.266670in}{1.357934in}}%
\pgfpathlineto{\pgfqpoint{2.266670in}{0.716028in}}%
\pgfpathclose%
\pgfusepath{fill}%
\end{pgfscope}%
\begin{pgfscope}%
\pgfpathrectangle{\pgfqpoint{0.592068in}{0.716028in}}{\pgfqpoint{5.395940in}{3.517843in}}%
\pgfusepath{clip}%
\pgfsetbuttcap%
\pgfsetmiterjoin%
\definecolor{currentfill}{rgb}{0.000000,0.000000,1.000000}%
\pgfsetfillcolor{currentfill}%
\pgfsetlinewidth{0.000000pt}%
\definecolor{currentstroke}{rgb}{0.000000,0.000000,0.000000}%
\pgfsetstrokecolor{currentstroke}%
\pgfsetdash{}{0pt}%
\pgfpathmoveto{\pgfqpoint{2.452737in}{0.716028in}}%
\pgfpathlineto{\pgfqpoint{2.638804in}{0.716028in}}%
\pgfpathlineto{\pgfqpoint{2.638804in}{1.273154in}}%
\pgfpathlineto{\pgfqpoint{2.452737in}{1.273154in}}%
\pgfpathlineto{\pgfqpoint{2.452737in}{0.716028in}}%
\pgfpathclose%
\pgfusepath{fill}%
\end{pgfscope}%
\begin{pgfscope}%
\pgfpathrectangle{\pgfqpoint{0.592068in}{0.716028in}}{\pgfqpoint{5.395940in}{3.517843in}}%
\pgfusepath{clip}%
\pgfsetbuttcap%
\pgfsetmiterjoin%
\definecolor{currentfill}{rgb}{0.000000,0.000000,1.000000}%
\pgfsetfillcolor{currentfill}%
\pgfsetlinewidth{0.000000pt}%
\definecolor{currentstroke}{rgb}{0.000000,0.000000,0.000000}%
\pgfsetstrokecolor{currentstroke}%
\pgfsetdash{}{0pt}%
\pgfpathmoveto{\pgfqpoint{2.638804in}{0.716028in}}%
\pgfpathlineto{\pgfqpoint{2.824871in}{0.716028in}}%
\pgfpathlineto{\pgfqpoint{2.824871in}{1.270126in}}%
\pgfpathlineto{\pgfqpoint{2.638804in}{1.270126in}}%
\pgfpathlineto{\pgfqpoint{2.638804in}{0.716028in}}%
\pgfpathclose%
\pgfusepath{fill}%
\end{pgfscope}%
\begin{pgfscope}%
\pgfpathrectangle{\pgfqpoint{0.592068in}{0.716028in}}{\pgfqpoint{5.395940in}{3.517843in}}%
\pgfusepath{clip}%
\pgfsetbuttcap%
\pgfsetmiterjoin%
\definecolor{currentfill}{rgb}{0.000000,0.000000,1.000000}%
\pgfsetfillcolor{currentfill}%
\pgfsetlinewidth{0.000000pt}%
\definecolor{currentstroke}{rgb}{0.000000,0.000000,0.000000}%
\pgfsetstrokecolor{currentstroke}%
\pgfsetdash{}{0pt}%
\pgfpathmoveto{\pgfqpoint{2.824871in}{0.716028in}}%
\pgfpathlineto{\pgfqpoint{3.010937in}{0.716028in}}%
\pgfpathlineto{\pgfqpoint{3.010937in}{1.129331in}}%
\pgfpathlineto{\pgfqpoint{2.824871in}{1.129331in}}%
\pgfpathlineto{\pgfqpoint{2.824871in}{0.716028in}}%
\pgfpathclose%
\pgfusepath{fill}%
\end{pgfscope}%
\begin{pgfscope}%
\pgfpathrectangle{\pgfqpoint{0.592068in}{0.716028in}}{\pgfqpoint{5.395940in}{3.517843in}}%
\pgfusepath{clip}%
\pgfsetbuttcap%
\pgfsetmiterjoin%
\definecolor{currentfill}{rgb}{0.000000,0.000000,1.000000}%
\pgfsetfillcolor{currentfill}%
\pgfsetlinewidth{0.000000pt}%
\definecolor{currentstroke}{rgb}{0.000000,0.000000,0.000000}%
\pgfsetstrokecolor{currentstroke}%
\pgfsetdash{}{0pt}%
\pgfpathmoveto{\pgfqpoint{3.010937in}{0.716028in}}%
\pgfpathlineto{\pgfqpoint{3.197004in}{0.716028in}}%
\pgfpathlineto{\pgfqpoint{3.197004in}{0.971882in}}%
\pgfpathlineto{\pgfqpoint{3.010937in}{0.971882in}}%
\pgfpathlineto{\pgfqpoint{3.010937in}{0.716028in}}%
\pgfpathclose%
\pgfusepath{fill}%
\end{pgfscope}%
\begin{pgfscope}%
\pgfpathrectangle{\pgfqpoint{0.592068in}{0.716028in}}{\pgfqpoint{5.395940in}{3.517843in}}%
\pgfusepath{clip}%
\pgfsetbuttcap%
\pgfsetmiterjoin%
\definecolor{currentfill}{rgb}{0.000000,0.000000,1.000000}%
\pgfsetfillcolor{currentfill}%
\pgfsetlinewidth{0.000000pt}%
\definecolor{currentstroke}{rgb}{0.000000,0.000000,0.000000}%
\pgfsetstrokecolor{currentstroke}%
\pgfsetdash{}{0pt}%
\pgfpathmoveto{\pgfqpoint{3.197004in}{0.716028in}}%
\pgfpathlineto{\pgfqpoint{3.383071in}{0.716028in}}%
\pgfpathlineto{\pgfqpoint{3.383071in}{1.044551in}}%
\pgfpathlineto{\pgfqpoint{3.197004in}{1.044551in}}%
\pgfpathlineto{\pgfqpoint{3.197004in}{0.716028in}}%
\pgfpathclose%
\pgfusepath{fill}%
\end{pgfscope}%
\begin{pgfscope}%
\pgfpathrectangle{\pgfqpoint{0.592068in}{0.716028in}}{\pgfqpoint{5.395940in}{3.517843in}}%
\pgfusepath{clip}%
\pgfsetbuttcap%
\pgfsetmiterjoin%
\definecolor{currentfill}{rgb}{0.000000,0.000000,1.000000}%
\pgfsetfillcolor{currentfill}%
\pgfsetlinewidth{0.000000pt}%
\definecolor{currentstroke}{rgb}{0.000000,0.000000,0.000000}%
\pgfsetstrokecolor{currentstroke}%
\pgfsetdash{}{0pt}%
\pgfpathmoveto{\pgfqpoint{3.383071in}{0.716028in}}%
\pgfpathlineto{\pgfqpoint{3.569138in}{0.716028in}}%
\pgfpathlineto{\pgfqpoint{3.569138in}{0.991563in}}%
\pgfpathlineto{\pgfqpoint{3.383071in}{0.991563in}}%
\pgfpathlineto{\pgfqpoint{3.383071in}{0.716028in}}%
\pgfpathclose%
\pgfusepath{fill}%
\end{pgfscope}%
\begin{pgfscope}%
\pgfpathrectangle{\pgfqpoint{0.592068in}{0.716028in}}{\pgfqpoint{5.395940in}{3.517843in}}%
\pgfusepath{clip}%
\pgfsetbuttcap%
\pgfsetmiterjoin%
\definecolor{currentfill}{rgb}{0.000000,0.000000,1.000000}%
\pgfsetfillcolor{currentfill}%
\pgfsetlinewidth{0.000000pt}%
\definecolor{currentstroke}{rgb}{0.000000,0.000000,0.000000}%
\pgfsetstrokecolor{currentstroke}%
\pgfsetdash{}{0pt}%
\pgfpathmoveto{\pgfqpoint{3.569138in}{0.716028in}}%
\pgfpathlineto{\pgfqpoint{3.755205in}{0.716028in}}%
\pgfpathlineto{\pgfqpoint{3.755205in}{0.993077in}}%
\pgfpathlineto{\pgfqpoint{3.569138in}{0.993077in}}%
\pgfpathlineto{\pgfqpoint{3.569138in}{0.716028in}}%
\pgfpathclose%
\pgfusepath{fill}%
\end{pgfscope}%
\begin{pgfscope}%
\pgfpathrectangle{\pgfqpoint{0.592068in}{0.716028in}}{\pgfqpoint{5.395940in}{3.517843in}}%
\pgfusepath{clip}%
\pgfsetbuttcap%
\pgfsetmiterjoin%
\definecolor{currentfill}{rgb}{0.000000,0.000000,1.000000}%
\pgfsetfillcolor{currentfill}%
\pgfsetlinewidth{0.000000pt}%
\definecolor{currentstroke}{rgb}{0.000000,0.000000,0.000000}%
\pgfsetstrokecolor{currentstroke}%
\pgfsetdash{}{0pt}%
\pgfpathmoveto{\pgfqpoint{3.755205in}{0.716028in}}%
\pgfpathlineto{\pgfqpoint{3.941272in}{0.716028in}}%
\pgfpathlineto{\pgfqpoint{3.941272in}{0.894672in}}%
\pgfpathlineto{\pgfqpoint{3.755205in}{0.894672in}}%
\pgfpathlineto{\pgfqpoint{3.755205in}{0.716028in}}%
\pgfpathclose%
\pgfusepath{fill}%
\end{pgfscope}%
\begin{pgfscope}%
\pgfpathrectangle{\pgfqpoint{0.592068in}{0.716028in}}{\pgfqpoint{5.395940in}{3.517843in}}%
\pgfusepath{clip}%
\pgfsetbuttcap%
\pgfsetmiterjoin%
\definecolor{currentfill}{rgb}{0.000000,0.000000,1.000000}%
\pgfsetfillcolor{currentfill}%
\pgfsetlinewidth{0.000000pt}%
\definecolor{currentstroke}{rgb}{0.000000,0.000000,0.000000}%
\pgfsetstrokecolor{currentstroke}%
\pgfsetdash{}{0pt}%
\pgfpathmoveto{\pgfqpoint{3.941272in}{0.716028in}}%
\pgfpathlineto{\pgfqpoint{4.127339in}{0.716028in}}%
\pgfpathlineto{\pgfqpoint{4.127339in}{0.923437in}}%
\pgfpathlineto{\pgfqpoint{3.941272in}{0.923437in}}%
\pgfpathlineto{\pgfqpoint{3.941272in}{0.716028in}}%
\pgfpathclose%
\pgfusepath{fill}%
\end{pgfscope}%
\begin{pgfscope}%
\pgfpathrectangle{\pgfqpoint{0.592068in}{0.716028in}}{\pgfqpoint{5.395940in}{3.517843in}}%
\pgfusepath{clip}%
\pgfsetbuttcap%
\pgfsetmiterjoin%
\definecolor{currentfill}{rgb}{0.000000,0.000000,1.000000}%
\pgfsetfillcolor{currentfill}%
\pgfsetlinewidth{0.000000pt}%
\definecolor{currentstroke}{rgb}{0.000000,0.000000,0.000000}%
\pgfsetstrokecolor{currentstroke}%
\pgfsetdash{}{0pt}%
\pgfpathmoveto{\pgfqpoint{4.127339in}{0.716028in}}%
\pgfpathlineto{\pgfqpoint{4.313406in}{0.716028in}}%
\pgfpathlineto{\pgfqpoint{4.313406in}{0.994591in}}%
\pgfpathlineto{\pgfqpoint{4.127339in}{0.994591in}}%
\pgfpathlineto{\pgfqpoint{4.127339in}{0.716028in}}%
\pgfpathclose%
\pgfusepath{fill}%
\end{pgfscope}%
\begin{pgfscope}%
\pgfpathrectangle{\pgfqpoint{0.592068in}{0.716028in}}{\pgfqpoint{5.395940in}{3.517843in}}%
\pgfusepath{clip}%
\pgfsetbuttcap%
\pgfsetmiterjoin%
\definecolor{currentfill}{rgb}{0.000000,0.000000,1.000000}%
\pgfsetfillcolor{currentfill}%
\pgfsetlinewidth{0.000000pt}%
\definecolor{currentstroke}{rgb}{0.000000,0.000000,0.000000}%
\pgfsetstrokecolor{currentstroke}%
\pgfsetdash{}{0pt}%
\pgfpathmoveto{\pgfqpoint{4.313406in}{0.716028in}}%
\pgfpathlineto{\pgfqpoint{4.499473in}{0.716028in}}%
\pgfpathlineto{\pgfqpoint{4.499473in}{1.044551in}}%
\pgfpathlineto{\pgfqpoint{4.313406in}{1.044551in}}%
\pgfpathlineto{\pgfqpoint{4.313406in}{0.716028in}}%
\pgfpathclose%
\pgfusepath{fill}%
\end{pgfscope}%
\begin{pgfscope}%
\pgfpathrectangle{\pgfqpoint{0.592068in}{0.716028in}}{\pgfqpoint{5.395940in}{3.517843in}}%
\pgfusepath{clip}%
\pgfsetbuttcap%
\pgfsetmiterjoin%
\definecolor{currentfill}{rgb}{0.000000,0.000000,1.000000}%
\pgfsetfillcolor{currentfill}%
\pgfsetlinewidth{0.000000pt}%
\definecolor{currentstroke}{rgb}{0.000000,0.000000,0.000000}%
\pgfsetstrokecolor{currentstroke}%
\pgfsetdash{}{0pt}%
\pgfpathmoveto{\pgfqpoint{4.499473in}{0.716028in}}%
\pgfpathlineto{\pgfqpoint{4.685539in}{0.716028in}}%
\pgfpathlineto{\pgfqpoint{4.685539in}{1.197458in}}%
\pgfpathlineto{\pgfqpoint{4.499473in}{1.197458in}}%
\pgfpathlineto{\pgfqpoint{4.499473in}{0.716028in}}%
\pgfpathclose%
\pgfusepath{fill}%
\end{pgfscope}%
\begin{pgfscope}%
\pgfpathrectangle{\pgfqpoint{0.592068in}{0.716028in}}{\pgfqpoint{5.395940in}{3.517843in}}%
\pgfusepath{clip}%
\pgfsetbuttcap%
\pgfsetmiterjoin%
\definecolor{currentfill}{rgb}{0.000000,0.000000,1.000000}%
\pgfsetfillcolor{currentfill}%
\pgfsetlinewidth{0.000000pt}%
\definecolor{currentstroke}{rgb}{0.000000,0.000000,0.000000}%
\pgfsetstrokecolor{currentstroke}%
\pgfsetdash{}{0pt}%
\pgfpathmoveto{\pgfqpoint{4.685539in}{0.716028in}}%
\pgfpathlineto{\pgfqpoint{4.871606in}{0.716028in}}%
\pgfpathlineto{\pgfqpoint{4.871606in}{1.330684in}}%
\pgfpathlineto{\pgfqpoint{4.685539in}{1.330684in}}%
\pgfpathlineto{\pgfqpoint{4.685539in}{0.716028in}}%
\pgfpathclose%
\pgfusepath{fill}%
\end{pgfscope}%
\begin{pgfscope}%
\pgfpathrectangle{\pgfqpoint{0.592068in}{0.716028in}}{\pgfqpoint{5.395940in}{3.517843in}}%
\pgfusepath{clip}%
\pgfsetbuttcap%
\pgfsetmiterjoin%
\definecolor{currentfill}{rgb}{0.000000,0.000000,1.000000}%
\pgfsetfillcolor{currentfill}%
\pgfsetlinewidth{0.000000pt}%
\definecolor{currentstroke}{rgb}{0.000000,0.000000,0.000000}%
\pgfsetstrokecolor{currentstroke}%
\pgfsetdash{}{0pt}%
\pgfpathmoveto{\pgfqpoint{4.871606in}{0.716028in}}%
\pgfpathlineto{\pgfqpoint{5.057673in}{0.716028in}}%
\pgfpathlineto{\pgfqpoint{5.057673in}{1.174749in}}%
\pgfpathlineto{\pgfqpoint{4.871606in}{1.174749in}}%
\pgfpathlineto{\pgfqpoint{4.871606in}{0.716028in}}%
\pgfpathclose%
\pgfusepath{fill}%
\end{pgfscope}%
\begin{pgfscope}%
\pgfpathrectangle{\pgfqpoint{0.592068in}{0.716028in}}{\pgfqpoint{5.395940in}{3.517843in}}%
\pgfusepath{clip}%
\pgfsetbuttcap%
\pgfsetmiterjoin%
\definecolor{currentfill}{rgb}{0.000000,0.000000,1.000000}%
\pgfsetfillcolor{currentfill}%
\pgfsetlinewidth{0.000000pt}%
\definecolor{currentstroke}{rgb}{0.000000,0.000000,0.000000}%
\pgfsetstrokecolor{currentstroke}%
\pgfsetdash{}{0pt}%
\pgfpathmoveto{\pgfqpoint{5.057673in}{0.716028in}}%
\pgfpathlineto{\pgfqpoint{5.243740in}{0.716028in}}%
\pgfpathlineto{\pgfqpoint{5.243740in}{0.999133in}}%
\pgfpathlineto{\pgfqpoint{5.057673in}{0.999133in}}%
\pgfpathlineto{\pgfqpoint{5.057673in}{0.716028in}}%
\pgfpathclose%
\pgfusepath{fill}%
\end{pgfscope}%
\begin{pgfscope}%
\pgfpathrectangle{\pgfqpoint{0.592068in}{0.716028in}}{\pgfqpoint{5.395940in}{3.517843in}}%
\pgfusepath{clip}%
\pgfsetbuttcap%
\pgfsetmiterjoin%
\definecolor{currentfill}{rgb}{0.000000,0.000000,1.000000}%
\pgfsetfillcolor{currentfill}%
\pgfsetlinewidth{0.000000pt}%
\definecolor{currentstroke}{rgb}{0.000000,0.000000,0.000000}%
\pgfsetstrokecolor{currentstroke}%
\pgfsetdash{}{0pt}%
\pgfpathmoveto{\pgfqpoint{5.243740in}{0.716028in}}%
\pgfpathlineto{\pgfqpoint{5.429807in}{0.716028in}}%
\pgfpathlineto{\pgfqpoint{5.429807in}{0.917381in}}%
\pgfpathlineto{\pgfqpoint{5.243740in}{0.917381in}}%
\pgfpathlineto{\pgfqpoint{5.243740in}{0.716028in}}%
\pgfpathclose%
\pgfusepath{fill}%
\end{pgfscope}%
\begin{pgfscope}%
\pgfpathrectangle{\pgfqpoint{0.592068in}{0.716028in}}{\pgfqpoint{5.395940in}{3.517843in}}%
\pgfusepath{clip}%
\pgfsetbuttcap%
\pgfsetmiterjoin%
\definecolor{currentfill}{rgb}{0.000000,0.000000,1.000000}%
\pgfsetfillcolor{currentfill}%
\pgfsetlinewidth{0.000000pt}%
\definecolor{currentstroke}{rgb}{0.000000,0.000000,0.000000}%
\pgfsetstrokecolor{currentstroke}%
\pgfsetdash{}{0pt}%
\pgfpathmoveto{\pgfqpoint{5.429807in}{0.716028in}}%
\pgfpathlineto{\pgfqpoint{5.615874in}{0.716028in}}%
\pgfpathlineto{\pgfqpoint{5.615874in}{0.962799in}}%
\pgfpathlineto{\pgfqpoint{5.429807in}{0.962799in}}%
\pgfpathlineto{\pgfqpoint{5.429807in}{0.716028in}}%
\pgfpathclose%
\pgfusepath{fill}%
\end{pgfscope}%
\begin{pgfscope}%
\pgfpathrectangle{\pgfqpoint{0.592068in}{0.716028in}}{\pgfqpoint{5.395940in}{3.517843in}}%
\pgfusepath{clip}%
\pgfsetbuttcap%
\pgfsetmiterjoin%
\definecolor{currentfill}{rgb}{0.000000,0.000000,1.000000}%
\pgfsetfillcolor{currentfill}%
\pgfsetlinewidth{0.000000pt}%
\definecolor{currentstroke}{rgb}{0.000000,0.000000,0.000000}%
\pgfsetstrokecolor{currentstroke}%
\pgfsetdash{}{0pt}%
\pgfpathmoveto{\pgfqpoint{5.615874in}{0.716028in}}%
\pgfpathlineto{\pgfqpoint{5.801941in}{0.716028in}}%
\pgfpathlineto{\pgfqpoint{5.801941in}{1.077857in}}%
\pgfpathlineto{\pgfqpoint{5.615874in}{1.077857in}}%
\pgfpathlineto{\pgfqpoint{5.615874in}{0.716028in}}%
\pgfpathclose%
\pgfusepath{fill}%
\end{pgfscope}%
\begin{pgfscope}%
\pgfpathrectangle{\pgfqpoint{0.592068in}{0.716028in}}{\pgfqpoint{5.395940in}{3.517843in}}%
\pgfusepath{clip}%
\pgfsetbuttcap%
\pgfsetmiterjoin%
\definecolor{currentfill}{rgb}{0.000000,0.000000,1.000000}%
\pgfsetfillcolor{currentfill}%
\pgfsetlinewidth{0.000000pt}%
\definecolor{currentstroke}{rgb}{0.000000,0.000000,0.000000}%
\pgfsetstrokecolor{currentstroke}%
\pgfsetdash{}{0pt}%
\pgfpathmoveto{\pgfqpoint{5.801941in}{0.716028in}}%
\pgfpathlineto{\pgfqpoint{5.988008in}{0.716028in}}%
\pgfpathlineto{\pgfqpoint{5.988008in}{2.128525in}}%
\pgfpathlineto{\pgfqpoint{5.801941in}{2.128525in}}%
\pgfpathlineto{\pgfqpoint{5.801941in}{0.716028in}}%
\pgfpathclose%
\pgfusepath{fill}%
\end{pgfscope}%
\begin{pgfscope}%
\pgfsetbuttcap%
\pgfsetroundjoin%
\definecolor{currentfill}{rgb}{0.000000,0.000000,0.000000}%
\pgfsetfillcolor{currentfill}%
\pgfsetlinewidth{0.200750pt}%
\definecolor{currentstroke}{rgb}{0.000000,0.000000,0.000000}%
\pgfsetstrokecolor{currentstroke}%
\pgfsetdash{}{0pt}%
\pgfsys@defobject{currentmarker}{\pgfqpoint{0.000000in}{-0.048611in}}{\pgfqpoint{0.000000in}{0.000000in}}{%
\pgfpathmoveto{\pgfqpoint{0.000000in}{0.000000in}}%
\pgfpathlineto{\pgfqpoint{0.000000in}{-0.048611in}}%
\pgfusepath{stroke,fill}%
}%
\begin{pgfscope}%
\pgfsys@transformshift{0.592068in}{0.716028in}%
\pgfsys@useobject{currentmarker}{}%
\end{pgfscope}%
\end{pgfscope}%
\begin{pgfscope}%
\definecolor{textcolor}{rgb}{0.000000,0.000000,0.000000}%
\pgfsetstrokecolor{textcolor}%
\pgfsetfillcolor{textcolor}%
\pgftext[x=0.592068in,y=0.618806in,,top]{\color{textcolor}{\rmfamily\fontsize{20.000000}{24.000000}\selectfont\catcode`\^=\active\def^{\ifmmode\sp\else\^{}\fi}\catcode`\%=\active\def
\end{pgfscope}%
\begin{pgfscope}%
\pgfsetbuttcap%
\pgfsetroundjoin%
\definecolor{currentfill}{rgb}{0.000000,0.000000,0.000000}%
\pgfsetfillcolor{currentfill}%
\pgfsetlinewidth{0.200750pt}%
\definecolor{currentstroke}{rgb}{0.000000,0.000000,0.000000}%
\pgfsetstrokecolor{currentstroke}%
\pgfsetdash{}{0pt}%
\pgfsys@defobject{currentmarker}{\pgfqpoint{0.000000in}{-0.048611in}}{\pgfqpoint{0.000000in}{0.000000in}}{%
\pgfpathmoveto{\pgfqpoint{0.000000in}{0.000000in}}%
\pgfpathlineto{\pgfqpoint{0.000000in}{-0.048611in}}%
\pgfusepath{stroke,fill}%
}%
\begin{pgfscope}%
\pgfsys@transformshift{1.671256in}{0.716028in}%
\pgfsys@useobject{currentmarker}{}%
\end{pgfscope}%
\end{pgfscope}%
\begin{pgfscope}%
\definecolor{textcolor}{rgb}{0.000000,0.000000,0.000000}%
\pgfsetstrokecolor{textcolor}%
\pgfsetfillcolor{textcolor}%
\pgftext[x=1.671256in,y=0.618806in,,top]{\color{textcolor}{\rmfamily\fontsize{20.000000}{24.000000}\selectfont\catcode`\^=\active\def^{\ifmmode\sp\else\^{}\fi}\catcode`\%=\active\def
\end{pgfscope}%
\begin{pgfscope}%
\pgfsetbuttcap%
\pgfsetroundjoin%
\definecolor{currentfill}{rgb}{0.000000,0.000000,0.000000}%
\pgfsetfillcolor{currentfill}%
\pgfsetlinewidth{0.200750pt}%
\definecolor{currentstroke}{rgb}{0.000000,0.000000,0.000000}%
\pgfsetstrokecolor{currentstroke}%
\pgfsetdash{}{0pt}%
\pgfsys@defobject{currentmarker}{\pgfqpoint{0.000000in}{-0.048611in}}{\pgfqpoint{0.000000in}{0.000000in}}{%
\pgfpathmoveto{\pgfqpoint{0.000000in}{0.000000in}}%
\pgfpathlineto{\pgfqpoint{0.000000in}{-0.048611in}}%
\pgfusepath{stroke,fill}%
}%
\begin{pgfscope}%
\pgfsys@transformshift{2.750444in}{0.716028in}%
\pgfsys@useobject{currentmarker}{}%
\end{pgfscope}%
\end{pgfscope}%
\begin{pgfscope}%
\definecolor{textcolor}{rgb}{0.000000,0.000000,0.000000}%
\pgfsetstrokecolor{textcolor}%
\pgfsetfillcolor{textcolor}%
\pgftext[x=2.750444in,y=0.618806in,,top]{\color{textcolor}{\rmfamily\fontsize{20.000000}{24.000000}\selectfont\catcode`\^=\active\def^{\ifmmode\sp\else\^{}\fi}\catcode`\%=\active\def
\end{pgfscope}%
\begin{pgfscope}%
\pgfsetbuttcap%
\pgfsetroundjoin%
\definecolor{currentfill}{rgb}{0.000000,0.000000,0.000000}%
\pgfsetfillcolor{currentfill}%
\pgfsetlinewidth{0.200750pt}%
\definecolor{currentstroke}{rgb}{0.000000,0.000000,0.000000}%
\pgfsetstrokecolor{currentstroke}%
\pgfsetdash{}{0pt}%
\pgfsys@defobject{currentmarker}{\pgfqpoint{0.000000in}{-0.048611in}}{\pgfqpoint{0.000000in}{0.000000in}}{%
\pgfpathmoveto{\pgfqpoint{0.000000in}{0.000000in}}%
\pgfpathlineto{\pgfqpoint{0.000000in}{-0.048611in}}%
\pgfusepath{stroke,fill}%
}%
\begin{pgfscope}%
\pgfsys@transformshift{3.829632in}{0.716028in}%
\pgfsys@useobject{currentmarker}{}%
\end{pgfscope}%
\end{pgfscope}%
\begin{pgfscope}%
\definecolor{textcolor}{rgb}{0.000000,0.000000,0.000000}%
\pgfsetstrokecolor{textcolor}%
\pgfsetfillcolor{textcolor}%
\pgftext[x=3.829632in,y=0.618806in,,top]{\color{textcolor}{\rmfamily\fontsize{20.000000}{24.000000}\selectfont\catcode`\^=\active\def^{\ifmmode\sp\else\^{}\fi}\catcode`\%=\active\def
\end{pgfscope}%
\begin{pgfscope}%
\pgfsetbuttcap%
\pgfsetroundjoin%
\definecolor{currentfill}{rgb}{0.000000,0.000000,0.000000}%
\pgfsetfillcolor{currentfill}%
\pgfsetlinewidth{0.200750pt}%
\definecolor{currentstroke}{rgb}{0.000000,0.000000,0.000000}%
\pgfsetstrokecolor{currentstroke}%
\pgfsetdash{}{0pt}%
\pgfsys@defobject{currentmarker}{\pgfqpoint{0.000000in}{-0.048611in}}{\pgfqpoint{0.000000in}{0.000000in}}{%
\pgfpathmoveto{\pgfqpoint{0.000000in}{0.000000in}}%
\pgfpathlineto{\pgfqpoint{0.000000in}{-0.048611in}}%
\pgfusepath{stroke,fill}%
}%
\begin{pgfscope}%
\pgfsys@transformshift{4.908820in}{0.716028in}%
\pgfsys@useobject{currentmarker}{}%
\end{pgfscope}%
\end{pgfscope}%
\begin{pgfscope}%
\definecolor{textcolor}{rgb}{0.000000,0.000000,0.000000}%
\pgfsetstrokecolor{textcolor}%
\pgfsetfillcolor{textcolor}%
\pgftext[x=4.908820in,y=0.618806in,,top]{\color{textcolor}{\rmfamily\fontsize{20.000000}{24.000000}\selectfont\catcode`\^=\active\def^{\ifmmode\sp\else\^{}\fi}\catcode`\%=\active\def
\end{pgfscope}%
\begin{pgfscope}%
\pgfsetbuttcap%
\pgfsetroundjoin%
\definecolor{currentfill}{rgb}{0.000000,0.000000,0.000000}%
\pgfsetfillcolor{currentfill}%
\pgfsetlinewidth{0.200750pt}%
\definecolor{currentstroke}{rgb}{0.000000,0.000000,0.000000}%
\pgfsetstrokecolor{currentstroke}%
\pgfsetdash{}{0pt}%
\pgfsys@defobject{currentmarker}{\pgfqpoint{0.000000in}{-0.048611in}}{\pgfqpoint{0.000000in}{0.000000in}}{%
\pgfpathmoveto{\pgfqpoint{0.000000in}{0.000000in}}%
\pgfpathlineto{\pgfqpoint{0.000000in}{-0.048611in}}%
\pgfusepath{stroke,fill}%
}%
\begin{pgfscope}%
\pgfsys@transformshift{5.988008in}{0.716028in}%
\pgfsys@useobject{currentmarker}{}%
\end{pgfscope}%
\end{pgfscope}%
\begin{pgfscope}%
\definecolor{textcolor}{rgb}{0.000000,0.000000,0.000000}%
\pgfsetstrokecolor{textcolor}%
\pgfsetfillcolor{textcolor}%
\pgftext[x=5.988008in,y=0.618806in,,top]{\color{textcolor}{\rmfamily\fontsize{20.000000}{24.000000}\selectfont\catcode`\^=\active\def^{\ifmmode\sp\else\^{}\fi}\catcode`\%=\active\def
\end{pgfscope}%
\begin{pgfscope}%
\definecolor{textcolor}{rgb}{0.000000,0.000000,0.000000}%
\pgfsetstrokecolor{textcolor}%
\pgfsetfillcolor{textcolor}%
\pgftext[x=3.290038in,y=0.307183in,,top]{\color{textcolor}{\rmfamily\fontsize{26.000000}{31.200000}\selectfont\catcode`\^=\active\def^{\ifmmode\sp\else\^{}\fi}\catcode`\%=\active\def
\end{pgfscope}%
\begin{pgfscope}%
\pgfsetbuttcap%
\pgfsetroundjoin%
\definecolor{currentfill}{rgb}{0.000000,0.000000,0.000000}%
\pgfsetfillcolor{currentfill}%
\pgfsetlinewidth{0.200750pt}%
\definecolor{currentstroke}{rgb}{0.000000,0.000000,0.000000}%
\pgfsetstrokecolor{currentstroke}%
\pgfsetdash{}{0pt}%
\pgfsys@defobject{currentmarker}{\pgfqpoint{-0.048611in}{0.000000in}}{\pgfqpoint{-0.000000in}{0.000000in}}{%
\pgfpathmoveto{\pgfqpoint{-0.000000in}{0.000000in}}%
\pgfpathlineto{\pgfqpoint{-0.048611in}{0.000000in}}%
\pgfusepath{stroke,fill}%
}%
\begin{pgfscope}%
\pgfsys@transformshift{0.592068in}{0.716028in}%
\pgfsys@useobject{currentmarker}{}%
\end{pgfscope}%
\end{pgfscope}%
\begin{pgfscope}%
\definecolor{textcolor}{rgb}{0.000000,0.000000,0.000000}%
\pgfsetstrokecolor{textcolor}%
\pgfsetfillcolor{textcolor}%
\pgftext[x=0.362738in, y=0.616009in, left, base]{\color{textcolor}{\rmfamily\fontsize{20.000000}{24.000000}\selectfont\catcode`\^=\active\def^{\ifmmode\sp\else\^{}\fi}\catcode`\%=\active\def
\end{pgfscope}%
\begin{pgfscope}%
\pgfsetbuttcap%
\pgfsetroundjoin%
\definecolor{currentfill}{rgb}{0.000000,0.000000,0.000000}%
\pgfsetfillcolor{currentfill}%
\pgfsetlinewidth{0.200750pt}%
\definecolor{currentstroke}{rgb}{0.000000,0.000000,0.000000}%
\pgfsetstrokecolor{currentstroke}%
\pgfsetdash{}{0pt}%
\pgfsys@defobject{currentmarker}{\pgfqpoint{-0.048611in}{0.000000in}}{\pgfqpoint{-0.000000in}{0.000000in}}{%
\pgfpathmoveto{\pgfqpoint{-0.000000in}{0.000000in}}%
\pgfpathlineto{\pgfqpoint{-0.048611in}{0.000000in}}%
\pgfusepath{stroke,fill}%
}%
\begin{pgfscope}%
\pgfsys@transformshift{0.592068in}{1.669125in}%
\pgfsys@useobject{currentmarker}{}%
\end{pgfscope}%
\end{pgfscope}%
\begin{pgfscope}%
\definecolor{textcolor}{rgb}{0.000000,0.000000,0.000000}%
\pgfsetstrokecolor{textcolor}%
\pgfsetfillcolor{textcolor}%
\pgftext[x=0.362738in, y=1.569106in, left, base]{\color{textcolor}{\rmfamily\fontsize{20.000000}{24.000000}\selectfont\catcode`\^=\active\def^{\ifmmode\sp\else\^{}\fi}\catcode`\%=\active\def
\end{pgfscope}%
\begin{pgfscope}%
\pgfsetbuttcap%
\pgfsetroundjoin%
\definecolor{currentfill}{rgb}{0.000000,0.000000,0.000000}%
\pgfsetfillcolor{currentfill}%
\pgfsetlinewidth{0.200750pt}%
\definecolor{currentstroke}{rgb}{0.000000,0.000000,0.000000}%
\pgfsetstrokecolor{currentstroke}%
\pgfsetdash{}{0pt}%
\pgfsys@defobject{currentmarker}{\pgfqpoint{-0.048611in}{0.000000in}}{\pgfqpoint{-0.000000in}{0.000000in}}{%
\pgfpathmoveto{\pgfqpoint{-0.000000in}{0.000000in}}%
\pgfpathlineto{\pgfqpoint{-0.048611in}{0.000000in}}%
\pgfusepath{stroke,fill}%
}%
\begin{pgfscope}%
\pgfsys@transformshift{0.592068in}{2.622222in}%
\pgfsys@useobject{currentmarker}{}%
\end{pgfscope}%
\end{pgfscope}%
\begin{pgfscope}%
\definecolor{textcolor}{rgb}{0.000000,0.000000,0.000000}%
\pgfsetstrokecolor{textcolor}%
\pgfsetfillcolor{textcolor}%
\pgftext[x=0.362738in, y=2.522203in, left, base]{\color{textcolor}{\rmfamily\fontsize{20.000000}{24.000000}\selectfont\catcode`\^=\active\def^{\ifmmode\sp\else\^{}\fi}\catcode`\%=\active\def
\end{pgfscope}%
\begin{pgfscope}%
\pgfsetbuttcap%
\pgfsetroundjoin%
\definecolor{currentfill}{rgb}{0.000000,0.000000,0.000000}%
\pgfsetfillcolor{currentfill}%
\pgfsetlinewidth{0.200750pt}%
\definecolor{currentstroke}{rgb}{0.000000,0.000000,0.000000}%
\pgfsetstrokecolor{currentstroke}%
\pgfsetdash{}{0pt}%
\pgfsys@defobject{currentmarker}{\pgfqpoint{-0.048611in}{0.000000in}}{\pgfqpoint{-0.000000in}{0.000000in}}{%
\pgfpathmoveto{\pgfqpoint{-0.000000in}{0.000000in}}%
\pgfpathlineto{\pgfqpoint{-0.048611in}{0.000000in}}%
\pgfusepath{stroke,fill}%
}%
\begin{pgfscope}%
\pgfsys@transformshift{0.592068in}{3.575319in}%
\pgfsys@useobject{currentmarker}{}%
\end{pgfscope}%
\end{pgfscope}%
\begin{pgfscope}%
\definecolor{textcolor}{rgb}{0.000000,0.000000,0.000000}%
\pgfsetstrokecolor{textcolor}%
\pgfsetfillcolor{textcolor}%
\pgftext[x=0.362738in, y=3.475300in, left, base]{\color{textcolor}{\rmfamily\fontsize{20.000000}{24.000000}\selectfont\catcode`\^=\active\def^{\ifmmode\sp\else\^{}\fi}\catcode`\%=\active\def
\end{pgfscope}%
\begin{pgfscope}%
\definecolor{textcolor}{rgb}{0.000000,0.000000,0.000000}%
\pgfsetstrokecolor{textcolor}%
\pgfsetfillcolor{textcolor}%
\pgftext[x=0.307183in,y=2.474950in,,bottom,rotate=90.000000]{\color{textcolor}{\rmfamily\fontsize{26.000000}{31.200000}\selectfont\catcode`\^=\active\def^{\ifmmode\sp\else\^{}\fi}\catcode`\%=\active\def
\end{pgfscope}%
\begin{pgfscope}%
\pgfpathrectangle{\pgfqpoint{0.592068in}{0.716028in}}{\pgfqpoint{5.395940in}{3.517843in}}%
\pgfusepath{clip}%
\pgfsetbuttcap%
\pgfsetroundjoin%
\pgfsetlinewidth{2.007500pt}%
\definecolor{currentstroke}{rgb}{0.501961,0.501961,0.501961}%
\pgfsetstrokecolor{currentstroke}%
\pgfsetdash{{7.400000pt}{3.200000pt}}{0.000000pt}%
\pgfpathmoveto{\pgfqpoint{0.592068in}{1.669125in}}%
\pgfpathlineto{\pgfqpoint{5.988008in}{1.669125in}}%
\pgfusepath{stroke}%
\end{pgfscope}%
\begin{pgfscope}%
\pgfsetrectcap%
\pgfsetmiterjoin%
\pgfsetlinewidth{0.803000pt}%
\definecolor{currentstroke}{rgb}{0.000000,0.000000,0.000000}%
\pgfsetstrokecolor{currentstroke}%
\pgfsetdash{}{0pt}%
\pgfpathmoveto{\pgfqpoint{0.592068in}{0.716028in}}%
\pgfpathlineto{\pgfqpoint{0.592068in}{4.233871in}}%
\pgfusepath{stroke}%
\end{pgfscope}%
\begin{pgfscope}%
\pgfsetrectcap%
\pgfsetmiterjoin%
\pgfsetlinewidth{0.803000pt}%
\definecolor{currentstroke}{rgb}{0.000000,0.000000,0.000000}%
\pgfsetstrokecolor{currentstroke}%
\pgfsetdash{}{0pt}%
\pgfpathmoveto{\pgfqpoint{5.988008in}{0.716028in}}%
\pgfpathlineto{\pgfqpoint{5.988008in}{4.233871in}}%
\pgfusepath{stroke}%
\end{pgfscope}%
\begin{pgfscope}%
\pgfsetrectcap%
\pgfsetmiterjoin%
\pgfsetlinewidth{0.803000pt}%
\definecolor{currentstroke}{rgb}{0.000000,0.000000,0.000000}%
\pgfsetstrokecolor{currentstroke}%
\pgfsetdash{}{0pt}%
\pgfpathmoveto{\pgfqpoint{0.592068in}{0.716028in}}%
\pgfpathlineto{\pgfqpoint{5.988008in}{0.716028in}}%
\pgfusepath{stroke}%
\end{pgfscope}%
\begin{pgfscope}%
\pgfsetrectcap%
\pgfsetmiterjoin%
\pgfsetlinewidth{0.803000pt}%
\definecolor{currentstroke}{rgb}{0.000000,0.000000,0.000000}%
\pgfsetstrokecolor{currentstroke}%
\pgfsetdash{}{0pt}%
\pgfpathmoveto{\pgfqpoint{0.592068in}{4.233871in}}%
\pgfpathlineto{\pgfqpoint{5.988008in}{4.233871in}}%
\pgfusepath{stroke}%
\end{pgfscope}%
\begin{pgfscope}%
\definecolor{textcolor}{rgb}{0.000000,0.000000,0.000000}%
\pgfsetstrokecolor{textcolor}%
\pgfsetfillcolor{textcolor}%
\pgftext[x=3.290038in,y=4.317204in,,base]{\color{textcolor}{\rmfamily\fontsize{25.000000}{30.000000}\selectfont\catcode`\^=\active\def^{\ifmmode\sp\else\^{}\fi}\catcode`\%=\active\def
\end{pgfscope}%
\begin{pgfscope}%
\pgfsetbuttcap%
\pgfsetmiterjoin%
\definecolor{currentfill}{rgb}{1.000000,1.000000,1.000000}%
\pgfsetfillcolor{currentfill}%
\pgfsetfillopacity{0.800000}%
\pgfsetlinewidth{1.003750pt}%
\definecolor{currentstroke}{rgb}{0.800000,0.800000,0.800000}%
\pgfsetstrokecolor{currentstroke}%
\pgfsetstrokeopacity{0.800000}%
\pgfsetdash{}{0pt}%
\pgfpathmoveto{\pgfqpoint{4.251661in}{3.616692in}}%
\pgfpathlineto{\pgfqpoint{5.793563in}{3.616692in}}%
\pgfpathquadraticcurveto{\pgfqpoint{5.849119in}{3.616692in}}{\pgfqpoint{5.849119in}{3.672248in}}%
\pgfpathlineto{\pgfqpoint{5.849119in}{4.039427in}}%
\pgfpathquadraticcurveto{\pgfqpoint{5.849119in}{4.094982in}}{\pgfqpoint{5.793563in}{4.094982in}}%
\pgfpathlineto{\pgfqpoint{4.251661in}{4.094982in}}%
\pgfpathquadraticcurveto{\pgfqpoint{4.196106in}{4.094982in}}{\pgfqpoint{4.196106in}{4.039427in}}%
\pgfpathlineto{\pgfqpoint{4.196106in}{3.672248in}}%
\pgfpathquadraticcurveto{\pgfqpoint{4.196106in}{3.616692in}}{\pgfqpoint{4.251661in}{3.616692in}}%
\pgfpathlineto{\pgfqpoint{4.251661in}{3.616692in}}%
\pgfpathclose%
\pgfusepath{stroke,fill}%
\end{pgfscope}%
\begin{pgfscope}%
\pgfsetbuttcap%
\pgfsetroundjoin%
\pgfsetlinewidth{2.007500pt}%
\definecolor{currentstroke}{rgb}{0.501961,0.501961,0.501961}%
\pgfsetstrokecolor{currentstroke}%
\pgfsetdash{{7.400000pt}{3.200000pt}}{0.000000pt}%
\pgfpathmoveto{\pgfqpoint{4.307217in}{3.881055in}}%
\pgfpathlineto{\pgfqpoint{4.584995in}{3.881055in}}%
\pgfpathlineto{\pgfqpoint{4.862772in}{3.881055in}}%
\pgfusepath{stroke}%
\end{pgfscope}%
\begin{pgfscope}%
\definecolor{textcolor}{rgb}{0.000000,0.000000,0.000000}%
\pgfsetstrokecolor{textcolor}%
\pgfsetfillcolor{textcolor}%
\pgftext[x=5.084995in,y=3.783833in,left,base]{\color{textcolor}{\rmfamily\fontsize{20.000000}{24.000000}\selectfont\catcode`\^=\active\def^{\ifmmode\sp\else\^{}\fi}\catcode`\%=\active\def
\end{pgfscope}%
\end{pgfpicture}%
\makeatother%
\endgroup%

%% file: figures/evaluation/Repair/Repair_PIT_event_elapsed_norm_4layer.pgf
\begingroup%
\makeatletter%
\begin{pgfpicture}%
\pgfpathrectangle{\pgfpointorigin}{\pgfqpoint{6.167556in}{4.557174in}}%
\pgfusepath{use as bounding box, clip}%
\begin{pgfscope}%
\pgfsetbuttcap%
\pgfsetmiterjoin%
\definecolor{currentfill}{rgb}{1.000000,1.000000,1.000000}%
\pgfsetfillcolor{currentfill}%
\pgfsetlinewidth{0.000000pt}%
\definecolor{currentstroke}{rgb}{1.000000,1.000000,1.000000}%
\pgfsetstrokecolor{currentstroke}%
\pgfsetdash{}{0pt}%
\pgfpathmoveto{\pgfqpoint{0.000000in}{0.000000in}}%
\pgfpathlineto{\pgfqpoint{6.167556in}{0.000000in}}%
\pgfpathlineto{\pgfqpoint{6.167556in}{4.557174in}}%
\pgfpathlineto{\pgfqpoint{0.000000in}{4.557174in}}%
\pgfpathlineto{\pgfqpoint{0.000000in}{0.000000in}}%
\pgfpathclose%
\pgfusepath{fill}%
\end{pgfscope}%
\begin{pgfscope}%
\pgfsetbuttcap%
\pgfsetmiterjoin%
\definecolor{currentfill}{rgb}{1.000000,1.000000,1.000000}%
\pgfsetfillcolor{currentfill}%
\pgfsetlinewidth{0.000000pt}%
\definecolor{currentstroke}{rgb}{0.000000,0.000000,0.000000}%
\pgfsetstrokecolor{currentstroke}%
\pgfsetstrokeopacity{0.000000}%
\pgfsetdash{}{0pt}%
\pgfpathmoveto{\pgfqpoint{0.802523in}{0.716028in}}%
\pgfpathlineto{\pgfqpoint{5.996275in}{0.716028in}}%
\pgfpathlineto{\pgfqpoint{5.996275in}{4.233871in}}%
\pgfpathlineto{\pgfqpoint{0.802523in}{4.233871in}}%
\pgfpathlineto{\pgfqpoint{0.802523in}{0.716028in}}%
\pgfpathclose%
\pgfusepath{fill}%
\end{pgfscope}%
\begin{pgfscope}%
\pgfpathrectangle{\pgfqpoint{0.802523in}{0.716028in}}{\pgfqpoint{5.193752in}{3.517843in}}%
\pgfusepath{clip}%
\pgfsetbuttcap%
\pgfsetmiterjoin%
\definecolor{currentfill}{rgb}{0.000000,0.000000,1.000000}%
\pgfsetfillcolor{currentfill}%
\pgfsetlinewidth{0.000000pt}%
\definecolor{currentstroke}{rgb}{0.000000,0.000000,0.000000}%
\pgfsetstrokecolor{currentstroke}%
\pgfsetdash{}{0pt}%
\pgfpathmoveto{\pgfqpoint{0.802523in}{0.716028in}}%
\pgfpathlineto{\pgfqpoint{0.981618in}{0.716028in}}%
\pgfpathlineto{\pgfqpoint{0.981618in}{0.716028in}}%
\pgfpathlineto{\pgfqpoint{0.802523in}{0.716028in}}%
\pgfpathlineto{\pgfqpoint{0.802523in}{0.716028in}}%
\pgfpathclose%
\pgfusepath{fill}%
\end{pgfscope}%
\begin{pgfscope}%
\pgfpathrectangle{\pgfqpoint{0.802523in}{0.716028in}}{\pgfqpoint{5.193752in}{3.517843in}}%
\pgfusepath{clip}%
\pgfsetbuttcap%
\pgfsetmiterjoin%
\definecolor{currentfill}{rgb}{0.000000,0.000000,1.000000}%
\pgfsetfillcolor{currentfill}%
\pgfsetlinewidth{0.000000pt}%
\definecolor{currentstroke}{rgb}{0.000000,0.000000,0.000000}%
\pgfsetstrokecolor{currentstroke}%
\pgfsetdash{}{0pt}%
\pgfpathmoveto{\pgfqpoint{0.981618in}{0.716028in}}%
\pgfpathlineto{\pgfqpoint{1.160713in}{0.716028in}}%
\pgfpathlineto{\pgfqpoint{1.160713in}{1.109912in}}%
\pgfpathlineto{\pgfqpoint{0.981618in}{1.109912in}}%
\pgfpathlineto{\pgfqpoint{0.981618in}{0.716028in}}%
\pgfpathclose%
\pgfusepath{fill}%
\end{pgfscope}%
\begin{pgfscope}%
\pgfpathrectangle{\pgfqpoint{0.802523in}{0.716028in}}{\pgfqpoint{5.193752in}{3.517843in}}%
\pgfusepath{clip}%
\pgfsetbuttcap%
\pgfsetmiterjoin%
\definecolor{currentfill}{rgb}{0.000000,0.000000,1.000000}%
\pgfsetfillcolor{currentfill}%
\pgfsetlinewidth{0.000000pt}%
\definecolor{currentstroke}{rgb}{0.000000,0.000000,0.000000}%
\pgfsetstrokecolor{currentstroke}%
\pgfsetdash{}{0pt}%
\pgfpathmoveto{\pgfqpoint{1.160713in}{0.716028in}}%
\pgfpathlineto{\pgfqpoint{1.339808in}{0.716028in}}%
\pgfpathlineto{\pgfqpoint{1.339808in}{2.486187in}}%
\pgfpathlineto{\pgfqpoint{1.160713in}{2.486187in}}%
\pgfpathlineto{\pgfqpoint{1.160713in}{0.716028in}}%
\pgfpathclose%
\pgfusepath{fill}%
\end{pgfscope}%
\begin{pgfscope}%
\pgfpathrectangle{\pgfqpoint{0.802523in}{0.716028in}}{\pgfqpoint{5.193752in}{3.517843in}}%
\pgfusepath{clip}%
\pgfsetbuttcap%
\pgfsetmiterjoin%
\definecolor{currentfill}{rgb}{0.000000,0.000000,1.000000}%
\pgfsetfillcolor{currentfill}%
\pgfsetlinewidth{0.000000pt}%
\definecolor{currentstroke}{rgb}{0.000000,0.000000,0.000000}%
\pgfsetstrokecolor{currentstroke}%
\pgfsetdash{}{0pt}%
\pgfpathmoveto{\pgfqpoint{1.339808in}{0.716028in}}%
\pgfpathlineto{\pgfqpoint{1.518902in}{0.716028in}}%
\pgfpathlineto{\pgfqpoint{1.518902in}{3.510284in}}%
\pgfpathlineto{\pgfqpoint{1.339808in}{3.510284in}}%
\pgfpathlineto{\pgfqpoint{1.339808in}{0.716028in}}%
\pgfpathclose%
\pgfusepath{fill}%
\end{pgfscope}%
\begin{pgfscope}%
\pgfpathrectangle{\pgfqpoint{0.802523in}{0.716028in}}{\pgfqpoint{5.193752in}{3.517843in}}%
\pgfusepath{clip}%
\pgfsetbuttcap%
\pgfsetmiterjoin%
\definecolor{currentfill}{rgb}{0.000000,0.000000,1.000000}%
\pgfsetfillcolor{currentfill}%
\pgfsetlinewidth{0.000000pt}%
\definecolor{currentstroke}{rgb}{0.000000,0.000000,0.000000}%
\pgfsetstrokecolor{currentstroke}%
\pgfsetdash{}{0pt}%
\pgfpathmoveto{\pgfqpoint{1.518902in}{0.716028in}}%
\pgfpathlineto{\pgfqpoint{1.697997in}{0.716028in}}%
\pgfpathlineto{\pgfqpoint{1.697997in}{4.066355in}}%
\pgfpathlineto{\pgfqpoint{1.518902in}{4.066355in}}%
\pgfpathlineto{\pgfqpoint{1.518902in}{0.716028in}}%
\pgfpathclose%
\pgfusepath{fill}%
\end{pgfscope}%
\begin{pgfscope}%
\pgfpathrectangle{\pgfqpoint{0.802523in}{0.716028in}}{\pgfqpoint{5.193752in}{3.517843in}}%
\pgfusepath{clip}%
\pgfsetbuttcap%
\pgfsetmiterjoin%
\definecolor{currentfill}{rgb}{0.000000,0.000000,1.000000}%
\pgfsetfillcolor{currentfill}%
\pgfsetlinewidth{0.000000pt}%
\definecolor{currentstroke}{rgb}{0.000000,0.000000,0.000000}%
\pgfsetstrokecolor{currentstroke}%
\pgfsetdash{}{0pt}%
\pgfpathmoveto{\pgfqpoint{1.697997in}{0.716028in}}%
\pgfpathlineto{\pgfqpoint{1.877092in}{0.716028in}}%
\pgfpathlineto{\pgfqpoint{1.877092in}{3.959775in}}%
\pgfpathlineto{\pgfqpoint{1.697997in}{3.959775in}}%
\pgfpathlineto{\pgfqpoint{1.697997in}{0.716028in}}%
\pgfpathclose%
\pgfusepath{fill}%
\end{pgfscope}%
\begin{pgfscope}%
\pgfpathrectangle{\pgfqpoint{0.802523in}{0.716028in}}{\pgfqpoint{5.193752in}{3.517843in}}%
\pgfusepath{clip}%
\pgfsetbuttcap%
\pgfsetmiterjoin%
\definecolor{currentfill}{rgb}{0.000000,0.000000,1.000000}%
\pgfsetfillcolor{currentfill}%
\pgfsetlinewidth{0.000000pt}%
\definecolor{currentstroke}{rgb}{0.000000,0.000000,0.000000}%
\pgfsetstrokecolor{currentstroke}%
\pgfsetdash{}{0pt}%
\pgfpathmoveto{\pgfqpoint{1.877092in}{0.716028in}}%
\pgfpathlineto{\pgfqpoint{2.056187in}{0.716028in}}%
\pgfpathlineto{\pgfqpoint{2.056187in}{3.765150in}}%
\pgfpathlineto{\pgfqpoint{1.877092in}{3.765150in}}%
\pgfpathlineto{\pgfqpoint{1.877092in}{0.716028in}}%
\pgfpathclose%
\pgfusepath{fill}%
\end{pgfscope}%
\begin{pgfscope}%
\pgfpathrectangle{\pgfqpoint{0.802523in}{0.716028in}}{\pgfqpoint{5.193752in}{3.517843in}}%
\pgfusepath{clip}%
\pgfsetbuttcap%
\pgfsetmiterjoin%
\definecolor{currentfill}{rgb}{0.000000,0.000000,1.000000}%
\pgfsetfillcolor{currentfill}%
\pgfsetlinewidth{0.000000pt}%
\definecolor{currentstroke}{rgb}{0.000000,0.000000,0.000000}%
\pgfsetstrokecolor{currentstroke}%
\pgfsetdash{}{0pt}%
\pgfpathmoveto{\pgfqpoint{2.056187in}{0.716028in}}%
\pgfpathlineto{\pgfqpoint{2.235282in}{0.716028in}}%
\pgfpathlineto{\pgfqpoint{2.235282in}{2.921776in}}%
\pgfpathlineto{\pgfqpoint{2.056187in}{2.921776in}}%
\pgfpathlineto{\pgfqpoint{2.056187in}{0.716028in}}%
\pgfpathclose%
\pgfusepath{fill}%
\end{pgfscope}%
\begin{pgfscope}%
\pgfpathrectangle{\pgfqpoint{0.802523in}{0.716028in}}{\pgfqpoint{5.193752in}{3.517843in}}%
\pgfusepath{clip}%
\pgfsetbuttcap%
\pgfsetmiterjoin%
\definecolor{currentfill}{rgb}{0.000000,0.000000,1.000000}%
\pgfsetfillcolor{currentfill}%
\pgfsetlinewidth{0.000000pt}%
\definecolor{currentstroke}{rgb}{0.000000,0.000000,0.000000}%
\pgfsetstrokecolor{currentstroke}%
\pgfsetdash{}{0pt}%
\pgfpathmoveto{\pgfqpoint{2.235282in}{0.716028in}}%
\pgfpathlineto{\pgfqpoint{2.414377in}{0.716028in}}%
\pgfpathlineto{\pgfqpoint{2.414377in}{2.805928in}}%
\pgfpathlineto{\pgfqpoint{2.235282in}{2.805928in}}%
\pgfpathlineto{\pgfqpoint{2.235282in}{0.716028in}}%
\pgfpathclose%
\pgfusepath{fill}%
\end{pgfscope}%
\begin{pgfscope}%
\pgfpathrectangle{\pgfqpoint{0.802523in}{0.716028in}}{\pgfqpoint{5.193752in}{3.517843in}}%
\pgfusepath{clip}%
\pgfsetbuttcap%
\pgfsetmiterjoin%
\definecolor{currentfill}{rgb}{0.000000,0.000000,1.000000}%
\pgfsetfillcolor{currentfill}%
\pgfsetlinewidth{0.000000pt}%
\definecolor{currentstroke}{rgb}{0.000000,0.000000,0.000000}%
\pgfsetstrokecolor{currentstroke}%
\pgfsetdash{}{0pt}%
\pgfpathmoveto{\pgfqpoint{2.414377in}{0.716028in}}%
\pgfpathlineto{\pgfqpoint{2.593472in}{0.716028in}}%
\pgfpathlineto{\pgfqpoint{2.593472in}{2.203518in}}%
\pgfpathlineto{\pgfqpoint{2.414377in}{2.203518in}}%
\pgfpathlineto{\pgfqpoint{2.414377in}{0.716028in}}%
\pgfpathclose%
\pgfusepath{fill}%
\end{pgfscope}%
\begin{pgfscope}%
\pgfpathrectangle{\pgfqpoint{0.802523in}{0.716028in}}{\pgfqpoint{5.193752in}{3.517843in}}%
\pgfusepath{clip}%
\pgfsetbuttcap%
\pgfsetmiterjoin%
\definecolor{currentfill}{rgb}{0.000000,0.000000,1.000000}%
\pgfsetfillcolor{currentfill}%
\pgfsetlinewidth{0.000000pt}%
\definecolor{currentstroke}{rgb}{0.000000,0.000000,0.000000}%
\pgfsetstrokecolor{currentstroke}%
\pgfsetdash{}{0pt}%
\pgfpathmoveto{\pgfqpoint{2.593472in}{0.716028in}}%
\pgfpathlineto{\pgfqpoint{2.772567in}{0.716028in}}%
\pgfpathlineto{\pgfqpoint{2.772567in}{2.120107in}}%
\pgfpathlineto{\pgfqpoint{2.593472in}{2.120107in}}%
\pgfpathlineto{\pgfqpoint{2.593472in}{0.716028in}}%
\pgfpathclose%
\pgfusepath{fill}%
\end{pgfscope}%
\begin{pgfscope}%
\pgfpathrectangle{\pgfqpoint{0.802523in}{0.716028in}}{\pgfqpoint{5.193752in}{3.517843in}}%
\pgfusepath{clip}%
\pgfsetbuttcap%
\pgfsetmiterjoin%
\definecolor{currentfill}{rgb}{0.000000,0.000000,1.000000}%
\pgfsetfillcolor{currentfill}%
\pgfsetlinewidth{0.000000pt}%
\definecolor{currentstroke}{rgb}{0.000000,0.000000,0.000000}%
\pgfsetstrokecolor{currentstroke}%
\pgfsetdash{}{0pt}%
\pgfpathmoveto{\pgfqpoint{2.772567in}{0.716028in}}%
\pgfpathlineto{\pgfqpoint{2.951662in}{0.716028in}}%
\pgfpathlineto{\pgfqpoint{2.951662in}{1.939384in}}%
\pgfpathlineto{\pgfqpoint{2.772567in}{1.939384in}}%
\pgfpathlineto{\pgfqpoint{2.772567in}{0.716028in}}%
\pgfpathclose%
\pgfusepath{fill}%
\end{pgfscope}%
\begin{pgfscope}%
\pgfpathrectangle{\pgfqpoint{0.802523in}{0.716028in}}{\pgfqpoint{5.193752in}{3.517843in}}%
\pgfusepath{clip}%
\pgfsetbuttcap%
\pgfsetmiterjoin%
\definecolor{currentfill}{rgb}{0.000000,0.000000,1.000000}%
\pgfsetfillcolor{currentfill}%
\pgfsetlinewidth{0.000000pt}%
\definecolor{currentstroke}{rgb}{0.000000,0.000000,0.000000}%
\pgfsetstrokecolor{currentstroke}%
\pgfsetdash{}{0pt}%
\pgfpathmoveto{\pgfqpoint{2.951662in}{0.716028in}}%
\pgfpathlineto{\pgfqpoint{3.130756in}{0.716028in}}%
\pgfpathlineto{\pgfqpoint{3.130756in}{1.828170in}}%
\pgfpathlineto{\pgfqpoint{2.951662in}{1.828170in}}%
\pgfpathlineto{\pgfqpoint{2.951662in}{0.716028in}}%
\pgfpathclose%
\pgfusepath{fill}%
\end{pgfscope}%
\begin{pgfscope}%
\pgfpathrectangle{\pgfqpoint{0.802523in}{0.716028in}}{\pgfqpoint{5.193752in}{3.517843in}}%
\pgfusepath{clip}%
\pgfsetbuttcap%
\pgfsetmiterjoin%
\definecolor{currentfill}{rgb}{0.000000,0.000000,1.000000}%
\pgfsetfillcolor{currentfill}%
\pgfsetlinewidth{0.000000pt}%
\definecolor{currentstroke}{rgb}{0.000000,0.000000,0.000000}%
\pgfsetstrokecolor{currentstroke}%
\pgfsetdash{}{0pt}%
\pgfpathmoveto{\pgfqpoint{3.130756in}{0.716028in}}%
\pgfpathlineto{\pgfqpoint{3.309851in}{0.716028in}}%
\pgfpathlineto{\pgfqpoint{3.309851in}{1.832804in}}%
\pgfpathlineto{\pgfqpoint{3.130756in}{1.832804in}}%
\pgfpathlineto{\pgfqpoint{3.130756in}{0.716028in}}%
\pgfpathclose%
\pgfusepath{fill}%
\end{pgfscope}%
\begin{pgfscope}%
\pgfpathrectangle{\pgfqpoint{0.802523in}{0.716028in}}{\pgfqpoint{5.193752in}{3.517843in}}%
\pgfusepath{clip}%
\pgfsetbuttcap%
\pgfsetmiterjoin%
\definecolor{currentfill}{rgb}{0.000000,0.000000,1.000000}%
\pgfsetfillcolor{currentfill}%
\pgfsetlinewidth{0.000000pt}%
\definecolor{currentstroke}{rgb}{0.000000,0.000000,0.000000}%
\pgfsetstrokecolor{currentstroke}%
\pgfsetdash{}{0pt}%
\pgfpathmoveto{\pgfqpoint{3.309851in}{0.716028in}}%
\pgfpathlineto{\pgfqpoint{3.488946in}{0.716028in}}%
\pgfpathlineto{\pgfqpoint{3.488946in}{1.860607in}}%
\pgfpathlineto{\pgfqpoint{3.309851in}{1.860607in}}%
\pgfpathlineto{\pgfqpoint{3.309851in}{0.716028in}}%
\pgfpathclose%
\pgfusepath{fill}%
\end{pgfscope}%
\begin{pgfscope}%
\pgfpathrectangle{\pgfqpoint{0.802523in}{0.716028in}}{\pgfqpoint{5.193752in}{3.517843in}}%
\pgfusepath{clip}%
\pgfsetbuttcap%
\pgfsetmiterjoin%
\definecolor{currentfill}{rgb}{0.000000,0.000000,1.000000}%
\pgfsetfillcolor{currentfill}%
\pgfsetlinewidth{0.000000pt}%
\definecolor{currentstroke}{rgb}{0.000000,0.000000,0.000000}%
\pgfsetstrokecolor{currentstroke}%
\pgfsetdash{}{0pt}%
\pgfpathmoveto{\pgfqpoint{3.488946in}{0.716028in}}%
\pgfpathlineto{\pgfqpoint{3.668041in}{0.716028in}}%
\pgfpathlineto{\pgfqpoint{3.668041in}{1.902313in}}%
\pgfpathlineto{\pgfqpoint{3.488946in}{1.902313in}}%
\pgfpathlineto{\pgfqpoint{3.488946in}{0.716028in}}%
\pgfpathclose%
\pgfusepath{fill}%
\end{pgfscope}%
\begin{pgfscope}%
\pgfpathrectangle{\pgfqpoint{0.802523in}{0.716028in}}{\pgfqpoint{5.193752in}{3.517843in}}%
\pgfusepath{clip}%
\pgfsetbuttcap%
\pgfsetmiterjoin%
\definecolor{currentfill}{rgb}{0.000000,0.000000,1.000000}%
\pgfsetfillcolor{currentfill}%
\pgfsetlinewidth{0.000000pt}%
\definecolor{currentstroke}{rgb}{0.000000,0.000000,0.000000}%
\pgfsetstrokecolor{currentstroke}%
\pgfsetdash{}{0pt}%
\pgfpathmoveto{\pgfqpoint{3.668041in}{0.716028in}}%
\pgfpathlineto{\pgfqpoint{3.847136in}{0.716028in}}%
\pgfpathlineto{\pgfqpoint{3.847136in}{1.976455in}}%
\pgfpathlineto{\pgfqpoint{3.668041in}{1.976455in}}%
\pgfpathlineto{\pgfqpoint{3.668041in}{0.716028in}}%
\pgfpathclose%
\pgfusepath{fill}%
\end{pgfscope}%
\begin{pgfscope}%
\pgfpathrectangle{\pgfqpoint{0.802523in}{0.716028in}}{\pgfqpoint{5.193752in}{3.517843in}}%
\pgfusepath{clip}%
\pgfsetbuttcap%
\pgfsetmiterjoin%
\definecolor{currentfill}{rgb}{0.000000,0.000000,1.000000}%
\pgfsetfillcolor{currentfill}%
\pgfsetlinewidth{0.000000pt}%
\definecolor{currentstroke}{rgb}{0.000000,0.000000,0.000000}%
\pgfsetstrokecolor{currentstroke}%
\pgfsetdash{}{0pt}%
\pgfpathmoveto{\pgfqpoint{3.847136in}{0.716028in}}%
\pgfpathlineto{\pgfqpoint{4.026231in}{0.716028in}}%
\pgfpathlineto{\pgfqpoint{4.026231in}{2.069134in}}%
\pgfpathlineto{\pgfqpoint{3.847136in}{2.069134in}}%
\pgfpathlineto{\pgfqpoint{3.847136in}{0.716028in}}%
\pgfpathclose%
\pgfusepath{fill}%
\end{pgfscope}%
\begin{pgfscope}%
\pgfpathrectangle{\pgfqpoint{0.802523in}{0.716028in}}{\pgfqpoint{5.193752in}{3.517843in}}%
\pgfusepath{clip}%
\pgfsetbuttcap%
\pgfsetmiterjoin%
\definecolor{currentfill}{rgb}{0.000000,0.000000,1.000000}%
\pgfsetfillcolor{currentfill}%
\pgfsetlinewidth{0.000000pt}%
\definecolor{currentstroke}{rgb}{0.000000,0.000000,0.000000}%
\pgfsetstrokecolor{currentstroke}%
\pgfsetdash{}{0pt}%
\pgfpathmoveto{\pgfqpoint{4.026231in}{0.716028in}}%
\pgfpathlineto{\pgfqpoint{4.205326in}{0.716028in}}%
\pgfpathlineto{\pgfqpoint{4.205326in}{1.994991in}}%
\pgfpathlineto{\pgfqpoint{4.026231in}{1.994991in}}%
\pgfpathlineto{\pgfqpoint{4.026231in}{0.716028in}}%
\pgfpathclose%
\pgfusepath{fill}%
\end{pgfscope}%
\begin{pgfscope}%
\pgfpathrectangle{\pgfqpoint{0.802523in}{0.716028in}}{\pgfqpoint{5.193752in}{3.517843in}}%
\pgfusepath{clip}%
\pgfsetbuttcap%
\pgfsetmiterjoin%
\definecolor{currentfill}{rgb}{0.000000,0.000000,1.000000}%
\pgfsetfillcolor{currentfill}%
\pgfsetlinewidth{0.000000pt}%
\definecolor{currentstroke}{rgb}{0.000000,0.000000,0.000000}%
\pgfsetstrokecolor{currentstroke}%
\pgfsetdash{}{0pt}%
\pgfpathmoveto{\pgfqpoint{4.205326in}{0.716028in}}%
\pgfpathlineto{\pgfqpoint{4.384421in}{0.716028in}}%
\pgfpathlineto{\pgfqpoint{4.384421in}{2.101571in}}%
\pgfpathlineto{\pgfqpoint{4.205326in}{2.101571in}}%
\pgfpathlineto{\pgfqpoint{4.205326in}{0.716028in}}%
\pgfpathclose%
\pgfusepath{fill}%
\end{pgfscope}%
\begin{pgfscope}%
\pgfpathrectangle{\pgfqpoint{0.802523in}{0.716028in}}{\pgfqpoint{5.193752in}{3.517843in}}%
\pgfusepath{clip}%
\pgfsetbuttcap%
\pgfsetmiterjoin%
\definecolor{currentfill}{rgb}{0.000000,0.000000,1.000000}%
\pgfsetfillcolor{currentfill}%
\pgfsetlinewidth{0.000000pt}%
\definecolor{currentstroke}{rgb}{0.000000,0.000000,0.000000}%
\pgfsetstrokecolor{currentstroke}%
\pgfsetdash{}{0pt}%
\pgfpathmoveto{\pgfqpoint{4.384421in}{0.716028in}}%
\pgfpathlineto{\pgfqpoint{4.563516in}{0.716028in}}%
\pgfpathlineto{\pgfqpoint{4.563516in}{1.740125in}}%
\pgfpathlineto{\pgfqpoint{4.384421in}{1.740125in}}%
\pgfpathlineto{\pgfqpoint{4.384421in}{0.716028in}}%
\pgfpathclose%
\pgfusepath{fill}%
\end{pgfscope}%
\begin{pgfscope}%
\pgfpathrectangle{\pgfqpoint{0.802523in}{0.716028in}}{\pgfqpoint{5.193752in}{3.517843in}}%
\pgfusepath{clip}%
\pgfsetbuttcap%
\pgfsetmiterjoin%
\definecolor{currentfill}{rgb}{0.000000,0.000000,1.000000}%
\pgfsetfillcolor{currentfill}%
\pgfsetlinewidth{0.000000pt}%
\definecolor{currentstroke}{rgb}{0.000000,0.000000,0.000000}%
\pgfsetstrokecolor{currentstroke}%
\pgfsetdash{}{0pt}%
\pgfpathmoveto{\pgfqpoint{4.563516in}{0.716028in}}%
\pgfpathlineto{\pgfqpoint{4.742611in}{0.716028in}}%
\pgfpathlineto{\pgfqpoint{4.742611in}{1.475992in}}%
\pgfpathlineto{\pgfqpoint{4.563516in}{1.475992in}}%
\pgfpathlineto{\pgfqpoint{4.563516in}{0.716028in}}%
\pgfpathclose%
\pgfusepath{fill}%
\end{pgfscope}%
\begin{pgfscope}%
\pgfpathrectangle{\pgfqpoint{0.802523in}{0.716028in}}{\pgfqpoint{5.193752in}{3.517843in}}%
\pgfusepath{clip}%
\pgfsetbuttcap%
\pgfsetmiterjoin%
\definecolor{currentfill}{rgb}{0.000000,0.000000,1.000000}%
\pgfsetfillcolor{currentfill}%
\pgfsetlinewidth{0.000000pt}%
\definecolor{currentstroke}{rgb}{0.000000,0.000000,0.000000}%
\pgfsetstrokecolor{currentstroke}%
\pgfsetdash{}{0pt}%
\pgfpathmoveto{\pgfqpoint{4.742611in}{0.716028in}}%
\pgfpathlineto{\pgfqpoint{4.921705in}{0.716028in}}%
\pgfpathlineto{\pgfqpoint{4.921705in}{1.596474in}}%
\pgfpathlineto{\pgfqpoint{4.742611in}{1.596474in}}%
\pgfpathlineto{\pgfqpoint{4.742611in}{0.716028in}}%
\pgfpathclose%
\pgfusepath{fill}%
\end{pgfscope}%
\begin{pgfscope}%
\pgfpathrectangle{\pgfqpoint{0.802523in}{0.716028in}}{\pgfqpoint{5.193752in}{3.517843in}}%
\pgfusepath{clip}%
\pgfsetbuttcap%
\pgfsetmiterjoin%
\definecolor{currentfill}{rgb}{0.000000,0.000000,1.000000}%
\pgfsetfillcolor{currentfill}%
\pgfsetlinewidth{0.000000pt}%
\definecolor{currentstroke}{rgb}{0.000000,0.000000,0.000000}%
\pgfsetstrokecolor{currentstroke}%
\pgfsetdash{}{0pt}%
\pgfpathmoveto{\pgfqpoint{4.921705in}{0.716028in}}%
\pgfpathlineto{\pgfqpoint{5.100800in}{0.716028in}}%
\pgfpathlineto{\pgfqpoint{5.100800in}{1.652081in}}%
\pgfpathlineto{\pgfqpoint{4.921705in}{1.652081in}}%
\pgfpathlineto{\pgfqpoint{4.921705in}{0.716028in}}%
\pgfpathclose%
\pgfusepath{fill}%
\end{pgfscope}%
\begin{pgfscope}%
\pgfpathrectangle{\pgfqpoint{0.802523in}{0.716028in}}{\pgfqpoint{5.193752in}{3.517843in}}%
\pgfusepath{clip}%
\pgfsetbuttcap%
\pgfsetmiterjoin%
\definecolor{currentfill}{rgb}{0.000000,0.000000,1.000000}%
\pgfsetfillcolor{currentfill}%
\pgfsetlinewidth{0.000000pt}%
\definecolor{currentstroke}{rgb}{0.000000,0.000000,0.000000}%
\pgfsetstrokecolor{currentstroke}%
\pgfsetdash{}{0pt}%
\pgfpathmoveto{\pgfqpoint{5.100800in}{0.716028in}}%
\pgfpathlineto{\pgfqpoint{5.279895in}{0.716028in}}%
\pgfpathlineto{\pgfqpoint{5.279895in}{2.041330in}}%
\pgfpathlineto{\pgfqpoint{5.100800in}{2.041330in}}%
\pgfpathlineto{\pgfqpoint{5.100800in}{0.716028in}}%
\pgfpathclose%
\pgfusepath{fill}%
\end{pgfscope}%
\begin{pgfscope}%
\pgfpathrectangle{\pgfqpoint{0.802523in}{0.716028in}}{\pgfqpoint{5.193752in}{3.517843in}}%
\pgfusepath{clip}%
\pgfsetbuttcap%
\pgfsetmiterjoin%
\definecolor{currentfill}{rgb}{0.000000,0.000000,1.000000}%
\pgfsetfillcolor{currentfill}%
\pgfsetlinewidth{0.000000pt}%
\definecolor{currentstroke}{rgb}{0.000000,0.000000,0.000000}%
\pgfsetstrokecolor{currentstroke}%
\pgfsetdash{}{0pt}%
\pgfpathmoveto{\pgfqpoint{5.279895in}{0.716028in}}%
\pgfpathlineto{\pgfqpoint{5.458990in}{0.716028in}}%
\pgfpathlineto{\pgfqpoint{5.458990in}{2.370339in}}%
\pgfpathlineto{\pgfqpoint{5.279895in}{2.370339in}}%
\pgfpathlineto{\pgfqpoint{5.279895in}{0.716028in}}%
\pgfpathclose%
\pgfusepath{fill}%
\end{pgfscope}%
\begin{pgfscope}%
\pgfpathrectangle{\pgfqpoint{0.802523in}{0.716028in}}{\pgfqpoint{5.193752in}{3.517843in}}%
\pgfusepath{clip}%
\pgfsetbuttcap%
\pgfsetmiterjoin%
\definecolor{currentfill}{rgb}{0.000000,0.000000,1.000000}%
\pgfsetfillcolor{currentfill}%
\pgfsetlinewidth{0.000000pt}%
\definecolor{currentstroke}{rgb}{0.000000,0.000000,0.000000}%
\pgfsetstrokecolor{currentstroke}%
\pgfsetdash{}{0pt}%
\pgfpathmoveto{\pgfqpoint{5.458990in}{0.716028in}}%
\pgfpathlineto{\pgfqpoint{5.638085in}{0.716028in}}%
\pgfpathlineto{\pgfqpoint{5.638085in}{2.249857in}}%
\pgfpathlineto{\pgfqpoint{5.458990in}{2.249857in}}%
\pgfpathlineto{\pgfqpoint{5.458990in}{0.716028in}}%
\pgfpathclose%
\pgfusepath{fill}%
\end{pgfscope}%
\begin{pgfscope}%
\pgfpathrectangle{\pgfqpoint{0.802523in}{0.716028in}}{\pgfqpoint{5.193752in}{3.517843in}}%
\pgfusepath{clip}%
\pgfsetbuttcap%
\pgfsetmiterjoin%
\definecolor{currentfill}{rgb}{0.000000,0.000000,1.000000}%
\pgfsetfillcolor{currentfill}%
\pgfsetlinewidth{0.000000pt}%
\definecolor{currentstroke}{rgb}{0.000000,0.000000,0.000000}%
\pgfsetstrokecolor{currentstroke}%
\pgfsetdash{}{0pt}%
\pgfpathmoveto{\pgfqpoint{5.638085in}{0.716028in}}%
\pgfpathlineto{\pgfqpoint{5.817180in}{0.716028in}}%
\pgfpathlineto{\pgfqpoint{5.817180in}{2.143277in}}%
\pgfpathlineto{\pgfqpoint{5.638085in}{2.143277in}}%
\pgfpathlineto{\pgfqpoint{5.638085in}{0.716028in}}%
\pgfpathclose%
\pgfusepath{fill}%
\end{pgfscope}%
\begin{pgfscope}%
\pgfpathrectangle{\pgfqpoint{0.802523in}{0.716028in}}{\pgfqpoint{5.193752in}{3.517843in}}%
\pgfusepath{clip}%
\pgfsetbuttcap%
\pgfsetmiterjoin%
\definecolor{currentfill}{rgb}{0.000000,0.000000,1.000000}%
\pgfsetfillcolor{currentfill}%
\pgfsetlinewidth{0.000000pt}%
\definecolor{currentstroke}{rgb}{0.000000,0.000000,0.000000}%
\pgfsetstrokecolor{currentstroke}%
\pgfsetdash{}{0pt}%
\pgfpathmoveto{\pgfqpoint{5.817180in}{0.716028in}}%
\pgfpathlineto{\pgfqpoint{5.996275in}{0.716028in}}%
\pgfpathlineto{\pgfqpoint{5.996275in}{2.643740in}}%
\pgfpathlineto{\pgfqpoint{5.817180in}{2.643740in}}%
\pgfpathlineto{\pgfqpoint{5.817180in}{0.716028in}}%
\pgfpathclose%
\pgfusepath{fill}%
\end{pgfscope}%
\begin{pgfscope}%
\pgfsetbuttcap%
\pgfsetroundjoin%
\definecolor{currentfill}{rgb}{0.000000,0.000000,0.000000}%
\pgfsetfillcolor{currentfill}%
\pgfsetlinewidth{0.200750pt}%
\definecolor{currentstroke}{rgb}{0.000000,0.000000,0.000000}%
\pgfsetstrokecolor{currentstroke}%
\pgfsetdash{}{0pt}%
\pgfsys@defobject{currentmarker}{\pgfqpoint{0.000000in}{-0.048611in}}{\pgfqpoint{0.000000in}{0.000000in}}{%
\pgfpathmoveto{\pgfqpoint{0.000000in}{0.000000in}}%
\pgfpathlineto{\pgfqpoint{0.000000in}{-0.048611in}}%
\pgfusepath{stroke,fill}%
}%
\begin{pgfscope}%
\pgfsys@transformshift{0.802523in}{0.716028in}%
\pgfsys@useobject{currentmarker}{}%
\end{pgfscope}%
\end{pgfscope}%
\begin{pgfscope}%
\definecolor{textcolor}{rgb}{0.000000,0.000000,0.000000}%
\pgfsetstrokecolor{textcolor}%
\pgfsetfillcolor{textcolor}%
\pgftext[x=0.802523in,y=0.618806in,,top]{\color{textcolor}{\rmfamily\fontsize{20.000000}{24.000000}\selectfont\catcode`\^=\active\def^{\ifmmode\sp\else\^{}\fi}\catcode`\%=\active\def
\end{pgfscope}%
\begin{pgfscope}%
\pgfsetbuttcap%
\pgfsetroundjoin%
\definecolor{currentfill}{rgb}{0.000000,0.000000,0.000000}%
\pgfsetfillcolor{currentfill}%
\pgfsetlinewidth{0.200750pt}%
\definecolor{currentstroke}{rgb}{0.000000,0.000000,0.000000}%
\pgfsetstrokecolor{currentstroke}%
\pgfsetdash{}{0pt}%
\pgfsys@defobject{currentmarker}{\pgfqpoint{0.000000in}{-0.048611in}}{\pgfqpoint{0.000000in}{0.000000in}}{%
\pgfpathmoveto{\pgfqpoint{0.000000in}{0.000000in}}%
\pgfpathlineto{\pgfqpoint{0.000000in}{-0.048611in}}%
\pgfusepath{stroke,fill}%
}%
\begin{pgfscope}%
\pgfsys@transformshift{1.841273in}{0.716028in}%
\pgfsys@useobject{currentmarker}{}%
\end{pgfscope}%
\end{pgfscope}%
\begin{pgfscope}%
\definecolor{textcolor}{rgb}{0.000000,0.000000,0.000000}%
\pgfsetstrokecolor{textcolor}%
\pgfsetfillcolor{textcolor}%
\pgftext[x=1.841273in,y=0.618806in,,top]{\color{textcolor}{\rmfamily\fontsize{20.000000}{24.000000}\selectfont\catcode`\^=\active\def^{\ifmmode\sp\else\^{}\fi}\catcode`\%=\active\def
\end{pgfscope}%
\begin{pgfscope}%
\pgfsetbuttcap%
\pgfsetroundjoin%
\definecolor{currentfill}{rgb}{0.000000,0.000000,0.000000}%
\pgfsetfillcolor{currentfill}%
\pgfsetlinewidth{0.200750pt}%
\definecolor{currentstroke}{rgb}{0.000000,0.000000,0.000000}%
\pgfsetstrokecolor{currentstroke}%
\pgfsetdash{}{0pt}%
\pgfsys@defobject{currentmarker}{\pgfqpoint{0.000000in}{-0.048611in}}{\pgfqpoint{0.000000in}{0.000000in}}{%
\pgfpathmoveto{\pgfqpoint{0.000000in}{0.000000in}}%
\pgfpathlineto{\pgfqpoint{0.000000in}{-0.048611in}}%
\pgfusepath{stroke,fill}%
}%
\begin{pgfscope}%
\pgfsys@transformshift{2.880024in}{0.716028in}%
\pgfsys@useobject{currentmarker}{}%
\end{pgfscope}%
\end{pgfscope}%
\begin{pgfscope}%
\definecolor{textcolor}{rgb}{0.000000,0.000000,0.000000}%
\pgfsetstrokecolor{textcolor}%
\pgfsetfillcolor{textcolor}%
\pgftext[x=2.880024in,y=0.618806in,,top]{\color{textcolor}{\rmfamily\fontsize{20.000000}{24.000000}\selectfont\catcode`\^=\active\def^{\ifmmode\sp\else\^{}\fi}\catcode`\%=\active\def
\end{pgfscope}%
\begin{pgfscope}%
\pgfsetbuttcap%
\pgfsetroundjoin%
\definecolor{currentfill}{rgb}{0.000000,0.000000,0.000000}%
\pgfsetfillcolor{currentfill}%
\pgfsetlinewidth{0.200750pt}%
\definecolor{currentstroke}{rgb}{0.000000,0.000000,0.000000}%
\pgfsetstrokecolor{currentstroke}%
\pgfsetdash{}{0pt}%
\pgfsys@defobject{currentmarker}{\pgfqpoint{0.000000in}{-0.048611in}}{\pgfqpoint{0.000000in}{0.000000in}}{%
\pgfpathmoveto{\pgfqpoint{0.000000in}{0.000000in}}%
\pgfpathlineto{\pgfqpoint{0.000000in}{-0.048611in}}%
\pgfusepath{stroke,fill}%
}%
\begin{pgfscope}%
\pgfsys@transformshift{3.918774in}{0.716028in}%
\pgfsys@useobject{currentmarker}{}%
\end{pgfscope}%
\end{pgfscope}%
\begin{pgfscope}%
\definecolor{textcolor}{rgb}{0.000000,0.000000,0.000000}%
\pgfsetstrokecolor{textcolor}%
\pgfsetfillcolor{textcolor}%
\pgftext[x=3.918774in,y=0.618806in,,top]{\color{textcolor}{\rmfamily\fontsize{20.000000}{24.000000}\selectfont\catcode`\^=\active\def^{\ifmmode\sp\else\^{}\fi}\catcode`\%=\active\def
\end{pgfscope}%
\begin{pgfscope}%
\pgfsetbuttcap%
\pgfsetroundjoin%
\definecolor{currentfill}{rgb}{0.000000,0.000000,0.000000}%
\pgfsetfillcolor{currentfill}%
\pgfsetlinewidth{0.200750pt}%
\definecolor{currentstroke}{rgb}{0.000000,0.000000,0.000000}%
\pgfsetstrokecolor{currentstroke}%
\pgfsetdash{}{0pt}%
\pgfsys@defobject{currentmarker}{\pgfqpoint{0.000000in}{-0.048611in}}{\pgfqpoint{0.000000in}{0.000000in}}{%
\pgfpathmoveto{\pgfqpoint{0.000000in}{0.000000in}}%
\pgfpathlineto{\pgfqpoint{0.000000in}{-0.048611in}}%
\pgfusepath{stroke,fill}%
}%
\begin{pgfscope}%
\pgfsys@transformshift{4.957524in}{0.716028in}%
\pgfsys@useobject{currentmarker}{}%
\end{pgfscope}%
\end{pgfscope}%
\begin{pgfscope}%
\definecolor{textcolor}{rgb}{0.000000,0.000000,0.000000}%
\pgfsetstrokecolor{textcolor}%
\pgfsetfillcolor{textcolor}%
\pgftext[x=4.957524in,y=0.618806in,,top]{\color{textcolor}{\rmfamily\fontsize{20.000000}{24.000000}\selectfont\catcode`\^=\active\def^{\ifmmode\sp\else\^{}\fi}\catcode`\%=\active\def
\end{pgfscope}%
\begin{pgfscope}%
\pgfsetbuttcap%
\pgfsetroundjoin%
\definecolor{currentfill}{rgb}{0.000000,0.000000,0.000000}%
\pgfsetfillcolor{currentfill}%
\pgfsetlinewidth{0.200750pt}%
\definecolor{currentstroke}{rgb}{0.000000,0.000000,0.000000}%
\pgfsetstrokecolor{currentstroke}%
\pgfsetdash{}{0pt}%
\pgfsys@defobject{currentmarker}{\pgfqpoint{0.000000in}{-0.048611in}}{\pgfqpoint{0.000000in}{0.000000in}}{%
\pgfpathmoveto{\pgfqpoint{0.000000in}{0.000000in}}%
\pgfpathlineto{\pgfqpoint{0.000000in}{-0.048611in}}%
\pgfusepath{stroke,fill}%
}%
\begin{pgfscope}%
\pgfsys@transformshift{5.996275in}{0.716028in}%
\pgfsys@useobject{currentmarker}{}%
\end{pgfscope}%
\end{pgfscope}%
\begin{pgfscope}%
\definecolor{textcolor}{rgb}{0.000000,0.000000,0.000000}%
\pgfsetstrokecolor{textcolor}%
\pgfsetfillcolor{textcolor}%
\pgftext[x=5.996275in,y=0.618806in,,top]{\color{textcolor}{\rmfamily\fontsize{20.000000}{24.000000}\selectfont\catcode`\^=\active\def^{\ifmmode\sp\else\^{}\fi}\catcode`\%=\active\def
\end{pgfscope}%
\begin{pgfscope}%
\definecolor{textcolor}{rgb}{0.000000,0.000000,0.000000}%
\pgfsetstrokecolor{textcolor}%
\pgfsetfillcolor{textcolor}%
\pgftext[x=3.399399in,y=0.307183in,,top]{\color{textcolor}{\rmfamily\fontsize{26.000000}{31.200000}\selectfont\catcode`\^=\active\def^{\ifmmode\sp\else\^{}\fi}\catcode`\%=\active\def
\end{pgfscope}%
\begin{pgfscope}%
\pgfsetbuttcap%
\pgfsetroundjoin%
\definecolor{currentfill}{rgb}{0.000000,0.000000,0.000000}%
\pgfsetfillcolor{currentfill}%
\pgfsetlinewidth{0.200750pt}%
\definecolor{currentstroke}{rgb}{0.000000,0.000000,0.000000}%
\pgfsetstrokecolor{currentstroke}%
\pgfsetdash{}{0pt}%
\pgfsys@defobject{currentmarker}{\pgfqpoint{-0.048611in}{0.000000in}}{\pgfqpoint{-0.000000in}{0.000000in}}{%
\pgfpathmoveto{\pgfqpoint{-0.000000in}{0.000000in}}%
\pgfpathlineto{\pgfqpoint{-0.048611in}{0.000000in}}%
\pgfusepath{stroke,fill}%
}%
\begin{pgfscope}%
\pgfsys@transformshift{0.802523in}{0.716028in}%
\pgfsys@useobject{currentmarker}{}%
\end{pgfscope}%
\end{pgfscope}%
\begin{pgfscope}%
\definecolor{textcolor}{rgb}{0.000000,0.000000,0.000000}%
\pgfsetstrokecolor{textcolor}%
\pgfsetfillcolor{textcolor}%
\pgftext[x=0.362738in, y=0.616009in, left, base]{\color{textcolor}{\rmfamily\fontsize{20.000000}{24.000000}\selectfont\catcode`\^=\active\def^{\ifmmode\sp\else\^{}\fi}\catcode`\%=\active\def
\end{pgfscope}%
\begin{pgfscope}%
\pgfsetbuttcap%
\pgfsetroundjoin%
\definecolor{currentfill}{rgb}{0.000000,0.000000,0.000000}%
\pgfsetfillcolor{currentfill}%
\pgfsetlinewidth{0.200750pt}%
\definecolor{currentstroke}{rgb}{0.000000,0.000000,0.000000}%
\pgfsetstrokecolor{currentstroke}%
\pgfsetdash{}{0pt}%
\pgfsys@defobject{currentmarker}{\pgfqpoint{-0.048611in}{0.000000in}}{\pgfqpoint{-0.000000in}{0.000000in}}{%
\pgfpathmoveto{\pgfqpoint{-0.000000in}{0.000000in}}%
\pgfpathlineto{\pgfqpoint{-0.048611in}{0.000000in}}%
\pgfusepath{stroke,fill}%
}%
\begin{pgfscope}%
\pgfsys@transformshift{0.802523in}{1.480146in}%
\pgfsys@useobject{currentmarker}{}%
\end{pgfscope}%
\end{pgfscope}%
\begin{pgfscope}%
\definecolor{textcolor}{rgb}{0.000000,0.000000,0.000000}%
\pgfsetstrokecolor{textcolor}%
\pgfsetfillcolor{textcolor}%
\pgftext[x=0.362738in, y=1.380127in, left, base]{\color{textcolor}{\rmfamily\fontsize{20.000000}{24.000000}\selectfont\catcode`\^=\active\def^{\ifmmode\sp\else\^{}\fi}\catcode`\%=\active\def
\end{pgfscope}%
\begin{pgfscope}%
\pgfsetbuttcap%
\pgfsetroundjoin%
\definecolor{currentfill}{rgb}{0.000000,0.000000,0.000000}%
\pgfsetfillcolor{currentfill}%
\pgfsetlinewidth{0.200750pt}%
\definecolor{currentstroke}{rgb}{0.000000,0.000000,0.000000}%
\pgfsetstrokecolor{currentstroke}%
\pgfsetdash{}{0pt}%
\pgfsys@defobject{currentmarker}{\pgfqpoint{-0.048611in}{0.000000in}}{\pgfqpoint{-0.000000in}{0.000000in}}{%
\pgfpathmoveto{\pgfqpoint{-0.000000in}{0.000000in}}%
\pgfpathlineto{\pgfqpoint{-0.048611in}{0.000000in}}%
\pgfusepath{stroke,fill}%
}%
\begin{pgfscope}%
\pgfsys@transformshift{0.802523in}{2.244264in}%
\pgfsys@useobject{currentmarker}{}%
\end{pgfscope}%
\end{pgfscope}%
\begin{pgfscope}%
\definecolor{textcolor}{rgb}{0.000000,0.000000,0.000000}%
\pgfsetstrokecolor{textcolor}%
\pgfsetfillcolor{textcolor}%
\pgftext[x=0.362738in, y=2.144245in, left, base]{\color{textcolor}{\rmfamily\fontsize{20.000000}{24.000000}\selectfont\catcode`\^=\active\def^{\ifmmode\sp\else\^{}\fi}\catcode`\%=\active\def
\end{pgfscope}%
\begin{pgfscope}%
\pgfsetbuttcap%
\pgfsetroundjoin%
\definecolor{currentfill}{rgb}{0.000000,0.000000,0.000000}%
\pgfsetfillcolor{currentfill}%
\pgfsetlinewidth{0.200750pt}%
\definecolor{currentstroke}{rgb}{0.000000,0.000000,0.000000}%
\pgfsetstrokecolor{currentstroke}%
\pgfsetdash{}{0pt}%
\pgfsys@defobject{currentmarker}{\pgfqpoint{-0.048611in}{0.000000in}}{\pgfqpoint{-0.000000in}{0.000000in}}{%
\pgfpathmoveto{\pgfqpoint{-0.000000in}{0.000000in}}%
\pgfpathlineto{\pgfqpoint{-0.048611in}{0.000000in}}%
\pgfusepath{stroke,fill}%
}%
\begin{pgfscope}%
\pgfsys@transformshift{0.802523in}{3.008382in}%
\pgfsys@useobject{currentmarker}{}%
\end{pgfscope}%
\end{pgfscope}%
\begin{pgfscope}%
\definecolor{textcolor}{rgb}{0.000000,0.000000,0.000000}%
\pgfsetstrokecolor{textcolor}%
\pgfsetfillcolor{textcolor}%
\pgftext[x=0.362738in, y=2.908363in, left, base]{\color{textcolor}{\rmfamily\fontsize{20.000000}{24.000000}\selectfont\catcode`\^=\active\def^{\ifmmode\sp\else\^{}\fi}\catcode`\%=\active\def
\end{pgfscope}%
\begin{pgfscope}%
\pgfsetbuttcap%
\pgfsetroundjoin%
\definecolor{currentfill}{rgb}{0.000000,0.000000,0.000000}%
\pgfsetfillcolor{currentfill}%
\pgfsetlinewidth{0.200750pt}%
\definecolor{currentstroke}{rgb}{0.000000,0.000000,0.000000}%
\pgfsetstrokecolor{currentstroke}%
\pgfsetdash{}{0pt}%
\pgfsys@defobject{currentmarker}{\pgfqpoint{-0.048611in}{0.000000in}}{\pgfqpoint{-0.000000in}{0.000000in}}{%
\pgfpathmoveto{\pgfqpoint{-0.000000in}{0.000000in}}%
\pgfpathlineto{\pgfqpoint{-0.048611in}{0.000000in}}%
\pgfusepath{stroke,fill}%
}%
\begin{pgfscope}%
\pgfsys@transformshift{0.802523in}{3.772500in}%
\pgfsys@useobject{currentmarker}{}%
\end{pgfscope}%
\end{pgfscope}%
\begin{pgfscope}%
\definecolor{textcolor}{rgb}{0.000000,0.000000,0.000000}%
\pgfsetstrokecolor{textcolor}%
\pgfsetfillcolor{textcolor}%
\pgftext[x=0.362738in, y=3.672481in, left, base]{\color{textcolor}{\rmfamily\fontsize{20.000000}{24.000000}\selectfont\catcode`\^=\active\def^{\ifmmode\sp\else\^{}\fi}\catcode`\%=\active\def
\end{pgfscope}%
\begin{pgfscope}%
\definecolor{textcolor}{rgb}{0.000000,0.000000,0.000000}%
\pgfsetstrokecolor{textcolor}%
\pgfsetfillcolor{textcolor}%
\pgftext[x=0.307183in,y=2.474950in,,bottom,rotate=90.000000]{\color{textcolor}{\rmfamily\fontsize{26.000000}{31.200000}\selectfont\catcode`\^=\active\def^{\ifmmode\sp\else\^{}\fi}\catcode`\%=\active\def
\end{pgfscope}%
\begin{pgfscope}%
\pgfpathrectangle{\pgfqpoint{0.802523in}{0.716028in}}{\pgfqpoint{5.193752in}{3.517843in}}%
\pgfusepath{clip}%
\pgfsetbuttcap%
\pgfsetroundjoin%
\pgfsetlinewidth{2.007500pt}%
\definecolor{currentstroke}{rgb}{0.501961,0.501961,0.501961}%
\pgfsetstrokecolor{currentstroke}%
\pgfsetdash{{7.400000pt}{3.200000pt}}{0.000000pt}%
\pgfpathmoveto{\pgfqpoint{0.802523in}{2.244264in}}%
\pgfpathlineto{\pgfqpoint{5.996275in}{2.244264in}}%
\pgfusepath{stroke}%
\end{pgfscope}%
\begin{pgfscope}%
\pgfsetrectcap%
\pgfsetmiterjoin%
\pgfsetlinewidth{0.803000pt}%
\definecolor{currentstroke}{rgb}{0.000000,0.000000,0.000000}%
\pgfsetstrokecolor{currentstroke}%
\pgfsetdash{}{0pt}%
\pgfpathmoveto{\pgfqpoint{0.802523in}{0.716028in}}%
\pgfpathlineto{\pgfqpoint{0.802523in}{4.233871in}}%
\pgfusepath{stroke}%
\end{pgfscope}%
\begin{pgfscope}%
\pgfsetrectcap%
\pgfsetmiterjoin%
\pgfsetlinewidth{0.803000pt}%
\definecolor{currentstroke}{rgb}{0.000000,0.000000,0.000000}%
\pgfsetstrokecolor{currentstroke}%
\pgfsetdash{}{0pt}%
\pgfpathmoveto{\pgfqpoint{5.996275in}{0.716028in}}%
\pgfpathlineto{\pgfqpoint{5.996275in}{4.233871in}}%
\pgfusepath{stroke}%
\end{pgfscope}%
\begin{pgfscope}%
\pgfsetrectcap%
\pgfsetmiterjoin%
\pgfsetlinewidth{0.803000pt}%
\definecolor{currentstroke}{rgb}{0.000000,0.000000,0.000000}%
\pgfsetstrokecolor{currentstroke}%
\pgfsetdash{}{0pt}%
\pgfpathmoveto{\pgfqpoint{0.802523in}{0.716028in}}%
\pgfpathlineto{\pgfqpoint{5.996275in}{0.716028in}}%
\pgfusepath{stroke}%
\end{pgfscope}%
\begin{pgfscope}%
\pgfsetrectcap%
\pgfsetmiterjoin%
\pgfsetlinewidth{0.803000pt}%
\definecolor{currentstroke}{rgb}{0.000000,0.000000,0.000000}%
\pgfsetstrokecolor{currentstroke}%
\pgfsetdash{}{0pt}%
\pgfpathmoveto{\pgfqpoint{0.802523in}{4.233871in}}%
\pgfpathlineto{\pgfqpoint{5.996275in}{4.233871in}}%
\pgfusepath{stroke}%
\end{pgfscope}%
\begin{pgfscope}%
\definecolor{textcolor}{rgb}{0.000000,0.000000,0.000000}%
\pgfsetstrokecolor{textcolor}%
\pgfsetfillcolor{textcolor}%
\pgftext[x=3.399399in,y=4.317204in,,base]{\color{textcolor}{\rmfamily\fontsize{25.000000}{30.000000}\selectfont\catcode`\^=\active\def^{\ifmmode\sp\else\^{}\fi}\catcode`\%=\active\def
\end{pgfscope}%
\begin{pgfscope}%
\pgfsetbuttcap%
\pgfsetmiterjoin%
\definecolor{currentfill}{rgb}{1.000000,1.000000,1.000000}%
\pgfsetfillcolor{currentfill}%
\pgfsetfillopacity{0.800000}%
\pgfsetlinewidth{1.003750pt}%
\definecolor{currentstroke}{rgb}{0.800000,0.800000,0.800000}%
\pgfsetstrokecolor{currentstroke}%
\pgfsetstrokeopacity{0.800000}%
\pgfsetdash{}{0pt}%
\pgfpathmoveto{\pgfqpoint{4.259928in}{3.616692in}}%
\pgfpathlineto{\pgfqpoint{5.801830in}{3.616692in}}%
\pgfpathquadraticcurveto{\pgfqpoint{5.857386in}{3.616692in}}{\pgfqpoint{5.857386in}{3.672248in}}%
\pgfpathlineto{\pgfqpoint{5.857386in}{4.039427in}}%
\pgfpathquadraticcurveto{\pgfqpoint{5.857386in}{4.094982in}}{\pgfqpoint{5.801830in}{4.094982in}}%
\pgfpathlineto{\pgfqpoint{4.259928in}{4.094982in}}%
\pgfpathquadraticcurveto{\pgfqpoint{4.204373in}{4.094982in}}{\pgfqpoint{4.204373in}{4.039427in}}%
\pgfpathlineto{\pgfqpoint{4.204373in}{3.672248in}}%
\pgfpathquadraticcurveto{\pgfqpoint{4.204373in}{3.616692in}}{\pgfqpoint{4.259928in}{3.616692in}}%
\pgfpathlineto{\pgfqpoint{4.259928in}{3.616692in}}%
\pgfpathclose%
\pgfusepath{stroke,fill}%
\end{pgfscope}%
\begin{pgfscope}%
\pgfsetbuttcap%
\pgfsetroundjoin%
\pgfsetlinewidth{2.007500pt}%
\definecolor{currentstroke}{rgb}{0.501961,0.501961,0.501961}%
\pgfsetstrokecolor{currentstroke}%
\pgfsetdash{{7.400000pt}{3.200000pt}}{0.000000pt}%
\pgfpathmoveto{\pgfqpoint{4.315484in}{3.881055in}}%
\pgfpathlineto{\pgfqpoint{4.593262in}{3.881055in}}%
\pgfpathlineto{\pgfqpoint{4.871039in}{3.881055in}}%
\pgfusepath{stroke}%
\end{pgfscope}%
\begin{pgfscope}%
\definecolor{textcolor}{rgb}{0.000000,0.000000,0.000000}%
\pgfsetstrokecolor{textcolor}%
\pgfsetfillcolor{textcolor}%
\pgftext[x=5.093262in,y=3.783833in,left,base]{\color{textcolor}{\rmfamily\fontsize{20.000000}{24.000000}\selectfont\catcode`\^=\active\def^{\ifmmode\sp\else\^{}\fi}\catcode`\%=\active\def
\end{pgfscope}%
\end{pgfpicture}%
\makeatother%
\endgroup%

%% file: figures/evaluation/Repair/Repair_PIT_remaining_time_norm_4layer.pgf
\begingroup%
\makeatletter%
\begin{pgfpicture}%
\pgfpathrectangle{\pgfpointorigin}{\pgfqpoint{6.167556in}{4.557174in}}%
\pgfusepath{use as bounding box, clip}%
\begin{pgfscope}%
\pgfsetbuttcap%
\pgfsetmiterjoin%
\definecolor{currentfill}{rgb}{1.000000,1.000000,1.000000}%
\pgfsetfillcolor{currentfill}%
\pgfsetlinewidth{0.000000pt}%
\definecolor{currentstroke}{rgb}{1.000000,1.000000,1.000000}%
\pgfsetstrokecolor{currentstroke}%
\pgfsetdash{}{0pt}%
\pgfpathmoveto{\pgfqpoint{0.000000in}{0.000000in}}%
\pgfpathlineto{\pgfqpoint{6.167556in}{0.000000in}}%
\pgfpathlineto{\pgfqpoint{6.167556in}{4.557174in}}%
\pgfpathlineto{\pgfqpoint{0.000000in}{4.557174in}}%
\pgfpathlineto{\pgfqpoint{0.000000in}{0.000000in}}%
\pgfpathclose%
\pgfusepath{fill}%
\end{pgfscope}%
\begin{pgfscope}%
\pgfsetbuttcap%
\pgfsetmiterjoin%
\definecolor{currentfill}{rgb}{1.000000,1.000000,1.000000}%
\pgfsetfillcolor{currentfill}%
\pgfsetlinewidth{0.000000pt}%
\definecolor{currentstroke}{rgb}{0.000000,0.000000,0.000000}%
\pgfsetstrokecolor{currentstroke}%
\pgfsetstrokeopacity{0.000000}%
\pgfsetdash{}{0pt}%
\pgfpathmoveto{\pgfqpoint{0.802523in}{0.716028in}}%
\pgfpathlineto{\pgfqpoint{5.996275in}{0.716028in}}%
\pgfpathlineto{\pgfqpoint{5.996275in}{4.233871in}}%
\pgfpathlineto{\pgfqpoint{0.802523in}{4.233871in}}%
\pgfpathlineto{\pgfqpoint{0.802523in}{0.716028in}}%
\pgfpathclose%
\pgfusepath{fill}%
\end{pgfscope}%
\begin{pgfscope}%
\pgfpathrectangle{\pgfqpoint{0.802523in}{0.716028in}}{\pgfqpoint{5.193752in}{3.517843in}}%
\pgfusepath{clip}%
\pgfsetbuttcap%
\pgfsetmiterjoin%
\definecolor{currentfill}{rgb}{0.000000,0.000000,1.000000}%
\pgfsetfillcolor{currentfill}%
\pgfsetlinewidth{0.000000pt}%
\definecolor{currentstroke}{rgb}{0.000000,0.000000,0.000000}%
\pgfsetstrokecolor{currentstroke}%
\pgfsetdash{}{0pt}%
\pgfpathmoveto{\pgfqpoint{0.802523in}{0.716028in}}%
\pgfpathlineto{\pgfqpoint{0.981618in}{0.716028in}}%
\pgfpathlineto{\pgfqpoint{0.981618in}{2.482492in}}%
\pgfpathlineto{\pgfqpoint{0.802523in}{2.482492in}}%
\pgfpathlineto{\pgfqpoint{0.802523in}{0.716028in}}%
\pgfpathclose%
\pgfusepath{fill}%
\end{pgfscope}%
\begin{pgfscope}%
\pgfpathrectangle{\pgfqpoint{0.802523in}{0.716028in}}{\pgfqpoint{5.193752in}{3.517843in}}%
\pgfusepath{clip}%
\pgfsetbuttcap%
\pgfsetmiterjoin%
\definecolor{currentfill}{rgb}{0.000000,0.000000,1.000000}%
\pgfsetfillcolor{currentfill}%
\pgfsetlinewidth{0.000000pt}%
\definecolor{currentstroke}{rgb}{0.000000,0.000000,0.000000}%
\pgfsetstrokecolor{currentstroke}%
\pgfsetdash{}{0pt}%
\pgfpathmoveto{\pgfqpoint{0.981618in}{0.716028in}}%
\pgfpathlineto{\pgfqpoint{1.160713in}{0.716028in}}%
\pgfpathlineto{\pgfqpoint{1.160713in}{4.066355in}}%
\pgfpathlineto{\pgfqpoint{0.981618in}{4.066355in}}%
\pgfpathlineto{\pgfqpoint{0.981618in}{0.716028in}}%
\pgfpathclose%
\pgfusepath{fill}%
\end{pgfscope}%
\begin{pgfscope}%
\pgfpathrectangle{\pgfqpoint{0.802523in}{0.716028in}}{\pgfqpoint{5.193752in}{3.517843in}}%
\pgfusepath{clip}%
\pgfsetbuttcap%
\pgfsetmiterjoin%
\definecolor{currentfill}{rgb}{0.000000,0.000000,1.000000}%
\pgfsetfillcolor{currentfill}%
\pgfsetlinewidth{0.000000pt}%
\definecolor{currentstroke}{rgb}{0.000000,0.000000,0.000000}%
\pgfsetstrokecolor{currentstroke}%
\pgfsetdash{}{0pt}%
\pgfpathmoveto{\pgfqpoint{1.160713in}{0.716028in}}%
\pgfpathlineto{\pgfqpoint{1.339808in}{0.716028in}}%
\pgfpathlineto{\pgfqpoint{1.339808in}{3.824210in}}%
\pgfpathlineto{\pgfqpoint{1.160713in}{3.824210in}}%
\pgfpathlineto{\pgfqpoint{1.160713in}{0.716028in}}%
\pgfpathclose%
\pgfusepath{fill}%
\end{pgfscope}%
\begin{pgfscope}%
\pgfpathrectangle{\pgfqpoint{0.802523in}{0.716028in}}{\pgfqpoint{5.193752in}{3.517843in}}%
\pgfusepath{clip}%
\pgfsetbuttcap%
\pgfsetmiterjoin%
\definecolor{currentfill}{rgb}{0.000000,0.000000,1.000000}%
\pgfsetfillcolor{currentfill}%
\pgfsetlinewidth{0.000000pt}%
\definecolor{currentstroke}{rgb}{0.000000,0.000000,0.000000}%
\pgfsetstrokecolor{currentstroke}%
\pgfsetdash{}{0pt}%
\pgfpathmoveto{\pgfqpoint{1.339808in}{0.716028in}}%
\pgfpathlineto{\pgfqpoint{1.518902in}{0.716028in}}%
\pgfpathlineto{\pgfqpoint{1.518902in}{3.006477in}}%
\pgfpathlineto{\pgfqpoint{1.339808in}{3.006477in}}%
\pgfpathlineto{\pgfqpoint{1.339808in}{0.716028in}}%
\pgfpathclose%
\pgfusepath{fill}%
\end{pgfscope}%
\begin{pgfscope}%
\pgfpathrectangle{\pgfqpoint{0.802523in}{0.716028in}}{\pgfqpoint{5.193752in}{3.517843in}}%
\pgfusepath{clip}%
\pgfsetbuttcap%
\pgfsetmiterjoin%
\definecolor{currentfill}{rgb}{0.000000,0.000000,1.000000}%
\pgfsetfillcolor{currentfill}%
\pgfsetlinewidth{0.000000pt}%
\definecolor{currentstroke}{rgb}{0.000000,0.000000,0.000000}%
\pgfsetstrokecolor{currentstroke}%
\pgfsetdash{}{0pt}%
\pgfpathmoveto{\pgfqpoint{1.518902in}{0.716028in}}%
\pgfpathlineto{\pgfqpoint{1.697997in}{0.716028in}}%
\pgfpathlineto{\pgfqpoint{1.697997in}{2.613488in}}%
\pgfpathlineto{\pgfqpoint{1.518902in}{2.613488in}}%
\pgfpathlineto{\pgfqpoint{1.518902in}{0.716028in}}%
\pgfpathclose%
\pgfusepath{fill}%
\end{pgfscope}%
\begin{pgfscope}%
\pgfpathrectangle{\pgfqpoint{0.802523in}{0.716028in}}{\pgfqpoint{5.193752in}{3.517843in}}%
\pgfusepath{clip}%
\pgfsetbuttcap%
\pgfsetmiterjoin%
\definecolor{currentfill}{rgb}{0.000000,0.000000,1.000000}%
\pgfsetfillcolor{currentfill}%
\pgfsetlinewidth{0.000000pt}%
\definecolor{currentstroke}{rgb}{0.000000,0.000000,0.000000}%
\pgfsetstrokecolor{currentstroke}%
\pgfsetdash{}{0pt}%
\pgfpathmoveto{\pgfqpoint{1.697997in}{0.716028in}}%
\pgfpathlineto{\pgfqpoint{1.877092in}{0.716028in}}%
\pgfpathlineto{\pgfqpoint{1.877092in}{2.145077in}}%
\pgfpathlineto{\pgfqpoint{1.697997in}{2.145077in}}%
\pgfpathlineto{\pgfqpoint{1.697997in}{0.716028in}}%
\pgfpathclose%
\pgfusepath{fill}%
\end{pgfscope}%
\begin{pgfscope}%
\pgfpathrectangle{\pgfqpoint{0.802523in}{0.716028in}}{\pgfqpoint{5.193752in}{3.517843in}}%
\pgfusepath{clip}%
\pgfsetbuttcap%
\pgfsetmiterjoin%
\definecolor{currentfill}{rgb}{0.000000,0.000000,1.000000}%
\pgfsetfillcolor{currentfill}%
\pgfsetlinewidth{0.000000pt}%
\definecolor{currentstroke}{rgb}{0.000000,0.000000,0.000000}%
\pgfsetstrokecolor{currentstroke}%
\pgfsetdash{}{0pt}%
\pgfpathmoveto{\pgfqpoint{1.877092in}{0.716028in}}%
\pgfpathlineto{\pgfqpoint{2.056187in}{0.716028in}}%
\pgfpathlineto{\pgfqpoint{2.056187in}{2.041868in}}%
\pgfpathlineto{\pgfqpoint{1.877092in}{2.041868in}}%
\pgfpathlineto{\pgfqpoint{1.877092in}{0.716028in}}%
\pgfpathclose%
\pgfusepath{fill}%
\end{pgfscope}%
\begin{pgfscope}%
\pgfpathrectangle{\pgfqpoint{0.802523in}{0.716028in}}{\pgfqpoint{5.193752in}{3.517843in}}%
\pgfusepath{clip}%
\pgfsetbuttcap%
\pgfsetmiterjoin%
\definecolor{currentfill}{rgb}{0.000000,0.000000,1.000000}%
\pgfsetfillcolor{currentfill}%
\pgfsetlinewidth{0.000000pt}%
\definecolor{currentstroke}{rgb}{0.000000,0.000000,0.000000}%
\pgfsetstrokecolor{currentstroke}%
\pgfsetdash{}{0pt}%
\pgfpathmoveto{\pgfqpoint{2.056187in}{0.716028in}}%
\pgfpathlineto{\pgfqpoint{2.235282in}{0.716028in}}%
\pgfpathlineto{\pgfqpoint{2.235282in}{1.930720in}}%
\pgfpathlineto{\pgfqpoint{2.056187in}{1.930720in}}%
\pgfpathlineto{\pgfqpoint{2.056187in}{0.716028in}}%
\pgfpathclose%
\pgfusepath{fill}%
\end{pgfscope}%
\begin{pgfscope}%
\pgfpathrectangle{\pgfqpoint{0.802523in}{0.716028in}}{\pgfqpoint{5.193752in}{3.517843in}}%
\pgfusepath{clip}%
\pgfsetbuttcap%
\pgfsetmiterjoin%
\definecolor{currentfill}{rgb}{0.000000,0.000000,1.000000}%
\pgfsetfillcolor{currentfill}%
\pgfsetlinewidth{0.000000pt}%
\definecolor{currentstroke}{rgb}{0.000000,0.000000,0.000000}%
\pgfsetstrokecolor{currentstroke}%
\pgfsetdash{}{0pt}%
\pgfpathmoveto{\pgfqpoint{2.235282in}{0.716028in}}%
\pgfpathlineto{\pgfqpoint{2.414377in}{0.716028in}}%
\pgfpathlineto{\pgfqpoint{2.414377in}{1.942629in}}%
\pgfpathlineto{\pgfqpoint{2.235282in}{1.942629in}}%
\pgfpathlineto{\pgfqpoint{2.235282in}{0.716028in}}%
\pgfpathclose%
\pgfusepath{fill}%
\end{pgfscope}%
\begin{pgfscope}%
\pgfpathrectangle{\pgfqpoint{0.802523in}{0.716028in}}{\pgfqpoint{5.193752in}{3.517843in}}%
\pgfusepath{clip}%
\pgfsetbuttcap%
\pgfsetmiterjoin%
\definecolor{currentfill}{rgb}{0.000000,0.000000,1.000000}%
\pgfsetfillcolor{currentfill}%
\pgfsetlinewidth{0.000000pt}%
\definecolor{currentstroke}{rgb}{0.000000,0.000000,0.000000}%
\pgfsetstrokecolor{currentstroke}%
\pgfsetdash{}{0pt}%
\pgfpathmoveto{\pgfqpoint{2.414377in}{0.716028in}}%
\pgfpathlineto{\pgfqpoint{2.593472in}{0.716028in}}%
\pgfpathlineto{\pgfqpoint{2.593472in}{1.803693in}}%
\pgfpathlineto{\pgfqpoint{2.414377in}{1.803693in}}%
\pgfpathlineto{\pgfqpoint{2.414377in}{0.716028in}}%
\pgfpathclose%
\pgfusepath{fill}%
\end{pgfscope}%
\begin{pgfscope}%
\pgfpathrectangle{\pgfqpoint{0.802523in}{0.716028in}}{\pgfqpoint{5.193752in}{3.517843in}}%
\pgfusepath{clip}%
\pgfsetbuttcap%
\pgfsetmiterjoin%
\definecolor{currentfill}{rgb}{0.000000,0.000000,1.000000}%
\pgfsetfillcolor{currentfill}%
\pgfsetlinewidth{0.000000pt}%
\definecolor{currentstroke}{rgb}{0.000000,0.000000,0.000000}%
\pgfsetstrokecolor{currentstroke}%
\pgfsetdash{}{0pt}%
\pgfpathmoveto{\pgfqpoint{2.593472in}{0.716028in}}%
\pgfpathlineto{\pgfqpoint{2.772567in}{0.716028in}}%
\pgfpathlineto{\pgfqpoint{2.772567in}{1.883085in}}%
\pgfpathlineto{\pgfqpoint{2.593472in}{1.883085in}}%
\pgfpathlineto{\pgfqpoint{2.593472in}{0.716028in}}%
\pgfpathclose%
\pgfusepath{fill}%
\end{pgfscope}%
\begin{pgfscope}%
\pgfpathrectangle{\pgfqpoint{0.802523in}{0.716028in}}{\pgfqpoint{5.193752in}{3.517843in}}%
\pgfusepath{clip}%
\pgfsetbuttcap%
\pgfsetmiterjoin%
\definecolor{currentfill}{rgb}{0.000000,0.000000,1.000000}%
\pgfsetfillcolor{currentfill}%
\pgfsetlinewidth{0.000000pt}%
\definecolor{currentstroke}{rgb}{0.000000,0.000000,0.000000}%
\pgfsetstrokecolor{currentstroke}%
\pgfsetdash{}{0pt}%
\pgfpathmoveto{\pgfqpoint{2.772567in}{0.716028in}}%
\pgfpathlineto{\pgfqpoint{2.951662in}{0.716028in}}%
\pgfpathlineto{\pgfqpoint{2.951662in}{1.990264in}}%
\pgfpathlineto{\pgfqpoint{2.772567in}{1.990264in}}%
\pgfpathlineto{\pgfqpoint{2.772567in}{0.716028in}}%
\pgfpathclose%
\pgfusepath{fill}%
\end{pgfscope}%
\begin{pgfscope}%
\pgfpathrectangle{\pgfqpoint{0.802523in}{0.716028in}}{\pgfqpoint{5.193752in}{3.517843in}}%
\pgfusepath{clip}%
\pgfsetbuttcap%
\pgfsetmiterjoin%
\definecolor{currentfill}{rgb}{0.000000,0.000000,1.000000}%
\pgfsetfillcolor{currentfill}%
\pgfsetlinewidth{0.000000pt}%
\definecolor{currentstroke}{rgb}{0.000000,0.000000,0.000000}%
\pgfsetstrokecolor{currentstroke}%
\pgfsetdash{}{0pt}%
\pgfpathmoveto{\pgfqpoint{2.951662in}{0.716028in}}%
\pgfpathlineto{\pgfqpoint{3.130756in}{0.716028in}}%
\pgfpathlineto{\pgfqpoint{3.130756in}{2.077595in}}%
\pgfpathlineto{\pgfqpoint{2.951662in}{2.077595in}}%
\pgfpathlineto{\pgfqpoint{2.951662in}{0.716028in}}%
\pgfpathclose%
\pgfusepath{fill}%
\end{pgfscope}%
\begin{pgfscope}%
\pgfpathrectangle{\pgfqpoint{0.802523in}{0.716028in}}{\pgfqpoint{5.193752in}{3.517843in}}%
\pgfusepath{clip}%
\pgfsetbuttcap%
\pgfsetmiterjoin%
\definecolor{currentfill}{rgb}{0.000000,0.000000,1.000000}%
\pgfsetfillcolor{currentfill}%
\pgfsetlinewidth{0.000000pt}%
\definecolor{currentstroke}{rgb}{0.000000,0.000000,0.000000}%
\pgfsetstrokecolor{currentstroke}%
\pgfsetdash{}{0pt}%
\pgfpathmoveto{\pgfqpoint{3.130756in}{0.716028in}}%
\pgfpathlineto{\pgfqpoint{3.309851in}{0.716028in}}%
\pgfpathlineto{\pgfqpoint{3.309851in}{1.922781in}}%
\pgfpathlineto{\pgfqpoint{3.130756in}{1.922781in}}%
\pgfpathlineto{\pgfqpoint{3.130756in}{0.716028in}}%
\pgfpathclose%
\pgfusepath{fill}%
\end{pgfscope}%
\begin{pgfscope}%
\pgfpathrectangle{\pgfqpoint{0.802523in}{0.716028in}}{\pgfqpoint{5.193752in}{3.517843in}}%
\pgfusepath{clip}%
\pgfsetbuttcap%
\pgfsetmiterjoin%
\definecolor{currentfill}{rgb}{0.000000,0.000000,1.000000}%
\pgfsetfillcolor{currentfill}%
\pgfsetlinewidth{0.000000pt}%
\definecolor{currentstroke}{rgb}{0.000000,0.000000,0.000000}%
\pgfsetstrokecolor{currentstroke}%
\pgfsetdash{}{0pt}%
\pgfpathmoveto{\pgfqpoint{3.309851in}{0.716028in}}%
\pgfpathlineto{\pgfqpoint{3.488946in}{0.716028in}}%
\pgfpathlineto{\pgfqpoint{3.488946in}{2.049807in}}%
\pgfpathlineto{\pgfqpoint{3.309851in}{2.049807in}}%
\pgfpathlineto{\pgfqpoint{3.309851in}{0.716028in}}%
\pgfpathclose%
\pgfusepath{fill}%
\end{pgfscope}%
\begin{pgfscope}%
\pgfpathrectangle{\pgfqpoint{0.802523in}{0.716028in}}{\pgfqpoint{5.193752in}{3.517843in}}%
\pgfusepath{clip}%
\pgfsetbuttcap%
\pgfsetmiterjoin%
\definecolor{currentfill}{rgb}{0.000000,0.000000,1.000000}%
\pgfsetfillcolor{currentfill}%
\pgfsetlinewidth{0.000000pt}%
\definecolor{currentstroke}{rgb}{0.000000,0.000000,0.000000}%
\pgfsetstrokecolor{currentstroke}%
\pgfsetdash{}{0pt}%
\pgfpathmoveto{\pgfqpoint{3.488946in}{0.716028in}}%
\pgfpathlineto{\pgfqpoint{3.668041in}{0.716028in}}%
\pgfpathlineto{\pgfqpoint{3.668041in}{1.791785in}}%
\pgfpathlineto{\pgfqpoint{3.488946in}{1.791785in}}%
\pgfpathlineto{\pgfqpoint{3.488946in}{0.716028in}}%
\pgfpathclose%
\pgfusepath{fill}%
\end{pgfscope}%
\begin{pgfscope}%
\pgfpathrectangle{\pgfqpoint{0.802523in}{0.716028in}}{\pgfqpoint{5.193752in}{3.517843in}}%
\pgfusepath{clip}%
\pgfsetbuttcap%
\pgfsetmiterjoin%
\definecolor{currentfill}{rgb}{0.000000,0.000000,1.000000}%
\pgfsetfillcolor{currentfill}%
\pgfsetlinewidth{0.000000pt}%
\definecolor{currentstroke}{rgb}{0.000000,0.000000,0.000000}%
\pgfsetstrokecolor{currentstroke}%
\pgfsetdash{}{0pt}%
\pgfpathmoveto{\pgfqpoint{3.668041in}{0.716028in}}%
\pgfpathlineto{\pgfqpoint{3.847136in}{0.716028in}}%
\pgfpathlineto{\pgfqpoint{3.847136in}{2.041868in}}%
\pgfpathlineto{\pgfqpoint{3.668041in}{2.041868in}}%
\pgfpathlineto{\pgfqpoint{3.668041in}{0.716028in}}%
\pgfpathclose%
\pgfusepath{fill}%
\end{pgfscope}%
\begin{pgfscope}%
\pgfpathrectangle{\pgfqpoint{0.802523in}{0.716028in}}{\pgfqpoint{5.193752in}{3.517843in}}%
\pgfusepath{clip}%
\pgfsetbuttcap%
\pgfsetmiterjoin%
\definecolor{currentfill}{rgb}{0.000000,0.000000,1.000000}%
\pgfsetfillcolor{currentfill}%
\pgfsetlinewidth{0.000000pt}%
\definecolor{currentstroke}{rgb}{0.000000,0.000000,0.000000}%
\pgfsetstrokecolor{currentstroke}%
\pgfsetdash{}{0pt}%
\pgfpathmoveto{\pgfqpoint{3.847136in}{0.716028in}}%
\pgfpathlineto{\pgfqpoint{4.026231in}{0.716028in}}%
\pgfpathlineto{\pgfqpoint{4.026231in}{2.033929in}}%
\pgfpathlineto{\pgfqpoint{3.847136in}{2.033929in}}%
\pgfpathlineto{\pgfqpoint{3.847136in}{0.716028in}}%
\pgfpathclose%
\pgfusepath{fill}%
\end{pgfscope}%
\begin{pgfscope}%
\pgfpathrectangle{\pgfqpoint{0.802523in}{0.716028in}}{\pgfqpoint{5.193752in}{3.517843in}}%
\pgfusepath{clip}%
\pgfsetbuttcap%
\pgfsetmiterjoin%
\definecolor{currentfill}{rgb}{0.000000,0.000000,1.000000}%
\pgfsetfillcolor{currentfill}%
\pgfsetlinewidth{0.000000pt}%
\definecolor{currentstroke}{rgb}{0.000000,0.000000,0.000000}%
\pgfsetstrokecolor{currentstroke}%
\pgfsetdash{}{0pt}%
\pgfpathmoveto{\pgfqpoint{4.026231in}{0.716028in}}%
\pgfpathlineto{\pgfqpoint{4.205326in}{0.716028in}}%
\pgfpathlineto{\pgfqpoint{4.205326in}{1.966446in}}%
\pgfpathlineto{\pgfqpoint{4.026231in}{1.966446in}}%
\pgfpathlineto{\pgfqpoint{4.026231in}{0.716028in}}%
\pgfpathclose%
\pgfusepath{fill}%
\end{pgfscope}%
\begin{pgfscope}%
\pgfpathrectangle{\pgfqpoint{0.802523in}{0.716028in}}{\pgfqpoint{5.193752in}{3.517843in}}%
\pgfusepath{clip}%
\pgfsetbuttcap%
\pgfsetmiterjoin%
\definecolor{currentfill}{rgb}{0.000000,0.000000,1.000000}%
\pgfsetfillcolor{currentfill}%
\pgfsetlinewidth{0.000000pt}%
\definecolor{currentstroke}{rgb}{0.000000,0.000000,0.000000}%
\pgfsetstrokecolor{currentstroke}%
\pgfsetdash{}{0pt}%
\pgfpathmoveto{\pgfqpoint{4.205326in}{0.716028in}}%
\pgfpathlineto{\pgfqpoint{4.384421in}{0.716028in}}%
\pgfpathlineto{\pgfqpoint{4.384421in}{1.728271in}}%
\pgfpathlineto{\pgfqpoint{4.205326in}{1.728271in}}%
\pgfpathlineto{\pgfqpoint{4.205326in}{0.716028in}}%
\pgfpathclose%
\pgfusepath{fill}%
\end{pgfscope}%
\begin{pgfscope}%
\pgfpathrectangle{\pgfqpoint{0.802523in}{0.716028in}}{\pgfqpoint{5.193752in}{3.517843in}}%
\pgfusepath{clip}%
\pgfsetbuttcap%
\pgfsetmiterjoin%
\definecolor{currentfill}{rgb}{0.000000,0.000000,1.000000}%
\pgfsetfillcolor{currentfill}%
\pgfsetlinewidth{0.000000pt}%
\definecolor{currentstroke}{rgb}{0.000000,0.000000,0.000000}%
\pgfsetstrokecolor{currentstroke}%
\pgfsetdash{}{0pt}%
\pgfpathmoveto{\pgfqpoint{4.384421in}{0.716028in}}%
\pgfpathlineto{\pgfqpoint{4.563516in}{0.716028in}}%
\pgfpathlineto{\pgfqpoint{4.563516in}{1.605214in}}%
\pgfpathlineto{\pgfqpoint{4.384421in}{1.605214in}}%
\pgfpathlineto{\pgfqpoint{4.384421in}{0.716028in}}%
\pgfpathclose%
\pgfusepath{fill}%
\end{pgfscope}%
\begin{pgfscope}%
\pgfpathrectangle{\pgfqpoint{0.802523in}{0.716028in}}{\pgfqpoint{5.193752in}{3.517843in}}%
\pgfusepath{clip}%
\pgfsetbuttcap%
\pgfsetmiterjoin%
\definecolor{currentfill}{rgb}{0.000000,0.000000,1.000000}%
\pgfsetfillcolor{currentfill}%
\pgfsetlinewidth{0.000000pt}%
\definecolor{currentstroke}{rgb}{0.000000,0.000000,0.000000}%
\pgfsetstrokecolor{currentstroke}%
\pgfsetdash{}{0pt}%
\pgfpathmoveto{\pgfqpoint{4.563516in}{0.716028in}}%
\pgfpathlineto{\pgfqpoint{4.742611in}{0.716028in}}%
\pgfpathlineto{\pgfqpoint{4.742611in}{1.521853in}}%
\pgfpathlineto{\pgfqpoint{4.563516in}{1.521853in}}%
\pgfpathlineto{\pgfqpoint{4.563516in}{0.716028in}}%
\pgfpathclose%
\pgfusepath{fill}%
\end{pgfscope}%
\begin{pgfscope}%
\pgfpathrectangle{\pgfqpoint{0.802523in}{0.716028in}}{\pgfqpoint{5.193752in}{3.517843in}}%
\pgfusepath{clip}%
\pgfsetbuttcap%
\pgfsetmiterjoin%
\definecolor{currentfill}{rgb}{0.000000,0.000000,1.000000}%
\pgfsetfillcolor{currentfill}%
\pgfsetlinewidth{0.000000pt}%
\definecolor{currentstroke}{rgb}{0.000000,0.000000,0.000000}%
\pgfsetstrokecolor{currentstroke}%
\pgfsetdash{}{0pt}%
\pgfpathmoveto{\pgfqpoint{4.742611in}{0.716028in}}%
\pgfpathlineto{\pgfqpoint{4.921705in}{0.716028in}}%
\pgfpathlineto{\pgfqpoint{4.921705in}{1.486127in}}%
\pgfpathlineto{\pgfqpoint{4.742611in}{1.486127in}}%
\pgfpathlineto{\pgfqpoint{4.742611in}{0.716028in}}%
\pgfpathclose%
\pgfusepath{fill}%
\end{pgfscope}%
\begin{pgfscope}%
\pgfpathrectangle{\pgfqpoint{0.802523in}{0.716028in}}{\pgfqpoint{5.193752in}{3.517843in}}%
\pgfusepath{clip}%
\pgfsetbuttcap%
\pgfsetmiterjoin%
\definecolor{currentfill}{rgb}{0.000000,0.000000,1.000000}%
\pgfsetfillcolor{currentfill}%
\pgfsetlinewidth{0.000000pt}%
\definecolor{currentstroke}{rgb}{0.000000,0.000000,0.000000}%
\pgfsetstrokecolor{currentstroke}%
\pgfsetdash{}{0pt}%
\pgfpathmoveto{\pgfqpoint{4.921705in}{0.716028in}}%
\pgfpathlineto{\pgfqpoint{5.100800in}{0.716028in}}%
\pgfpathlineto{\pgfqpoint{5.100800in}{1.815602in}}%
\pgfpathlineto{\pgfqpoint{4.921705in}{1.815602in}}%
\pgfpathlineto{\pgfqpoint{4.921705in}{0.716028in}}%
\pgfpathclose%
\pgfusepath{fill}%
\end{pgfscope}%
\begin{pgfscope}%
\pgfpathrectangle{\pgfqpoint{0.802523in}{0.716028in}}{\pgfqpoint{5.193752in}{3.517843in}}%
\pgfusepath{clip}%
\pgfsetbuttcap%
\pgfsetmiterjoin%
\definecolor{currentfill}{rgb}{0.000000,0.000000,1.000000}%
\pgfsetfillcolor{currentfill}%
\pgfsetlinewidth{0.000000pt}%
\definecolor{currentstroke}{rgb}{0.000000,0.000000,0.000000}%
\pgfsetstrokecolor{currentstroke}%
\pgfsetdash{}{0pt}%
\pgfpathmoveto{\pgfqpoint{5.100800in}{0.716028in}}%
\pgfpathlineto{\pgfqpoint{5.279895in}{0.716028in}}%
\pgfpathlineto{\pgfqpoint{5.279895in}{1.752089in}}%
\pgfpathlineto{\pgfqpoint{5.100800in}{1.752089in}}%
\pgfpathlineto{\pgfqpoint{5.100800in}{0.716028in}}%
\pgfpathclose%
\pgfusepath{fill}%
\end{pgfscope}%
\begin{pgfscope}%
\pgfpathrectangle{\pgfqpoint{0.802523in}{0.716028in}}{\pgfqpoint{5.193752in}{3.517843in}}%
\pgfusepath{clip}%
\pgfsetbuttcap%
\pgfsetmiterjoin%
\definecolor{currentfill}{rgb}{0.000000,0.000000,1.000000}%
\pgfsetfillcolor{currentfill}%
\pgfsetlinewidth{0.000000pt}%
\definecolor{currentstroke}{rgb}{0.000000,0.000000,0.000000}%
\pgfsetstrokecolor{currentstroke}%
\pgfsetdash{}{0pt}%
\pgfpathmoveto{\pgfqpoint{5.279895in}{0.716028in}}%
\pgfpathlineto{\pgfqpoint{5.458990in}{0.716028in}}%
\pgfpathlineto{\pgfqpoint{5.458990in}{1.359100in}}%
\pgfpathlineto{\pgfqpoint{5.279895in}{1.359100in}}%
\pgfpathlineto{\pgfqpoint{5.279895in}{0.716028in}}%
\pgfpathclose%
\pgfusepath{fill}%
\end{pgfscope}%
\begin{pgfscope}%
\pgfpathrectangle{\pgfqpoint{0.802523in}{0.716028in}}{\pgfqpoint{5.193752in}{3.517843in}}%
\pgfusepath{clip}%
\pgfsetbuttcap%
\pgfsetmiterjoin%
\definecolor{currentfill}{rgb}{0.000000,0.000000,1.000000}%
\pgfsetfillcolor{currentfill}%
\pgfsetlinewidth{0.000000pt}%
\definecolor{currentstroke}{rgb}{0.000000,0.000000,0.000000}%
\pgfsetstrokecolor{currentstroke}%
\pgfsetdash{}{0pt}%
\pgfpathmoveto{\pgfqpoint{5.458990in}{0.716028in}}%
\pgfpathlineto{\pgfqpoint{5.638085in}{0.716028in}}%
\pgfpathlineto{\pgfqpoint{5.638085in}{1.327344in}}%
\pgfpathlineto{\pgfqpoint{5.458990in}{1.327344in}}%
\pgfpathlineto{\pgfqpoint{5.458990in}{0.716028in}}%
\pgfpathclose%
\pgfusepath{fill}%
\end{pgfscope}%
\begin{pgfscope}%
\pgfpathrectangle{\pgfqpoint{0.802523in}{0.716028in}}{\pgfqpoint{5.193752in}{3.517843in}}%
\pgfusepath{clip}%
\pgfsetbuttcap%
\pgfsetmiterjoin%
\definecolor{currentfill}{rgb}{0.000000,0.000000,1.000000}%
\pgfsetfillcolor{currentfill}%
\pgfsetlinewidth{0.000000pt}%
\definecolor{currentstroke}{rgb}{0.000000,0.000000,0.000000}%
\pgfsetstrokecolor{currentstroke}%
\pgfsetdash{}{0pt}%
\pgfpathmoveto{\pgfqpoint{5.638085in}{0.716028in}}%
\pgfpathlineto{\pgfqpoint{5.817180in}{0.716028in}}%
\pgfpathlineto{\pgfqpoint{5.817180in}{1.116956in}}%
\pgfpathlineto{\pgfqpoint{5.638085in}{1.116956in}}%
\pgfpathlineto{\pgfqpoint{5.638085in}{0.716028in}}%
\pgfpathclose%
\pgfusepath{fill}%
\end{pgfscope}%
\begin{pgfscope}%
\pgfpathrectangle{\pgfqpoint{0.802523in}{0.716028in}}{\pgfqpoint{5.193752in}{3.517843in}}%
\pgfusepath{clip}%
\pgfsetbuttcap%
\pgfsetmiterjoin%
\definecolor{currentfill}{rgb}{0.000000,0.000000,1.000000}%
\pgfsetfillcolor{currentfill}%
\pgfsetlinewidth{0.000000pt}%
\definecolor{currentstroke}{rgb}{0.000000,0.000000,0.000000}%
\pgfsetstrokecolor{currentstroke}%
\pgfsetdash{}{0pt}%
\pgfpathmoveto{\pgfqpoint{5.817180in}{0.716028in}}%
\pgfpathlineto{\pgfqpoint{5.996275in}{0.716028in}}%
\pgfpathlineto{\pgfqpoint{5.996275in}{1.402766in}}%
\pgfpathlineto{\pgfqpoint{5.817180in}{1.402766in}}%
\pgfpathlineto{\pgfqpoint{5.817180in}{0.716028in}}%
\pgfpathclose%
\pgfusepath{fill}%
\end{pgfscope}%
\begin{pgfscope}%
\pgfsetbuttcap%
\pgfsetroundjoin%
\definecolor{currentfill}{rgb}{0.000000,0.000000,0.000000}%
\pgfsetfillcolor{currentfill}%
\pgfsetlinewidth{0.200750pt}%
\definecolor{currentstroke}{rgb}{0.000000,0.000000,0.000000}%
\pgfsetstrokecolor{currentstroke}%
\pgfsetdash{}{0pt}%
\pgfsys@defobject{currentmarker}{\pgfqpoint{0.000000in}{-0.048611in}}{\pgfqpoint{0.000000in}{0.000000in}}{%
\pgfpathmoveto{\pgfqpoint{0.000000in}{0.000000in}}%
\pgfpathlineto{\pgfqpoint{0.000000in}{-0.048611in}}%
\pgfusepath{stroke,fill}%
}%
\begin{pgfscope}%
\pgfsys@transformshift{0.802523in}{0.716028in}%
\pgfsys@useobject{currentmarker}{}%
\end{pgfscope}%
\end{pgfscope}%
\begin{pgfscope}%
\definecolor{textcolor}{rgb}{0.000000,0.000000,0.000000}%
\pgfsetstrokecolor{textcolor}%
\pgfsetfillcolor{textcolor}%
\pgftext[x=0.802523in,y=0.618806in,,top]{\color{textcolor}{\rmfamily\fontsize{20.000000}{24.000000}\selectfont\catcode`\^=\active\def^{\ifmmode\sp\else\^{}\fi}\catcode`\%=\active\def
\end{pgfscope}%
\begin{pgfscope}%
\pgfsetbuttcap%
\pgfsetroundjoin%
\definecolor{currentfill}{rgb}{0.000000,0.000000,0.000000}%
\pgfsetfillcolor{currentfill}%
\pgfsetlinewidth{0.200750pt}%
\definecolor{currentstroke}{rgb}{0.000000,0.000000,0.000000}%
\pgfsetstrokecolor{currentstroke}%
\pgfsetdash{}{0pt}%
\pgfsys@defobject{currentmarker}{\pgfqpoint{0.000000in}{-0.048611in}}{\pgfqpoint{0.000000in}{0.000000in}}{%
\pgfpathmoveto{\pgfqpoint{0.000000in}{0.000000in}}%
\pgfpathlineto{\pgfqpoint{0.000000in}{-0.048611in}}%
\pgfusepath{stroke,fill}%
}%
\begin{pgfscope}%
\pgfsys@transformshift{1.841273in}{0.716028in}%
\pgfsys@useobject{currentmarker}{}%
\end{pgfscope}%
\end{pgfscope}%
\begin{pgfscope}%
\definecolor{textcolor}{rgb}{0.000000,0.000000,0.000000}%
\pgfsetstrokecolor{textcolor}%
\pgfsetfillcolor{textcolor}%
\pgftext[x=1.841273in,y=0.618806in,,top]{\color{textcolor}{\rmfamily\fontsize{20.000000}{24.000000}\selectfont\catcode`\^=\active\def^{\ifmmode\sp\else\^{}\fi}\catcode`\%=\active\def
\end{pgfscope}%
\begin{pgfscope}%
\pgfsetbuttcap%
\pgfsetroundjoin%
\definecolor{currentfill}{rgb}{0.000000,0.000000,0.000000}%
\pgfsetfillcolor{currentfill}%
\pgfsetlinewidth{0.200750pt}%
\definecolor{currentstroke}{rgb}{0.000000,0.000000,0.000000}%
\pgfsetstrokecolor{currentstroke}%
\pgfsetdash{}{0pt}%
\pgfsys@defobject{currentmarker}{\pgfqpoint{0.000000in}{-0.048611in}}{\pgfqpoint{0.000000in}{0.000000in}}{%
\pgfpathmoveto{\pgfqpoint{0.000000in}{0.000000in}}%
\pgfpathlineto{\pgfqpoint{0.000000in}{-0.048611in}}%
\pgfusepath{stroke,fill}%
}%
\begin{pgfscope}%
\pgfsys@transformshift{2.880024in}{0.716028in}%
\pgfsys@useobject{currentmarker}{}%
\end{pgfscope}%
\end{pgfscope}%
\begin{pgfscope}%
\definecolor{textcolor}{rgb}{0.000000,0.000000,0.000000}%
\pgfsetstrokecolor{textcolor}%
\pgfsetfillcolor{textcolor}%
\pgftext[x=2.880024in,y=0.618806in,,top]{\color{textcolor}{\rmfamily\fontsize{20.000000}{24.000000}\selectfont\catcode`\^=\active\def^{\ifmmode\sp\else\^{}\fi}\catcode`\%=\active\def
\end{pgfscope}%
\begin{pgfscope}%
\pgfsetbuttcap%
\pgfsetroundjoin%
\definecolor{currentfill}{rgb}{0.000000,0.000000,0.000000}%
\pgfsetfillcolor{currentfill}%
\pgfsetlinewidth{0.200750pt}%
\definecolor{currentstroke}{rgb}{0.000000,0.000000,0.000000}%
\pgfsetstrokecolor{currentstroke}%
\pgfsetdash{}{0pt}%
\pgfsys@defobject{currentmarker}{\pgfqpoint{0.000000in}{-0.048611in}}{\pgfqpoint{0.000000in}{0.000000in}}{%
\pgfpathmoveto{\pgfqpoint{0.000000in}{0.000000in}}%
\pgfpathlineto{\pgfqpoint{0.000000in}{-0.048611in}}%
\pgfusepath{stroke,fill}%
}%
\begin{pgfscope}%
\pgfsys@transformshift{3.918774in}{0.716028in}%
\pgfsys@useobject{currentmarker}{}%
\end{pgfscope}%
\end{pgfscope}%
\begin{pgfscope}%
\definecolor{textcolor}{rgb}{0.000000,0.000000,0.000000}%
\pgfsetstrokecolor{textcolor}%
\pgfsetfillcolor{textcolor}%
\pgftext[x=3.918774in,y=0.618806in,,top]{\color{textcolor}{\rmfamily\fontsize{20.000000}{24.000000}\selectfont\catcode`\^=\active\def^{\ifmmode\sp\else\^{}\fi}\catcode`\%=\active\def
\end{pgfscope}%
\begin{pgfscope}%
\pgfsetbuttcap%
\pgfsetroundjoin%
\definecolor{currentfill}{rgb}{0.000000,0.000000,0.000000}%
\pgfsetfillcolor{currentfill}%
\pgfsetlinewidth{0.200750pt}%
\definecolor{currentstroke}{rgb}{0.000000,0.000000,0.000000}%
\pgfsetstrokecolor{currentstroke}%
\pgfsetdash{}{0pt}%
\pgfsys@defobject{currentmarker}{\pgfqpoint{0.000000in}{-0.048611in}}{\pgfqpoint{0.000000in}{0.000000in}}{%
\pgfpathmoveto{\pgfqpoint{0.000000in}{0.000000in}}%
\pgfpathlineto{\pgfqpoint{0.000000in}{-0.048611in}}%
\pgfusepath{stroke,fill}%
}%
\begin{pgfscope}%
\pgfsys@transformshift{4.957524in}{0.716028in}%
\pgfsys@useobject{currentmarker}{}%
\end{pgfscope}%
\end{pgfscope}%
\begin{pgfscope}%
\definecolor{textcolor}{rgb}{0.000000,0.000000,0.000000}%
\pgfsetstrokecolor{textcolor}%
\pgfsetfillcolor{textcolor}%
\pgftext[x=4.957524in,y=0.618806in,,top]{\color{textcolor}{\rmfamily\fontsize{20.000000}{24.000000}\selectfont\catcode`\^=\active\def^{\ifmmode\sp\else\^{}\fi}\catcode`\%=\active\def
\end{pgfscope}%
\begin{pgfscope}%
\pgfsetbuttcap%
\pgfsetroundjoin%
\definecolor{currentfill}{rgb}{0.000000,0.000000,0.000000}%
\pgfsetfillcolor{currentfill}%
\pgfsetlinewidth{0.200750pt}%
\definecolor{currentstroke}{rgb}{0.000000,0.000000,0.000000}%
\pgfsetstrokecolor{currentstroke}%
\pgfsetdash{}{0pt}%
\pgfsys@defobject{currentmarker}{\pgfqpoint{0.000000in}{-0.048611in}}{\pgfqpoint{0.000000in}{0.000000in}}{%
\pgfpathmoveto{\pgfqpoint{0.000000in}{0.000000in}}%
\pgfpathlineto{\pgfqpoint{0.000000in}{-0.048611in}}%
\pgfusepath{stroke,fill}%
}%
\begin{pgfscope}%
\pgfsys@transformshift{5.996275in}{0.716028in}%
\pgfsys@useobject{currentmarker}{}%
\end{pgfscope}%
\end{pgfscope}%
\begin{pgfscope}%
\definecolor{textcolor}{rgb}{0.000000,0.000000,0.000000}%
\pgfsetstrokecolor{textcolor}%
\pgfsetfillcolor{textcolor}%
\pgftext[x=5.996275in,y=0.618806in,,top]{\color{textcolor}{\rmfamily\fontsize{20.000000}{24.000000}\selectfont\catcode`\^=\active\def^{\ifmmode\sp\else\^{}\fi}\catcode`\%=\active\def
\end{pgfscope}%
\begin{pgfscope}%
\definecolor{textcolor}{rgb}{0.000000,0.000000,0.000000}%
\pgfsetstrokecolor{textcolor}%
\pgfsetfillcolor{textcolor}%
\pgftext[x=3.399399in,y=0.307183in,,top]{\color{textcolor}{\rmfamily\fontsize{26.000000}{31.200000}\selectfont\catcode`\^=\active\def^{\ifmmode\sp\else\^{}\fi}\catcode`\%=\active\def
\end{pgfscope}%
\begin{pgfscope}%
\pgfsetbuttcap%
\pgfsetroundjoin%
\definecolor{currentfill}{rgb}{0.000000,0.000000,0.000000}%
\pgfsetfillcolor{currentfill}%
\pgfsetlinewidth{0.200750pt}%
\definecolor{currentstroke}{rgb}{0.000000,0.000000,0.000000}%
\pgfsetstrokecolor{currentstroke}%
\pgfsetdash{}{0pt}%
\pgfsys@defobject{currentmarker}{\pgfqpoint{-0.048611in}{0.000000in}}{\pgfqpoint{-0.000000in}{0.000000in}}{%
\pgfpathmoveto{\pgfqpoint{-0.000000in}{0.000000in}}%
\pgfpathlineto{\pgfqpoint{-0.048611in}{0.000000in}}%
\pgfusepath{stroke,fill}%
}%
\begin{pgfscope}%
\pgfsys@transformshift{0.802523in}{0.716028in}%
\pgfsys@useobject{currentmarker}{}%
\end{pgfscope}%
\end{pgfscope}%
\begin{pgfscope}%
\definecolor{textcolor}{rgb}{0.000000,0.000000,0.000000}%
\pgfsetstrokecolor{textcolor}%
\pgfsetfillcolor{textcolor}%
\pgftext[x=0.362738in, y=0.616009in, left, base]{\color{textcolor}{\rmfamily\fontsize{20.000000}{24.000000}\selectfont\catcode`\^=\active\def^{\ifmmode\sp\else\^{}\fi}\catcode`\%=\active\def
\end{pgfscope}%
\begin{pgfscope}%
\pgfsetbuttcap%
\pgfsetroundjoin%
\definecolor{currentfill}{rgb}{0.000000,0.000000,0.000000}%
\pgfsetfillcolor{currentfill}%
\pgfsetlinewidth{0.200750pt}%
\definecolor{currentstroke}{rgb}{0.000000,0.000000,0.000000}%
\pgfsetstrokecolor{currentstroke}%
\pgfsetdash{}{0pt}%
\pgfsys@defobject{currentmarker}{\pgfqpoint{-0.048611in}{0.000000in}}{\pgfqpoint{-0.000000in}{0.000000in}}{%
\pgfpathmoveto{\pgfqpoint{-0.000000in}{0.000000in}}%
\pgfpathlineto{\pgfqpoint{-0.048611in}{0.000000in}}%
\pgfusepath{stroke,fill}%
}%
\begin{pgfscope}%
\pgfsys@transformshift{0.802523in}{1.370598in}%
\pgfsys@useobject{currentmarker}{}%
\end{pgfscope}%
\end{pgfscope}%
\begin{pgfscope}%
\definecolor{textcolor}{rgb}{0.000000,0.000000,0.000000}%
\pgfsetstrokecolor{textcolor}%
\pgfsetfillcolor{textcolor}%
\pgftext[x=0.362738in, y=1.270579in, left, base]{\color{textcolor}{\rmfamily\fontsize{20.000000}{24.000000}\selectfont\catcode`\^=\active\def^{\ifmmode\sp\else\^{}\fi}\catcode`\%=\active\def
\end{pgfscope}%
\begin{pgfscope}%
\pgfsetbuttcap%
\pgfsetroundjoin%
\definecolor{currentfill}{rgb}{0.000000,0.000000,0.000000}%
\pgfsetfillcolor{currentfill}%
\pgfsetlinewidth{0.200750pt}%
\definecolor{currentstroke}{rgb}{0.000000,0.000000,0.000000}%
\pgfsetstrokecolor{currentstroke}%
\pgfsetdash{}{0pt}%
\pgfsys@defobject{currentmarker}{\pgfqpoint{-0.048611in}{0.000000in}}{\pgfqpoint{-0.000000in}{0.000000in}}{%
\pgfpathmoveto{\pgfqpoint{-0.000000in}{0.000000in}}%
\pgfpathlineto{\pgfqpoint{-0.048611in}{0.000000in}}%
\pgfusepath{stroke,fill}%
}%
\begin{pgfscope}%
\pgfsys@transformshift{0.802523in}{2.025169in}%
\pgfsys@useobject{currentmarker}{}%
\end{pgfscope}%
\end{pgfscope}%
\begin{pgfscope}%
\definecolor{textcolor}{rgb}{0.000000,0.000000,0.000000}%
\pgfsetstrokecolor{textcolor}%
\pgfsetfillcolor{textcolor}%
\pgftext[x=0.362738in, y=1.925149in, left, base]{\color{textcolor}{\rmfamily\fontsize{20.000000}{24.000000}\selectfont\catcode`\^=\active\def^{\ifmmode\sp\else\^{}\fi}\catcode`\%=\active\def
\end{pgfscope}%
\begin{pgfscope}%
\pgfsetbuttcap%
\pgfsetroundjoin%
\definecolor{currentfill}{rgb}{0.000000,0.000000,0.000000}%
\pgfsetfillcolor{currentfill}%
\pgfsetlinewidth{0.200750pt}%
\definecolor{currentstroke}{rgb}{0.000000,0.000000,0.000000}%
\pgfsetstrokecolor{currentstroke}%
\pgfsetdash{}{0pt}%
\pgfsys@defobject{currentmarker}{\pgfqpoint{-0.048611in}{0.000000in}}{\pgfqpoint{-0.000000in}{0.000000in}}{%
\pgfpathmoveto{\pgfqpoint{-0.000000in}{0.000000in}}%
\pgfpathlineto{\pgfqpoint{-0.048611in}{0.000000in}}%
\pgfusepath{stroke,fill}%
}%
\begin{pgfscope}%
\pgfsys@transformshift{0.802523in}{2.679739in}%
\pgfsys@useobject{currentmarker}{}%
\end{pgfscope}%
\end{pgfscope}%
\begin{pgfscope}%
\definecolor{textcolor}{rgb}{0.000000,0.000000,0.000000}%
\pgfsetstrokecolor{textcolor}%
\pgfsetfillcolor{textcolor}%
\pgftext[x=0.362738in, y=2.579720in, left, base]{\color{textcolor}{\rmfamily\fontsize{20.000000}{24.000000}\selectfont\catcode`\^=\active\def^{\ifmmode\sp\else\^{}\fi}\catcode`\%=\active\def
\end{pgfscope}%
\begin{pgfscope}%
\pgfsetbuttcap%
\pgfsetroundjoin%
\definecolor{currentfill}{rgb}{0.000000,0.000000,0.000000}%
\pgfsetfillcolor{currentfill}%
\pgfsetlinewidth{0.200750pt}%
\definecolor{currentstroke}{rgb}{0.000000,0.000000,0.000000}%
\pgfsetstrokecolor{currentstroke}%
\pgfsetdash{}{0pt}%
\pgfsys@defobject{currentmarker}{\pgfqpoint{-0.048611in}{0.000000in}}{\pgfqpoint{-0.000000in}{0.000000in}}{%
\pgfpathmoveto{\pgfqpoint{-0.000000in}{0.000000in}}%
\pgfpathlineto{\pgfqpoint{-0.048611in}{0.000000in}}%
\pgfusepath{stroke,fill}%
}%
\begin{pgfscope}%
\pgfsys@transformshift{0.802523in}{3.334309in}%
\pgfsys@useobject{currentmarker}{}%
\end{pgfscope}%
\end{pgfscope}%
\begin{pgfscope}%
\definecolor{textcolor}{rgb}{0.000000,0.000000,0.000000}%
\pgfsetstrokecolor{textcolor}%
\pgfsetfillcolor{textcolor}%
\pgftext[x=0.362738in, y=3.234290in, left, base]{\color{textcolor}{\rmfamily\fontsize{20.000000}{24.000000}\selectfont\catcode`\^=\active\def^{\ifmmode\sp\else\^{}\fi}\catcode`\%=\active\def
\end{pgfscope}%
\begin{pgfscope}%
\pgfsetbuttcap%
\pgfsetroundjoin%
\definecolor{currentfill}{rgb}{0.000000,0.000000,0.000000}%
\pgfsetfillcolor{currentfill}%
\pgfsetlinewidth{0.200750pt}%
\definecolor{currentstroke}{rgb}{0.000000,0.000000,0.000000}%
\pgfsetstrokecolor{currentstroke}%
\pgfsetdash{}{0pt}%
\pgfsys@defobject{currentmarker}{\pgfqpoint{-0.048611in}{0.000000in}}{\pgfqpoint{-0.000000in}{0.000000in}}{%
\pgfpathmoveto{\pgfqpoint{-0.000000in}{0.000000in}}%
\pgfpathlineto{\pgfqpoint{-0.048611in}{0.000000in}}%
\pgfusepath{stroke,fill}%
}%
\begin{pgfscope}%
\pgfsys@transformshift{0.802523in}{3.988879in}%
\pgfsys@useobject{currentmarker}{}%
\end{pgfscope}%
\end{pgfscope}%
\begin{pgfscope}%
\definecolor{textcolor}{rgb}{0.000000,0.000000,0.000000}%
\pgfsetstrokecolor{textcolor}%
\pgfsetfillcolor{textcolor}%
\pgftext[x=0.362738in, y=3.888860in, left, base]{\color{textcolor}{\rmfamily\fontsize{20.000000}{24.000000}\selectfont\catcode`\^=\active\def^{\ifmmode\sp\else\^{}\fi}\catcode`\%=\active\def
\end{pgfscope}%
\begin{pgfscope}%
\definecolor{textcolor}{rgb}{0.000000,0.000000,0.000000}%
\pgfsetstrokecolor{textcolor}%
\pgfsetfillcolor{textcolor}%
\pgftext[x=0.307183in,y=2.474950in,,bottom,rotate=90.000000]{\color{textcolor}{\rmfamily\fontsize{26.000000}{31.200000}\selectfont\catcode`\^=\active\def^{\ifmmode\sp\else\^{}\fi}\catcode`\%=\active\def
\end{pgfscope}%
\begin{pgfscope}%
\pgfpathrectangle{\pgfqpoint{0.802523in}{0.716028in}}{\pgfqpoint{5.193752in}{3.517843in}}%
\pgfusepath{clip}%
\pgfsetbuttcap%
\pgfsetroundjoin%
\pgfsetlinewidth{2.007500pt}%
\definecolor{currentstroke}{rgb}{0.501961,0.501961,0.501961}%
\pgfsetstrokecolor{currentstroke}%
\pgfsetdash{{7.400000pt}{3.200000pt}}{0.000000pt}%
\pgfpathmoveto{\pgfqpoint{0.802523in}{2.025169in}}%
\pgfpathlineto{\pgfqpoint{5.996275in}{2.025169in}}%
\pgfusepath{stroke}%
\end{pgfscope}%
\begin{pgfscope}%
\pgfsetrectcap%
\pgfsetmiterjoin%
\pgfsetlinewidth{0.803000pt}%
\definecolor{currentstroke}{rgb}{0.000000,0.000000,0.000000}%
\pgfsetstrokecolor{currentstroke}%
\pgfsetdash{}{0pt}%
\pgfpathmoveto{\pgfqpoint{0.802523in}{0.716028in}}%
\pgfpathlineto{\pgfqpoint{0.802523in}{4.233871in}}%
\pgfusepath{stroke}%
\end{pgfscope}%
\begin{pgfscope}%
\pgfsetrectcap%
\pgfsetmiterjoin%
\pgfsetlinewidth{0.803000pt}%
\definecolor{currentstroke}{rgb}{0.000000,0.000000,0.000000}%
\pgfsetstrokecolor{currentstroke}%
\pgfsetdash{}{0pt}%
\pgfpathmoveto{\pgfqpoint{5.996275in}{0.716028in}}%
\pgfpathlineto{\pgfqpoint{5.996275in}{4.233871in}}%
\pgfusepath{stroke}%
\end{pgfscope}%
\begin{pgfscope}%
\pgfsetrectcap%
\pgfsetmiterjoin%
\pgfsetlinewidth{0.803000pt}%
\definecolor{currentstroke}{rgb}{0.000000,0.000000,0.000000}%
\pgfsetstrokecolor{currentstroke}%
\pgfsetdash{}{0pt}%
\pgfpathmoveto{\pgfqpoint{0.802523in}{0.716028in}}%
\pgfpathlineto{\pgfqpoint{5.996275in}{0.716028in}}%
\pgfusepath{stroke}%
\end{pgfscope}%
\begin{pgfscope}%
\pgfsetrectcap%
\pgfsetmiterjoin%
\pgfsetlinewidth{0.803000pt}%
\definecolor{currentstroke}{rgb}{0.000000,0.000000,0.000000}%
\pgfsetstrokecolor{currentstroke}%
\pgfsetdash{}{0pt}%
\pgfpathmoveto{\pgfqpoint{0.802523in}{4.233871in}}%
\pgfpathlineto{\pgfqpoint{5.996275in}{4.233871in}}%
\pgfusepath{stroke}%
\end{pgfscope}%
\begin{pgfscope}%
\definecolor{textcolor}{rgb}{0.000000,0.000000,0.000000}%
\pgfsetstrokecolor{textcolor}%
\pgfsetfillcolor{textcolor}%
\pgftext[x=3.399399in,y=4.317204in,,base]{\color{textcolor}{\rmfamily\fontsize{25.000000}{30.000000}\selectfont\catcode`\^=\active\def^{\ifmmode\sp\else\^{}\fi}\catcode`\%=\active\def
\end{pgfscope}%
\begin{pgfscope}%
\pgfsetbuttcap%
\pgfsetmiterjoin%
\definecolor{currentfill}{rgb}{1.000000,1.000000,1.000000}%
\pgfsetfillcolor{currentfill}%
\pgfsetfillopacity{0.800000}%
\pgfsetlinewidth{1.003750pt}%
\definecolor{currentstroke}{rgb}{0.800000,0.800000,0.800000}%
\pgfsetstrokecolor{currentstroke}%
\pgfsetstrokeopacity{0.800000}%
\pgfsetdash{}{0pt}%
\pgfpathmoveto{\pgfqpoint{4.259928in}{3.616692in}}%
\pgfpathlineto{\pgfqpoint{5.801830in}{3.616692in}}%
\pgfpathquadraticcurveto{\pgfqpoint{5.857386in}{3.616692in}}{\pgfqpoint{5.857386in}{3.672248in}}%
\pgfpathlineto{\pgfqpoint{5.857386in}{4.039427in}}%
\pgfpathquadraticcurveto{\pgfqpoint{5.857386in}{4.094982in}}{\pgfqpoint{5.801830in}{4.094982in}}%
\pgfpathlineto{\pgfqpoint{4.259928in}{4.094982in}}%
\pgfpathquadraticcurveto{\pgfqpoint{4.204373in}{4.094982in}}{\pgfqpoint{4.204373in}{4.039427in}}%
\pgfpathlineto{\pgfqpoint{4.204373in}{3.672248in}}%
\pgfpathquadraticcurveto{\pgfqpoint{4.204373in}{3.616692in}}{\pgfqpoint{4.259928in}{3.616692in}}%
\pgfpathlineto{\pgfqpoint{4.259928in}{3.616692in}}%
\pgfpathclose%
\pgfusepath{stroke,fill}%
\end{pgfscope}%
\begin{pgfscope}%
\pgfsetbuttcap%
\pgfsetroundjoin%
\pgfsetlinewidth{2.007500pt}%
\definecolor{currentstroke}{rgb}{0.501961,0.501961,0.501961}%
\pgfsetstrokecolor{currentstroke}%
\pgfsetdash{{7.400000pt}{3.200000pt}}{0.000000pt}%
\pgfpathmoveto{\pgfqpoint{4.315484in}{3.881055in}}%
\pgfpathlineto{\pgfqpoint{4.593262in}{3.881055in}}%
\pgfpathlineto{\pgfqpoint{4.871039in}{3.881055in}}%
\pgfusepath{stroke}%
\end{pgfscope}%
\begin{pgfscope}%
\definecolor{textcolor}{rgb}{0.000000,0.000000,0.000000}%
\pgfsetstrokecolor{textcolor}%
\pgfsetfillcolor{textcolor}%
\pgftext[x=5.093262in,y=3.783833in,left,base]{\color{textcolor}{\rmfamily\fontsize{20.000000}{24.000000}\selectfont\catcode`\^=\active\def^{\ifmmode\sp\else\^{}\fi}\catcode`\%=\active\def
\end{pgfscope}%
\end{pgfpicture}%
\makeatother%
\endgroup%

%% file: main.bbl
\begin{thebibliography}{28}
\providecommand{\natexlab}[1]{#1}
\providecommand{\url}[1]{\texttt{#1}}
\expandafter\ifx\csname urlstyle\endcsname\relax
  \providecommand{\doi}[1]{doi: #1}\else
  \providecommand{\doi}{doi: \begingroup \urlstyle{rm}\Url}\fi

\bibitem[Abdar et~al.(2021)Abdar, Pourpanah, Hussain, Rezazadegan, Liu, Ghavamzadeh, Fieguth, Cao, Khosravi, Acharya, Makarenkov, and Nahavandi]{abdar}
Moloud Abdar, Farhad Pourpanah, Sadiq Hussain, Dana Rezazadegan, Li~Liu, Mohammad Ghavamzadeh, Paul~W. Fieguth, Xiaochun Cao, Abbas Khosravi, U.~Rajendra Acharya, Vladimir Makarenkov, and Saeid Nahavandi.
\newblock A review of uncertainty quantification in deep learning: Techniques, applications and challenges.
\newblock \emph{Inf. Fusion}, 76:\penalty0 243--297, 2021.
\newblock \doi{10.1016/J.INFFUS.2021.05.008}.

\bibitem[Bishop(1994)]{bishop_1994_mixture_density}
Christopher~M Bishop.
\newblock Mixture density networks.
\newblock 1994.

\bibitem[Camargo et~al.(2019)Camargo, Dumas, and Rojas]{camargo_2019_lstm_bpm}
Manuel Camargo, Marlon Dumas, and Oscar~Gonz{\'{a}}lez Rojas.
\newblock Learning accurate {LSTM} models of business processes.
\newblock In \emph{Business Process Management - BPM}, pages 286--302, 2019.
\newblock \doi{10.1007/978-3-030-26619-6\_19}.

\bibitem[Ceravolo et~al.(2024)Ceravolo, Comuzzi, De~Weerdt, Di~Francescomarino, and Maggi]{ceravolo_2024_ppm_trends}
Paolo Ceravolo, Marco Comuzzi, Jochen De~Weerdt, Chiara Di~Francescomarino, and Fabrizio~Maria Maggi.
\newblock Predictive process monitoring: concepts, challenges, and future research directions.
\newblock \emph{Process Science}, 1\penalty0 (1):\penalty0 2, 2024.

\bibitem[Chen et~al.(2018)Chen, Badrinarayanan, Lee, and Rabinovich]{chen_2018_gradnorm}
Zhao Chen, Vijay Badrinarayanan, Chen-Yu Lee, and Andrew Rabinovich.
\newblock Gradnorm: Gradient normalization for adaptive loss balancing in deep multitask networks.
\newblock In \emph{International Conference on Machine Learning - ICML}, pages 794--803, 2018.

\bibitem[Evermann et~al.(2017)Evermann, Rehse, and Fettke]{evermann_2017_lstm_suffix}
Joerg Evermann, Jana{-}Rebecca Rehse, and Peter Fettke.
\newblock Predicting process behaviour using deep learning.
\newblock \emph{Decis. Support Syst.}, 100:\penalty0 129--140, 2017.
\newblock \doi{10.1016/J.DSS.2017.04.003}.

\bibitem[Gal and Ghahramani(2016{\natexlab{a}})]{gal_2016_dropout}
Yarin Gal and Zoubin Ghahramani.
\newblock Dropout as a bayesian approximation: Representing model uncertainty in deep learning.
\newblock In \emph{International Conference on Machine Learning - ICML}, pages 1050--1059, 2016{\natexlab{a}}.
\newblock URL \url{http://proceedings.mlr.press/v48/gal16.html}.

\bibitem[Gal and Ghahramani(2016{\natexlab{b}})]{gal_2016_dropout_rnn}
Yarin Gal and Zoubin Ghahramani.
\newblock A theoretically grounded application of dropout in recurrent neural networks.
\newblock In \emph{Advances in Neural Information Processing Systems - NeurIPS}, pages 1019--1027, 2016{\natexlab{b}}.
\newblock URL \url{https://proceedings.neurips.cc/paper/2016/hash/076a0c97d09cf1a0ec3e19c7f2529f2b-Abstract.html}.

\bibitem[Gunnarsson et~al.(2023)Gunnarsson, vanden Broucke, and Weerdt]{gunnarsson_2023_}
Bj{\"{o}}rn~Rafn Gunnarsson, Seppe vanden Broucke, and Jochen~De Weerdt.
\newblock A direct data aware {LSTM} neural network architecture for complete remaining trace and runtime prediction.
\newblock \emph{{IEEE} Trans. Serv. Comput.}, 16\penalty0 (4):\penalty0 2330--2342, 2023.
\newblock \doi{10.1109/TSC.2023.3245726}.

\bibitem[Hochreiter and Schmidhuber(1997)]{hochreiter_1997_lstm}
Sepp Hochreiter and Jürgen Schmidhuber.
\newblock Long {Short}-{Term} {Memory}.
\newblock \emph{Neural Computation}, 9\penalty0 (8):\penalty0 1735--1780, 1997.
\newblock \doi{10.1162/neco.1997.9.8.1735}.

\bibitem[H{\"{u}}llermeier and Waegeman(2021)]{hullermeier_2021_alea_epi_unc}
Eyke H{\"{u}}llermeier and Willem Waegeman.
\newblock Aleatoric and epistemic uncertainty in machine learning: an introduction to concepts and methods.
\newblock \emph{Mach. Learn.}, 110\penalty0 (3):\penalty0 457--506, 2021.
\newblock \doi{10.1007/S10994-021-05946-3}.

\bibitem[Kendall and Gal(2017)]{kendall_2017_caeu}
Alex Kendall and Yarin Gal.
\newblock What uncertainties do we need in bayesian deep learning for computer vision?
\newblock In \emph{Advances in Neural Information Processing Systems - NeurIPS}, pages 5574--5584, 2017.
\newblock URL \url{https://proceedings.neurips.cc/paper/2017/hash/2650d6089a6d640c5e85b2b88265dc2b-Abstract.html}.

\bibitem[Ketyk{\'{o}} et~al.(2022)Ketyk{\'{o}}, Mannhardt, Hassani, and van Dongen]{ketyko_2022_lstm_suffix}
Istv{\'{a}}n Ketyk{\'{o}}, Felix Mannhardt, Marwan Hassani, and Boudewijn~F. van Dongen.
\newblock What averages do not tell: predicting real life processes with sequential deep learning.
\newblock In \emph{{ACM/SIGAPP} Symposium on Applied Computing - {SAC}}, pages 1128--1131, 2022.
\newblock \doi{10.1145/3477314.3507179}.

\bibitem[Klein(2024)]{klein_distributional}
Nadja Klein.
\newblock Distributional regression for data analysis.
\newblock \emph{Annual Review of Statistics and Its Application}, 11, 2024.
\newblock \doi{10.1146/annurev-statistics-040722-053607}.

\bibitem[Lin et~al.(2019)Lin, Wen, and Wang]{lin_2019_ppm}
Li~Lin, Lijie Wen, and Jianmin Wang.
\newblock Mm-pred: {A} deep predictive model for multi attribute event sequence.
\newblock In \emph{{SIAM} International Conference on Data Mining - {SDM}}, pages 118--126, 2019.
\newblock \doi{10.1137/1.9781611975673.14}.

\bibitem[Maggi et~al.(2014)Maggi, Francescomarino, Dumas, and Ghidini]{maggi_2014_ppm}
Fabrizio~Maria Maggi, Chiara~Di Francescomarino, Marlon Dumas, and Chiara Ghidini.
\newblock Predictive monitoring of business processes.
\newblock In \emph{Advanced Information Systems Engineering - CAiSE}, pages 457--472. Springer, 2014.
\newblock \doi{10.1007/978-3-319-07881-6\_31}.

\bibitem[Mehdiyev et~al.(2025)Mehdiyev, Majlatow, and Fettke]{mehdiyev_2025_post_hoc_unc_ppm}
Nijat Mehdiyev, Maxim Majlatow, and Peter Fettke.
\newblock Augmenting post-hoc explanations for predictive process monitoring with uncertainty quantification via conformalized monte carlo dropout.
\newblock \emph{Data Knowl. Eng.}, 156:\penalty0 102402, 2025.
\newblock \doi{10.1016/J.DATAK.2024.102402}.

\bibitem[Pasquadibisceglie et~al.(2024)Pasquadibisceglie, Appice, and Malerba]{Pasquadibisceglie_2024_lupin}
Vincenzo Pasquadibisceglie, Annalisa Appice, and Donato Malerba.
\newblock {LUPIN:} {A} {LLM} approach for activity suffix prediction in business process event logs.
\newblock In \emph{International Conference on Process Mining - ICPM}, pages 1--8, 2024.
\newblock \doi{10.1109/ICPM63005.2024.10680620}.

\bibitem[Pauwels and Calders(2020)]{pauwels_2020_bn_ppm}
Stephen Pauwels and Toon Calders.
\newblock Bayesian network based predictions of business processes.
\newblock In \emph{Business Process Management Forum - {BPM} Forum}, pages 159--175. Springer, 2020.
\newblock \doi{10.1007/978-3-030-58638-6\_10}.

\bibitem[Portolani et~al.(2022)Portolani, Brusaferri, Ballarino, and Matteucci]{portolani_2022_unc_ppm}
Pietro Portolani, Alessandro Brusaferri, Andrea Ballarino, and Matteo Matteucci.
\newblock Uncertainty in predictive process monitoring.
\newblock In \emph{Information Processing and Management of Uncertainty in Knowledge-Based Systems - IPMU}, pages 547--559, 2022.
\newblock \doi{10.1007/978-3-031-08974-9\_44}.

\bibitem[Rama{-}Maneiro et~al.(2024)Rama{-}Maneiro, Vidal, Lama, and Monteagudo{-}Lago]{rama_maneiro_2024_exp_rnn}
Efr{\'{e}}n Rama{-}Maneiro, Juan~Carlos Vidal, Manuel Lama, and Pablo Monteagudo{-}Lago.
\newblock Exploiting recurrent graph neural networks for suffix prediction in predictive monitoring.
\newblock \emph{Computing}, 106\penalty0 (9):\penalty0 3085--3111, 2024.
\newblock \doi{10.1007/S00607-024-01315-9}.

\bibitem[Rauch et~al.(2024)Rauch, Frey, Zellner, and Seidl]{rauch_2024_pa_bayesian}
Simon Rauch, Christian M.~M. Frey, Ludwig Zellner, and Thomas Seidl.
\newblock Process-aware bayesian networks for sequential event log queries.
\newblock In \emph{International Conference on Process Mining - ICPM}, pages 161--168, 2024.
\newblock \doi{10.1109/ICPM63005.2024.10680678}.

\bibitem[Salinas et~al.(2020)Salinas, Flunkert, Gasthaus, and Januschowski]{salinas}
David Salinas, Valentin Flunkert, Jan Gasthaus, and Tim Januschowski.
\newblock Deepar: Probabilistic forecasting with autoregressive recurrent networks.
\newblock \emph{International Journal of Forecasting}, 36\penalty0 (3):\penalty0 1181--1191, 2020.
\newblock \doi{0.1016/j.ijforecast.2019.07.001}.

\bibitem[Tax et~al.(2017)Tax, Verenich, Rosa, and Dumas]{tax_2017_ppm_lstm}
Niek Tax, Ilya Verenich, Marcello~La Rosa, and Marlon Dumas.
\newblock Predictive business process monitoring with {LSTM} neural networks.
\newblock In \emph{Advanced Information Systems Engineering - CAiSE}, pages 477--492, 2017.
\newblock \doi{10.1007/978-3-319-59536-8\_30}.

\bibitem[Taymouri et~al.(2021)Taymouri, Rosa, and Erfani]{taymouri_2021_deep_adv_model}
Farbod Taymouri, Marcello~La Rosa, and Sarah~M. Erfani.
\newblock A deep adversarial model for suffix and remaining time prediction of event sequences.
\newblock In \emph{{SIAM} International Conference on Data Mining - {SDM}}, pages 522--530, 2021.
\newblock \doi{10.1137/1.9781611976700.59}.

\bibitem[Weytjens and Weerdt(2022)]{weytjens_2022_unc_ppm}
Hans Weytjens and Jochen~De Weerdt.
\newblock Learning uncertainty with artificial neural networks for predictive process monitoring.
\newblock \emph{Appl. Soft Comput.}, 125:\penalty0 109134, 2022.
\newblock \doi{10.1016/J.ASOC.2022.109134}.

\bibitem[Wuyts et~al.(2024)Wuyts, vanden Broucke, and Weerdt]{wuyts_2024_sutran}
Brecht Wuyts, Seppe K. L.~M. vanden Broucke, and Jochen~De Weerdt.
\newblock Sutran: an encoder-decoder transformer for full-context-aware suffix prediction of business processes.
\newblock In \emph{International Conference on Process Mining - ICPM}, pages 17--24, 2024.
\newblock \doi{10.1109/ICPM63005.2024.10680671}.

\bibitem[Zhu and Laptev(2017)]{zhu_2017_seq2seq_unc}
Lingxue Zhu and Nikolay Laptev.
\newblock Deep and confident prediction for time series at uber.
\newblock In \emph{{IEEE} International Conference on Data Mining Workshops - ICDM}, pages 103--110, 2017.
\newblock \doi{10.1109/ICDMW.2017.19}.

\end{thebibliography}
